\documentclass[11pt]{ucthesis}

\usepackage{epsfig}
\usepackage{graphicx}
\usepackage{soul}

\usepackage{lsalike}
\usepackage{tree-dvips}
\usepackage{pst-jftree}
\usepackage{pstricks}
\usepackage{pst-xkey}
\usepackage{pst-jtree}
\usepackage{pst-node}
\usepackage{amssymb}

\def\dsp{\def\baselinestretch{2.0}\large\normalsize}
\dsp
\begin{document}


\title{Using the Web as an Implicit Training Set:\\Application to Noun Compound Syntax and Semantics}
\author{Preslav Ivanov Nakov}
\degreeyear{2007}
\degreesemester{Fall}
\degree{Doctor of Philosophy}
\chairone{Professor Marti Hearst}
\chairtwo{Professor Dan Klein}
\othermembers{Professor Jerome Feldman\\
Professor Lynn Nichols}
\numberofmembers{4}
\prevdegrees{Magistar (Sofia University, Bulgaria) 2001}
\field{Computer Science}
\campus{Berkeley}

\maketitle
\approvalpage
\copyrightpage

\begin{abstract}

An important characteristic of English written text is the abundance
of noun compounds -- sequences of nouns acting as a single noun,
e.g., {\it colon cancer tumor suppressor protein}.
While eventually mastered by domain experts, their interpretation
poses a major challenge for automated analysis.
Understanding noun compounds' syntax and semantics is important
for many natural language applications,
including question answering, machine translation, information retrieval, and information extraction.
For example, a question answering system might need to know whether
`{\it protein acting as a tumor suppressor}' is an acceptable paraphrase
of the noun compound {\it tumor suppressor protein},
and an information extraction system might need to decide if the terms
{\it neck vein thrombosis} and {\it neck thrombosis} can possibly co-refer
when used in the same document.
Similarly, a phrase-based machine translation system facing the unknown phrase
{\it WTO Geneva headquarters}, could benefit from being able to paraphrase it as
{\it Geneva headquarters of the WTO} or {\it WTO headquarters located in Geneva}.
Given a query like {\it migraine treatment}, an information retrieval system
could use paraphrasing verbs like {\it relieve} and {\it prevent}
for page ranking and query refinement.

I address the problem of noun compounds syntax
by means of novel, highly accurate unsupervised and lightly supervised algorithms
using the Web as a corpus and search engines as interfaces to that corpus.
Traditionally the Web has been viewed as a source of page hit counts,
used as an estimate for $n$-gram word frequencies.
I extend this approach by introducing novel surface features
and paraphrases, which yield state-of-the-art results for the task of noun compound bracketing.
I also show how these kinds of features can be applied to other structural ambiguity problems,
like prepositional phrase attachment and noun phrase coordination.
I address noun compound semantics by automatically generating paraphrasing verbs
and prepositions that make explicit the hidden semantic relations between the nouns
in a noun compound.
I also demonstrate how these paraphrasing verbs can be used to solve various relational similarity problems,
and how paraphrasing noun compounds can improve machine translation.



\abstractsignature
\end{abstract}

\begin{frontmatter}

\begin{dedication}
\null\vfil
{\large
\begin{center}
To Petya.
\end{center}}
\vfil\null
\end{dedication}

\tableofcontents
\begin{acknowledgements}

First, I would like to thank Marti Hearst for being such a fantastic advisor:
for helping and encouraging me throughout my graduate studies,
and for giving me the freedom to pursue my ideas,
but also for pushing me to get things done in time when it was necessary.

Next, I would like to thank all members of my dissertation committee,
Marti Hearst, Dan Klein, Jerome Feldman, and Lynn Nichols
for their comments, which were invaluable for shaping
the final version of the present thesis.
Many additional comments on particular thesis topics are due to
Roxana Girju, Elena Paskaleva, Zornitza Kozareva and Ayla Algar.
I am also very grateful to Michael Jordan
for having served in the committee for my qualifying exam,
and to Galia Angelova, my master's thesis advisor,
who was the first to introduce me to the field of computational linguistics.

Thanks to all present and former members of the BioText group at Berkeley:
I will really miss you guys!
I am especially grateful to Ariel Schwartz, Anna Divoli, and Barbara Rosario
for the many fruitful discussions and collaborations throughout the years.
Many thanks to all other (former) members of the group:
Gaurav Bhalotia, Janice Hamer, Emilia Stoica, Mike Wooldridge, Brian Wolf,
Jerry Ye, Harendra Guturu, Rowena Luk, Tingting Zhang,
Alex Ksikes and Megan Richardson.

Last but not least, I would like to express my deepest gratitude for the love and support of my family --
to my wonderful wife Petya, to my mother Ivanka, and to my brother Svetlin --
for their support and encouragement throughout the long years of my graduate studies.
\end{acknowledgements}

\end{frontmatter}

\chapter{Introduction}


\vspace{12pt}
\begin{quotation}
``\emph{Recent studies identify the
\underline{colon cancer tumor suppressor protein}
adenomatous polyposis coli (APC) as a
\underline{core organizer} of excitatory nicotinic synapses.}''
\begin{flushright}
\texttt{http://www.tufts.edu/sackler/tcvr/overview.htm}
\end{flushright}
\end{quotation}
\vspace{12pt}

%

An important characteristic of technical literature is the
abundance of long sequences of nouns acting as a single noun,
which are known as noun compounds.
While eventually mastered by domain experts,
noun compound interpretation poses major challenge for
automated analysis.
For example, what is the internal syntactic structure of
{\it colon cancer tumor suppressor protein}:
`{\it $[$colon cancer$]$ $[[$tumor suppressor$]$ protein$]$}' or
`{\it $[[$colon cancer$]$ $[$tumor suppressor$]]$ protein}' or
`{\it $[[[$colon cancer$]$ tumor$]$ suppressor$]$ protein}', etc.?
Can {\it colon cancer} be paraphrased as `{\it cancer that occurs in the column}'?
Or as `{\it cancer in the column}'?
What is the relationship between {\it colon cancer} and {\it tumor suppressor protein}?
Between {\it colon} and {\it cancer}?
Is a {\it tumor suppressor protein} a kind/type of {\it tumor suppressor}?
Is it a kind of {\it suppressor}?

Understanding noun compounds' syntax and semantics is important
for many natural language applications
including question answering, machine translation, information retrieval, and information extraction.
A question answering system might need to know whether
`{\it protein acting as a tumor suppressor}' is an acceptable paraphrase
of the noun compound {\it tumor suppressor protein},
and an information extraction system might need to decide if the terms
{\it neck vein thrombosis} and {\it neck thrombosis} could possibly co-refer
when used in the same document.
Similarly, a phrase-based machine translation system facing the unknown phrase
{\it WTO Geneva headquarters} could benefit from being able to paraphrase it as
{\it Geneva headquarters of the WTO} or {\it WTO headquarters located in Geneva}.
Given a query like {\it migraine treatment}, an information retrieval system
could use suitable paraphrasing verbs like {\it relieve} and {\it prevent}
for page ranking and query refinement.

This gives rise to the following problems:

\begin{itemize}

    \item {\bf Syntax}: What is the internal syntactic structure of the noun compound?
        E.g., is it {\it $[[$neck vein$]$ thrombosis$]$} or {\it $[$neck $[$vein thrombosis$]]$}?

    \item {\bf Paraphrasing}: Which verbs can paraphrase the noun compound?
    E.g., `{\it thrombosis that \underline{blocks} the neck vein}' is a good paraphrase for {\it neck vein thrombosis}.

    \item {\bf Semantics}: What are the semantics of the noun compounds?

    \item {\bf Semantic entailment}: Dropping which nouns yields a hypernym
 of the noun compound? E.g., is {\it neck vein thrombosis} a kind of {\it neck thrombosis}?
 See Appendix \ref{chapter:sem:entailment:nc}.

\end{itemize}
\vspace{6pt}

I focus on the first three problems, which I address with novel,
highly accurate algorithms using the Web as a corpus and search engines as interfaces to that corpus.

Traditionally in natural language processing (NLP) research,
the Web has been used as a source of page hit counts,
which are used as an estimate for $n$-gram word frequencies.
I extend this approach with novel surface features and paraphrases
which go beyond simple $n$-gram counts and prove highly effective,
yielding a statistically significant improvement over the previous
state-of-the-art results for noun compound bracketing.

I further address noun compound semantics by automatically generating paraphrasing verbs
and prepositions that make explicit the hidden semantic relations between the nouns in a
noun compound.
Using them as features in a classifier,
I demonstrate state-of-the-art results on various relational similarity problems:
mapping noun-modifier pairs to abstract relations like \texttt{TIME}, \texttt{LOCATION} and \texttt{CONTAINER},
classifying relation between nominals and solving SAT verbal analogy questions.

Finally, I demonstrate the potential of these techniques by applying them to
machine translation, prepositional phrase attachment, and noun phrase coordination.

The remainder of the thesis is organized as follows:

\begin{itemize}
  \item In chapter \ref{chapter:nc},
    I discuss noun compounds from a linguistic point of view:
    First, I describe the process of compounding in general,
    as a mechanism of producing a new word by putting two or more existing ones together.
    Then, I focus on the special case of {\it noun compounds}:
    I discuss their properties, I provide several definitions
    (including one of my own to be used throughout the rest of the thesis),
    and I describe some interesting variations across language families.
    Finally, I explore some important theories,
    including \quotecite{levi:1978} theory of recoverably deletable predicates,
    \quotecite{lauer:1995:thesis} prepositional semantics,
    and \quotecite{Rosario:al:2002:descent} descent of hierarchy.

  \item In chapter \ref{chapter:NC:bracketing},
    I describe a novel, highly accurate lightly supervised method for
    making left vs. right bracketing decisions for three-word noun compounds.
    For example, {\it $[[$tumor suppressor$]$ protein$]$} is left-bracketed,
    while {\it $[$world $[$food production$]]$} is right-bracketed.
    Traditionally, the problem has been addressed using unigram and bigram frequency
    estimates to compute adjacency- and dependency-based word association scores.
    I extend this approach by introducing novel surface features and paraphrases
    extracted from the Web. Using combinations of these features,
    I demonstrate state-of-the-art results on two separate collections,
    one consisting of terms drawn from encyclopedia text,
    and another one extracted from bioscience text.

  \item In chapter \ref{chapter:NC:semantics},
    I present a novel, simple, unsupervised method
    for characterizing the semantic relations that hold between nouns in noun-noun
    compounds. The main idea is to look for {\it predicates} that make
    explicit the hidden relations between the nouns.  This is
    accomplished by writing Web search engine queries that restate the
    noun compound as a relative clause containing a wildcard character
    to be filled in with a verb.  A comparison to results from the
    literature and to human-generated verb paraphrases
    suggests this is a promising approach.
    Using these verbs as features in classifiers,
    I demonstrate state-of-the-art results on various relational similarity problems:
    mapping noun-modifier pairs to abstract relations like \texttt{TIME} and \texttt{LOCATION},
    classifying relation between nominals, and solving SAT verbal analogy problems.

    \item Chapter \ref{chapter:MT} describes an application
    of the methods developed in chapters \ref{chapter:NC:bracketing} and \ref{chapter:NC:semantics}
    for noun compound paraphrasing to an important real-world task: {\it machine translation}.
    I propose a novel monolingual paraphrasing method
    based on syntactic transformations at the NP-level,
    which augments the training data with nearly equivalent
    sentence-level syntactic paraphrases of the original corpus,
    focused on the noun compounds.
    The idea is to recursively generate sentence variants
    where noun compounds are paraphrased using suitable prepositions,
    and vice-versa -- NPs with an internal PP-attachment are turned into noun compounds.
    The evaluation results show an improvement equivalent to 33\%-–50\% of that
    of doubling the amount of training data.

    \item
    Chapter \ref{chapter:appl} describes applications
    of the methods developed in chapter \ref{chapter:NC:bracketing}
    to two important structural ambiguity problems a syntactic parser faces --
    {\it prepositional phrase attachment} and {\it NP coordination}.
    Using word-association scores, Web-derived surface features and paraphrases,
    I achieve results that are on par with the state-of-the-art.

  \item Web search engines provide easy access for NLP researchers to
    world's biggest corpus, but this does not come without drawbacks.
    In chapter \ref{chapter:instability},
    I point to some problems and limitations of using
    search engine page hits as a proxy for $n$-gram frequency estimates.
    I further describe a study on the stability of such estimates
    across search engines and over time,
    as well as on the impact of using word inflections and of limiting the queries to English pages.
    Using the task of noun compound bracketing and 14 different $n$-gram based models,
    I illustrate that while sometimes causing sizable fluctuations,
    variability's impact generally is not statistically significant.

    \item Finally, chapter \ref{chapter:conclusion}
    lists my contributions and points to directions for future work.

\end{itemize}

I believe these efforts constitute a significant step towards the goal
of automatic interpretation of English noun compounds. 

\chapter{Noun Compounds}
\label{chapter:nc}

In this chapter, I discuss noun compounds from a linguistic point of view:
First, I describe the process of compounding in general,
as a mechanism of producing a new word by putting two or more existing ones together.
Then, I focus on the special case of {\it noun compounds}:
I discuss their properties, I provide several definitions
(including one of my own to be used throughout the rest of the thesis),
and I describe some interesting variations across language families.
Finally, I explore some important linguistic theories,
including \quotecite{levi:1978} theory of recoverably deletable predicates,
\quotecite{lauer:1995:thesis} prepositional semantics,
and \quotecite{Rosario:al:2002:descent} descent of hierarchy.

\section{The Process of Compounding}
\label{sec:compounding:process}

The {\it Dictionary of Grammatical Terms in Linguistics}
defines the process of {\it compounding} as follows \cite{Trask:1993}:

\begin{quote}
``The process of forming a word by combining two or more existing
words: {\it newspaper}, {\it paper-thin}, {\it babysit}, {\it video
game}.''
\end{quote}

As the provided examples show, the words forming an English
compound can appear orthographically separated,
connected with a hyphen, or concatenated;
often the same compound can be written in different ways,
e.g., {\it health care}, {\it health-care}, {\it healthcare}.

Since the process of compounding constructs new words,
these words can in turn combine with other words to form longer compounds,
and this process can be repeated indefinitely, e.g.,
{\it orange juice},
{\it orange juice company},
{\it orange juice company homepage},
{\it orange juice company homepage logo},
{\it orange juice company homepage logo update}, etc.

Some authors impose further restrictions, e.g.,
\namecite{Liberman:Sproat:1992:NP} consider `true' compounds only
the ones representing lexical objects.
Under their definition, {\it honeymoon} would be a compound,
but {\it honey production} would not be.


\namecite{Chomsky:Halle:1991} give a phonological definition:
the words preceding a noun form a {\it compound} with it
if they receive the primary stress. Therefore,
{\it blackboard} is a compound ({\it black} gets the primary stress),
but {\it black board} is not (equal stress).
This definition is appealing since it stems from a standard linguistic test
of whether a sequence of lexemes represents a single word,
and is consistent with \quotecite{Trask:1993} definition above,
which views the process of compounding as a new word formation process.
However, it is problematic since pronunciation can vary across different speakers.
It is also of limited use for most computational approaches
to language analysis, which work with written text,
where information about the stress is not available.

On the other hand, knowing whether a sequence of words represents
a compound might be useful for a speech synthesis system,
where using the wrong stress can convey a meaning that is different from what is intended.
For example, using a fronted stress would make {\it French teacher} a compound with the
meaning of {\it `teacher who teaches French'}, while a double
stress would convey the non-compound meaning of `{\it teacher who is French}'
\cite{levi:1978}.
For compounds longer than two words,
the correct pronunciation also depends on their internal syntactic structure,
which makes a noun compound parser
an indispensable component of an ``ideal'' speech synthesis system.
If the internal syntactic structure is known,
the stress assignment algorithm of \namecite{Chomsky:Halle:Lukoff:1956}
can be used, which follows the constituent structure from the inside out,
making the stress of the first constituent primary and reducing the rest by one degree.
For example, if {\it black board} is not a compound, each word would get a
primary stress {\it $[$black/1$]$ $[$board/1$]$},
but if it is a compound then the stress of the second word would be reduced:
{\it $[$black/1 board/2$]$}. Now, if {\it $[$black/1 board/2$]$}
is a sub-constituent of a longer compound, e.g.,
{\it black board eraser}, the rule would be applied one more time,
yielding {\it $[[$black/1 board/3$]$ eraser/2$]$}, etc.

Although I am interested only in noun compounds,
in English the process of compounding can put together words belonging to
various parts of speech:
noun+noun (e.g., {\it silkworm}, {\it honey bee}, {\it bee honey}, {\it stem cell}),
adjective+noun (e.g., {\it hot dog}, {\it shortlist}, {\it white collar}, {\it highlife}),
verb+noun (e.g., {\it pickpocket}, {\it cutthroat}, {\it know-nothing}),
preposition+noun (e.g., {\it counter-attack}, {\it underwater}, {\it indoor}, {\it upper-class}),
noun+adjective (e.g., {\it trigger-happy}, {\it army strong}, {\it bulletproof}, {\it dog tired}, {\it waterfull}, {\it English-specific}, {\it brand-new}),
adjective+adjective (e.g., {\it dark-green}, {\it south-west}, {\it dry-clean}, {\it leftmost}),
verb+adjective (e.g., {\it timbledown}),
preposition+adjective (e.g., {\it overeager}, {\it over-ripe}),
noun+verb (e.g., {\it hand wash}, {\it finger-point}, {\it taperecord}),
adjective+verb (e.g., {\it highlight}, {\it broadcast}, {\it quick-freeze}),
verb+verb (e.g., {\it freeze-dry}),
preposition+verb (e.g., {\it overestimate}, {\it withdraw}, {\it upgrade}, {\it withhold}),
noun+preposition (e.g., {\it love-in}, {\it timeout}), {\it breakup}),
verb+preposition (e.g., {\it countdown}, {\it stand-by}, {\it cut-off}, {\it castaway}),
adjective+preposition (e.g., {\it blackout}, {\it forthwith}),
preposition+preposition (e.g., {\it within}, {\it without}, {\it into}, {\it onto}).
In a typical compound, the second word is the head, and the first one is a modifier,
which modifies or attributes a property to the head.
The part of speech of the compound is the same as that of the head.
Some authors also allow complex structures like
{\it state-of-the-art}, {\it part-of-speech} or {\it over-the-counter eye drop}.


\section{Defining the Notion of \emph{Noun Compound}}
\label{sec:compounding:notion}

There is little agreement in the research community on how to define
the notion of {\it noun compound}.\footnote{Under most definitions,
the term {\it noun compound} itself is an example of a noun compound.}
Different authors have different definitions, focusing
on different aspects, and often use different terms in order to emphasize
particular distinctions. \namecite{lauer:1995:thesis} provides the
following list of closely related notions, used in the literature:
{\it complex nominal}, {\it compound}, {\it compound nominal}, {\it
compound noun}, {\it nominal compound}, {\it nominalization}, {\it
noun compound}, {\it noun premodifier}, {\it noun sequence}, {\it
noun-noun compound}, {\it noun+noun compound}.
While some of these terms are broader and some are narrower in scope,
most of them refer to objects that are syntactically analyzable as nouns \cite{Chomsky:Halle:1991,Jackendoff:1975:lexicon}.
Some of their building elements however can belong to other parts-of-speech,
e.g., {\it complex nominals} allow for adjectival modifiers (see below).

Since noun compounds are syntactically analyzed as nouns,
and since most of them exhibit some degree of lexicalization,
it might be tempting to make them all lexical entries.
Unfortunately, since the process of compounding can
theoretically produce an unlimited number of noun compounds,
this would effectively open the door to an infinite lexicon,
which would create more problems than it would
solve.
\namecite{levi:1978} proposes the following solution for English
(which does not apply to all kinds of noun compounds though):
``\emph{Complex nominals are all derived from an underlying NP structure
    containing a head noun and a full S in either a relative clause
    or NP complement construction; on the surface, however,
    the complex nominal is dominated by a node label of N.}''

\namecite{downing:1977:nc:sem} defines a {\it nominal compound}
as a sequence of nouns which function as a single noun,
e.g., {\it orange juice}, {\it company tax policy}, {\it law enforcement officer},
{\it colon cancer tumor suppressor protein},
etc. While being more restrictive than most
alternatives, this definition is both simple and relatively
unambiguous, which makes it the preferred choice for computational purposes.

\namecite{Quirk:Greenbaum:Leech:Svartvik:1985} allow for any constituent
to form a {\it premodified noun} with the following noun,
e.g., {\it out-of-the-box solution}.
Unfortunately, such a liberal definition makes it hard to distinguish
a noun premodifier from an adjectival modification.

\namecite{levi:1978} introduces the concept of
{\it complex nominals}, which groups three
partially overlapping classes: {\it nominal compounds} (e.g.,
{\it doghouse}, {\it deficiency disease}, {\it apple cake}), {\it nominalizations}
(e.g., {\it American attack}, {\it presidential refusal}, {\it dream analysis})
and {\it nonpredicate NPs}\footnote{Nonpredicate NPs contain a modifying adjective which
cannot be used in predicate position with the same meaning. For
example, the NP {\it solar generator} could not be paraphrased as
*`{\it generator which is solar}'.} (e.g., {\it electric shock}, {\it
electrical engineering}, {\it musical critism}).

None of the above definitions can clearly distinguish
between noun compounds and other nominal phrases.
\namecite{levi:1978} lists the following three basic criteria that have been proposed
for this purpose in the literature:
{\it fronted stress}, {\it permanent aspect}, and {\it semantic specialization}.
While each of them captures some useful characteristics that are valid for most noun compounds,
they are of limited use as tests for distinguishing a noun compound from other nominals.

The first criterion (that I have already introduced above),
states that noun compounds should have a fronted stress.
For example, {\it baby photographer}
would be a noun compound only if pronounced with a fronted stress,
(in which case it means `photographer who photographs babies'),
but not with a normal stress
(in which case it would would mean `photographer who is a baby').
This criterion is problematic since stress could differ across dialects
and even across speakers of the same dialect. It also
separates semantically parallel examples, e.g.,
it accepts {\it apple cake} as a noun compound, but not {\it apple pie},
which is undesirable.

The second criterion requires that the words forming a noun compound
be in a permanent or at least habitual relationship, e.g., {\it desert rat}
can only refer to rats that are strongly associated with a desert, e.g.,
living in or around it. Unfortunately, this criterion rejects examples like
{\it heart attack}, {\it car accident}, and {\it birth trauma}.

The last criterion asks that noun compounds be at least partially lexicalized.
Unfortunately, lexicalization is a matter of personal judgement and interpretation.
For example, while {\it birdbrain} is an accepted lexicalization with a specialized meaning,
`{\it ham sandwich}' (referring to a person ordering a ham sandwich)
is an innovative one with a more general meaning,
many variations on which can be constructed on the fly.
Lexicalization is also a matter of degree which makes it problematic to use as a test.
Some noun compounds are completely lexicalized and therefore would be considered
single lexical item/single lexical entries, e.g., {\it honeymoon} has nothing
to do with {\it honey} or a {\it moon}.
Other could be argued to be productively composed, e.g., {\it orange peel}.
Many other lie in the continuum between, e.g.,
{\it boy friend} and {\it healthcare} exhibit a low degree of lexicalization.
Towards the other end lie the metaphorical
{\it ladyfinger} and {\it birdbrain}, which are highly lexicalized,
but are still partially transparent, e.g., {\it ladyfinger} is a pastry
that resembles a lady finger, and {\it birdbrain} is a person whose brain
supposedly has the size of a bird's brain.

The partial lexicalization can sometimes be seen by the variation in spelling,
e.g., both {\it health care} and {\it healthcare} are commonly used,
and the latter seems to suggest the author's belief of a higher degree of lexicalization
compared to the space-separated version.
Similarly, the concatenated {\it bathroom} is more lexicalized than {\it game room}.
In some cases, a high degree of lexicalization can be signalled by
spelling changes in the compounded form,
e.g., {\it \underline{do}ugh} + {\it \underline{nut}} = {\it donut}.

While most of the theories of the syntax and semantics of noun compounds
focus primarily on nonlexicalized or lightly lexicalized noun compounds,
for the purposes of my study, I do not try to distinguish or to treat differently
lexicalized vs. nonlexicalized noun compounds. Note however,
that my definition for a noun compound in section \ref{sec:compounding:my:def}
does not allow for word concatenations,
which readily eliminates most of the highly lexicalized noun compounds.

Lexicalization and transparency are related, but different notions.
Taking for example the days of the week, {\it Sunday} is highly lexicalized,
but is quite transparent: one can easily see `{\it sun} + {\it day}'.
{\it Friday} is just as highly lexicalized, but is much less transparent in English.
For comparison, the German for Friday, {\it Freitag} looks quite transparent `{\it frei} + {\it Tag'},
i.e., `{\it free day}', but this interpretation is actually wrong:
both {\it Friday} and {\it Freitag} are historically derived
from a translation of the Latin {\it dies Veneris}, which means `{\it day of Venus}'.
Words that are borrowed as readily-constructed compounds would often be opaque to English speakers,
e.g., {\it hippopotamus} comes from the Greek {\it hippopotamos},
which is an alteration of {\it hippos} {\it potamios}, i.e., `{\it riverine horse}'.

Noun compounds are sub-divided with respect to transparency
into {\it endocentric} and {\it exocentric} ones, defined by
the {\it Lexicon of Linguistics\footnote{\texttt{http://www2.let.uu.nl/Uil-OTS/Lexicon/}}}
as follows:

\vspace{12pt}
\begin{quote}

{\it Endocentric compound}: a type of compound in which one member
functions as the head and the other as its modifier, attributing a
property to the head. The relation between the members of an
endocentric compound can be schematized as `$AB$ is (a) $B$'.
Example: the English compound {\it steamboat} as compared with {\it
boat} is a modified, expanded version of boat with its range of
usage restricted, so that {\it steamboat} will be found in basically
the same semantic contexts as the noun {\it boat}. The compound also
retains the primary syntactic features of {\it boat}, since both are
nouns. Hence, a {\it steamboat} is a particular type of {\it boat},
where the class of {\it steamboats} is a subclass of the class of
{\it boats}.

{\it Exocentric compound}: a term used to refer to a particular type
of compound, viz. compounds that lack a head. Often these compounds
refer to pejorative properties of human beings. A Dutch compound
such as {\it wijsneus} `{\it wise guy}' ({\it lit.} `{\it wise-nose}') (in normal
usage) does not refer to a nose that is wise. In fact, it does not
even refer to a nose, but to a human being with a particular
property. An alternative term used for compounds such as
{\it wijsneus} is {\it bahuvrihi} compound.

\end{quote}

In general, English noun compounds are right-headed,
but this is not always the case,
e.g., {\it vitamin D} and {\it interferon alpha} are left-headed,
and some noun compounds like {\it sofa-bed}, {\it coach-player} and {\it programmer analyst} are headless.
The latter are known as coordinative, copulative or {\it dvandva} compounds;
they combine nouns with similar meaning,
and the resulting compound may be a generalization rather than a specialization.
English also allows for reduplication, e.g., {\it house house}.

Finally, for noun compounds of length three or longer, there can be multiple readings
due to structural ambiguity. For example, {\it plastic water bottle} is ambiguous between
a left- and a right-bracketing:

\begin{center}
\begin{tabular}{llll}
(1) & $[$ $[$ {\it plastic water} $]$ {\it bottle} $]$ & (left bracketing)\\
(2) & $[$ {\it plastic} $[$ {\it water bottle} $]$  $]$  & (right bracketing)
\end{tabular}
\end{center}

The correct interpretation is (2), with the meaning of a `{\it water bottle that is made of plastic}';
in (1) we have a {\it bottle} that has something to do with ``{\it plastic water}''
(which does not exist).
As I mentioned in section \ref{sec:compounding:process} above,
in spoken English, most of the time,
the correct interpretation will be signalled by the stress pattern used by the speaker.
Another way to signal the structure at speech time
is by putting a pause at the appropriate position:
`{\it plastic water \_\_ bottle}' vs. `{\it plastic \_\_ water bottle}'.

As I will show below, other languages can have more accurate mechanisms
for resolving such structural ambiguities.

\section{Noun Compounds in Other Languages}
\label{sec:nc:other:lang}

In this section, I briefly describe some interesting characteristics
of noun compounds in other languages and language families,
including
Germanic languages, whose noun compounds are written without separating blanks.
Romance languages, whose noun phrase (NP) structure is predominantly left-headed,
Slavic languages (Russian, which forms noun compounds using various grammatical cases, and
Bulgarian, which respects the original head-modifier order when borrowing foreign noun compounds),
and Turkic languages, which mark the noun compound head with a special possessive suffix.

\subsection{Germanic Languages}

As I mentioned above, noun compounds in English are typically written
with blank separators between the words,
except for short and established lexicalizations like
{\it textbook}, {\it newspaper}, {\it homework}, {\it housewife}, {\it Sunday}.
In some cases, hyphenated forms are used,
mainly for {\it dvandva} compounds, e.g., {\it coach-player}, {\it member-state},
and in compounds of length three or longer, e.g., {\it law-enforcement officer}.
In the other Germanic languages, however,
noun compounds are almost exclusively concatenated, e.g.,
{\it nagellackborttagningsmedel} (Swedish, `{\it nail polish remover}'),
{\it sannsynlighetsmaksimeringsestimator} (Norwegian, `{\it maximum likelihood estimator}'),
{\it kvindeh{\aa}ndboldlandsholdet} (Danish, `{\it the female handball national team}'),
{\it Sprachgruppe} (German, `{\it language group}'),
{\it wapenstilstandsonderhandeling} (Dutch, `{\it ceasefire negotiation}').
Concatenated noun compounds are pronounced with a fronted stress,
which is lost if the words are written separately.
Therefore, concatenation is very important; for example, in Norwegian
the concatenated front-stressed form {\it sm{\o}rbr{\o}d} means `{\it sandwich}',
while the normal stress form {\it sm{\o}r br{\o}d} means `{\it butter bread}'.
This is similar to the way stress changes meaning in English;
see the {\it English teacher} example in section \ref{sec:compounding:process} above.

Because of the concatenated spelling, noun compounds can be the source of very long words.
The classic example in German is
{\it Donaudampfschiffahrtsgesellschaftskapit\"{a}n},
which stands for `{\it Danube steamship company captain}'.
However, since there are no theoretical limits on noun compounds' length,
even longer ones have been coined and used, e.g.,
{\it Donaudampfschiffahrtselektrizit\"{a}tenhauptbetriebswerkbauunterbeamtengesellschaft},
meaning `{\it association of subordinate officials of
the head office management of the Danube steamboat electrical services}',
which is the name of a pre-war club in Vienna.


\subsection{Romance Languages}

Unlike in Germanic languages, in Romance languages noun phrases
are generally left-headed, i.e., the modifier generally follows the head it modifies,
e.g., {\it ur\^{a}nio enriquecido} (Portuguese, lit. `{\it uranium enriched}', i.e. `{\it enriched uranium}').
The same principle naturally extends from adjectives-noun modification to noun compounds,
e.g., {\it estado miembro} (Spanish, lit. `{\it state member}', meaning `{\it member-state}'),
or {\it legge quadro} (Italian, lit. `{\it law framework}', meaning `{\it framework law}').
In some cases, a linking element could be incorporated before the post-modifier,
e.g., {\it chemin-de-fer} (French, lit. `{\it road of iron}', i.e., `{\it railway}').

Romanian, the only contemporary Romance language with grammatical cases,
also allows for noun compounds formed using genitive,
which is assigned by the definite article of the
first noun to the second one, and is realized as a suffix
if the second noun is preceded by the definite article,
e.g., {\it frumuse\c{t}ea fete\underline{i}} (lit. `{\it beauty-the girl-gen}',
meaning `{\it the beauty of the girl}').
\cite{Romalo:2005,Girju:2006:cross:ling,girju:2007:LAW,girju:2007:ACLMain}

\subsection{Slavic Languages}

Like in English, the NP structure of the Slavic languages is predominantly right-headed
and adjectives precede the nouns they modify, e.g., {\it zelenaya trava} (Russian, `{\it green grass}').
Lexicalized noun compounds are written concatenated or with hyphens
and are generally right-headed as well,
e.g., {\it gorod-geroy} (Russian, `{\it city-hero}').
Nonlexicalized noun compounds, however, are left-headed, e.g.,
{\it uchebnik istorii} (Russian, lit. `{\it book history-gen}', i.e. `{\it history book}').
They are also case-inflected since Slavic languages are synthetic
(except for the analytical Bulgarian and Macedonian\footnote{There
is a heated linguistic (and political) debate about
whether Macedonian is a separate language or is a
regional literary form of Bulgarian. Since no clear criteria exist
for distinguishing a dialect from a language,
linguists remain divided on that issue.}).

While here I use the term noun compound for both lexicalized and nonlexicalized noun compounds,
traditionally, Russian grammar books use that term for neither of them;
{\it complex words} and {\it word combinations} are used instead
\cite{Rozental:1967,Rozental:1970,Rozental:1977,Gochev:al:1987,Dzhambazov:al:1994,Atanassova:2001}.

\subsubsection{Russian}

Russian forms lexicalized right-headed compounds,
which are typically included in a dictionary as separate lexical entries,
and are considered {\it complex words}.
They can be written as a single word, possibly with a connecting vowel,
e.g., {\it kinoteatr} (`{\it cinema theater}') and
{\it krov\underline{o}obrashchenie} (`{\it blood circulation}'),
or as two words connected with a hyphen,
e.g., {\it vagon-restoran}, (lit. `{\it car-restaurant}', meaning `{\it dining car (in a train)}').
In the former two cases, inflections are applied to the second word only,
but in the latter case there is a choice between inflecting both nouns (preferred in writing)
or the second noun only (preferred in spoken language).
Russian is also very productive in right-headed lexicalized noun compounds of type {\it portmanteaux},
where one or more of the nouns is shortened.
For example, in {\it zamdekan} (`{\it deputy dean}'),
the modifier {\it zamestnik} (`{\it deputy}') is shortened to {\it zam}.

Much more interesting are the nonlexicalized noun compounds,
considered by the grammars as {\it word combinations}.
They are right-headed and the modifier is typically in {\it genitive}
case\footnote{Such genitive structures are common in many other languages,
including German, Greek, and Latin.},
which can express a relation of possession of the head by the modifier,
e.g., {\it kniga uchenik\underline{a}} is formed out of {\it kniga} + {\it uchenik} + {\it a}-gen
(lit. `{\it book student-gen}', i.e. `{\it student's book}').
Genitive is also used in part-whole relations
(e.g., {\it krysha dom\underline{a}}, lit. `{\it roof house-gen}', i.e. `{\it house's roof}'),
in subject nominalizations
(e.g., {\it lay sobak\underline{i}}, lit. `{\it barking dog-gen}', i.e. `{\it dog's barking}'),
in object nominalizations
(e.g., {\it reshenie zadach\underline{i}}, lit. `{\it solution problem-gen}', i.e. `{\it problem's solution}'),
for measurements
(e.g., {\it butylka vodk\underline{i}}, lit. `{\it bottle vodka-gen}', i.e. `{\it bottle of vodka}'),
and for sets
(e.g., {\it gruppa det\underline{ey}}, lit. `{\it group children-gen}', i.e. `{\it group of children}'),

While genitive is the most frequent case for the modifier,
other cases are also possible, depending on the grammatical function
and the semantic relation between the head and the modifier.

For example, {\it dative} case is used when the head is an object
and the modifier is the recipient of that object,
e.g., {\it pamyatnik partizan\underline{am}},
(lit. `{\it monument partisans-dat}', i.e. `{\it monument to the partisans}').
It is also used in slogans, e.g.,
{\it slava geroy\underline{am}},
(lit. `{\it glory heroes-dat}', i.e. `{\it glory to the heroes}').

{\it Instrumental} case is possible as well:
to do an comparison
(e.g., {\it nos kartoshk\underline{oy}}, lit. `{\it nose potato-instr}', i.e. `{\it nose that looks like a potato}'),
to express an instrument applied to the head
(e.g., {\it porez nozh\underline{om}}, lit. `{\it cut knife-instr}', i.e. `{\it cut with a knife}'),
to express a mean of transportation
(e.g., {\it poezdka avtobus\underline{om}}, lit. `{\it travel bus-instr}', i.e. `{\it travel with a bus}'),
to describe circumstances
(e.g., {\it ezda pole\underline{m}}, lit. `{\it ride field-instr}', i.e. `{\it ride in the field}'),
in subject nominalizations
(e.g., {\it proverka uchitel\underline{em}}, lit. `{\it check teacher-instr}', i.e. `{\it checking by the teacher}'),
and in object nominalizations
(e.g., {\it zanyatie balet\underline{om}}, lit. `{\it occupation ballet-instr}', i.e. `{\it occupation with ballet}').
In the latter case, instrumental focuses on the action, while genitive focuses on the theme; compare
{\it komandovanie polk\underline{om}} (lit. `{\it commanding regiment-instr}', i.e. `{\it the action of commanding the regiment}')
and
{\it komandovanie polk\underline{a}} (lit. `{\it command regiment-gen}', i.e. `{\it the people who command the regiment}').

In all above examples, the head noun is in nominative.
If a noun compound has to be inflected by case,
the corresponding inflectional changes are applied to the head only (the first noun),
e.g., {\it kniga uchenik\underline{a}} in accusative becomes {\it knig\underline{u} uchenik\underline{a}}.

As in English, noun compounds can be much longer than two words. For example,
{\it obsuzhdenie kriteri\underline{ev} ocenk\underline{i} kachestv\underline{a} obucheni\underline{ya}}
(lit. `{\it discussion criteria-gen evaluation-gen quality-gen education-gen}',
meaning `{\it discussion on the criteria for the evaluation of the quality of education}'),
where all modifiers are in genitive. From a syntactic viewpoint, this is a right-bracketed noun compound:
\begin{center}
{\it $[$ obsuzhdenie $[$ kriteri\underline{ev} $[$ ocenk\underline{i} $[$ kachestv\underline{a} obucheni\underline{ya} $]$ $]$ $]$ $]$}
\end{center}

\noindent In Russian, different case patterns correspond to different syntactic structures:

\begin{itemize}
  \item {\bf head+gen+gen} is generally right-bracketed (as we saw above).
  \item {\bf head+instr+gen} is right-bracketed. For example, {\it voshishchenie zelen\underline{'yu} pole\underline{y}}, lit. `{\it admiration greenery-instr fields-gen}',
      meaning `{\it admiration for the greenery of the fields}'.
  \item {\bf head+gen+instr} is left-bracketed. For example, {\it zahvat vlast\underline{i} kontrrevolyucioner\underline{ami}}, lit. `{\it grab power-gen counter-revolutionary-instr}', meaning `{\it power grab by the counter-revolutionary}'.
\end{itemize}

Note that in some cases, the pattern {\bf head+gen+gen} can be ambiguous.
For example, {\it kniga istori\underline{i} uchitel\underline{ya}} (lit. `{\it book (hi)story-gen teacher-gen}')
can be interpreted as `{\it book about the teacher's story}' (right bracketing)
or as `{\it history book of the teacher}' (left bracketing; preferred reading).
As in English, in spoken language, stress and pause can help disambiguate between the two readings.
However, {\it kniga uchitel\underline{ya} istori\underline{i}} is not ambiguous;
it has a right-bracketing interpretation only, meaning `{\it book of the history teacher}'.

\subsubsection{Bulgarian}

As in Russian, the highly lexicalized noun compounds
in Bulgarian are right-headed and are written concatenated (often with a linking vowel), e.g.,
{\it kinosalon} (lit. `{\it cinema saloon}', meaning `{\it cinema hall}')
and {\it grad\underline{o}nachalnik} (lit. `{\it city chief}', meaning `{\it mayor}').
\cite{Boyadzhiev:al:1998,Osenova:Simov:2007}

However, unlike Russian, which is synthetic, Bulgarian is an analytical language;
therefore, head-modifier relations between nouns that are expressible
with cases in Russian (i.e., following the pattern $N_{head}$ + $N_{mod\_cased}$)
would typically be expressed using prepositions
in Bulgarian (i.e., following the pattern $N_{head}$ + preposition + $N_{mod}$),
which severely limits the number of nonlexicalized noun compounds in Bulgarian compared to Russian.

Nonlexicalized noun compounds in Bulgarian are pre-dominantly left-headed
if written as two-words (often connected with a dash), e.g.,
{\it strana-chlenka} (lit. `{\it state-member}', meaning `{\it member-state}')
and {\it ochi-chereshi} (lit. `{\it eyes-cherries}', meaning `{\it eyes that look like cherries}'),
but there are also right-headed ones, e.g.,
{\it maystor gotvach} (lit. `{\it master cook}', meaning `{\it chef}')
and {\it kandidat-student} (lit. `{\it candidate student}',
meaning `{\it person who applies in an university to become a student}').
This distinction can be seen in definite forms,
since the definite article in Bulgarian is a suffix and attaches to the head of the noun compound.
For example, in {\it strana\underline{ta} chlenka} and {\it ochi\underline{te}-chereshi}
it is left-attached,
but in {\it maystor gotvach\underline{yt}} and {\it kandidat-student\underline{yt}} it is right-attached.

Interestingly, borrowed noun compounds keep the attachment from the language they are borrowed from.
For example,
{\it shkembe chorba} (`{\it tripe soup}'), which comes from the Turkish {\it i\c{s}kembe \c{c}orbas{\i}},
and the translations from English
{\it ofis paket} (`{\it office package}'),
{\it biznes oferta} (`{\it business offer}'),
{\it Internet dostavchik} (`{\it Internet provider}'),
{\it onlayn usluga} (`{\it online service}'),
{\it general-mayor} (`general-major') and
{\it ski avtomat} (`{\it ski automaton}')
are all right-attached, which can be seen in the definite forms:
{\it shkembe chorba\underline{ta}},
{\it ofis paket\underline{yt}},
{\it biznes oferta\underline{ta}},
{\it Internet dostavchik\underline{yt}},
{\it onlayn usluga\underline{ta}},
{\it general-mayor\underline{yt}},
and {\it ski avotmat\underline{yt}}.


\subsection{Turkic Languages}

In Turkish, the head in a noun compound is marked with a possessive suffix, e.g.,
{\it g\"{o}bek dans\underline{\i}} (`{\it belly dance}') is formed out of
{\it g\"{o}bek} + {\it dans} + {\it {\i}}-poss.
The possessive suffix is subject to vowel harmony constraints,
and in some cases it can alter the last consonant of the word it attaches to,
e.g., in {\it su barda\underline{\v{g}{\i}}} (`{\it water glass}'),
the original head word is {\it bardak}.

It is important to note that unlike genitive in Romanian or Russian,
the possessive suffix in Turkish marks the head (the possessed),
rather than the modifier (the possessor).
Turkish also has a genitive case, which can be used to mark the possessor
when necessary, just like English uses genitive in compounds like
{\it Peter's book}. A corresponding example in Turkish would be
{\it Ay\c{s}e\underline{'nin} kitab\underline{{\i}}}
(lit. `{\it Ayshe-gen book-poss}', meaning `{\it Ayshe's book}').
As in English, the modifier does not have to be a person's name;
it can be a regular noun as well, e.g.,
{\it manav\underline{{\i}n} merak\underline{{\i}}},
which is formed out of
{\it manav} (`{\it greengrocer'}) + {\it merak} (`{\it curiosity}'),
meaning `{\it greengrocer's curiosity}'.

A noun compound like {\it g\"{o}bek dans\underline{\i}}
can in turn be used as a modifier of another noun, e.g.,
{\it kurs} (`{\it course}'), which yields the left-bracketed double-marked noun compound
{\it g\"{o}bek dans\underline{\i} kurs\underline{u}} (`{\it belly dance course}').
By contrast, a right-bracketed noun compound
like {\it Berkeley mangal parti\underline{si}} (`{\it Berkeley BBQ party}')
has only one possessive suffix. Its derivation first puts together {\it mangal} and {\it parti}
to form {\it mangal parti\underline{si}}, which in turn is modified by {\it Berkeley}.
Note that {\it parti} does not acquire a second suffix in the process
since Turkish does not allow double possessive suffixes on the same word;
in some cases, this can create ambiguity for longer compounds, e.g.,
{\it Berkeley g\"{o}bek dans\underline{\i} kurs\underline{u}}
is structurally ambiguous between bracketings (3) and (4):

\begin{center}
\begin{tabular}{ll@{ }l}
(3) & {\it $[$ $[$ Berkeley $[$ g\"{o}bek dans\underline{\i} $]$ $]$ kurs\underline{u} $]$}\\
(4) & {\it $[$ Berkeley $[$ $[$ g\"{o}bek dans\underline{\i} $]$ kurs\underline{u} $]$ $]$}
\end{tabular}
\end{center}

\section{Noun Compounds: My Definition, Scope Restrictions}
\label{sec:compounding:my:def}

 {\bf Definition.} \emph{Following \namecite{downing:1977:nc:sem},
 I define a \emph{noun compound} as a sequence of nouns which function as a single noun.
 I further require that all nouns in a noun compound be spelled as separate words.
 Occasionally, I will use the term \emph{noun-noun compound} for noun compounds consisting of two nouns.}
\vspace{6pt}

Note that, according to my definition, not all sequences of nouns represent noun compounds;
they have to function as a single noun.
For example, {\it cat food} is not a noun compound in the sentence
``{\it I gave my \underline{cat} \underline{food}.}'', which is perfectly reasonable:
the two nouns are two separate arguments of the verb, a direct and an indirect object,
and therefore they do not function as a single noun.
Of course, {\it cat food} can be a noun compound in other contexts,
e.g., in the sentence ``{\it I would never allow my dog eat \underline{cat food}.}''

The last requirement of my definition is computational rather than linguistic:
I prefer to think of a noun compound as a sequence of words,
each of which represents a noun, and I do not want to look inside the words.
Therefore, {\it silkworm}, {\it textbook} and {\it headache} are single words for me,
rather than noun compounds. However, they are noun compounds under most linguistic theories,
and the fact that they are spelled as single words is rather a convention
(but also a way to suggest a higher degree of lexicalization compared to noun compounds that are spelled separately).
For example, {\it healthcare} and {\it health care} represent the same noun compound
under most linguistic theories; the former probably expresses the writer's belief
in a higher degree of lexicalization. However, under my definition, {\it healthcare}
is not a noun compound at all while {\it health care} is, which creates some inconsistency,
that I need to accept for the sake of simplicity of my experiments.

On the positive side, the requirement for separate spelling effectively eliminates
many of the problematic classes of noun compounds
that other researchers typically exclude explicitly \cite{levi:1978},
e.g., {\it silverfish}, which is a metaphorical name, where the fish is not really {\it silver};
the synecdochic {\it birdbrain}, which actually refers to a quality of a person;
and coordinate structures like {\it speaker-listener}.
All these examples are spelled as single words and therefore do not represent noun compounds.
Genitive constructions like {\it cat's food} are automatically excluded as well:
while arguably being a separate token, the genitive marker is not a noun.
Other commonly excluded noun compounds include names like
{\it Union Square}, {\it Bush Street},  {\it Mexico City}, or {\it Stanford University}.
While not formally excluded by my definition, they are not present in the datasets I use in my experiments.

\section{Linguistic Theories}

\subsection{Levi's Recoverably Deletable Predicates}
\label{nc:ling:theory:levi}


One of the most important theoretical linguistic theories
of the syntax and semantics of noun compounds is that of \namecite{levi:1978}.
The theory targets the more general class of {\it complex nominals},
a concept grouping together\footnote{Levi's theory is limited to {\it endocentric} complex nominals
and excludes {\it exocentric} ones like
metaphorical names (e.g., {\it silverfish}), synecdochic (e.g., {\it birdbrain}),
coordinate structures (e.g., {\it speaker-listener}),
and names (e.g., {\it San Francisco}).} the partially
overlapping classes of {\it nominal compounds}\footnote{Levi's {\it nominal compounds}
roughly correspond to my noun compounds, with some minor differences,
e.g., she allows for word concatenations as in {\it silkworm}.}
(e.g., {\it peanut butter}, {\it mountain temperature}, {\it doghouse}, {\it silkworm}), {\it
nominalizations} (e.g., {\it American attack}, {\it presidential
refusal}, {\it dream analysis}) and {\it nonpredicate
noun phrases}\footnote{Nonpredicate NPs contain a modifying adjective that
cannot be used in predicate position with the same meaning. For
example, the NP {\it a solar generator} cannot be paraphrased as
`{\it *a generator which is solar}'.} (e.g., {\it electric shock}, {\it
electrical engineering}, {\it musical critism}).
The head of a complex nominal is always a noun, but the modifier
is allowed to be either a noun or an adjective.
In Levi's theory, the three subclasses
share important syntactic and semantic properties. For example, the
nominal compound {\it language difficulties}
is synonymous with the nonpredicate NP {\it linguistic difficulties}: 
despite the surface morphological differences,
they share the same semantic structure since
the adjectival modifier of a complex nominal is derived from an underlying noun.

Levi focuses on the syntactic and semantic properties of the complex nominals
and proposes detailed derivations within a theory of generative semantics.
The derivations are based on two basic processes:
{\it predicate deletion} and {\it predicate nominalization}.
Given a two-argument predicate, {\it predicate deletion} gets rid of the predicate,
retaining its arguments only (e.g., `{\it pie made of apples}' $\rightarrow$ {\it apple pie}),
while {\it predicate nominalization} forms a complex nominal whose head is
a nominalization of the underlying predicate and
the modifier is either the subject or the object of
that predicate (e.g., `{\it the President refused general MacArthur's request}' $\rightarrow$ {\it presidential refusal}).

\subsubsection{Predicate Deletion}

\begin{table}
  \centering
\begin{tabular}{llcl}
  \multicolumn{1}{c}{\bf RDP} & \multicolumn{1}{c}{\bf Example} &
  \multicolumn{1}{l}{\bf Subj./Obj.} & \multicolumn{1}{l}{\bf Traditional Name}\\
  \hline
  \texttt{CAUSE$_1$} & {\it tear gas} & object & causative\\
  \texttt{CAUSE$_2$} & {\it drug deaths} & subject & causative\\
  \texttt{HAVE$_1$}  & {\it apple cake} & object & possessive/dative\\
  \texttt{HAVE$_2$}  & {\it lemon peel} & subject & possessive/dative\\
  \texttt{MAKE$_1$}  & {\it silkworm} & object & productive; constitutive, compositional\\
  \texttt{MAKE$_2$}  & {\it snowball} & subject & productive; constitutive, compositional\\
  \texttt{USE}       & {\it steam iron} & object & instrumental\\
  \texttt{BE}        & {\it soldier ant} & object & essive/appositional\\
  \texttt{IN}        & {\it field mouse} & object & locative [spatial or temporal]\\
  \texttt{FOR}       & {\it horse doctor} & object & purposive/benefactive\\
  \texttt{FROM}      & {\it olive oil} & object & source/ablative\\
  \texttt{ABOUT}     & {\it price war} & object & topic\\
  \hline
\end{tabular}
  \caption{{\bf Levi's recoverably deletable predicates (RDPs).}
  The third column indicates whether the modifier was the subject or the object
  of the underlying relative clause.}
  \label{table:RDPs}
\end{table}

According to Levi, there exists a very limited number of
{\it Recoverably Deletable Predicates} ({\it RDPs}) that can be deleted
in the process of transformation of an underlying relative clause
into a complex nominal: five verbs (\texttt{CAUSE}, \texttt{HAVE}, \texttt{MAKE},
\texttt{USE} and \texttt{BE}) and four prepositions (\texttt{IN},
\texttt{FOR}, \texttt{FROM} and \texttt{ABOUT}). See Table
\ref{table:RDPs} for examples and alternative names for each predicate.
While typically the modifier is derived from the object of the underlying relative clause,
the first three verbs also allow for it to be derived from the subject.

Below I give an eight-step derivation of the complex nominal {\it musical clock},
using Levi's original notation. Step
\texttt{b} forms a compound adjective, step \texttt{c} inserts a
copula, step \texttt{d} forms a relative clause, step \texttt{e}
deletes WH-{\it be}, step \texttt{f} performs a predicate preposing,
step \texttt{g} deletes \texttt{MAKE$_1$}, and step
\texttt{h} performs a morphological adjectivalization. This last
step reveals the connection between two subclasses of complex nominals:
nominal compounds and nonpredicate NPs.

\vspace{12pt}
\begin{center}
\parbox{2.75in}{
\begin{small}
\texttt{a. clock \#\# clock make music}

\texttt{b. clock \#\# clock music-making}

\texttt{c. clock \#\# clock be music-making}

\texttt{d. clock \#\# which \#\# be music-making}

\texttt{e. clock music-making}

\texttt{f. music-making clock}

\texttt{g. music clock}

\texttt{h. musical clock}
\end{small}
} 
\end{center}
\vspace{12pt}

Note that the names of Levi's RDPs in Table \ref{table:RDPs}
are capitalized, which is to stress that what is important
is the semantic structure rather than the
presence of the particular predicate. For example, \texttt{IN}
refers to a generalized location which can be spatial or temporal, concrete or
abstract, and the RDP deletion operation can recognize as instances of
\texttt{IN} not only {\it in}, but also preposition like
{\it on}, {\it at}, {\it near}, etc.,
(e.g., {\it terrestrial life} means `{\it life on earth}',
{\it polar climate} is `{\it climate near the pole}',
and {\it night flight} is `{\it flight at night}'),
or any verb expressing that kind of locative relation
(e.g., {\it desert rat} means `{\it rat inhabiting the desert}',
{\it water lilies} means `{\it lilies growing in water}', etc.).
Below I briefly describe each of the predicates:

{\bf \texttt{CAUSE$\mathbf{_1}$}} means `$N_2$ causes $N_1$'. It is
derived by deletion of the present participle {\it causing} (from
`{\it $N_1$-causing $N_2$}'), e.g., {\it tear gas} is obtained from
`{\it tear-causing gas}' (and earlier in the derivation, from
`{\it gas that causes tears}').

{\bf \texttt{CAUSE$\mathbf{_2}$}} means `$N_2$ is caused by $N_1$'.
It is derived by deletion of the past participle {\it caused}
(from `{\it $N_1$-caused $N_2$}'), e.g., {\it drug deaths} is
obtained from `{\it drug-caused deaths}' (and earlier in the
derivation, from `{\it deaths that are caused by drugs}').

{\bf \texttt{HAVE$\mathbf{_1}$}} means `$N_2$ has $N_1$'.
For example, {\it apple cake} can be derived from the paraphrase `{\it cake with apples}'.
Note however that the following intermediate derivation steps are not possible:
*`{\it apple-having cake}' and *`{\it cake that is apple having}'.
The verb can be replaced by a genitive marker,
by prepositions like {\it of} and {\it with}, etc.

{\bf \texttt{HAVE$\mathbf{_2}$}} means `$N_1$ has $N_2$'.
For example, {\it lemon peel} can be derived from the paraphrase
`{\it peel of a lemon}' or `{\it lemon's peel}', but the following intermediate derivation steps
are not possible: *`{\it lemon-had peel}' and *`{\it peel that is
lemon-had}'. The verb can be replaced by a genitive marker,
by prepositions like {\it of} and {\it with}, etc.

{\bf \texttt{FROM}}. In this RDP, the modifier is a source for the head.
Possible paraphrases include `{\it $N_2$ from $N_1$}', `{\it $N_2$ derived
from $N_1$}', etc. The modifier is typically a natural object such
as a vegetable (as in {\it cane sugar}) or an animal (as in {\it
pork suet}), and the head denotes a product or a
by-product obtained by processing the object named by the modifier.
Another big subgroup is exemplified by {\it country visitor}, where
the modifier denotes a previous location; it is ambiguous
with a derivation from past tense + \texttt{IN} deletion. A third
subgroup contains plants or animals separated by their biological
source, like {\it peach pit}; it is ambiguous with a derivation
by means of \texttt{HAVE$_2$} deletion.

{\bf \texttt{MAKE$\mathbf{_1}$}} means `$N_2$ physically produces,
causes to come into existence $N_1$'. It is derived by deletion of
the present participle {\it making} (from `{\it $N_1$-making
$N_2$}'), e.g., {\it silkworm} is obtained from {\it silk-making
worm}. Some forms analyzed as \texttt{MAKE$_1$} deletion
can have an alternative analysis as \texttt{FOR} deletion,
e.g., {\it music box}. These different analyses correspond
to a semantic difference.

{\bf \texttt{MAKE$\mathbf{_2}$}} means
`$N_2$ made up/out of $N_1$'. It is derived
by deletion of the past participle {\it made} from `{\it $N_1$-made $N_2$}'.
There are three subtypes of this RDP: (a) the modifier is a
unit and the head is a configuration, e.g., {\it root system}; (b)
the modifier represents a material and the head represents a mass or
an artefact, e.g., {\it chocolate bar}; (c) the head represents human
collectives and the modifier specifies their membership, e.g., {\it
worker teams}. All members of subtype (b) have an alternative analysis as
\texttt{BE} deletion, e.g., {\it bar chocolate} can be analyzed as
`{\it bar which is made of chocolate}' or as `{\it bar which is chocolate}'.

{\bf \texttt{BE}}. This RDP covers multiple semantic classes,
including: (a) compositional, e.g., {\it snowball}; (b)
genus-species, e.g., {\it pine tree}; (c) metaphorical, e.g., {\it
queen bee}; (d) coordinate, e.g., {\it secretary-treasure}; and (e)
reduplicated, e.g., {\it house house}. The last two
classes are exocentric and thus excluded from Levi's theory.

{\bf \texttt{IN}}.
Possible paraphrases for this RDP include `{\it $N_2$ be located at
$N_1$}', `{\it $N_2$ in $N_1$}', `{\it $N_2$ on $N_1$}', `{\it $N_2$
at $N_1$}', `{\it $N_2$ during $N_1$}', etc.
It refers to a concrete (e.g., {\it desert mouse}), an abstract (e.g., {\it
professional specialization}) or a temporal location (e.g., {\it night flight}).


{\bf \texttt{ABOUT}}. This RDP can be paraphrased as
`{\it $N_2$ about $N_1$}', `{\it $N_2$ concerned with $N_1$}',
`{\it $N_2$ concerning $N_1$}', `{\it $N_2$
dealing with $N_1$}', `{\it $N_2$ pertaining to $N_1$}', `{\it $N_2$
on the subject of $N_1$}', `{\it $N_2$ on $N_1$}', '{\it $N_2$ over
$N_1$}', '{\it $N_2$ on the subject of $N_1$}', '{\it $N_2$ whose
subject is $N_1$}', etc. For example {\it tax law}, {\it sports magazine},
{\it border crisis}. This is a very homogeneous RDP.

{\bf \texttt{USE}}. The verb {\it use} can represent two different
lexical items: one with an instrumental and another one with an
agentive meaning. Only the former can be used in a \texttt{USE}
deletion, e.g., `{\it clock using electricity}' (instrumental)
can give rise to the nominal compound {\it electrical clock},
but `{\it villagers using electricity}' (agentive)
are not *{\it electrical villagers}. If the head
represents an activity then `{\it $N_2$ by means of $N_1$}' is also possible,
in addition to `{\it $N_2$ using $N_1$}', e.g.,
{\it shock treatment} can be paraphrased as both
`{\it treatment using shock}' and `{\it treatment by means of shock}'. If the head represents
an object, e.g., {\it steam iron}, a longer-form paraphrase might be
required, such as `{\it $N_2$ using $N_1$ in order to function}' or
`{\it $N_2$ functioning by means of $N_1$}'.

{\bf \texttt{FOR}}. This RDP expresses purpose.
The preposition {\it for} can be \texttt{FOR} deleted in the case of
`{\it spray for bugs}' ({\it bug spray}, i.e., that kills them), and
`{\it spray for pets}' ({\it pet spray}, i.e., that helps them),
but this is blocked when it means favor, as in `{\it women for peace}' (*{\it peace women}),
or in case of semantically empty object marker, as
in `{\it appeal for money}' (*{\it money appeal}).
Possible paraphrases include `{\it $N_2$ be for $N_1$}',
`{\it $N_2$ be intended for $N_1$}', `{\it $N_2$ for $N_1$}', `{\it $N_2$ be used for $N_1$}',
`{\it $N_2$ be for V-ing $N_1$}', etc. The last
one is exemplified by {\it administrative office}, which means
`{\it office for handling administration}'. The intervening verb {\it V} is
often predictable from the meaning of the head, but not always. In
some rare cases, the paraphrase could be `{\it $N_2$ is good for
$N_1$}', e.g., {\it beach weather}.

\subsubsection{Predicate Nominalization}

The second operation in Levi's theory that produces complex nominals is {\it predicate
nominalization}.
The resulting complex nominals have a nominalized verb as their head,
and a modifier derived from either the subject or the object of the
underlying predicate.

Multi-modifier nominalizations retaining
both the subject and the object as modifiers are possible as well.
Therefore, there are three types of nominalizations depending on the
modifier, which are combined with the following four types of
nominalizations the head can represent: {\it act}, {\it product},
{\it agent} and {\it patient}. See Table \ref{table:NOMs} for examples.


\begin{table}
  \centering
\begin{tabular}{|l|l|l|l|}
  \hline
   & {\bf Subjective} & {\bf Objective} & {\bf Multi-modifier}\\
  \hline
  {\bf Act} & {\it parental refusal} & {\it dream analysis} & {\it
  city land acquisition}\\
  {\bf Product} & {\it clerical errors} & {\it musical critique} & {\it
  student course ratings}\\
  {\bf Agent} & \multicolumn{1}{c|}{---} & {\it city planner} & \multicolumn{1}{c|}{---}\\
  {\bf Patient} & {\it student inventions} & \multicolumn{1}{c|}{---} & \multicolumn{1}{c|}{---}\\
  \hline
\end{tabular}
  \caption{{\bf Levi's nominalization types with examples.}}
  \label{table:NOMs}
\end{table}

\subsubsection{Discussion}

Levi's theory is one of the most sound and detailed theories on the syntax and semantics of noun compounds,
but is not without drawbacks. First, it excludes many types of noun compounds
(but it is normal for a theory to be limited in scope in order to stay focused).
Second, despite the strong linguistic evidence for her decision
to allow for adjectival modifiers in a noun compound, this remains controversial and is not widely accepted.
Third, while being useful from a generative semantics point of view,
her recoverably deletable predicates are quite abstract,
which limits their value from a computational linguistics point of view,
where `{\it lives in}' would arguably be more useful than \texttt{IN}.
Finally, the dataset she built her theory on is quite heterogeneous
and the noun compounds are of various degrees of lexicalization
(compare {\it lemon peel} vs. {\it silkworm});
she also treated as two-word long some noun compounds
which are actually composed of three or even four words, e.g.,
{\it wastebasket category}, {\it hairpin turn}, {\it headache pills},
{\it basketball season}, {\it testtube baby}, {\it beehive hairdo}.

\subsection{Lauer's Prepositional Semantics}

\begin{table}[h]
  \centering
\begin{tabular}{lll}
  \multicolumn{1}{l}{\bf Preposition} & \multicolumn{1}{l}{\bf Example} & \multicolumn{1}{l}{\bf Meaning}\\
  \hline
  \texttt{OF}    & {\it state laws} & laws of the state\\
  \texttt{FOR}   & {\it baby chair} & chair for babies\\
  \texttt{IN}    & {\it morning prayers} & prayers in the morning\\
  \texttt{AT}    & {\it airport food} & food at the airport\\
  \texttt{ON}    & {\it Sunday television} & television on Sunday\\
  \texttt{FROM}  & {\it reactor waste} & waste from a reactor\\
  \texttt{WITH}  & {\it gun men} & men with guns\\
  \texttt{ABOUT} & {\it war story} & story about war\\
  \hline
\end{tabular}
  \caption{{\bf Lauer's prepositions with examples.}}
  \label{table:Lauer:sem:prep}
\end{table}

\namecite{lauer:1995:thesis}
proposes that the semantics of a noun compound can be expressed
by the following eight prepositions:
{\it of}, {\it for}, {\it in}, {\it at}, {\it
on}, {\it from}, {\it with} and {\it about}.
See Table \ref{table:Lauer:sem:prep}.
While being simple, this semantics is
problematic since the same preposition can indicate several different
relations, and conversely, the same relation can be paraphrased by
several different prepositions. For example, {\it in}, {\it on}, and
{\it at}, all can refer to both location and time.


\subsection{Descent of Hierarchy}
\label{sec:descent:hierarchy}

\namecite{Rosario:al:2002:descent} assume a head-modifier relationship between
the nouns in a noun-noun compound and an argument structure for the head,
similar to the argument structure of the verbs and related to the {\it qualia structure}
in the theory of the generative lexicon of \namecite{Pustejovsky:1995}.
Under this interpretation, the meaning of the head determines
what can be done to it, what it is made of, what it is a part of, and so on.
For example, for the word {\it knife}, the possible relations
(with example noun-noun compounds) include the following:

\begin{itemize}
    \item (Used-in): \emph{kitchen knife}, \emph{hunting knife}
    \item (Made-of): \emph{steel knife}, \emph{plastic knife}
    \item (Instrument-for): \emph{carving knife}
    \item (Used-on): \emph{meat knife}, \emph{putty knife}
    \item (Used-by): \emph{chef's knife}, \emph{butcher's knife}
\end{itemize}

Some relations are specific for limited classes of nouns,
while other are more general and apply to larger classes.
Building on this idea, \namecite{Rosario:al:2002:descent}
propose a semi-supervised approach for characterizing the relation
between the nouns in a bioscience noun-noun compound based on the
semantic category in a lexical hierarchy each of the nouns belongs to.
They extract all noun-noun compounds from a very large corpus (one million MEDLINE abstracts),
and they make observations on which pairs of semantic categories the nouns tend to belong to.
Based on these observations, they manually label the relations,
thus avoiding the need to decide on a particular set of relations in advance.

They used the MeSH\footnote{http://www.nlm.nih.gov/mesh} (Medical Subject Heading)
lexical hierarchy, where each concept is assigned a unique identifier
(e.g., \emph{Eye} is D005123) and one or more descriptor codes
corresponding to particular positions in the hierarchy. For
example, A (\emph{Anatomy}), A01 (\emph{Body Regions}), A01.456
(\emph{Head}), A01.456.505 (\emph{Face}), A01.456.505.420
(\emph{Eye}). \emph{Eye} is ambiguous and
has a second code: A09.371 (A09 is \emph{Sense Organs}).

The authors manually selected different levels of generalization,
and then tested them on new data, reporting about 90\% accuracy.
For example, all noun-noun compounds in which the first noun is classified
under the A01 sub-hierarchy ({\it Body Regions}), and the second one falls into A07
({\it Cardiovascular System}), are hypothesized to express the same
relation, e.g., \emph{limb vein}, \emph{scalp arteries}, \emph{shoulder artery},
\emph{forearm arteries}, \emph{finger capillary},
\emph{heel capillary},
\emph{leg veins}, \emph{eyelid capillary}, \emph{ankle artery},
\emph{hand vein}, \emph{forearm microcirculation}, \emph{forearm veins},
\emph{limb arteries}, \emph{thigh vein}, \emph{foot vein}, etc.

The authors empirically found that the majority
of the observed noun-noun compounds fall within a limited number
of semantic category pairs corresponding to the top levels in the lexical hierarchy,
e.g., A01-A07; most of the remaining ones require descending one or two levels
down the hierarchy for at least one of the nouns
in order to arrive at the appropriate level of generalization
of the relation. 
For example, the relation is not homogeneous
when the modifier falls under A01 ({\it Body Regions})
and the head is under M01 ({\it Persons}), e.g.,
\emph{abdomen patients}, \emph{arm amputees}, \emph{chest physicians},
\emph{eye patients}, \emph{skin donor};
it depends on whether the person is a patient,
a physician, or a donor. Making this distinction requires descending
one level down the hierarchy for the head:
\vspace{12pt}

A01-M01.643 ({\it Patients}):
\emph{abdomen patients}, \emph{ankle inpatient}, \emph{eye outpatient}

A01-M01.526 ({\it Occupational Groups}):
\emph{chest physician}, \emph{eye nurse}, \emph{eye physician}

A01-M01.898 ({\it Donors}): \emph{eye donor}, \emph{skin donor}

A01-M01.150 ({\it Disabled Persons}): \emph{arm amputees}, \emph{knee amputees}
\vspace{6pt}

The idea of the descent of hierarchy is appealing
and the demonstrated accuracy is very high, but it is not without limitations.
First, the classification is not automated; it is performed manually.
Second, the coverage is limited by the lexical hierarchy, most likely to specific domains.
Third, there are problems caused by lexical and relational ambiguities.
Finally, the approach does not propose explicit names for the assigned relations.

The descent of hierarchy is similar in spirit to the work of \namecite{Li:Abe:1998:case:frames},
who apply the {\it Minimum Description Length} ({\it MDL})
principle to the task of acquiring case frame patterns for verbs.
The principle is a theoretically sound approach to model selection,
introduced by \namecite{Rissanen:1978:shortest:descr}. Its main idea is that
learning represents a form of data compression: the better a model
captures the principles underlying a given data sample, the more
efficiently it can describe it. While a more complex model has
a better chance of fitting the data well, being too specific
it could also overfit it and therefore miss some important generalizations.
The MDL principle states that the optimal balance between the
complexity of the model and the fit to the data is achieved by
minimizing the sum of the number of bits needed to encode the model
({\it model length}) and the number of bits needed to encode the
data under that model ({\it data description length}). See
\cite{Quinlan:Rivest:1989:MDL} for details.
\namecite{Li:Abe:1998:case:frames} apply the MDL principle to
acquiring case frame patterns for verbs.
Given a verb, e.g., {\it fly} and a set of example arguments with corresponding
corpus frequencies, e.g., {\it crow}:2, {\it eagle}:2, {\it bird}:4, {\it bee}:2
for one of its slots, e.g., the direct object, they try to find the best level of
generalization over the possible values of that slot in terms of {\it WordNet} categories.
For example, in the case of the verb {\it fly}, suitable categories include
\texttt{BIRD} and \texttt{INSECT}, but not the more general \texttt{ANIMAL}.
\namecite{Abe:Li:1996:word:assoc} and \namecite{Li:Abe:1999:frame:slots}
learn dependencies between case frame slots in a similar manner.
\namecite{Wagner:2005:learning:thematic:roles} extends their work and
applies it to the task of learning thematic role relations at the
appropriate level of generalization, e.g., \texttt{FOOD} is a suitable {\it patient}
for the verb {\it eat}, while \texttt{CAKE} and \texttt{PHYSICAL\_OBJECT} are not.

\chapter{Parsing Noun Compounds}
\label{chapter:NC:bracketing}

In this chapter,
    I describe a novel, highly accurate lightly supervised method for
    making left vs. right bracketing decisions for three-word noun compounds.
    For example, {\it $[[$tumor suppressor$]$ protein$]$} is left-bracketed,
    while {\it $[$world $[$food production$]]$} is right-bracketed.
    Traditionally, the problem has been addressed using unigram and bigram frequency
    estimates used to compute adjacency- and dependency-based word association scores.
    I extend this approach by introducing novel surface features and paraphrases
    extracted from the Web. Using combinations of these features,
    I demonstrate state-of-the-art results on two separate collections,
    one consisting of terms drawn from encyclopedia text,
    and another one extracted from bioscience text.

    These experiments were reported in abbreviated form
    in \cite{nakov:hearst:2005:CoNLL} and \cite{nakov:2005:lql:biolink}.

\section{Introduction}
\label{NCSyntax:intro}

The semantic interpretation of noun compounds of length three or more
requires that their syntactic structure be determined first.
Consider for example the following contrastive pair of noun compounds:

\begin{center}
\begin{tabular}{ll}
(1) & {\it liver cell antibody} \\
(2) & {\it liver cell line}
\end{tabular}
\end{center}

\begin{figure}[htb]
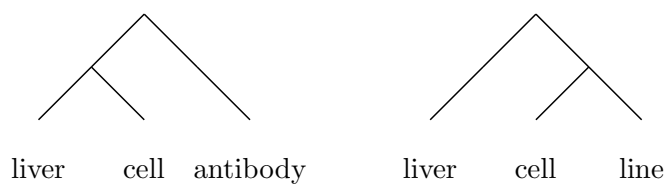

    \vspace{24pt}
    \centerline{
    \mbox{\jftree[unit=20pt]
    \!\0{\omit}\medleft\2\medright\4\cr
    \!\2{\omit}\medleft{liver}\medright{cell}\cr \!\4{\omit}\medright\6\cr
    \!\6{antibody}\cr
    \endjftree}
    \mbox{  }\mbox{  }\mbox{  }\mbox{  }\mbox{  }\mbox{  }\mbox{  }\mbox{  }
    \mbox{\jftree[unit=20pt] \!\0{\omit}\medleft\4\medright\2\cr
    \!\2{\omit}\medleft{cell}\medright{line}\cr \!\4{\omit}\medleft\6\cr
    \!\6{liver}\cr
    \endjftree}}
  \caption{{\bf Left vs. right bracketing:} constituency trees.}\label{fig:left:right:constituency}
\end{figure}

\noindent In example (1), there is an {\it antibody} that targets
a {\it liver cell}, while example (2) refers to a {\it cell line} which is derived from
the {\it liver}.  In order to make these semantic distinctions
accurately, it can be useful to begin with the correct grouping of
terms, since choosing a particular syntactic structure limits the options
left for semantics.  Although equivalent at the part of speech (POS)
level, the above two noun compounds have different constituency trees,
as Figure \ref{fig:left:right:constituency} shows.

The above trees can be represented using brackets,
which gives the name of the problem, {\it noun compound bracketing}:

\vspace{6pt}
\begin{center}
\begin{tabular}{ll@{ }l@{ }@{ }l}
(1b) & $[$ $[$ {\it liver cell} $]$ & {\it antibody} $]$ & (left bracketing)\\
(2b) & $[$ {\it liver} & $[${\it cell line}$]$  $]$  & (right bracketing)
\end{tabular}
\end{center}
\vspace{6pt}

Longer noun compounds like {\it colon cancer tumor suppressor protein}
are rarely dealt with, since it is believed that
parsing them can be reduced to similar left/right-bracketing decisions
for triples of nouns.
For example, suppose we have decided that {\it $[$ colon \underline{cancer} $]$}
and {\it $[$ tumor \underline{suppressor} $]$} are noun compounds and are used as subunits
in the bracketing: {\it $[$ colon \underline{cancer} $]$ $[$ tumor \underline{suppressor} $]$ \underline{protein}}.
Assuming a noun compound behaves like its head,
we end up with a bracketing problem for the compound {\it cancer suppressor protein}.
If we decide on a right bracketing for that compound, we end up with the following overall structure:
{\it $[$ $[$ colon cancer $]$ $[$ $[$ tumor suppressor $]$ protein $]$ $]$}.

Parsing the noun compound is a necessary step towards semantic interpretation 
since the syntactic structure reveals the sub-parts between which relations need to be assigned, e.g.,
for the above example, we have the following semantic representation:

\begin{center}
\framebox{
\parbox{5.4in}{
$[$\emph{tumor suppressor protein}$]$ which is \underline{implicated in} $[$\emph{colon cancer}$]$ (\texttt{IN}; \texttt{LOCATION})

$[$\emph{protein}$]$ that \underline{acts as} $[$\emph{tumor suppressor}$]$ (\texttt{IS}; \texttt{AGENT})

$[$\emph{suppressor}$]$ that \underline{inhibits} $[$\emph{tumor(s)}$]$ (\texttt{OF}; \texttt{PURPOSE})

$[$\emph{cancer}$]$ that \underline{occurs in} $[$\emph{(the) colon}$]$ (\texttt{OF}; \texttt{IN}; \texttt{LOCATION})

}}
\end{center}

Below I describe a method for parsing three-word noun compounds
in written text.
In spoken text, in most situations there will be differences
in stress or pauses that would help speakers to pick the correct interpretation
(as I already mentioned in section \ref{sec:compounding:process}).
I further limit the present discussion to English;
other languages, like Russian and Turkish, would use explicit case markers
to help noun compound parsing (see section \ref{sec:nc:other:lang} for details).
Finally, while I focus on noun compounds,
similar ambiguities occur for adjectival modifiers,
e.g., {\it \underline{graduate} student parent},
{\it \underline{real} world application}, {\it \underline{ancient} \underline{Greek} corpus},
{\it \underline{American} Heart Association}.

\section{Related Work}
\label{chapter:related:work}

\label{sec:related:NC:syntax}

\subsection{Unsupervised Models}

The best known early work on automated unsupervised noun-compound bracketing
is that of \namecite{lauer:1995:thesis},
who introduces the probabilistic {\it dependency model}
and argues against the {\it adjacency model}\footnote{See
sections \ref{sec:adj}, \ref{sec:dep} and \ref{sec:adj:vs:dep}
for more details and a comparison between the two models.},
used by \namecite{marcus80}, \namecite{pustejovsky93} and
\namecite{Resnik:1993:thesis}. Given a three-word noun compound $w_1w_2w_3$,
the adjacency model compares the strengths of association $Assoc(w_1,w_2)$
and $Assoc(\underline{w_2},w_3)$, while the dependency model
compares $Assoc(w_1,w_2)$ and $Assoc(\underline{w_1},w_3)$.

Lauer collects $n$-gram statistics from
{\it Grolier's encyclopedia}\footnote{\texttt{http://go.grolier.com}},
which contain about eight million words (in 1995).
To overcome data sparsity issues, he estimates probabilities over
conceptual categories $t_i$ in {\it Roget's thesaurus}\footnote{\texttt{http://thesaurus.reference.com/}}
rather than for individual words $w_i$.
Lauer defines the {\it adjacency} and the {\it dependency} models,
respectively, as the following ratios:
\begin{eqnarray}
 R_{adj} &=& \frac{\sum_{t_i \in cats(w_i)} \mathrm{Pr}(t_1
 \rightarrow t_2|t_2)}{\sum_{t_i \in cats(w_i)} \mathrm{Pr}(t_2
 \rightarrow t_3|t_3)}\\
 \nonumber\\
 R_{dep} &=& \frac{\sum_{t_i \in cats(w_i)} \mathrm{Pr}(t_1
 \rightarrow t_2|t_2) \mathrm{Pr}(t_2 \rightarrow t_3|t_3)}{\sum_{t_i
 \in cats(w_i)} \mathrm{Pr}(t_1 \rightarrow t_3|t_3) \mathrm{Pr}(t_2
 \rightarrow t_3|t_3)}
\end{eqnarray}

\noindent where $cats(w_i)$ denotes the set of conceptual categories
from {\it Grolier's encyclopedia} the word $w_i$ can belong to,
and $\mathrm{Pr}(t_i \rightarrow t_j|t_j)$ is
the probability that the term $t_i$ modifies a given term $t_j$, and is
estimated as follows:
\begin{equation}
    \mathrm{Pr}(t_i \rightarrow t_j|t_j) = \frac{\#(t_i,t_j)}{\#(t_j)}
\end{equation}
\noindent where $\#(t_i,t_j)$ and $\#(t_j)$ are the corresponding bigram
and unigram frequencies calculated for the text of {\it Grolier's encyclopedia}.

\begin{table}
\begin{center}
\begin{small}
\begin{tabular}{lcc}
    \multicolumn{1}{l}{\bf Model} & \multicolumn{1}{c}{\bf AltaVista} & \multicolumn{1}{c}{\bf BNC}\\
  \hline
    Baseline & 63.93 & 63.93\\
    $\#(n_1,n_2)$ : $\#(n_2,n_3)$ & 77.86 & 66.39\\
    $\#(n_1,n_2)$ : $\#(n_1,n_3)$ & {\bf 78.68} & 65.57\\
    $\#(n_1,n_2)/\#(n_1)$ : $\#(n_2,n_3)/\#(n_2)$ & 68.85 & 65.57\\
    $\#(n_1,n_2)/\#(n_2)$ : $\#(n_2,n_3)/\#(n_3)$ & 70.49 & 63.11\\
    $\#(n_1,n_2)/\#(n_2)$ : $\#(n_1,n_3)/\#(n_3)$ & 80.32 & 66.39\\
    $\#(n_1,n_2)$ : $\#(n_2,n_3)$ (NEAR) & 68.03 & 63.11\\
    $\#(n_1,n_2)$ : $\#(n_1,n_3)$ (NEAR) & 71.31 & 67.21\\
    $\#(n_1,n_2)/\#(n_1)$ : $\#(n_2,n_3)/\#(n_2)$ (NEAR) & 61.47 & 62.29\\
    $\#(n_1,n_2)/\#(n_2)$ : $\#(n_2,n_3)/\#(n_3)$ (NEAR) & 65.57 & 57.37\\
    $\#(n_1,n_2)/\#(n_2)$ : $\#(n_1,n_3)/\#(n_3)$ (NEAR) & 75.40 & {\bf 68.03}\\
  \hline
\end{tabular}
\end{small}
\caption{{\bf Noun compound bracketing experiments of Lapata \& Keller (2004):}
Accuracy in \%s for {\it AltaVista} vs. {\it BNC} counts. Here $\#(n_i)$ and $\#(n_i,n_j)$
are estimates for word and bigram frequencies, respectively,
and (NEAR) means that the words co-occur in a ten-word window.
Shown in bold are the best performing models on the development set.}
\label{table:lauer:keller:lapata}
\end{center}
\end{table}

These models predict a left bracketing if the ratio ($R_{adj}$ or $R_{dep}$)
is greater than one, and right bracketing if it is less than one.
Lauer evaluates them on a testing dataset of 244 unambiguous noun compounds
derived from {\it Grolier's encyclopedia}
(see section \ref{dataset:lauer} for a detailed description
of the dataset and Table \ref{table:dataset:lauer} for some examples).
Below I will refer to that dataset as {\it Lauer's dataset}.
The baseline of always predicting a left bracketing yields 66.8\% accuracy,
and the adjacency model
is not statistically significantly different, obtaining only 68.9\% accuracy.
The dependency model achieves significantly better results, at 77.5\% accuracy.
Adding part-of-speech information in a more sophisticated {\it tuned} model
allowed Lauer to improve the results to 80.70\%.

More recently, \namecite{Lapata:Keller:04} and \namecite{Lapata:Keller:05:Web:based:Models}
improve on Lauer's dependency results
utilizing very simple word and bigram counts estimated using exact phrase queries
against {\it AltaVista} or the \texttt{NEAR} operator\footnote{In {\it AltaVista},
a query for \underline{\texttt{x NEAR y}} forces \texttt{x} to occur
within ten word before or after \texttt{y}.},
with all possible word inflections.
They try several different models, and therefore reserve
half of {\it Lauer's dataset} for model selection (tuning set),
and test on the remaining 122 examples (testing set),
which changes the accuracy of their left-bracketing baseline to 63.93\%.
Table \ref{table:lauer:keller:lapata} shows the models they experiment
with and the corresponding results.
The model that performs best on the tuning set is inspired from
Lauer's dependency model and compares the bigram frequencies
for $\#(n_1,n_2)$ and for $\#(n_1,n_3)$.
On the test set, this model achieves 78.68\% accuracy,
which is a small improvement compared to the 77.50\% for the dependency model
in Lauer's experiments, but is worse compared to Lauer's {\it tuned} model.
For comparison purposes, \namecite{Lapata:Keller:04} also try using frequency estimates
from the {\it British National Corpus} ({\it BNC}), which represents 100M words
(compared to 8 million in Lauer's experiments) of carefully edited, balanced English text.
As Table \ref{table:lauer:keller:lapata} shows, the results when using {\it BNC} are much worse
compared to using {\it AltaVista}: the best model that uses {\it BNC} only achieves 68.03\% accuracy,
which is statistically worse compared to the best model using {\it AltaVista} (78.68\%).
This result confirms the observation of \namecite{Banko:Brill:2001:verylarge}
that using orders of magnitude more data can lead to significant improvements:
the Web is orders of magnitude bigger than {\it BNC}.

\subsection{Supervised Models}

\namecite{girju:2005:on:the:semantics} propose a {\it supervised} model
for noun compound bracketing, based on 15 semantic features:
five distinct semantic features calculated for each of the three nouns.
The features require the correct {\it WordNet} sense for each noun to be provided:

\vspace{6pt}
\begin{enumerate}
  \item {\bf \emph{WordNet} derivationally related form}.
      Specifies if that sense of the noun is related to a verb in {\it WordNet}.
      For example, in ``{\it coffee maker industry}'',
      the correct sense of the second noun is {\it maker\#3}, which is related to the verb to {\it make\#6}.
  \item {\bf \emph{WordNet} top semantic class of the noun}.
      For example, in ``{\it coffee maker industry}'', {\it maker\#3} is a {\it \{group, grouping\}\#1}.
  \item {\bf \emph{WordNet} second top semantic class of the noun}. For example, in ``{\it coffee maker industry}'',
      {\it maker\#3} is a {\it social\_group\#1}.
  \item {\bf \emph{WordNet} third top semantic class of the noun.} For example, in``{\it coffee maker industry}'',
      {\it maker\#3} is {\it organizational\#1}.
  \item {\bf Nominalization.} Indicates if the noun is a nominalization\footnote{A noun
        is considered a nominalization if it is listed in the
        {\it NomLex}\footnote{\texttt{http://nlp.cs.nyu.edu/nomlex/index.html}}
        dictionary of nominalizations \cite{Macleod:al:1998:Nomlex},
        or if it is an {\it event} or an {\it action} in {\it WordNet}.}.
      For example, in ``{\it coffee maker industry}'', {\it maker} is a nominalization.
\end{enumerate}
\vspace{6pt}

Since their model is supervised, \namecite{girju:2005:on:the:semantics} need training data.
Therefore they assemble 49,208 sentences from {\it Wall Street Journal} articles from
the Question Answering track of
TREC-9\footnote{TExt Retrieval Conference, (\texttt{http://www.trec.nist.gov/}).} in 2000.
Then they use \quotecite{lauer:1995:thesis} heuristic to extract candidate
three-word noun compounds from that text, looking for sequences of three nouns not preceded
and not followed by other nouns. The extracted candidates are checked
for errors and manually bracketed in context by two Ph.D. students in Computational Semantics.
This procedure yields 362 examples after agreement between the annotators has been reached.

\namecite{girju:2005:on:the:semantics} use these 362 examples as training data,
and the original {\it Lauer's dataset} of 244 examples as test data.
Due to major differences between the training and the test datasets,
they only achieve 73.10\% accuracy, which is worse than the best unsupervised results
of \namecite{lauer:1995:thesis} and \namecite{Lapata:Keller:04}. Therefore, \namecite{girju:2005:on:the:semantics}
mix the training and the testing set and then randomly create new {\it shuffled}
training and test sets of the same sizes as before: 362 training and 244 testing examples.
Using these {\it shuffled} datasets, they achieve 83.10\% accuracy
with the C5.0 decision tree classifier \cite{Quinlan:1993},
which represents an improvement over \namecite{lauer:1995:thesis} and \namecite{Lapata:Keller:04}.

For comparison purposes, \namecite{girju:2005:on:the:semantics} repeat the Web-based
experiments of \namecite{Lapata:Keller:04} on the {\it shuffled} test dataset (using {\it Google}),
which yields 77.36\% accuracy for the {\it dependency} model,
and 73.45\% for the {\it adjacency} model.
Table \ref{table:lauer:comparisons} allows for an easy interpretation
of these supervised experiments in comparison with the unsupervised
experiments of \namecite{lauer:1995:thesis} and \namecite{Lapata:Keller:04}.

\begin{table}
\begin{center}
\begin{small}
\begin{tabular}{l@{ }lllr}
\multicolumn{1}{l}{\bf Publication} & \multicolumn{1}{l}{\bf Training} & \multicolumn{1}{l}{\bf Testing} & \multicolumn{1}{l}{\bf Model} & \multicolumn{1}{c}{\bf Acc.}\\
  \hline
    \cite{lauer:1995:thesis} & & Lauer: 244 & baseline ({\it left}) & 66.80\\
    & & & adjacency & 68.90\\
    & & & dependency & {\bf 77.50}\\
    & & & tuned & {\bf 80.70}\\
  \hline
    \cite{Lapata:Keller:04} & & Lauer: 122 & baseline ({\it left}) & 63.93\\
    & & & best {\it BNC} & 68.03\\
    & & & best {\it AltaVista} & {\bf 78.68}\\
  \hline
    \cite{girju:2005:on:the:semantics} & additional: 362 & Lauer: 244 & baseline ({\it left}) & 66.80\\
    & & & C5.0   & $^\star${\bf 73.10}\\
    & & & C5.0 (no WSD) & $^\star$72.80\\
  \hline
    \cite{girju:2005:on:the:semantics} & shuffled: 362 & shuffled: 244 & baseline ({\it left}) & 66.80\\
    & & & C5.0   & $^\star${\bf 83.10}\\
    & & & C5.0 (no WSD) & $^\star$74.40\\
    & & & Google adjacency & 73.45\\
    & & & Google dependency & 77.36\\
  \hline
\end{tabular}
\end{small} \caption{{\bf Noun compound bracketing experiments on the \emph{Lauer's dataset}:}
accuracy in \%s.
Lauer (1995) uses no training nor tuning data, and tests on all 244 examples.
Lapata \& Keller (2004) use 122 of the examples for tuning and the remaining 122 ones for testing.
Girju {\it et al.} (2005) present supervised models which are trained on 362 additional examples;
the shuffled datasets are mixes of \emph{Lauer's dataset} and these additional examples.
The results for the supervised models are marked with an asterisk.}
\label{table:lauer:comparisons}
\end{center}
\end{table}

Since having the correct {\it WordNet} sense might be unrealistic,
\namecite{girju:2005:on:the:semantics} also try using the first {\it WordNet} sense instead of
the manually assigned one. As Table \ref{table:lauer:comparisons} shows,
this causes a significant drop in accuracy on the {\it shuffled} dataset -- from 83.10\% to 74.40\% --
which suggests the algorithm may be sensitive to whether the provided {\it WordNet}
sense is correct or not.

\subsection{Web as a Corpus}

In 2001, \namecite{Banko:Brill:2001:verylarge} advocated the creative
use of very large text collections as an alternative to sophisticated
algorithms and hand-built resources.  They demonstrated the idea
on a lexical disambiguation problem for which labeled examples
were available ``for free''.  The problem was to choose which of
two to three commonly confused words,
e.g., {\it \{principle, principal\}},
are appropriate for a given context. The labeled
data was ``free'' since the authors could safely assume that,
in the carefully edited text in their training set,
the words are usually used correctly. They demonstrated that even using a very simple
algorithm, the results continued to improve log-linearly with
more training data, even out to a billion words.
They also found that even the worst algorithm, with little training data,
performed well when given orders of magnitude more data.
Therefore, they concluded that obtaining more training data
may be more effective overall than devising more sophisticated algorithms.

The question then arises about how and whether to apply this idea more
generally for a wide range of natural language processing tasks.
Today, the obvious answer is to use the Web.

Using the Web as a training and testing corpus is attracting ever-increasing attention.
In 2003, the journal {\it Computational Linguistics} had a special issue
on the Web as a corpus \cite{Kilgariff:Grefenstette:2003:CL:web:corpus}.
In 2005, the Corpus Linguistics conference hosted the first workshop
on the {\it Web as Corpus} ({\it WAC}). In 2006, there was a 
WAC workshop in conjunction with the 11$^{\mathrm{th}}$ Conference of the European Chapter of
the Association for Computational Linguistics, and in 2007 there was a 
WAC workshop, incorporating CleanEval. 

The Web has been used as a corpus for a variety of NLP tasks including, but not limited to:
machine translation
\cite{Grefenstette:1999:web,Resnik:1999:mining,Nagata:al:2001:web,Cao:Li:2002:baseNP,Way:Gough:2003:webmt},
anaphora resolution \cite{Modjeska:all:2003:anaphora},
prepositional phrase attachment
\cite{Olteanu:Moldovan:2005:PPattach,Volk:2001:www:PPattach,Calvo:Gelbukh:2003:PPattach},
question answering \cite{Soricut:Brill:2006,Dumais:all:2002:webQA},
extraction of semantic relations
\cite{Chklovski:Pantel:2004:verbocean,Shinzato:Torisawa:2004:web,Szpektor:all:2004:entailment},
language modeling \cite{Zhu:Rosenfeld:2001:LM,Keller:Lapata:03:web:unseen:bigrams},
word sense disambiguation \cite{Mihalcea:Moldovan:1999:wsd,Santamaria:2003:wsd,Zahariev:2004:acronyms}.

Despite the variability of applications, the most popular use of
the Web as a corpus is as a source of page hit counts, which are used
as an estimate for $n$-gram word frequencies.
\namecite{Keller:Lapata:03:web:unseen:bigrams} have demonstrated a high correlation between
page hits and corpus bigram frequencies as well as between page
hits and plausibility judgments.

In a related strand of work, \namecite{Lapata:Keller:05:Web:based:Models}
show that computing $n$-gram statistics over very large corpora yields
results that are competitive with if not better than the best
supervised and knowledge-based approaches on a wide range of NLP
tasks.
For example, they showed that for the problems of
machine translation candidate selection,
noun compound interpretation and
article generation,
the performance of an $n$-gram based model
computed using search engine statistics was significantly better
than the best supervised algorithm.
For other problems, like noun compound bracketing and adjective ordering,
they achieve results that are not significantly different from
the best supervised algorithms,
while for spelling correction,
countability detection and
prepositional phrase attachment
they could not match the performance of the supervised state-of-the-art models.
Therefore, they concluded that
Web-based $n$-gram statistics should be used as the baseline to beat.

I feel the potential of these ideas is not yet fully realized,
and I am interested in finding ways to further exploit the
availability of enormous Web corpora as implicit training data.
This is especially important for structural ambiguity problems
in which the decisions must be made on the basis of the behavior
of individual lexical items. The problem is to figure out how to
use information that is latent in the Web as a corpus, and Web
search engines as query interfaces to that corpus.

\section{Adjacency and Dependency}

\subsection{Adjacency Model}
\label{sec:adj}

According to the bracketing representation introduced above in section \ref{NCSyntax:intro},
given a three-word noun compound $w_1w_2w_3$, the task is to decide whether $w_2$
is more closely associated with $w_1$ or with $w_3$.
Therefore, we need to compare the strength of association between the first two and
the last two words, which is the {\bf adjacency model} \cite{lauer:1995:thesis}:

\begin{itemize}
  \item if $Assoc(w_1,w_2) < Assoc(w_2,w_3)$, predict {\it right bracketing};
  \item if $Assoc(w_1,w_2) = Assoc(w_2,w_3)$, make no prediction;
  \item if $Assoc(w_1,w_2) > Assoc(w_2,w_3)$, predict {\it left bracketing}.
\end{itemize}

\subsection{Dependency Model}
\label{sec:dep}

\begin{figure}[htb]
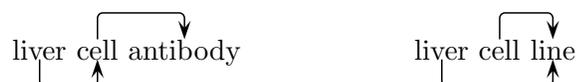

    \vspace{24pt}
    \centerline{
    \rnode{M1}{liver}
    \rnode{M2}{cell}
    \rnode{M3}{antibody}
    \mbox{ }\mbox{ }\mbox{ }\mbox{ }\mbox{ }\mbox{ }\mbox{ }\mbox{ }\mbox{ }\mbox{ }\mbox{ }\mbox{ }\mbox{ }\mbox{ }\mbox{ }\mbox{ }
    \rnode{N1}{liver}
    \rnode{N2}{cell}
    \rnode{N3}{line}
    \psset{linearc=2pt}
    \ncbar[angle=-90]{->}{N1}{N3}
    \ncbar[angle=90]{->}{N2}{N3}
    \ncbar[angle=-90]{->}{M1}{M2}
    \ncbar[angle=90]{->}{M2}{M3}
    }
  \caption{{\bf Left vs. right bracketing:} dependency structures.}\label{fig:left:right:dependency}
\end{figure}

\namecite{lauer:1995:thesis} proposes an alternative {\it dependency model},
where there is an arc pointing from the modifier to the head it depends on,
as shown in Figure \ref{fig:left:right:dependency}.

In this representation,
both the left and the right dependency structures contain a link $w_2 \rightarrow w_3$,
but differ because of $w_1 \rightarrow w_2$ and $w_1 \rightarrow w_3$, respectively.\footnote{In my examples, the arcs always point to the right, i.e. the head always follows the modifier. While this is
the typical case for English, it is not always the case,
e.g., in {\it hepatitis b} the head is {\it hepatitis}.
Since such cases are rare, below we will make
the simplifying assumption that the modifier always precedes the head.}
Therefore, the {\bf dependency model} focuses not on $w_1$, rather than on $w_2$:

\begin{itemize}
  \item if $Assoc(w_1,w_2) < Assoc(\underline{w_1},w_3)$, predict {\it right bracketing};
  \item if $Assoc(w_1,w_2) = Assoc(\underline{w_1},w_3)$, make no prediction;
  \item if $Assoc(w_1,w_2) > Assoc(\underline{w_1},w_3)$, predict {\it left bracketing}.
\end{itemize}

\subsection{Adjacency vs. Dependency}
\label{sec:adj:vs:dep}

\begin{figure}[htb]
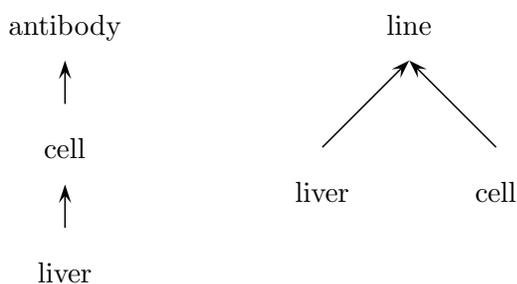

    \centerline{
    \mbox{\jtree[xunit=3em,yunit=3em]
    \! = {antibody}
     <vert>[scaleby=1 0.5,arrows=<-]{cell}
     <vert>[scaleby=1 0.5,arrows=<-]{liver}.
    \endjtree}
    \mbox{ }\mbox{ }\mbox{ }\mbox{ }\mbox{ }\mbox{ }\mbox{ }\mbox{ }\mbox{ }\mbox{ }\mbox{ }\mbox{ }\mbox{ }\mbox{ }\mbox{ }\mbox{ }
    \mbox{\jtree[xunit=3em,yunit=3em]
    \! = {line}
    <left>[arrows=<-]{liver} ^<right>[arrows=<-]{cell}.
    \endjtree}
    }
  \caption{{\bf Left vs. right bracketing:} dependency trees.}\label{fig:left:right:dependency:trees}
\end{figure}

Figure \ref{fig:left:right:dependency:trees}
shows the {\it dependency trees} corresponding to the dependency structures
from Figure \ref{fig:left:right:dependency}.
Note the structural difference from the constituency trees
in Figure \ref{fig:left:right:constituency}.
First, the constituency trees contain words in the leaves only,
while the dependency trees have words in the internal nodes as well.
Second, the constituency trees are {\it ordered binary trees}: each internal node
has exactly two {\it ordered} descendants, one left and one right,
while there is no such ordering for the dependency trees.


Consider examples (3) and (4): both are right-bracketed,
but the order of the first two words is switched.

\begin{center}
\begin{tabular}{ll@{ }l@{ }@{ }l}
(3) & $[$ {\it adult} & $[${\it male rat}$]$  $]$  & (right bracketing)\\
(4) & $[$ {\it male} & $[${\it adult rat}$]$  $]$  & (right bracketing)
\end{tabular}
\end{center}

\begin{figure}[htb]
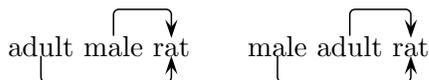

    \vspace{24pt}
    \centerline{
    \rnode{A1}{adult}
    \rnode{A2}{male}
    \rnode{A3}{rat}
    \mbox{ }\mbox{ }\mbox{ }\mbox{ }
    \rnode{B1}{male}
    \rnode{B2}{adult}
    \rnode{B3}{rat}
    \psset{linearc=2pt}
    \ncbar[angle=-90]{->}{A1}{A3}
    \ncbar[angle=90]{->}{A2}{A3}
    \ncbar[angle=-90]{->}{B1}{B3}
    \ncbar[angle=90]{->}{B2}{B3}
    }
  \caption{Dependency structures for {\it adult male rat} and for {\it male adult rat}.}\label{fig:left:right:dependency:adultmalerat:struc}
\end{figure}

\begin{figure}[htb]
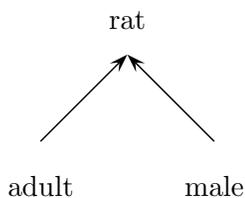

    \centerline{
    \mbox{\jtree[xunit=3em,yunit=3em]
    \! = {rat}
    <left>[arrows=<-]{adult} ^<right>[arrows=<-]{male}.
    \endjtree}
    }
  \caption{Shared dependency tree for {\it adult male rat} and for {\it male adult rat}.}\label{fig:left:right:dependency:adultmalerat}
\end{figure}

Despite (3) and (4) being different,
the corresponding dependency structures are equivalent as
Figures \ref{fig:left:right:dependency:adultmalerat:struc}
and \ref{fig:left:right:dependency:adultmalerat} show:
there are no dependency arcs between {\it adult} and {\it male},
and therefore changing their linear order does not alter
the dependency structure.

The adjacency and the dependency models target different kinds
of right-bracketed structures.
Consider for example {\it home health care}: {\it health care} is a compound,
which in turn is modified by {\it home} \underline{as a whole},
which can easily be seen in the alternative spelling
{\it home healthcare}, where the last two words are concatenated.
This is different from {\it adult male rat},
where the two modifiers {\it adult} and {\it male} can be switched freely,
which suggests that {\it adult} and {\it male} \underline{independently} modify {\it rat}.
We can conclude that, given a three-word noun compound $w_1w_2w_3$,
there are two reasons it may take on right bracketing, $[w_1[w_2w_3]]$:

\begin{enumerate}
  \renewcommand{\labelenumi}{(\alph{enumi})}
  \item $w_2w_3$ is a compound, which is modified by $w_1$;
  \item $w_1$ and $w_2$ independently modify $w_3$.
\end{enumerate}

Let us now look closely at the {\it adjacency}
and the {\it dependency} models:

\begin{itemize}
  \item {\bf adjacency model:} compare $Assoc(w_1,w_2)$ and $Assoc(\underline{w_2},w_3)$;
  \item {\bf dependency model:} compare $Assoc(w_1,w_2)$ and $Assoc(\underline{w_1},w_3)$.
\end{itemize}

Therefore the adjacency model checks (a) -- whether $w_2w_3$ is a compound,
while the dependency model checks (b) -- whether $w_1$ modifies $w_3$.

Note that there is only a modificational choice in case of left bracketing.
If $w_1$ modifies $w_2$, then $w_1w_2$ is a noun compound,
which now acts as a single noun to modify $w_3$.
Consider for example {\it law enforcement agent}: {\it law} is a modifier of {\it enforcement},
together they form a noun compound {\it law enforcement},
which in turn, as a whole, modifies {\it agent}.

\section{Frequency-based Association Scores}
\label{sec:assoc:scores}

Below I describe different ways to estimate the strength of association between a pair of words
$(w_i,w_j)$ -- frequencies, probabilities, pointwise mutual information, and $\chi^2$.

\subsection{Frequency}

The simplest association score I use is frequency-based:
\begin{equation}
    Assoc(w_i,w_j) = \#(w_i,w_j)
\end{equation}

\noindent where $\#(w_i,w_j)$ is a bigram frequency ($1 \leq i < j \leq 3$),
which can be approximated as the number of page hits returned by a search
engine for the exact phrase query ``$w_i$ $w_j$''.

In the experiments below, I sum the frequencies for
all inflected forms $infl(w_j)$ for the word $w_j$:
\begin{equation}
    Assoc(w_i,w_j) = \sum_{t_j \in infl(w_j)}{\#(w_i,t_j)}
\end{equation}

For example, the calculation of the strength of association
between {\it stem} and {\it cell} would add up
\#({\it stem}, {\it cell}) and \#({\it stem}, {\it cells}).

\subsection{Conditional Probability}
\label{sec:assoc:scores:prob}

I also use conditional probabilities as an association score:
\begin{equation}\label{bracketing:probs}
    Assoc(w_i,w_j) = \mathrm{Pr}(w_i \rightarrow w_j|w_j)
\end{equation}

\noindent where $\mathrm{Pr}(w_i \rightarrow w_j|w_j)$ is the
probability that $w_i$ modifies the head ($w_j$, $1 \leq i < j \leq 3$).
This probability can be estimated as follows:
\begin{equation}
    \mathrm{Pr}(w_i \rightarrow w_j|w_j) = \frac{\#(w_i,w_j)}{\#(w_j)}
\end{equation}
\noindent where $\#(w_i,w_j)$ and $\#(w_j)$ are the corresponding bigram
and unigram frequencies, which
can be approximated as the number of page hits returned by a search
engine, as demonstrated by \namecite{Keller:Lapata:03:web:unseen:bigrams}.
See chapter \ref{chapter:instability} for a discussion on the stability of such estimates.

In my experiments, I sum the frequencies for all inflected forms of the word $w_j$:
\begin{equation}
    Assoc(w_i,w_j) = \frac{\sum_{t_j \in infl(w_j)}{\#(w_i,t_j)}}{\sum_{t_j \in infl(w_j)}{\#(t_j)}}
\end{equation}


\subsection{Pointwise Mutual Information}

Another association score I use is the {\it pointwise mutual information} (PMI):
\begin{equation}
    Assoc(w_i,w_j) = \mathrm{PMI}(w_i,w_j)
\end{equation}

\noindent which is defined as follows:
\begin{equation} \label{PMI:def}
\mathrm{PMI}(w_i,w_j) = \log
\frac{\mathrm{Pr}(w_i,w_j)}{\mathrm{Pr}(w_i)\mathrm{Pr}(w_j)}
\end{equation}

Here $\mathrm{Pr}(w_i,w_j)$ is the probability of the sequence ``$w_i$ $w_j$'',
and $\mathrm{Pr}(w)$ is the probability of the word $w$ on the Web.
Let $N$ be the total number of words/bigrams on the Web, which
I estimate to be about 8 trillion: {\it Google} indexes about 8
billion pages and I hypothesize that each one contains about 1,000 words/bigrams on average.
Then the probabilities can be estimated as follows:
\begin{equation}
\mathrm{Pr}(w_i,w_j) = \frac{\#(w_i,w_j)}{N}
\end{equation}
\begin{equation}
\mathrm{Pr}(w_i) = \frac{\#(w_i)}{N}
\end{equation}

And therefore for PMI we obtain:
\begin{equation}
\mathrm{PMI}(w_i,w_j) = \log
\frac{N \times \#(w_i,w_j)}{\#(w_i)\#(w_j)}
\end{equation}

Again, in my experiments, I sum the frequencies for all inflected forms:
\begin{equation}
    Assoc(w_i,w_j) = \log \frac{N \times \sum_{t_j \in infl(w_j)}{\#(w_i,t_j)}}{\sum_{t_i \in infl(w_i)}{\#(t_i)} \sum_{t_j \in infl(w_j)}{\#(t_j)}}
\end{equation}

Using the definition of conditional probability, we can re-write eq. \ref{PMI:def}, as follows:
\begin{equation}
\mathrm{PMI}(w_i,w_j)
= \log \frac{\mathrm{Pr}(w_i \rightarrow w_j | w_j)}{\mathrm{Pr}(w_i)}
\end{equation}

The dependency model compares $\mathrm{PMI}(w_1,w_2)$ and $\mathrm{PMI}(w_1,w_3)$.
We have:
\begin{equation}
\mathrm{PMI}(w_1,w_2)
= \log \frac{\mathrm{Pr}(w_1 \rightarrow w_2 | w_2)}{\mathrm{Pr}(w_1)}
= \log \mathrm{Pr}(w_1 \rightarrow w_2 | w_2) - \log \mathrm{Pr}(w_1)
\end{equation}
\begin{equation}
\mathrm{PMI}(w_1,w_3)
= \log \frac{\mathrm{Pr}(w_1 \rightarrow w_3 | w_3)}{\mathrm{Pr}(w_1)}
= \log \mathrm{Pr}(w_1 \rightarrow w_3 | w_3) - \log \mathrm{Pr}(w_1)
\end{equation}

Since $\log \mathrm{Pr}(w_1)$ is subtracted from both expressions, we can ignore it
and directly compare $\log \mathrm{Pr}(w_1 \rightarrow w_2 | w_2)$ to $\log \mathrm{Pr}(w_1 \rightarrow w_3 | w_3)$.
Since the logarithm is a monotonic function, we can simply compare
$\mathrm{Pr}(w_1 \rightarrow w_2 | w_2)$ to $\mathrm{Pr}(w_1 \rightarrow w_3 | w_3)$,
i.e. for the {\it dependency model}, comparing PMIs is equivalent to comparing conditional probabilities.
This is not true for the {\it adjacency model} though.

\subsection{Chi-square}
\label{sec:chi2:score}

The last association score is {\it Chi-square} ($\chi^2$),
which uses the following statistics:

\begin{itemize}
\itemsep -3pt
\item[$A$ =] $\#(w_i,w_j)$;
\item[$B$ =] $\#(w_i,\overline{w_j})$, number of
bigrams where $w_i$ is followed by a word other than $w_j$;
\item[$C$ =] $\#(\overline{w_i},w_j)$, number of bigrams,
ending in $w_j$, whose first word is other than $w_i$;
\item[$D$ =] $\#(\overline{w_i},\overline{w_j})$, number of bigrams
with first word that is not $w_i$ and second is not $w_j$.
\end{itemize}

\noindent These statistics are combined in the following formula:
\begin{equation}
\chi^2 = \frac{N (AD-BC)^2}{(A+C) (B+D) (A+B) (C+D)}
\end{equation}

Here $N=A+B+C+D$ is the total number of bigrams on the Web (estimated as 8 trillion),
$B = \#(w_i) - \#(w_i,w_j)$ and $C=\#(w_j) - \#(w_i,w_j)$.  While it is hard to
estimate $D$ directly, it can be calculated as $D=N-A-B-C$.
Finally, as for the other association scores,
I sum the unigram and bigram frequencies for the inflected word forms:
$\#(w_i) = \sum_{t_i \in infl(w_i)}{\#(t_i)}$, and $\#(w_i,w_j) = \sum_{t_j \in infl(w_j)}{\#(w_i,t_j)}$.

\section{Web-Derived Surface Features}
\label{sec:bracketing:web:derived:features}

Consciously or not, authors sometimes disambiguate the noun compounds they
write by using surface-level markers to suggest the correct meaning.
For example, in some cases an author could write {\it law enforcement officer}
with a dash, as {\it law-enforcement officer}, thus suggesting a left bracketing interpretation.
I have found that exploiting such markers, when they occur, can
be very helpful for making bracketing predictions.  The enormous size of the Web
facilitates finding them frequently enough to be useful.

Unfortunately, Web search engines ignore punctuation characters,
thus preventing querying directly for terms containing hyphens,
brackets, apostrophes, etc. For example, querying for
\texttt{"brain-stem cell"} would produce the same result as for \texttt{"brain stem cell"},
which makes the number of page hits unusable for counting the occurrences of hyphenation.
Therefore, I look for the features indirectly by issuing
an exact phrase query for the noun compound and then post-processing the
results (up to 1,000 per query).
I collect the returned text snippets (typically 1-2 sentences or pars of sentences)
and then I search for the surface patterns using regular expressions over the text.
Each match increases the score for left or right bracketing, depending on which one the pattern favors.

While some of the features may be more reliable than others,
I do not try to weight them. Given a noun compound,
I make a bracketing decision by comparing the total number of
left-predicting surface feature instances (regardless of their type)
to the total number of right-predicting feature instances. This appears as {\it Surface features
(sum)} in Tables \ref{table:lauer} and \ref{table:bio}.
The surface features used are described below.

\subsection{Dash}

One very productive feature is the {\it dash} (hyphen).  Consider
for example {\it cell cycle analysis}. If we can find a version of
it in which a dash occurs between the first two words, {\it
cell-cycle}, this suggests left bracketing for the full compound.
Similarly, the dash in {\it donor T-cell} favors right bracketing.
Dashes predicting right bracketing are less reliable though, as their scope is
ambiguous. For example, in {\it fiber optics-system}, hyphen's scope is
not {\it optics}, but {\it fiber optics}. Finally, multiple dashes are unusable,
e.g., {\it t-cell-depletion}.

\subsection{Genitive Marker}

The genitive ending, or {\it genitive marker} is another useful
feature. Finding the phrase {\it brain\underline{'s} stem cells}
suggests right bracketing for {\it brain stem cells},
while finding {\it brain stem\underline{'s} cells} suggests left bracketing.
Note that these features can also occur combined, as in {\it brain's stem-cells},
where we have both a {\it dash} and a {\it genitive marker}.

\subsection{Internal Capitalization}

{\it Internal capitalization} is another highly reliable feature.
For example, the capitalized spellings
{\it Plasmodium \underline{v}ivax \underline{M}alaria} and
{\it plasmodium \underline{v}ivax \underline{M}alaria} both suggest
left bracketing for the noun compound {\it plasmodium vivax malaria},
while {\it \underline{b}rain \underline{S}tem cells} and
{\it \underline{b}rain \underline{S}tem Cells} both favor right bracketing
for {\it brain stem cells}.

This feature is disabled on Roman digits and single-letter words
in order to prevent problems with terms like {\it vitamin D deficiency},
which is right-bracketed and where the capitalization is just a convention.

\subsection{Embedded Slash}

I also make use of {\it embedded slashes}. For example, in
{\it leukemia/lymphoma cell}, the first word is an alternative of the second one and
therefore cannot modify it, which suggests right bracketing for {\it leukemia lymphoma cell}.

\subsection{Parentheses}

In some cases we can find instances of the noun compound in which one or more
words are enclosed in {\it parentheses}, e.g., {\it growth factor (beta)}
or {\it (growth factor) beta}, both of which suggest left bracketing
for {\it growth factor beta}; or {\it (brain) stem cells} and {\it brain (stem cells)},
which favor right bracketing for {\it brain stem cells}.

\subsection{External Dash}

{\it External dashes}, i.e. dashes to words outside the target compound
can be used as weak indicators. For example, {\it \underline{mouse-}brain stem cells}
favors right bracketing for {\it brain stem cells}, while
{\it tumor necrosis factor\underline{-alpha}} suggests
left bracketing for {\it tumor necrosis factor}.

\subsection{Other Punctuation}

Even a comma, a period or a colon (or any special character)
can act as an indicator. For example, ``{\it health care, provider}'' or ``{\it
lung cancer: patients}'' are weak predictors of left bracketing,
showing that the author chose to keep two of the words together,
separating out the third one, while ``{\it home. health care}'' and
``{\it adult, male rat}'' suggest right bracketing.

\section{Other Web-Derived Features}

Some features can be obtained directly as the number of page hits
for a suitably formulated query.
As these counts are derived from the entire Web, not just from the top 1,000 summaries,
they are of different magnitude and therefore
should not be added to the frequencies of the above surface features;
I use them in separate models.

\subsection{Genitive Marker -- Querying Directly}

In some cases, it is possible to query for {\it genitive markers} directly:
while search engines do not index apostrophes, since they index no punctuation,
the {\it s} in a genitive marker is indexed.
This makes it possible to query for ``{\it brain stem's cell}''
indirectly by querying for ``{\it brain stem s cell}''.
An {\it s} in this context is very likely to have been part of a genitive marker,
and if we assume so, then it is possible to make a bracketing decision
by comparing the number of times an {\it s} appears following
the second word vs. the first word.
For example, a query for ``{\it brain s stem cell}'' returns 4 results,
and for ``{\it brain s stem cells}'' returns 281 results, i.e. there are
a total of 285 page hits supporting left bracketing for {\it brain stem cell}.
On the other hand, the right-predicting queries ``{\it brain stem s cell}'' and ``{\it brain stem s cells}''
return 2 and 3 results, respectively.
Thus there are only 5 right-predicting page hits vs. 285 left-predicting,
and therefore the model makes a left bracketing prediction.

{\bf Note:} I conducted the original experiments in 2005, when {\it Google}
seemed to ignore punctuation. It appears that it can handle some of it now.
For example, at the time of writing,
``{\it brain stem's cell}'' return 701 results, all having the genitive marker {\it 's}.

\subsection{Abbreviation}

{\it Abbreviations} are another very reliable feature. For example, {\it
``tumor \underline{n}ecrosis \underline{f}actor (\underline{NF})''} suggests a right bracketing, while
{\it ``\underline{t}umor \underline{n}ecrosis (\underline{TN}) factor''}
favors a left bracketing for {\it ``tumor necrosis factor''}.
Since search engines ignore parentheses and capitalization,
a query for {\it ``tumor necrosis factor (NF)''} is equivalent to a query for
{\it ``tumor necrosis factor nf''}. Again, I compare the number of page hits
where the abbreviation follows the second word to the number of page hits
where the abbreviation follows the first word (using inflections).
In general, this model is very accurate, but errors may occur when the
abbreviation is an existing word (e.g., {\it me}), a Roman digit
(e.g., {\it IV}), a state name abbreviation (e.g., {\it CA}), etc.

\subsection{Concatenation}

\subsubsection{Concatenation Adjacency Model}

Some noun compounds are often written concatenated.
I exploit this in adjacency and dependency models.
Let $concat(w_i,w_j)$ be the concatenation of the words $w_i$ and $w_j$.
Given a three-word noun compound $w_1w_2w_3$,
the {\it concatenation adjacency model} compares the number of page hits
for $concat(w_1,w_2)$ and $concat(w_2,w_3)$,
using all possible inflections of $w_2$ and $w_3$, respectively.
It predicts a left bracketing if the former is greater, a right bracketing if the latter is greater,
and makes no prediction otherwise.

Consider for example the noun compound {\it health care reform}.
{\it Google} returns 98,600,000 page hits for {\it healthcare}, and 33,000 more for {\it healthcares}:
a total of 98,633,000.
On the other hand, there are only 471 page hits for {\it carereform}, and 27 more for {\it carereforms}:
a total of 498. Therefore, the model predicts a left bracketing.

\subsubsection{Concatenation Dependency Model}

Given a three-word noun compound $w_1w_2w_3$,
the {\it concatenation dependency model} compares the number of page hits
for $concat(w_1,w_2)$ and $concat(w_1,w_3)$,
using all possible inflections of $w_2$ and $w_3$, respectively.
It predicts a left bracketing if the former is greater, a right bracketing if the latter is greater,
and makes no prediction otherwise.

As we have seen above, {\it Google} returns a total of 98,633,000 for {\it healthcare} and {\it healthcares}.
On the other hand, there are only 27,800 page hits for {\it healthreform}, and 82 more for {\it healthreforms}:
a total of 27,882. Therefore, the model predicts a left bracketing for the noun compound {\it health care reform}.

\subsubsection{Concatenation Adjacency Triples}

Given a three-word noun compound $w_1w_2w_3$,
the {\it concatenation adjacency triples} model compares the number of page hits
for the exact phrase {\it Google} queries: ``$concat(w_1,w_2)$ $w_3$'' and ``$w_1$ $concat(w_2,w_3)$
using all possible inflections of $w_2$ and $w_3$, and for $w_1$ and $w_3$, respectively.
It predicts a left bracketing if the former is greater, a right bracketing if the latter is greater,
and makes no prediction otherwise.

At the time of writing, {\it Google} returns
668,000 page hits for {\it ``healthcare reform''},
77,700 page hits for {\it ``healthcare reforms''},
and no page hits for {\it ``healthcares reform''} or {\it ``healthcares reforms''}:
a total of 745,700.
On the other hand, there are only
289 page hits for {\it ``health carereform''},
15 page hits for {\it ``health carereforms''},
and no page hits for {\it ``health carereforms''} or {\it ``healths carereforms''}:
a total of 304.
This supports a left bracketing prediction for the noun compound {\it health care reform}.

\subsection{Wildcard}

I also make use of the {\it Google}'s support for ``*'', a wildcard substituting
a single word.\footnote{While it should substitute a single word, {\it Google}'s ``*'' operator often substitutes multiple words.}
The idea is to see how often two of the words follow each other and are separated from the third word by some
other word(s). This implicitly tries to capture paraphrases involving the two
sub-concepts making up the whole.
I propose wildcards adjacency and dependency models,
using up to three asterisks, possibly with a re-ordering. 

\subsubsection{Wildcard Adjacency Model}

Given a three-word noun compound $w_1w_2w_3$,
the {\it wildcards adjacency model} compares the number of page hits
for ``{\it $w_1$ $w_2$ * $w_3$}'' and ``{\it $w_1$ * $w_2$ $w_3$}'',
using all possible inflections of $w_2$ and $w_3$, and for $w_1$ and $w_3$, respectively.
It predicts a left bracketing if the former is greater, a right bracketing if the latter is greater,
and makes no prediction otherwise.
I also use a version of the model with two and with three asterisks.

Consider the following example, where only one asterisk is used.
At the time of writing, {\it Google} returns
556,000 page hits for {\it ``health care * reform''},
79,700 page hits for {\it ``health care * reforms''},
1 page hit for {\it ``health cares * reform''},
and no page hits for {\it ``health cares * reforms''}:
a total of 635,701.
On the other hand, there are only
255,000 page hits for {\it ``health * care reform''},
17,600 page hits for {\it ``health * care reforms''},
1 page hit for {\it ``health * care reforms''},
and no page hits for {\it ``healths * care reforms''}:
a total of 272,601.
This supports a left bracketing prediction for the noun compound {\it health care reform}.

\subsubsection{Wildcard Dependency Model}

Given a three-word noun compound $w_1w_2w_3$,
the {\it wildcards dependency model} compares the number of page hits
for ``{\it $w_1$ $w_2$ * $w_3$}'' and ``{\it $w_2$ * $w_1$ $w_3$}'',
using all possible inflections of $w_2$ and $w_3$.
It predicts a left bracketing if the former is greater, a right bracketing if the latter is greater,
and makes no prediction otherwise.
I try using up to three asterisks.

Consider the following example, where only one asterisk is used.
At the time of writing, {\it Google} returns
556,000 page hits for {\it ``health care * reform''},
79,700 page hits for {\it ``health care * reforms''},
1 page hit for {\it ``health cares * reform''},
and no page hits for {\it ``health cares * reforms''}:
a total of 635,701.
On the other hand, there are only
198,000 page hits for {\it ``care * health reform''},
12,700 page hits for {\it ``care * health reforms''},
122,000 page hit for {\it ``cares * health reform''},
and no page hits for {\it ``cares * health reforms''}:
a total of 332,700.
This supports a left bracketing prediction for the noun compound {\it health care reform}.

\subsubsection{Wildcard Reversed Adjacency Model}

Given a three-word noun compound $w_1w_2w_3$,
the {\it wildcards reversed adjacency model} compares the number of page hits
for ``{\it $w_3$ * $w_1$ $w_2$}'' and ``{\it $w_2$ $w_3$ * $w_1$}'',
using inflections for $w_2$ and $w_3$, and for $w_1$ and $w_3$, respectively.
It predicts a left bracketing if the former is greater, a right bracketing if the latter is greater,
and makes no prediction otherwise.

Consider the following example, where only one asterisk is used.
At the time of writing, {\it Google} returns
824,000 page hits for {\it ``reform * health care''},
3 page hits for {\it ``reform * health cares''},
119,000 page hits for {\it ``reforms * health care''},
and 2 page hits for {\it ``reforms * health cares''}:
a total of 943,005.
On the other hand, there are only
255,000 page hits for {\it ``care reform * health''},
1 page hit for {\it ``care reform * healths''},
13,900 page hit for {\it ``care reforms * health''},
and no page hits for {\it ``care reforms * healths''}:
a total of 268,901.
This supports a left bracketing prediction for the noun compound {\it health care reform}.

\subsubsection{Wildcard Reversed Dependency Model}

Given a three-word noun compound $w_1w_2w_3$,
the {\it wildcards reversed dependency model} compares the number of page hits
for ``{\it $w_3$ * $w_1$ $w_2$}'' and ``{\it $w_1$ $w_3$ * $w_2$}'',
using all possible inflections of $w_2$ and $w_3$.
It predicts a left bracketing if the former is greater, a right bracketing if the latter is greater,
and makes no prediction otherwise. I try using up to three asterisks.

Consider the following example, where only one asterisk is used.
At the time of writing, {\it Google} returns
824,000 page hits for {\it ``reform * health care''},
3 page hits for {\it ``reform * health cares''},
119,000 page hits for {\it ``reforms * health care''},
and 2 page hits for {\it ``reforms * health cares''}:
a total of 943,005.
On the other hand, there are only
273,000 page hits for {\it ``health reform * care''},
122,000 page hit for {\it ``health reform * cares''},
10,800 page hit for {\it ``health reforms * care''},
and no page hits for {\it ``health reforms * cares''}:
a total of 405,800.
This supports a left bracketing prediction for the noun compound {\it health care reform}.

\subsection{Reordering}

I also try a simple {\it reordering} without inserting asterisks.
Given a three-word noun compound $w_1w_2w_3$,
the {\it reordering model} compares the number of page hits
for ``{\it $w_3$ $w_1$ $w_2$}'' and ``{\it $w_2$ $w_3$ $w_1$}'',
using all possible inflections of $w_2$ and $w_3$, and for $w_1$ and $w_3$, respectively.
It predicts a left bracketing if the former is greater, a right bracketing if the latter is greater,
and makes no prediction otherwise.
For example, at the time of writing, {\it Google} returns
137,000 page hits for {\it ``reform health care''},
1,010 page hits for {\it ``reforms health care''},
and no page hits for {\it ``reform health cares''} or ``{\it reforms health cares}'':
a total of 138,010.
On the other hand, there are only
23,300 page hits for {\it ``care reform health''},
1,720 page hits for {\it ``care reforms health''},
and no page hits for {\it ``care reform healths''} or {\it ``care reforms healths''}:
a total of 25,020.
This supports a left bracketing prediction for {\it health care reform}.

\subsection{Internal Inflection Variability}

Further, I try to use the {\it internal inflection variability}.
Consider the three-word noun compound {\it ``tyrosine kinase activation''}.
Since it is left-bracketed, the first two nouns form a two-word noun compound
(which in turn modifies the third noun). Therefore, we could expect to see
this two-word noun compound inflected in the context of the three-word noun compound, i.e.
{\it ``tyrosine kinase\underline{s} activation''}. If it were right bracketed,
we would expect a possible inflection on the first noun: {\it ``tyrosine\underline{s} kinase activation''}.
The model compares the number of page hits for the internal inflections of the second noun vs. the first noun.
It predicts a left bracketing if the former is greater, a right bracketing if the latter is greater,
and makes no prediction otherwise.

For example, the left-predicting patterns for {\it ``tyrosine kinase activation''}
are {\it ``tyrosine kinase\underline{s} activation''} and
{\it ``tyrosine kinase\underline{s} activation\underline{s}''}, yielding
882 + 0 = 882 page hits.
On the other hand, there are no page hits for {\it ``tyrosine\underline{s} kinase activation''}
nor for {\it ``tyrosine\underline{s} kinase activation\underline{s}''}.
This supports a left bracketing prediction for the noun compound {\it tyrosine kinase activation}.

In case of plural nouns, I convert them to singular. Consider for example
the noun compound {\it math problems solutions}.
The left-predicting patterns are ``{\it math problem\underline{ } solutions}''
and ``{\it math problem\underline{ } solution\underline{ }}'', which yield 3,720 + 2,920 = 6,640 page hits.
The right-predicting patterns are ``{\it math\underline{s} problems solutions}''
and ``{\it math\underline{s} problems solution\underline{ }}'', which yield 2,990 + 0 = 2,990 page hits.
This supports a left bracketing prediction.

\subsection{Swapping the First Two Words}

Recall the {\it adult male rat} example, discussed in section \ref{sec:adj:vs:dep}.
This is a right-bracketed noun compound, where the first two words independently modify
the third one, and therefore we can swap them. For example, at the time of writing,
{\it Google} returns 78,200 page hits for {\it adult male rat} and 1,700 for {\it male adult rat}.
I exploit this literally: testing whether it is feasible to swap the first two words,
in which case I make a right bracketing prediction.

More formally, given a three-word noun compound $w_1w_2w_3$,
the {\it swapping the first two words model} checks the number of page hits
for ``{\it $w_2$ $w_1$ $w_3$}'', using all possible inflections for $w_1$, $w_3$, and $w_3$.
It predicts a left bracketing if the former is greater, a right bracketing if the latter is greater,
and makes no prediction otherwise.

\section{Paraphrases}
\label{sec:paraphrases}

Many researchers point out that the semantics of the relation between the nouns
in a noun compound can be made overt by a paraphrase \cite{warren:1978,lauer:1995:thesis}.
For example, an author describing the concept of {\it brain stem cell}
may choose to write it in a more expanded manner, paraphrasing it as {\it stem cells \underline{from} the brain}.
Since {\it stem} and {\it cells} are kept together in that paraphrase,
this suggests a right bracketing.
On the other hand, if the expansion groups the first two nouns together,
as in {\it cells \underline{from} the brain stem}, this would support a left bracketing.

Not all noun compounds can be readily expressed with a prepositional
paraphrase \cite{levi:1978}. For example, the noun compound {\it skyscraper office building}
is best expanded as a copula paraphrase, e.g., {\it office building \underline{that is} a skyscraper},
which predicts a right bracketing. Another option is to use verbal paraphrases like
{\it pain \underline{associated with} arthritis migraine},
which supports a left bracketing for {\it arthritis migraine pain}.

Some researchers use prepositional paraphrases as a proxy for
determining the semantic relation that holds between the nouns in a
{\it two-word} noun compound \cite{girju:2005:on:the:semantics,Lapata:Keller:04,lauer:1995:thesis}.
\namecite{lauer:1995:thesis} takes the idea literally,
formulating the semantic interpretation task as selecting one of the following prepositions\footnote{He excludes {\it like}, which is mentioned by \namecite{warren:1978}.}:
{\it of}, {\it for}, {\it in}, {\it at}, {\it on}, {\it from}, {\it with} and {\it about}.
For example, a {\it night flight} is paraphrased as a {\it flight \underline{at} night},
while {\it war story} is paraphrased as a {\it story \underline{about} war}.

In contrast, I use paraphrases in order to obtain evidence supporting
left or right {\it syntactic} bracketing decision for {\it three-word} noun compounds.
Rather than trying to decide on a single best paraphrase,
I issue queries using several different paraphrase patterns.
I find how often each one occurs on the Web, add separately
the number of page hits predicting a left versus right bracketing
and then compare the two sums. As before, the bigger sum wins,
and no decision is made in case of equal sums.
While some prepositions may be better predictors than other ones
(see \cite{girju:2005:on:the:semantics} for a frequency distribution),
I have just one feature for all prepositions.

Unfortunately, search engines lack linguistic annotations and
therefore they do not support typed queries like ``{\it stem cells VERB PREP DET brain}'',
where the uppercase placeholders stand for a verb, a preposition and a determiner, respectively.
This makes the general verbal paraphrases prohibitively expensive --
a separate query is needed for every combination of a verb,
a preposition and a determiner.

To overcome this problem, I use a fixed set of
{\it verbal prepositions}, i.e. passive verbs accompanied by a preposition
acting as a complex preposition, e.g., {\it used in} or {\it made of};
a full list is given below.

More formally, given a three-word noun compound $w_1w_2w_3$, I use the following generalized
left-predicting exact phrase queries which keep $w_1$ and $w_2$ together.
Note that $w_2$ and $w_3$ can be number-inflected for singular or plural,
which is indicated with the $^{\prime}$ sign in the patterns below:

\begin{enumerate}
  \item $w_3^{\prime}$ \texttt{PREP DET?} $w_1$ $w_2^{\prime}$
  \item $w_3^{\prime}$ \texttt{COMPL BE DET?} $w_1$ $w_2^{\prime}$
  \item $w_3^{\prime}$ \texttt{COMPL BE PREP DET?} $w_1$ $w_2^{\prime}$
\end{enumerate}

Similarly, the right-predicting paraphrases are generated using
corresponding generalized exact phrase queries that keep $w_2$ and $w_3$ together
($w_1$ and $w_3$ can be inflected):

\begin{enumerate}
  \item $w_2$ $w_3^{\prime}$ \texttt{PREP DET?} $w_1^{\prime}$
  \item $w_2$ $w_3^{\prime}$ \texttt{COMPL BE DET?} $w_1^{\prime}$
  \item $w_2$ $w_3^{\prime}$ \texttt{COMPL BE PREP DET?} $w_1^{\prime}$
\end{enumerate}

The above queries are instantiated in all possible ways
by enumerating all inflected forms for $w_1$, $w_2$ and $w_3$
(where indicated with the $^{\prime}$ sign)
and by substituting the placeholders \texttt{COMPL} (a complementizer),
\texttt{BE} (a form of the verb to be), \texttt{DET} (a determiner)
and \texttt{PREP} (a preposition) using the following
values\footnote{In the process of generation, I use {\it WordNet} to check the number agreement.
While wrong agreement does not generally harm the accuracy since it generates phrases
that simply return no results, I still try to eliminate it in order to reduce
the total number of necessary queries.}:

\begin{itemize}
  \item \texttt{PREP}:
    \begin{itemize}
      \item {\bf Nonverbal prepositions:} {\it about, across, after, against,
all over, along, alongside, amid, amidst, among, around,
as, as to, aside, at, before, behind, beside, besides, between, beyond,
by, close to, concerning, considering, down, due to, during, except, except for,
excluding, following, for, from, in, in addition to, in front of, including,
inside, instead of, into, like, near, of, off, on, onto, other than, out, out of,
outside, over, per, regarding, respecting, similar to, through, throughout, to,
toward, towards, under, underneath, unlike, until, up, upon, versus, via, with,
within, without};
      \item {\bf Verbal prepositions:} {\it associated with, caused by, contained in,
derived from, focusing on, found in, involved in, located at, located in, made of,
performed by, preventing, related to, used by, used in, used for}.
    \end{itemize}
  \item \texttt{DET} -- an optional element, which can be any of the following:
    \begin{itemize}
      \item {\bf Determiners:} {\it a, an, the};
      \item {\bf Quantifiers:} {\it all, each, every, some};
      \item {\bf Possessive pronouns:} {\it his, her, their};
      \item {\bf Demonstrative pronouns:} {\it this, these}.
    \end{itemize}
  \item \texttt{COMPL}
    \begin{itemize}
      \item {\bf Complementizers:} {\it that, which, who}.
    \end{itemize}
  \item \texttt{BE}
    \begin{itemize}
      \item {\bf Copula:} {\it are, is, was, were}.
    \end{itemize}
\end{itemize}

Some example left-predicting pattern instantiations
for the noun compound {\it bone marrow cells} follow:
{\it cell from the bone marrow,
cells in the bone marrow,
cells derived from the bone marrow,
cells that are from the bone marrows,
cell which is in a bone marrow}, etc.

The reason to include so many prepositions comes from my concern
that the coverage of Lauer's eight prepositions plus the copula paraphrases
might be insufficient. While the extended set of prepositions
did not improve the coverage in my experiments, it boosted the accuracy by more than 2\%.

One possible explanation is that the specialized biomedical
terms are paraphrased more often, probably in order to be explained to readers
that are new to the domain, while Lauer's noun compounds are extracted from
popular encyclopedia text and, therefore most of them are in everyday use.

\section{Datasets}

\subsection{Lauer's Dataset}
\label{dataset:lauer}

{\it Lauer's dataset} is considered the benchmark dataset for
the noun compound bracketing problem.
It is listed in the appendix of Lauer's thesis
\cite{lauer:1995:thesis}, and consists of 244 unambiguous three-word noun compounds
extracted from {\it Grolier's encyclopedia}.
In order to extract his noun compounds, Lauer POS-tagged the encyclopedia text and
then extracted three-word noun compound candidates $w_1w_2w_3$
by heuristically looking for five-word sequences $w_0w_1w_2w_3w_4$,
where $w_1$, $w_2$ and $w_3$ are nouns and $w_0$, $w_4$ are not nouns \cite{lauer:1995:thesis}.
This heuristic yielded 625 three-word noun compound candidate instances.
Since Lauer's learning models rely on a thesaurus, he discarded any
noun compound that contained a word that was not in {\it Roget's thesaurus},
which left him with 308 candidates.

\begin{table}
\begin{center}
\begin{small}
\begin{tabular}{|lll|c|}
  \multicolumn{3}{c}{\bf Noun Compound} & \multicolumn{1}{c}{\bf Bracketing}\\
  \hline
    minority & business & development & left \\
    satellite & data & systems & right \\
    disaster & relief & assistance & left \\
    county & extension & agents & right \\
    world & food & production & right \\
    granary & storage & baskets & right \\
    customs & enforcement & vehicles & left \\
    airport & security & improvements & left \\
    mountain & summit & areas & left \\
    law & enforcement & agencies & left \\
    college & commencement & poem & right \\
    health & education & institutions & left \\
    country & music & theme & left \\
    sea & transportation & hub & left \\
    army & ant & behavior & left \\
    missile & defense & systems & left \\
    world & petroleum & production & right \\
    arab & independence & movements & left \\
    speech & recognition & system & left \\
    production & passenger & vehicles & right \\
    revenue & ton & miles & right \\
    combustion & chemistry & technology & left \\
    science & fiction & novels & left \\
    missile & guidance & system & left \\
    sea & bass & species & left \\
    radiation & energy & conversion & left \\
    science & fiction & novels & left \\
    energy & distribution & properties & left \\
    science & fiction & writer & left \\
    science & fiction & themes & left \\
    breeder & technology & development & left \\
    landmark & majority & opinions & right \\
    community & college & system & left \\
    town & council & members & left \\
    war & crimes & prosecutor & left \\
    health & insurance & laws & left \\
    science & fiction & satire & left \\
    death & penalty & statutes & left \\
    calvinist & peasant & family & right \\
    exhibition & ballroom & dancers & right \\
  \hline
\end{tabular}
\end{small}
\caption{{\bf Lauer's dataset:} sample noun compounds and bracketing labels.}
\label{table:dataset:lauer}
\end{center}
\end{table}

Lauer manually investigated these 308 examples and tried to annotate each one
with a left or a right bracketing using as context the whole encyclopedia article
the noun compound occurred in. In the process, he discovered that 29 examples (i.e. 9\%)
represented extraction errors, e.g., ``{\it In \underline{monsoon regions} \underline{rainfall} does not $\ldots$}''.
He further could not unambiguously classify as left or right bracketed
another 35 examples (or 11\%), e.g., ``{\it Most advanced aircraft have \underline{precision navigation systems}.}''.
After having discarded these two categories, he was left with 244 unambiguous three-word noun compounds
to be used for testing: 163 left-bracketed and 81 right-bracketed.
His algorithms are unsupervised and need no training data.

Table \ref{table:dataset:lauer} shows some examples from this dataset.
Note that the resulting {\it Lauer's dataset} contains repetitions:
there are only 215 distinct examples out of all 244.

Lauer used his own left/right in-context judgments only.
However, he also performed an experiment with seven additional human annotators,
who were asked to annotate the three-word noun compounds alone, {\it out-of-context}.
Lauer split the 215 examples into two blocks, with three judges assigned to the first block
and four judges to the second block. On average the judges' judgments matched
Lauer's annotations 81.5\% of the time, and any two judges agreed between themselves
on between 73\% and 84\% of the instances.

Note that this 81.5\% agreement
cannot be considered as an ``upper
bound'' for his the test set: the actual test set contains 244 examples (with the repetitions),
while the inter-annotator agreement was measured over 215 examples only.
In addition, Lauer's programs had access to a text corpus and to a thesaurus,
while the human judges were given the noun compounds only, in isolation.
Therefore, it should not be surprising if some computer system manages to beat
the 81.5\% ``upper bound'' on accuracy.


\subsection{Biomedical Dataset}
\label{sec:dataset:bio}

I assembled a new dataset from the biomedical literature: the {\it Biomedical dataset}.
Unlike {\it Lauer's dataset}, this dataset is extracted using a more sophisticated
language technology, from a much larger collection (in the biomedical domain),
and has been annotated by two independent judges.
See section \ref{sec:dataset:bio:extract} in appendix \ref{app:dataset:bio}
for a detailed description of the process of extraction of the most frequent
three-word noun compounds in a 1.4 million abstracts subset of MEDLINE.

The top 500 extracted noun compounds are annotated independently by two judges:
myself and a student with a biological background.
The problematic cases are reconsidered by the two judges and,
after agreement is reached, the set contains 361 left-bracketed,
69 right-bracketed, and 70 ambiguous noun compounds. The latter group is not used
in the evaluation. The 430 unambiguous examples are shown in section \ref{appendix:bio:set}
in appendix \ref{app:dataset:bio}: the 70 ambiguous examples are omitted
since they are not used in the evaluation.

The inter-annotator agreement for the original annotator choices is 82\%.
In addition, I calculate the value of the Kappa statistics, which measures
the agreement between two \cite{Cohen:1960:kappa}
or more \cite{Fleiss:1981:kappa} annotators on categorial judgments,
taking into account the probability of agreement by chance:
\begin{equation}\label{kappa}
    K = \frac{\mathrm{Pr}(A) - \mathrm{Pr}(E)}{1 - \mathrm{Pr}(E)}
\end{equation}

\noindent where $\mathrm{Pr}(A)$ is the observed agreement, and
$\mathrm{Pr}(E)$ is the agreement by chance.

$K$ is 1 in case of complete agreement and is 0 if the agreement matches what is expected by chance.
The statistics can be less than 0 as well if the agreement is less than the expected by chance.
The values of $K$ can be interpreted as follows \cite{Siegel:Castellan:1988:kappa}:

\begin{itemize}
  \item 0.00–--0.20: slight agreement;
  \item 0.21--0.40: fair agreement
  \item 0.41--0.60: moderate agreement;
  \item 0.61--0.80: substantial agreement;
  \item 0.81--1.00: (almost) perfect agreement.
\end{itemize}

For the {\it Biomedical dataset} the value of the Kappa statistics was 0.442,
which corresponds to moderate inter-annotator agreement.
If we consider correct the examples marked as ambiguous by one of the annotators only,
the kappa statistics becomes .606, a substantial agreement.

\section{Evaluation}
\label{sec:bracketing:eval}

\subsection{Experimental Setup}

I experimented with {\it Lauer's dataset} (see section \ref{dataset:lauer})
in order to make my results comparable to those of previous researchers, e.g.,
\namecite{lauer:1995:thesis}, \namecite{Lapata:Keller:04}, etc.
I also experimented with the {\it Biomedical dataset},
described in section \ref{sec:dataset:bio} above.

I collected statistics for the $n$-grams, the surface features,
and the paraphrases by issuing exact phrase queries,
limiting the pages to English and requesting filtering
of the similar results. I used {\it MSN Search} (now {\it Live Search})
statistics for the $n$-grams and the paraphrases
(unless the pattern contains an asterisk, in which case I used {\it Google}),
and {\it Google} for the surface features. {\it MSN Search}
always returns exact numbers as page hits estimates,
while {\it Google} and {\it Yahoo!} return rounded estimates.
For example, at the time of writing, {\it Live Search} returns 60,\underline{219,609} page hits for {\it cell},
while {\it Google} and {\it Yahoo!} return 397,\underline{000,000} and 637,\underline{000,000}, respectively.
This rounding can be problematic, especially when comparing two ratios of page hits,
e.g., $\frac{\#(brain,stem)}{\#(stem)}$ and $\frac{\#(stem,cell)}{\#(cell)}$.

For each noun compound, I generated all possible word inflections
(e.g., {\it tumor} and {\it tumors}) and alternative
word variants (e.g., {\it tumor} and {\it tumour}).
For this purpose, for {\it Lauer's dataset}, I used {\it Carroll's morphological
tools}\footnote{\texttt{http://www.cogs.susx.ac.uk/lab/nlp/carroll/morph.html}} \cite{Minnen:Carroll:Pearce:2001}
and the {\it Java WordNet Library}\footnote{\texttt{http://sourceforge.net/projects/jwordnet}} ({\it JWNL}),
which provides programmatic access to {\it WordNet} \cite{Fellbaum:1998:wordnet}.
For the {\it Biomedical dataset},
I also used the {\it UMLS Specialist Lexicon}.\footnote{\texttt{http://www.nlm.nih.gov/pubs/factsheets/umlslex.html}}

For bigrams, I inflected the second word only. Similarly, for a prepositional
paraphrase, I generated the inflected forms for the two parts, before
and after the preposition.
See sections \ref{sec:assoc:scores} and \ref{sec:bracketing:web:derived:features}
for a detailed description of what kinds of inflections are used for each model
and how the page hits are summed up.

\begin{table}
\begin{center}
\begin{tabular}{|l|ccc|cr|}
\multicolumn{1}{l}{\bf Model} & {\bf Correct} & {\bf Wrong} & \multicolumn{1}{c}{\bf N/A} & {\bf Accuracy} & \multicolumn{1}{c}{\bf Cover.}\\
  \hline
\# adjacency & {\small 183} & {\small 61} & {\small 0} & {\small 75.00$\pm$5.79} & {\small 100.00} \\
$\mathrm{Pr}$ adjacency & {\small 180} & {\small 64} & {\small 0} & {\small 73.77$\pm$5.86} & {\small 100.00} \\
$\mathrm{PMI}$ adjacency & {\small 182} & {\small 62} & {\small 0} & {\small 74.59$\pm$5.81} & {\small 100.00} \\
{\bf $\mathbf{\chi^2}$ adjacency} & {\small 184} & {\small 60} & {\small 0} & {\small {\bf 75.41$\pm$5.77}} & {\small 100.00} \\
\hline
\# dependency & {\small 193} & {\small 50} & {\small 1} & {\small 79.42$\pm$5.52} & {\small 99.59} \\
$\mathrm{Pr}$ dependency (= PMI dep.) & {\small 194} & {\small 50} & {\small 0} & {\small 79.51$\pm$5.50} & {\small 100.00} \\
{\bf $\mathbf{\chi^2}$ dependency} & {\small 195} & {\small 49} & {\small 0} & {\small {\bf 79.92$\pm$5.47}} & {\small 100.00} \\
\hline
\# adjacency (*) & {\small 152} & {\small 41} & {\small 51} & {\small 78.76$\pm$6.30} & {\small 79.10}\\
\# adjacency (**) & {\small 162} & {\small 43} & {\small 39} & {\small 79.02$\pm$6.08} & {\small 84.02}\\
\# adjacency (***) & {\small 150} & {\small 51} & {\small 43} & {\small 74.63$\pm$6.44} & {\small 82.38}\\
\hline
\# adjacency (*, rev.) & {\small 163} & {\small 48} & {\small 33} & {\small 77.25$\pm$6.11} & {\small 86.47}\\
\# adjacency (**, rev.) & {\small 165} & {\small 51} & {\small 28} & {\small 76.39$\pm$6.09} & {\small 88.52}\\
\# adjacency (***, rev.) & {\small 156} & {\small 57} & {\small 31} & {\small 73.24$\pm$6.32} & {\small 87.30}\\
\hline
Concatenation adjacency & {\small 175} & {\small 48} & {\small 21} & {\small 78.48$\pm$5.85} & {\small 91.39}\\
{\bf Concatenation dependency} & {\small 167} & {\small 41} & {\small 36} & {\small {\bf 80.29$\pm$5.93}} & {\small 85.25}\\
{\bf Concatenation triples} & {\small 76} & {\small 3} & {\small 165} & {\small {\bf 96.20$\pm$6.78}} & {\small 32.38}\\
\hline
Inflection variability & {\small 69} & {\small 36} & {\small 139} & {\small 65.71$\pm$9.49} & {\small 43.03}\\
Swap first two words & {\small 66} & {\small 38} & {\small 140} & {\small 63.46$\pm$9.58} & {\small 42.62}\\
Reorder & {\small 112} & {\small 40} & {\small 92} & {\small 73.68$\pm$7.52} & {\small 62.30}\\
{\bf Abbreviations} & {\small 21} & {\small 3} & {\small 220} & {\small {\bf 87.50$\pm$18.50}} & {\small 9.84}\\
{\bf Genitive marker} & {\small 32} & {\small 4} & {\small 208} & {\small {\bf 88.89$\pm$14.20}} & {\small 14.75}\\
\hline
{\bf Paraphrases} & {\small 174} & {\small 38} & {\small 32} & {\small {\bf 82.08$\pm$5.72}} & {\small 86.89}\\
{\bf Surface features (sum)} & {\small 183} & {\small 31} & {\small 30} & {\small {\bf 85.51$\pm$5.34}} & {\small 87.70}\\
\hline
Majority vote & {\small 210} & {\small 22} & {\small 12} & {\small 90.52$\pm$4.46} & {\small 95.08}\\
\emph{Majority vote $\rightarrow$ `left'} & {\small \emph{218}} & {\small \emph{26}} & {\small \emph{0}} & {\small {\bf \emph{89.34$\pm$4.50}}} & {\small {\bf \emph{100.00}}}\\
\hline
{\bf Baseline} (choose `left') & {\small 163} & {\small 81} & {\small 0} & {\small 66.80$\pm$6.13} & {\small 100.00}\\
\hline
\end{tabular}
\caption{{\bf Bracketing results on \emph{Lauer's dataset}.}
For each model, the number of correctly classified,
wrongly classified, and non-classified examples is shown,
followed by the accuracy (in \%) and the coverage (\% examples for which the model makes prediction).}
\label{table:lauer}
\end{center}
\end{table}

\subsection{Results for \emph{Lauer's Dataset}}

The results of the evaluation for {\it Lauer's dataset} are shown in Table \ref{table:lauer}.
Note that the results for the accuracy are accompanied by confidence intervals,
which will be explained in section \ref{sec:conf:interval} below.
The left-bracketing baseline model yields 66.8\% accuracy.
The most accurate models are the {\it concatenations triple} (96.2\%),
{\it genitive} (88.89\%) and {\it abbreviation} (87.5\%),
which only make predictions for about 10-30\% of the examples.
The {\it surface features} and the {\it paraphrases} achieve 85.51\% and 82.08\%, respectively,
with almost 90\% coverage. All these models outperform the word association models
({\it adjacency} and {\it dependency}: frequency, probability, PMI, $\chi^2$),
whose accuracy is below 80\% (with 100\% coverage).
Among the word association models, the {\it dependency} models clearly outperform the {\it adjacency} models,
which is consistent with what other researchers previously have reported
\cite{lauer:1995:thesis,Lapata:Keller:04}.
Using patterns containing {\it Google}'s \* operator performs worse
than the dependency-based word association models:
the coverage is in the eighties (as opposed to 100\%),
and there is a 1-5\% absolute drop in accuracy.


The final bracketing decision is a majority vote combination of the predictions
of the models shown in bold in Table \ref{table:dataset:lauer}:
{\it $\chi^2$ adjacency}, {\it $\chi^2$ dependency}, {\it concatenation dependency},
{\it concatenation triples}, {\it genitive markers}, {\it abbreviations}, {\it paraphrases},
and {\it surface features}. This yields 90.52\% accuracy and 95.08\% coverage.
Defaulting the unassigned cases to left bracketing yields 89.34\% accuracy.


Table \ref{table:lauerSetComparisons} compares my results on {\it Lauer's dataset}
to the results of \namecite{lauer:1995:thesis}, \namecite{Lapata:Keller:04},
and \namecite{girju:2005:on:the:semantics}.
It is important to note that, while my results are {\it directly}
comparable to those of \namecite{lauer:1995:thesis},
the ones of \namecite{Lapata:Keller:04} are not
since they use half of {\it Lauer's dataset} for development (122 examples)
and the other half for testing\footnote{However, the differences are negligible;
their system achieves very similar results on the whole dataset (personal communication).}.
Following \namecite{lauer:1995:thesis}, I use the whole dataset for testing.
In addition, the Web-based experiments of \namecite{Lapata:Keller:04}
use the {\it AltaVista} search engine,
which no longer exists in its earlier form: it has been acquired by {\it Yahoo!}.

Table \ref{table:lauerSetComparisons}
also shows the results of \namecite{girju:2005:on:the:semantics},
who achieve 83.1\% accuracy, but use a {\it supervised} algorithm which
targets bracketing {\it in context}. They further ``shuffle''
{\it Lauer's dataset}, mixing it with additional data,
which makes it hard to compare directly to their results.
More details can be found in section \ref{sec:related:NC:syntax} above.

\begin{table}
\begin{center}
\begin{tabular}{|l|c|}
\multicolumn{1}{l}{\bf Model} & \multicolumn{1}{c}{\bf Accuracy (\%)}\\
  \hline
baseline ({\it `left'}) & 66.80$\pm$6.13\\
\cite{lauer:1995:thesis} adjacency & 68.90$\pm$6.07\\
\cite{lauer:1995:thesis} dependency & 77.50$\pm$5.65\\
{\it My $\chi^2$ dependency} & {\it 79.92$\pm$5.47}\\
\cite{lauer:1995:thesis} tuned & 80.70$\pm$5.41\\
{\bf \emph{My majority vote $\rightarrow$ left}} & {\bf \emph{89.34$\pm$4.50}}\\
  \hline
\cite{Lapata:Keller:04}: baseline (122 examples) & 63.93$\pm$8.83\\
\cite{Lapata:Keller:04}: best {\it BNC} (122 examples) & 68.03$\pm$8.72\\
\cite{Lapata:Keller:04}: best {\it AltaVista} (122 examples) & 78.68$\pm$8.08\\
  \hline
    $^{\star}$\cite{girju:2005:on:the:semantics}: best C5.0 (shuffled dataset) & 83.10$\pm$5.20\\
  \hline
\end{tabular}
\caption{{\bf Comparison to other unsupervised results on \emph{Lauer's dataset}.}
The results of Lapata \& Keller (2004) are on half of {\it Lauer's dataset}:
note the different baseline.
The model of Girju {\it et al.} (2005) is supervised;
it also mixes \emph{Lauer's dataset} with additional data.
See section \ref{sec:related:NC:syntax} for more details.}
\label{table:lauerSetComparisons}
\end{center}
\end{table}

\begin{table}
\begin{center}
\begin{tabular}{|l|ccc|cr|}
\multicolumn{1}{l}{\bf Model} & {\bf Correct} & {\bf Wrong} & \multicolumn{1}{c}{\bf N/A} & {\bf Accuracy} & \multicolumn{1}{c}{\bf Cover.}\\
  \hline
\# adjacency & {\small 374} & {\small 56} & {\small 0} & {\small 86.98$\pm$3.51} & {\small 100.00} \\
$\mathrm{Pr}$ adjacency & {\small 353} & {\small 77} & {\small 0} & {\small 82.09$\pm$3.90} & {\small 100.00} \\
$\mathrm{PMI}$ adjacency & {\small 372} & {\small 58} & {\small 0} & {\small 86.51$\pm$3.55} & {\small 100.00} \\
{\bf $\mathbf{\chi^2}$ adjacency} & {\small 379} & {\small 51} & {\small 0} & {\small {\bf 88.14$\pm$3.40}} & {\small 100.00} \\
\hline
\# dependency & {\small 374} & {\small 56} & {\small 0} & {\small 86.98$\pm$3.51} & {\small 100.00} \\
$\mathrm{Pr}$ dependency (= PMI dep.) & {\small 369} & {\small 61} & {\small 0} & {\small 85.81$\pm$3.62} & {\small 100.00} \\
{\bf $\mathbf{\chi^2}$ dependency} & {\small 380} & {\small 50} & {\small 0} & {\small {\bf 88.37$\pm$3.38}} & {\small 100.00} \\
\hline
\# adjacency (*) & {\small 373} & {\small 57} & {\small 0} & {\small 86.74$\pm$3.53} & {\small 100.00}\\
\# adjacency (**) & {\small 358} & {\small 72} & {\small 0} & {\small 83.26$\pm$3.82} & {\small 100.00}\\
\# adjacency (***) & {\small 334} & {\small 88} & {\small 8} & {\small 79.15$\pm$4.13} & {\small 98.14}\\
\hline
\# adjacency (*, rev.) & {\small 370} & {\small 59} & {\small 1} & {\small 86.25$\pm$3.58} & {\small 99.77}\\
\# adjacency (**, rev.) & {\small 367} & {\small 62} & {\small 1} & {\small 85.55$\pm$3.64} & {\small 99.77}\\
\# adjacency (***, rev.) & {\small 351} & {\small 79} & {\small 0} & {\small 81.63$\pm$3.93} & {\small 100.00}\\
\hline
Concatenation adjacency & {\small 370} & {\small 47} & {\small 13} & {\small 88.73$\pm$3.40} & {\small 96.98}\\
{\bf Concatenation dependency} & {\small 366} & {\small 43} & {\small 21} & {\small {\bf 89.49$\pm$3.35}} & {\small 95.12}\\
{\bf Concatenation triple} & {\small 238} & {\small 37} & {\small 155} & {\small {\bf 86.55$\pm$4.54}} & {\small 63.95}\\
\hline
Inflection variability & {\small 198} & {\small 49} & {\small 183} & {\small 80.16$\pm$5.42} & {\small 57.44}\\
Swap first two words & {\small 90} & {\small 18} & {\small 322} & {\small 83.33$\pm$8.15} & {\small 25.12}\\
Reorder & {\small 320} & {\small 78} & {\small 32} & {\small 80.40$\pm$4.18} & {\small 92.56}\\
{\bf Abbreviations} & {\small 133} & {\small 23} & {\small 274} & {\small {\bf 85.25$\pm$6.41}} & {\small 36.27}\\
{\bf Genitive markers} & {\small 48} & {\small 7} & {\small 375} & {\small {\bf 87.27$\pm$11.29}} & {\small 12.79}\\
\hline
{\bf Paraphrases} & {\small 383} & {\small 44} & {\small 3} & {\small {\bf 89.70$\pm$3.25}} & {\small 99.30}\\
{\bf Surface features (sum)} & {\small 382} & {\small 48} & {\small 0} & {\small {\bf 88.84$\pm$3.33}} & {\small 100.00}\\
\hline
Majority vote & {\small 403} & {\small 17} & {\small 10} & {\small 95.95$\pm$2.34} & {\small 97.67}\\
\emph{Majority vote $\rightarrow$ `right'} & {\small \emph{410}} & {\small \emph{20}} & {\small \emph{0}} & {\small {\bf \emph{95.35$\pm$2.42}}} & {\small {\bf \emph{100.00}}}\\
\hline
{\bf Baseline} (choose `left') & {\small 361} & {\small 69} & {\small 0} & {\small 83.95$\pm$3.77} & {\small 100.00}\\
\hline
\end{tabular}
\caption{{\bf Bracketing results on the \emph{Biomedical dataset} using the Web.}
For each model, the number of correctly classified,
wrongly classified, and non-classified examples is shown,
followed by the accuracy (in \%) and the coverage (\% examples for which the model makes prediction).}
\label{table:bio}
\end{center}
\end{table}

\subsection{Results for the \emph{Biomedical Dataset}}

The results for the {\it Biomedical dataset} are shown in Table \ref{table:bio}.
This dataset has a high left-bracketing baseline of almost 84\% accuracy.
The {\it adjacency} and the {\it dependency} word association models
yield very similar accuracy; in both cases $\chi^2$ is a clear winner with over 88\%,
outperforming the other word association models by 1.5-6\%.
Further, the {\it frequency} models outperform the {\it probability} models
due to the abundance of words with unreliable Web frequency estimates:
single-letter (e.g., {\it \underline{T} cell}, {\it vitamin \underline{D}}),
Roman digits (e.g., {\it ii}, {\it iii}), Greek letters (e.g., {\it alpha}, {\it beta}), etc.
These are used by the {\it probability} model, but not by the {\it frequency} model,
which uses bigrams only. While being problematic for most models,
these words work well with {\it concatenation dependency} (89.49\% accuracy, 95.12\% coverage),
e.g., {\it T cell} often can be found written as {\it Tcell}.

As before, the different models are combined in a majority vote,
using the same voting models as for {\it Lauer's dataset} above,
which yields 95.95\% accuracy and 97.67\% coverage.
Defaulting the unassigned cases to right bracketing yields 95.35\% accuracy overall.

\begin{table}
\begin{center}
\begin{tabular}{|lccc|}
\multicolumn{1}{l}{\bf Example} & \multicolumn{1}{l}{\bf Predicts} & \multicolumn{1}{l}{\bf Accuracy (\%)} & \multicolumn{1}{l}{\bf Coverage (\%)} \\
\hline
{\bf brain-stem cells} & left & {\bf 88.22} & {\bf 92.79}\\
{\bf brain stem's cells} & left & {\bf 91.43} & {\bf 16.28}\\
(brain stem) cells & left & 96.55 & 6.74\\
brain stem (cells) & left & 100.00 & 1.63\\
brain stem, cells & left & 96.13 & 42.09\\
brain stem: cells & left & 97.53 & 18.84\\
brain stem cells-death & left & 80.69 & 60.23\\
brain stem cells/tissues & left & 83.59 & 45.35\\
{\bf brain stem Cells} & left & {\bf 90.32} & {\bf 36.04}\\
brain stem/cells & left & 100.00 & 7.21\\
brain. stem cells & left & 97.58 & 38.37\\
\hline
\textbf{brain stem-cells} & right & \textbf{25.35} & \textbf{50.47}\\
\textbf{brain's stem cells} & right & \textbf{55.88} & {\bf 7.90}\\
(brain) stem cells & right & 46.67 & 3.49\\
brain (stem cells) & right & 0.00 & 0.23\\
brain, stem cells & right & 54.84 & 14.42\\
brain: stem cells & right & 44.44 & 6.28\\
rat-brain stem cells & right & 17.97 & 68.60\\
neural/brain stem cells & right & 16.36 & 51.16\\
brain Stem cells & right & 24.69 & 18.84\\
brain/stem cells & right & 53.33 & 3.49\\
brain stem. cells & right & 39.34 & 14.19\\
\hline
\end{tabular}
\caption{{\bf Surface features analysis (\%s)}, run over the {\it Biomedical dataset}.}
\label{table:surface}
\end{center}
\end{table}

Finally, Table \ref{table:surface} shows the performance of the surface features
on the {\it Biomedical dataset}. We can see that most of
the assumed right-predicting surface features actually correlate better with a
left bracketing.  Overall, the surface features are very good
at predicting left bracketing, but are unreliable for the right-bracketed examples
due to scope ambiguity 
e.g., in ``{\it brain stem-cell}'', {\it cell} could attach to {\it stem} only,
in which case it would correctly predict a right bracketing;
however, it could also target the compound {\it brain stem}, in case of left-bracketed compound.

\subsection{Confidence Intervals}
\label{sec:conf:interval}


Table \ref{table:lauer} shows that 
the accuracy for the {\it $\chi^2$ dependency} model is 79.92\%;
however it depends on the particular testing dataset
used in the evaluation: {\it Lauer's dataset}.
What if I repeated the experiments with another
dataset of 244 examples extracted in the same way from the same data source,
i.e. from {\it Grolier's encyclopedia}?
Probably, accuracy values like 79.6\% or 81.3\% should not be surprising,
while 50\% would be extremely rare. The question then becomes:
Which values are to be expected and which ones should be considered particularly unusual?

The most popular statistical answer to this question is to provide a whole interval
of likely values, rather than a single estimate,
and the most widely used frequentist approach is to construct a {\it confidence interval},
which is an interval of values calculated in a way
so that if the experiment is repeated with multiple samples (here datasets)
from the same population,
the calculated confidence interval (which would be different for each sample)
would contain the true population parameter
(here the accuracy for {\it $\chi^2$ dependency},
if all possible three-word noun compounds from {\it Grolier's encyclopedia}
were to be tested)
for a high proportion of these samples.

More formally, given an unobservable parameter $\theta$,
the random interval $(U, V)$ is called a {\it confidence interval for $\theta$},
if $U$ and $V$ are observable random variables
whose probability distribution depends on $\theta$,
and $\mathrm{Pr}(U < \theta < V|\theta) = x$.
The number $x$ ($0 \leq x \leq 1$) is called {\it confidence level}
and is usually given in per cents: the value of 95\% is typically used,
although 90\%, 99\% and 99.9\% are common as well,
with a higher confidence level corresponding to a wider interval.

For example, given a sample mean $\hat{\mu}$,
the confidence interval for the true mean $\mu$
of a normally-distributed random variable whose
population's standard deviation $\sigma_{\mu}$ is known can be calculated as follows:
\begin{equation}
    \hat{\mu} \pm z_{(1-\alpha/2)} \sigma_{\mu}
\end{equation}

\noindent where $z_{(1-\alpha/2)}$ is the $(1-\alpha/2)$ percentile of the standard normal distribution.
For a 95\% confidence level, we set $\alpha=0.05$.

In my case, I need to calculate a confidence interval for the accuracy,
which is a {\it confidence interval for a proportion}:
the proportion of correct examples out of all classified examples.
Since there is a fixed number of examples (244 for {\it Lauer's dataset}),
with two possible outcomes for each one (correct and wrong),
assuming statistically independent trials
and the same probability of success for each trial,
a {\it binomial distribution} can be assumed.
Using the {\it central limit theorem},
this binomial distribution can approximated with a normal distribution,
which yields the {\it Wald interval} \cite{Brown:Cai:DasGupta:2001:conf:interval}:

\begin{equation}\label{eq:wald}
    \hat{p} \pm z_{(1-\alpha/2)} \sqrt{\frac{\hat{p}(1-\hat{p})}{n}}
\end{equation}

\noindent where $n$ is the sample size, and $\hat{p}$ is an estimation for the proportion.

Here is an example:
According to Table \ref{table:lauer},
the {\it $\chi^2$ dependency} model makes
195 correct and 49 wrong classification decisions.
Therefore, $n=195+49=244$ and $\hat{p}=195/244=0.7992$.
A 95\% confidence level corresponds to $\alpha=0.05$, and
thus $z_{(1-\alpha/2)}$ is $z_{0.975}$, which is found
in a statistical table for the standard normal distribution to be 1.96.
Substituting these values in eq. \ref{eq:wald} yields $0.7992 \pm 0.05$.
Therefore, the 95\% confidence Wald interval
for {\it $\chi^2$ dependency} model's accuracy is [0.7492, 0.8492] or [74.92\%, 84.92\%].

While the Wald interval is simple to calculate,
there is a more accurate alternative, the {\it Wilson interval}
\cite{Wilson:1927:conf:interval}:

\vspace{3pt}
\begin{equation}\label{eq:wilson}
    \frac{\hat{p} + \frac{1}{2n}z^2_{(1-\alpha/2)} \pm z_{(1-\alpha/2)} \sqrt{\frac{1}{n}[\hat{p}(1-\hat{p}) + \frac{1}{4n}z^2_{(1-\alpha/2)}]}}    {1 + \frac{1}{n}z^2_{(1-\alpha/2)}}
\end{equation}
\vspace{3pt}

For the {\it $\chi^2$ dependency} model, the above formula yields
$79.92\% \pm 5.47\%$, which is the interval shown in Table \ref{table:lauer}.
Because of its higher accuracy \cite{Agresti:Coull:1998:conf:interval},
I use the Wilson interval for all calculations of confidence intervals for proportions
in this chapter, as well as elsewhere in the thesis, unless otherwise stated.

\subsection{Statistical Significance}
\label{sec:stat:significance}

Table \ref{table:lauerSetComparisons} shows that
the accuracy for my {\it $\chi^2$ dependency} model is 79.92\%,
while for \quotecite{lauer:1995:thesis} dependency model it is 77.5\%.
Apparently, the former model is better than the latter one,
but maybe this is due to chance alone?
What about the almost 9\% difference in accuracy between
my ``{\it Majority vote $\rightarrow$ left}'' (89.34\% accuracy)
and \quotecite{lauer:1995:thesis} tuned model (80.7\% accuracy):
could it have occurred by chance too?

These are questions about {\it statistical significance}:
an outcome is considered to be {\it statistically significant},
if it is unlikely to have occurred due to chance alone. The statistical significance of a result
is characterized by its $p$-value, which is the probability
of obtaining an outcome at least as extreme as the observed one.
A test for statistical significance calculates a $p$-value (under some assumptions)
and then compares it to a pre-specified {\it significance level} $\alpha$ (e.g., 0.05):
the result is considered statistically significant, if the $p$-value is less than $\alpha$.

\begin{table}
\begin{center}
\begin{tabular}{ccc}
     & {\bf Correct} & {\bf Wrong} \\
  \hline
   {\bf Model 1} & $A$ & $B$ \\
   {\bf Model 2} & $C$ & $D$ \\
  \hline
\end{tabular}
\caption{{\bf The contingency table:}
        used in a $\chi^2$ test for testing if the difference
        in performance between two models is statistically significant.}
\label{table:chi2:example}
\end{center}
\end{table}

There are various statistical tests depending on the kinds of observed events
and on the assumptions about them that could be made.
In particular, the most widely used test for
checking whether the difference in performance
between two models is statistically significant
is the {\it Pearson's $\chi^2$ test}.\footnote{Some other tests
are suitable for the purpose as well, e.g., the {\it Fisher's exact test}.}
The test makes use of a {\it $2 \times 2$ contingency table}
like the one shown in Table \ref{table:chi2:example}:
the rows are associated the models and the columns
with the number of correctly/wrongly classified examples.
It tries to determine whether the probability of an example
being classified correctly or wrongly depends on the model used,
i.e. it checks whether
the null hypothesis, that the columns of the table are independent of the rows,
can be rejected with a significance level $\alpha$.

Let us consider for example, how the {\it Pearson's $\chi^2$ test}
can be applied to comparing the {\it $\chi^2$ dependency} model (Model 1)
and the \quotecite{lauer:1995:thesis} dependency model (Model 2),
whose accuracies are 79.92\% and 77.5\%, respectively.
Since there are 244 testing examples in total,
we have $A=189$, $B=55$, $C=195$, $D=49$,
as shown in Table \ref{table:chi2:particular:one}.
Let $N$ be the total sum of all table entries, i.e. $N=A+B+C+D=488$.
Then the marginal probability of an example in this table
being classified correctly regardless of the model is:
\begin{equation}
    \mathrm{Pr}(correct) = \frac{A+C}{N} = \frac{189+195}{488} = 0.7869
\end{equation}
\vspace{6pt}

Similarly, the marginal probability of an example in this table
being classified by Model 1 is:
\begin{equation}
    \mathrm{Pr}(Model1) = \frac{A+B}{N} = \frac{189+55}{488} = 0.5
\end{equation}

\begin{table}
\begin{center}
\begin{tabular}{ccc}
     & {\bf Correct} & {\bf Wrong} \\
  \hline
   {\bf Model 1} & $A=189$ (192) & $B=55$ (52) \\
   {\bf Model 2} & $C=195$ (192) & $D=49$ (52) \\
  \hline
\end{tabular}
\caption{{\bf Sample contingency table:}
        the values in parentheses are the expected values
        under the null hypothesis that the rows and the columns are independent.}
\label{table:chi2:particular:one}
\end{center}
\end{table}

Under the null hypothesis that the rows and the columns of the table are independent,
the probability of an example being classified correctly by Model 1 would be
the product of the above marginal probabilities:
\begin{equation}
    \mathrm{Pr}(correct, Model 1) = \mathrm{Pr}(correct) \times \mathrm{Pr}(Model1)
\end{equation}

Therefore, the expected number of examples being classified correctly by Model 1 would be:
\begin{equation}
    E(correct, Model 1) = N \times \mathrm{Pr}(correct, Model 1) = 192
\end{equation}

The expected values for the remaining three table entries can be calculated in a similar manner:
\begin{equation}
    E(correct, Model 2) = N \times \mathrm{Pr}(correct, Model 2) = 192
\end{equation}
\begin{equation}
    E(wrong, Model 1) = N \times \mathrm{Pr}(correct, Model 2) = 52
\end{equation}
\begin{equation}
    E(wrong, Model 2) = N \times \mathrm{Pr}(correct, Model 2) = 52
\end{equation}

These expected values for the table entries are shown
in parentheses in Table \ref{table:chi2:particular:one}.
The $\chi^2$ score is calculated using the following formula:
\begin{equation}\label{eq:chi2:score}
    \chi^2 = \sum_{i}{\frac{(O_i-E_i)^2}{E_i}}
\end{equation}

\noindent where $O_i$ are the observed values (not in parentheses in Table \ref{table:chi2:particular:one})
and $E_i$ are the expected values (shown in parentheses in Table \ref{table:chi2:particular:one}).
The summation is over all table entries.

Substituting the values in Table \ref{table:chi2:particular:one}
in the above formula (\ref{eq:chi2:score}), we obtain:
\begin{equation}
    \chi^2 = \frac{(189-192)^2}{192} + \frac{(55-52)^2}{52} + \frac{(195-192)^2}{192} + \frac{(49-52)^2}{52}
\end{equation}

I then consult a table of the $\chi^2$ distribution with 1 degree of freedom\footnote{One
degree of freedom, since there are two outcomes, {\it correct} and {\it wrong}, whose sum is fixed.},
and I find that the probability of observing this or a more extreme difference,
under the null hypothesis is approximately 0.5072, i.e. the $p$-value equals 0.5072.
Since 0.5072 is quite a big probability, much bigger than $\alpha=0.05$,
the null hypothesis cannot be rejected.

As I already mentioned in section \ref{sec:chi2:score},
there is more direct way to calculate the $\chi^2$ score
from a $2 \times 2$ table (repeated here for completeness):
\begin{equation}
\chi^2 = \frac{N (AD-BC)^2}{(A+C) (B+D) (A+B) (C+D)}
\end{equation}

What about \quotecite{lauer:1995:thesis} tuned model
and my ``{\it Majority vote $\rightarrow$ left}'' model?
The accuracies are 80.7\% and 89.34\%, respectively,
and since there are a total of 244 examples then $A=197$, $B=47$, $C=218$, $D=26$.
The $\chi^2$ score is 7.104, and the corresponding $p$-value equals 0.0077,
which means that the null hypothesis can be rejected
with the standard significance level $\alpha=0.05$, and even with the much lower $\alpha=0.01$.
The difference between these models is considered very statistically significant.

Is the difference between my ``{\it Majority vote $\rightarrow$ left}''
and the best previously published result,
the supervised C5.0 model of \namecite{girju:2005:on:the:semantics}
significantly different?
The corresponding accuracies are 89.34\% and 83.1\%, respectively,
and thus $A=218$, $B=26$, $C=203$, and $D=41$.
The $\chi^2$ score equals 3.893, which corresponds to a $p$-value of 0.0485,
and therefore, the difference is statistically significant
with a significance level $\alpha=0.05$.
Thus, I can conclude that my result is statistically significantly better
(with a significance level $\alpha=0.05$),
on {\it Lauer's dataset} or a variation of it,
than any previously proposed algorithm: both supervised and unsupervised.

Note that in the last calculation the evaluation datasets are different (but overlapping):
the dataset of \namecite{girju:2005:on:the:semantics} still contains 244 examples,
but some of them are not from the original dataset (see section \ref{sec:related:NC:syntax}).
This is not a problem: the $\chi^2$ test is robust against such differences,
provided that the distribution does not change.
The test is also robust against different numbers of examples,
which allows me to compare my results or those of \namecite{lauer:1995:thesis},
which use all of {\it Lauer's dataset},
with the ones of \namecite{Lapata:Keller:04}, who only use half of it.

Finally, there is a connection between confidence intervals and statistical significance:
a confidence interval around a particular value includes all other values that
are not statistically significantly different from it.
For example, as I have calculated in section \ref{sec:conf:interval} above,
the 95\% confidence interval around the 79.92\% value for the accuracy
of the {\it $\chi^2$ dependency model} is [74.92\%, 84.92\%].
Therefore, the 80.7\% accuracy for \quotecite{lauer:1995:thesis} tuned model
is not statistically significantly different from 79.92\% since it is in the interval,
but the value of 89.34\% for my combined model is since it is outside of the interval.

\section{Using MEDLINE instead of the Web}

For comparison purposes, I experimented with the {\it Biomedical dataset}
using a domain-specific text corpus
with suitable linguistic annotations instead of the Web.
I use the Layered Query Language and architecture,
described in section \ref{sec:dataset:bio:extract} in appendix \ref{app:dataset:bio},
in order to acquire $n$-gram and paraphrase frequency statistics.

My corpus consists of about 1.4 million MEDLINE abstracts, each one being about 300 words long
on the average, which means about 420 million indexed words in total.
For comparison, {\it Google} indexes about eight billion pages;
if we assume that each one contains about 500 words on the average,
this yields about four trillion indexed words, which is about a million times
bigger than my corpus. Still, the subset of MEDLINE I use
is about four times bigger than the 100 million word {\it BNC}
used by \namecite{Lapata:Keller:04}. It is also more than fifty times bigger than
the eight million word {\it Grolier's encyclopedia} used by \namecite{lauer:1995:thesis}.

\begin{table}
\begin{center}
\begin{tabular}{|l|ccc|cr|}
\multicolumn{1}{l}{\bf Model} & {\bf Correct} & {\bf Wrong} & \multicolumn{1}{c}{\bf N/A} & {\bf Accuracy} & \multicolumn{1}{c}{\bf Cover.}\\
 \hline
\# adjacency & {\small 196} & {\small 36} & {\small 0} & {\small 84.48$\pm$5.22} & {\small 100.00} \\
Pr adjacency & {\small 173} & {\small 59} & {\small 0} & {\small 74.57$\pm$5.97} & {\small 100.00} \\
$\chi^2$ adjacency & {\small 200} & {\small 32} & {\small 0} & {\small 86.21$\pm$5.03} & {\small 100.00} \\
  \hline
\# dependency & {\small 195} & {\small 37} & {\small 0} & {\small 84.05$\pm$5.26} & {\small 100.00} \\
Pr dependency & {\small 193} & {\small 39} & {\small 0} & {\small 83.19$\pm$5.34} & {\small 100.00} \\
$\chi^2$ dependency & {\small 196} & {\small 36} & {\small 0} & {\small 84.48$\pm$5.22} & {\small 100.00} \\
\hline
PrepPar & {\small 181} & {\small 13} & {\small 38} & {\small 93.30$\pm$4.42} & {\small 83.62}\\
PP+$\chi^2$adj+$\chi^2$dep & {\small 207} & {\small 13} & {\small 12} & {\small 94.09$\pm$3.94} & {\small 94.83}\\
PP+$\chi^2$adj+$\chi^2$dep$\rightarrow$right  & {\small 214} & {\small 18} & {\small 0} & {\small {\bf 92.24$\pm$4.17}} & {\small 100.00}\\
\hline
{\bf Baseline}  (choose left) & {\small 193} & {\small 39} & {\small 0} & {\small 83.19$\pm$5.34} & {\small 100.00}\\
\hline
\end{tabular}
\caption{{\bf Bracketing results on the \emph{Biomedical dataset} using 1.4M MEDLINE abstracts.}
For each model, the number of correctly classified,
wrongly classified, and non-classified examples is shown,
followed by the accuracy (in \%) and the coverage (in \%).}
\label{table:database:eval}
\end{center}
\end{table}

The queries I used to collect $n$-gram and paraphrases counts
are described in section \ref{app:dataset:bio:lql} in appendix \ref{app:dataset:bio}.
The results are shown in Table \ref{table:database:eval}.
In addition to probabilities (Pr), I also use
counts (\#) and $\chi^2$ (with the dependency and the adjacency models).
The prepositional paraphrases are much more accurate: 93.3\%
(with 83.62\% coverage). By combining the paraphrases with the $\chi^2$
models in a majority vote, and by assigning the undecided cases to
right-bracketing, I achieve 92.24\% accuracy, which is slightly worse than 95.35\%
I achieved using the Web.
This difference is not statistically
significant\footnote{Note however that here I experiment with 232 of the 430 examples.},
which suggests that in some cases a big domain-specific corpus with suitable linguistic
annotations could be a possible alternative of the Web.
This is not true, however, for general domain compounds:
for example, my subset of MEDLINE can provide prepositional paraphrases
for only 23 of the 244 examples in {\it Lauer's dataset} (i.e. for less than 10\%),
and for 12 of them the predictions are wrong (i.e., the accuracy is below 50\%).

\section{Conclusion and Future Work}

I have described a novel, highly accurate lightly supervised approach to
noun compound bracketing which makes use of novel surface features
and paraphrases extracted from the Web and achieves a statistically significant
improvement over the previous state-of-the-art results for {\it Lauer's dataset}.
The proposed approach is more robust than the one proposed by \namecite{lauer:1995:thesis}
and more accurate than that of \namecite{Lapata:Keller:04}.
It does not require labeled training data, lexicons, ontologies,
thesauri, or parsers, which makes it promising for a wide range of other NLP tasks.

A simplification of the method has been used by \namecite{vadas:curran:2007:NP:structure:treebank}
in order to augment the NPs in the {\it Penn Treebank} with internal structure.
Some of the proposed features have been found useful for query segmentation.
by \namecite{bergsma:wang:2007:query:segm}.

An important direction for future work is in reducing the number
of queries to the search engine. One solution would be to perform
a careful analysis and to exclude the most expensive patterns
and those with a minor positive contribution to the overall accuracy.
Using {\it Google}'s {\it Web 1T 5-gram} dataset
of 5-gram that appear on the Web at least 40 times
and their corresponding frequencies is another promising direction to go.
However, it might be of limited use since many of my patterns
are longer than 5-grams and/or appear less than 40 times on the Web.

\chapter{Noun Compounds' Semantics and Relational Similarity}
\label{chapter:NC:semantics}

In this chapter, I present a novel, simple, unsupervised method
for characterizing the semantic relations that hold between nouns in noun-noun
compounds. The main idea is to look for {\it predicates} that make
explicit the hidden relations between the nouns.  This is
accomplished by writing Web search engine queries that restate the
noun compound as a relative clause containing a wildcard character
to be filled in with a verb.  A comparison to results from the
literature and to human-generated verb paraphrases
suggests this is a promising approach.
Using these verbs as features in classifiers,
I demonstrate state-of-the-art results on various relational similarity problems:
mapping noun-modifier pairs to abstract relations like \texttt{TIME}, \texttt{LOCATION} and \texttt{CONTAINER},
classifying relation between nominals, and solving SAT verbal analogy problems.
A preliminary version of some of these ideas appeared in \cite{Nakov:Hearst:2006:using:verbs}
and \cite{Nakov:Hearst:2007:WMT}.

\section{Introduction}

%

While currently there is no consensus as to what relations
can hold between nouns in a noun compound, most proposals make use
of small sets of abstract relations, typically less than fifty.
Some researchers, however, have proposed that an unlimited number of relations
is needed \cite{downing:1977:nc:sem}. I hold a similar position.

Many algorithms that perform semantic interpretation place heavy
reliance on the appearance of verbs, since they are the predicates
which act as the backbone of the assertion being made.  Noun
compounds are terse elisions of the predicate; their structure
assumes that the reader knows enough about the constituent nouns and
about the world at large to be able to infer what the relationship
between the words is.  Here I propose to try to uncover the
relationship between the noun pairs by, in essence,  rewriting or
paraphrasing the noun compound in such a way as to be able to
determine the predicate(s) holding between the nouns. Therefore,
I represent noun compounds semantics in terms of verbs, rather than a fixed number of
abstract predicates \cite{levi:1978} (e.g., \texttt{HAVE}, \texttt{MAKE}, \texttt{USE}),
relations \cite{girju:2005:on:the:semantics} (e.g., \texttt{LOCATION}, \texttt{INSTRUMENT}, \texttt{AGENT}),
or prepositions \cite{lauer:1995:thesis} (e.g., \texttt{OF}, \texttt{FOR}, \texttt{IN}),
as is traditional in the literature.
The idea is similar to the approach of \namecite{finin:1980:nc:sem:thesis},
who characterizes the relation in a noun-noun compound using
an inventory of all possible verbs that can link the noun constituents, e.g.,
{\it salt water} is interpreted using relations like {\it dissolved in}.


In the proposed approach, I pose paraphrases for a given noun compound by
rewriting it as a phrase that contains a wildcard where a verb
would go.  For example, I rewrite {\it neck vein} as \texttt{"vein
that * neck"}, I send this as a query to a Web search engine, and then
I parse the resulting snippets in order to find the verbs that appear in the
place of the wildcard.  For example, the most frequent verbs
(+ prepositions) I find for {\it neck vein} are: {\it emerge from}, {\it
pass through}, {\it be found in}, {\it be terminated at}, {\it be in},
{\it run from}, {\it descend in}.


\section{Related Work}

The syntax and semantics of noun compounds are active areas of research,
which are part of the broader research on multi-word expressions (MWEs):
there have been workshops on MWEs as part of the annual meetings of
the Association for Computational Linguistics (ACL) in 2003, 2004, 2006 and 2007,
and during the meeting of the European chapter of ACL (EACL) in 2006.
In 2007, the SemEval workshop on Semantic Evaluations (formerly {\it Senseval}),
co-located with the annual meeting of the ACL,
had a specialized task on {\it Classification of Semantic Relations between Nominals}
\cite{Girju:al:2007:semeval}.
Finally, there was a special issue on MWEs of the
{\it Journal of Computer Speech and Language} in 2005 \cite{Villavicencio:all:CSL:MWEs},
and there are two upcoming special issues of the
{\it International Journal of Language Resources and Evaluation} in 2008:
one on {\it Multi-word Expressions}, and another one on
`{\it Computational Semantic Analysis of Language: SemEval-2007 and Beyond}'.

\namecite{lauer:1995:thesis}
defines the semantic relation identification problem as one of predicting which
among the following eight prepositions is most likely to be associated with the
noun compound when rewritten:
{\it of}, {\it for}, {\it in}, {\it at},
{\it on}, {\it from}, {\it with} and {\it about} (see Table \ref{table:Lauer:sem:prep}).
This approach can be problematic since the semantics of the prepositions is vague,
e.g., {\it in}, {\it on}, and {\it at}, all can refer to both \texttt{LOCATION} and \texttt{TIME}.
Lauer builds a corpus-based model for predicting the correct preposition,
and achieves 40\% accuracy.
\namecite{Lapata:Keller:05:Web:based:Models} improve on these results
the Web to estimate trigram frequencies for ($noun_1$, $prep$, $noun_2$), achieving 55.71\% accuracy.

\namecite{Rosario:Hearst:2001:nc:sem} use their own dataset
of 18 abstract semantic classes and show that a discriminative
classifier can work quite well at assigning relations from a
pre-defined set if training data is supplied in a domain-specific
setting: 60\% accuracy.

\namecite{Rosario:al:2002:descent}
report 90\% accuracy using a simple ``descent of hierarchy''
approach, which characterizes the relation between two nouns in a
bioscience noun-noun compound based on the semantic category each of
the constituent nouns belongs to.
See section \ref{sec:descent:hierarchy} for a detailed description.

\namecite{Girju:al:2004:nom:SVM} present an SVM-based approach for the
automatic classification of semantic relations in nominalized noun
phrases, where either the head or the modifier has been derived from a
verb. Their classification schema consists of 35 abstract semantic
relations and has been also used by \namecite{Moldovan:al:2004:sem:NP} for the
semantic classification of noun phrases in general,
including complex nominals, genitives and adjectival noun phrases.

\namecite{girju:2005:on:the:semantics} apply both classic (SVM and decision trees)
and novel supervised models (semantic scattering and iterative semantic
specialization), using {\it WordNet}, word sense disambiguation, and a set
of linguistic features.  They test their system against both Lauer's
eight prepositional paraphrases and their own set of 21 semantic relations,
achieving up to 54\% accuracy on the latter.

\namecite{girju:2007:ACLMain} uses cross-linguistic evidence from a
set of five Romance languages, from which NP features are extracted
and used in an SVM classifier, which yields 77.9\% accuracy when
{\it Europarl} is used, and 74.31\% with {\it CLUVI}.
Additional details on the features and on the process of their extraction
are provided in \cite{girju:2007:LAW}.

\namecite{Lapata:2002:nom} focuses on the disambiguation of nominalizations
-- a particular class of noun compounds whose head is derived
from a verb and whose modifier was an argument of that verb.
Using partial parsing, sophisticated smoothing and contextual
information, she achieves 86.1\% accuracy (baseline 61.5\% accuracy) on the
binary decision task of whether the modifier used to be the subject
or the object of the nominalized verb, i.e. the head.

\namecite{kim:baldwin:2006:POS} try to characterize the
semantic relation in a noun-noun compound using the verbs
connecting the two nouns by comparing them to a predefined set of seed verbs,
using a memory-based classifier (TiMBL).
There are two problems with their approach: it is highly resource intensive
(uses {\it WordNet}, {\it CoreLex} and {\it Moby's thesaurus}),
and it is quite sensitive to the seed set of verbs: on a collection of 453
examples and 19 relations, they achieve 52.6\% accuracy with 84 seed
verbs, but only 46.66\% when only 57 seed verbs were used.

\namecite{Kim:Baldwin:2007:bootstrap}
describe a bootstrapping method for automatically tagging noun
compounds with their corresponding semantic relations.
Starting with 200 seed annotated noun compounds\,
they replaced one constituent of each noun compound with similar words that are derived
from synonyms, hypernyms and sister words,
achieving accuracy ranging between 64.72\% and 70.78\%.

\namecite{Kim:Baldwin:2007:wsd} apply word sense disambiguation techniques
to help supervised and unsupervised noun compound interpretation.

\namecite{OSeaghdha:Copestake:2007}
use grammatical relations as features with an SVM classifier
to characterize a noun-noun compound.

\namecite{Devereux:Costello:2006} propose
a vector-space model as representation model
for the relations used to interpret noun-noun compounds.
They use a fixed set of 19 head-modifier (H-M) relations like
{\it H causes M}, {\it M causes H}, {\it H for M}, {\it H by M}, {\it M is H}.
The meaning of a noun compound is hypothesized to be characterized by distribution
over these 19 dimensions, as opposed to be expressed by a single relation.
The model builds on the CARIN theory of conceptual combination \cite{Gagne:2002:carin}
and was evaluated by measuring people's reaction time.

\namecite{Turney:all:2003} describe 13 independent classifiers,
whose predictions are weighted and combined in order to solve SAT
verbal analogy problems. They assembled a dataset of 374 questions,
which were subsequently used by other researchers, e.g.,
\namecite{Veale:2004}, who achieved 43\% accuracy on the same problem,
using {\it WordNet} to calculate the similarity
between individual words involved in the target SAT analogy problem instance.

\namecite{Turney:Littman:2005:rel} introduce a vector space model (VSM),
and characterize the relation between two words, $X$ and $Y$, as a
fixed-length vector whose coordinates correspond to Web frequencies
for 128 phrases like ``$X$ for $Y$'', ``$Y$ for $X$'', etc., derived
from a fixed set of 64 joining terms (e.g. ``for'', ``such as'',
``not the'', ``is *'', etc.). These vectors are then used in a
nearest-neighbor classifier to solve SAT analogy problems,
achieving 47\% accuracy (random-guessing baseline is 20\%). In that
paper they have applied that approach to classifying noun-modifier
pairs into a fixed set of relations. Using a dataset of 600 examples
created by \namecite{Barker:Szpakowicz:1998:nc:sem}, with 30 target relations they
achieved an F-value of 26.5\% (random guessing: 3.3\%;
majority-class baseline: 8.17\%), and 43.2\% when these 30 relations
have been grouped into 5 course-grained relations (random guessing:
20\%; majority-class baseline: 43.33\%).

\namecite{Turney:2006:rel} presents an unsupervised algorithm
for mining the Web for patterns expressing implicit semantic
relations. For example, \texttt{CAUSE} (e.g., {\it cold virus}) is
best characterized by ``{\it Y * causes X}'',
and the best pattern for \texttt{TEMPORAL} (e.g., {\it morning frost})
is ``{\it Y in * early X}''.
This approach yields 50.2\% F-value with 5 classes.

\namecite{Turney:2005:LRA} introduces the latent relational
analysis (LRA), which extends the VSM model of
\namecite{Turney:Littman:2005:rel} by making use of automatically
generated synonyms, by automatically discovering useful patterns,
and by using a singular value decomposition in order to smooth the
frequencies. The actual algorithm is quite complex and consists of
12 steps, described in detail in \cite{Turney:2006:CL:rel:sim}. When applied
to the 374 SAT questions, it achieves the state-of-the-art accuracy
of 56\%. On the \namecite{Barker:Szpakowicz:1998:nc:sem} dataset, the
achieves an accuracy of 39.8\% with 30 classes, and 58\% with 5
classes.

\namecite{nulty:2007:SRW} uses a vector-space model representation
where the vector coordinates are a fixed set of 28 joining terms like {\it of},
{\it for}, {\it from}, {\it without}, {\it across}, etc.
The values of the coordinates are filled using Web $n$-gram frequencies.
Using an SVM, he achieves 50.1\% accuracy ona 20-fold cross-validation for the
5-class Barker\&Szpakowicz dataset.

Most other approaches to noun compound interpretation use
hand-coded rules for at least one component of the algorithm
\cite{finin:1980:nc:sem:thesis}, or rules combined with lexical resources
\cite{Vanderwende:1994:nc} (52\% accuracy, 13 relations).
\namecite{Barker:Szpakowicz:1998:nc:sem} make use of the identity of the two
nouns and a number of syntactic clues in a nearest-neighbor
classifier with 60-70\% accuracy.

\section{Using Verbs to Characterize Noun-Noun Relations}

Traditionally the semantics of
a noun compound have been represented as an abstract relation
drawn from a small closed set.
This is problematic for several reasons. First, it is unclear which is
the best set, and mapping between different sets has proven
challenging \cite{girju:2005:on:the:semantics}.
Second, being both abstract and limited, such sets capture only part of the semantics;
often multiple meanings are possible, and sometimes none of the
pre-defined meanings are suitable for a given example. Finally, it
is unclear how useful the proposed sets are, since researchers have
often fallen short of demonstrating practical uses.

I believe verbs have more expressive power and are better tailored
for the task of semantic representation: there is an infinite number of them 
and they can capture fine-grained aspects of the meaning. For example,
while {\it wrinkle treatment} and {\it migraine treatment} express
the same abstract relation \texttt{TREATMENT-FOR-DISEASE}, some
fine-grained differences can be shown by specific verbs, e.g., {\it
smooth} can paraphrase the former, but not the latter.

In many theories, verbs play an important role in the process of
noun compound derivation \cite{levi:1978}, and speakers frequently use them
in order to make the hidden relation overt.
This allows for simple extraction, but also for straightforward
uses of the verbs and paraphrases in NLP tasks like
machine translation, information retrieval, etc.

I further believe that a single verb often is not enough and that
the meaning is approximated better by a collection of verbs. For
example, while {\it malaria mosquito} can very well be characterized
as \texttt{CAUSE} (or {\it cause}), further aspects of the meaning,
can be captured by adding some additional verbs, e.g., {\it carry},
{\it spread}, {\it be responsible for},
{\it be infected with}, {\it transmit}, {\it pass on}, etc.


\section{Method}
\label{sec:that:verbs:method}

In a typical noun-noun compound ``$noun_1$ $noun_2$'', $noun_2$ is
the head and $noun_1$ is a modifier, attributing a property to it.
The main idea of the proposed method is to preserve
the head-modifier relation by substituting
the pre-modifier $noun_1$ with a suitable post-modifying relative
clause, e.g., ``{\it tear gas}'' can be transformed into
``{\it gas that causes tears}'', ``{\it gas that brings tears}'',
``{\it gas which produces tears}'', etc.
Using all possible inflections of $noun_1$ and $noun_2$  from {\it WordNet},
I issue exact phrase {\it Google} queries of the following type:

\begin{center}
    \texttt{"noun2 THAT * noun1"}
\end{center}

\noindent where \texttt{THAT} is one of the following complementizers:
{\it that}, {\it which} or {\it who}.
The {\it Google} \texttt{*} operator stands for a one-word wildcard substitution;
I issue queries with up to eight stars.

I collect the text snippets (summaries) from the search results pages
(up to 1,000 per query) and I only keep the ones for which the
sequence of words following \texttt{noun1} is non-empty and contains
at least one non-noun, thus ensuring the snippet includes the entire
noun phrase.  To help POS tagging and shallow parsing the snippet,
I further substitute the part before \texttt{noun2} by the fixed
phrase ``{\it We look at the}''. I then perform POS tagging
\cite{Toutanova:Manning:2000:pos} and
shallow parsing\footnote{OpenNLP tools: \texttt{http://opennlp.sourceforge.net}},
and I extract all verb forms, and the following preposition, if any,
between \texttt{THAT} and \texttt{noun1}.
I allow for adjectives and participles to fall
between the verb and the preposition, but not nouns; I ignore the
modal verbs and the auxiliaries, but I retain the passive {\it be},
and I make sure there is exactly one verb phrase
(thus disallowing complex paraphrases like
``{\it gas that makes the eyes fill with tears}'').
Finally, I lemmatize the main verb using {\it WordNet}.

The proposed method is similar to previous paraphrase acquisition
approaches which look for similar endpoints and collect the
intervening material. \namecite{Lin:Pantel:2001:par} extract
paraphrases from dependency tree paths whose end points contain similar
sets of words by generalizing over these ends, e.g.,
for ``{\it X solves Y}'', they extract paraphrases like
``{\it X resolves Y}'', ``{\it Y is resolved by X}'',
``{\it X finds a solution to Y}'', ``{\it X tries to solve Y}'', etc.
The idea is extended by \namecite{Shinyama:al:2002:par}, who use named
entities of matching semantic classes as anchors, e.g.,
\texttt{LOCATION}, \texttt{ORGANIZATION}, etc.
Unlike these approaches, whose goal is to create summarizing paraphrases,
I look for verbs that can characterize noun compound semantics.

\section{Semantic Interpretation}

\subsection{Verb-based Vector-Space Model}

As an illustration of the method,
consider the paraphrasing verbs (corresponding frequencies are shown in parentheses)
it extracts for two noun-noun compounds with the same modifier and closely related synonymous heads:
{\it cancer physician} and {\it cancer doctor}.
Note the high proportion of shared verbs (underlined):

\begin{itemize}
    \item {\bf ``cancer doctor''}: {\it \underline{specialize in}}(12), {\it \underline{treat}}(12), {\it \underline{deal with}}(6), {\it \underline{believe}}(5), {\it \underline{cure}}(4), {\it attack}(4), {\it \underline{get}}(4), {\it understand}(3), {\it find}(2), {\it miss}(2), {\it remove}(2), {\it \underline{study}}(2), {\it know about}(2), {\it suspect}(2), {\it use}(2), {\it fight}(2), {\it deal}(2), {\it \underline{have}}(1), {\it suggest}(1), {\it track}(1), {\it \underline{diagnose}}(1), {\it recover from}(1), {\it specialize}(1), {\it rule out}(1), {\it meet}(1), {\it be afflicted with}(1), {\it study}(1), {\it look for}(1), {\it \underline{die from}}(1), {\it cut}(1), {\it mention}(1), {\it cure}(1), {\it \underline{die of}}(1), {\it say}(1), {\it \underline{develop}}(1), {\it contract}(1)

    \item {\bf ``cancer physician''}: {\it \underline{specialize in}}(11), {\it \underline{treat}}(7), {\it \underline{have}}(5), {\it \underline{diagnose}}(4), {\it \underline{deal with}}(4), {\it screen for}(4), {\it take out}(2), {\it \underline{cure}}(2), {\it \underline{die from}}(2), {\it experience}(2), {\it \underline{believe}}(2), {\it include}(2), {\it \underline{study}}(2), {\it misdiagnose}(1), {\it be treated for}(1), {\it work on}(1), {\it \underline{die of}}(1), {\it survive}(1), {\it \underline{get}}(1), {\it be mobilized against}(1), {\it \underline{develop}}(1)

\end{itemize}

\noindent Now consider the following four different kinds of treatments:

\begin{itemize}
    \item {\bf ``cancer treatment''}: {\it prevent}(8), {\it treat}(6), {\it
cause}(6), {\it irradiate}(4), {\it change}(3), {\it help
eliminate}(3), {\it be}(3), {\it die of}(3), {\it eliminate}(3),
{\it fight}(3), {\it have}(2), {\it ask for}(2), {\it be specific
for}(2), {\it decrease}(2), {\it put}(2), {\it help fight}(2), {\it
die from}(2), {\it keep}(2), {\it be for}(2), {\it contain}(2), {\it
destroy}(2), {\it heal}(2), {\it attack}(2), {\it work against}(2),
{\it be effective against}(2), {\it be allowed for}(1), {\it
stop}(1), {\it work on}(1), {\it reverse}(1), {\it characterise}(1),
{\it turn}(1), {\it control}(1), {\it see}(1), {\it identify}(1),
{\it be successful against}(1), {\it stifle}(1), {\it advance}(1),
{\it pinpoint}(1), {\it fight against}(1), {\it burrow into}(1),
{\it eradicate}(1), {\it be advocated for}(1), {\it counteract}(1),
{\it render}(1), {\it kill}(1), {\it go with}(1)

    \item {\bf ``migraine treatment''}: {\it prevent}(5), {\it be given
for}(3), {\it be}(3), {\it help prevent}(2), {\it help reduce}(2),
{\it benefit}(2), {\it relieve}(1)

    \item {\bf ``wrinkle treatment''}: {\it reduce}(5), {\it improve}(4), {\it
make}(4), {\it smooth}(3), {\it remove}(3), {\it be on}(3), {\it tackle}(3),
{\it work perfect on}(3), {\it help smooth}(2), {\it be
super on}(2), {\it help reduce}(2), {\it fight}(2), {\it target}(2),
{\it contrast}(2), {\it smooth out}(2), {\it combat}(1), {\it
correct}(1), {\it soften}(1), {\it reverse}(1), {\it resist}(1),
{\it address}(1), {\it eliminate}(1), {\it be}(1)

    \item {\bf ``herb treatment''}: {\it contain}(19), {\it use}(8), {\it be
concentrated with}(6), {\it consist of}(4), {\it be composed of}(3),
{\it include}(3), {\it range from}(2), {\it incorporate}(1), {\it
feature}(1), {\it combine}(1), {\it utilize}(1), {\it employ}(1)
\end{itemize}

Table \ref{table:dynCompAnalysis} shows a subset of these verbs.
As expected, {\it herb treatment},
which is quite different from the other compounds,
shares no verbs with them: it {\it uses} and {\it contains} herb,
but does not {\it treat} it. Further, while migraine
and wrinkles cannot be {\it cured}, they can be {\it reduced}.
Migraines can also be {\it prevented}, and wrinkles can be {\it smoothed}.
Of course, these results are merely suggestive and
should not be taken as ground truth, especially the absence of
a verb. Still they seem to capture interesting fine-grained
semantic distinctions, which normally require deep knowledge of the
semantics of the two nouns and/or about the world.

The above examples suggest that paraphrasing verbs,
and the corresponding frequencies,
may be a good semantic representation from a computational linguistics
point of view, e.g., they can be used in a vector space model
in order to measure semantic similarity between noun-noun compounds.

I believe the paraphrasing verbs can be useful from
a traditional linguistic (e.g., lexical semantics) point of view as well;
I explore this idea below.

\subsection{Componential Analysis}

In lexical semantics,
componential analysis is often used to represent the meaning of a word
in terms of semantic primitives (features),
thus reducing the word's meaning to series of binary components
\cite{Katz:Fodor:1963,Jackendoff:1983,Saeed:2003}.
For example, {\it bachelor} is (i) human, (ii) male, and (iii) unmarried, which can be expressed
as [$+$HUMAN] [$+$MALE] [$-$MARRIED]. Similarly, {\it boy} can be analyzed
as [$+$ANIMATE] [$+$HUMAN] [$+$MALE] [$-$ADULT].
See Table \ref{table:compAnalysis} for more examples.

Componential analysis has been very successful in phonology, where the
sound system is limited and the contrast between different sounds is very important.
For example, /p/ is distinguished from /b/ by the role of
the vocal chords, and this distinction can be represented as a feature, e.g.,
/p/ is [$-$VOICED], while /b/ is [$+$VOICED].

In contemporary lexical semantics, componential analysis is considered
useful for making explicit important semantic relations like hyponymy, incompatibility, etc.,
but is criticized for the following reasons:
(1) it is unclear how the analysis decides on the particular features/components to include;
and (2) it cannot really capture the full meaning of a given word.

\subsection{Dynamic Componential Analysis}

\begin{table}
\begin{center}
\begin{tabular}{lc|c|c|c}
        & \verb!  !{\bf man} \verb!   !& \verb! ! {\bf woman} \verb! ! & \verb!   !{\bf boy}\verb!   ! & \verb!  !{\bf bull} \verb!  !\\
\hline
ANIMATE & $+$   & $+$   & $+$   & $+$\\
HUMAN   & $+$   & $+$   & $+$   & $-$\\
MALE    & $+$   & $-$   & $+$   & $+$\\
ADULT   & $+$   & $+$   & $-$   & $+$\\
\hline
\end{tabular}
\caption{\textbf{Example componential analysis:} {\it man}, {\it woman},
{\it boy} and {\it bull}.} \label{table:compAnalysis}
\end{center}
\end{table}

\begin{table}
\begin{center}
\begin{tabular}{lc|c|c|c}
        & {\bf cancer} & {\bf migraine} & {\bf wrinkle} & {\bf herb} \\
        & {\bf treatment} & {\bf treatment} & {\bf treatment} & {\bf treatment} \\
\hline
 {\it treat}   & $+$   & $+$   & $+$   & $-$\\
 {\it prevent} & $+$   & $+$   & $-$   & $-$\\
 {\it cure}    & $+$   & $-$   & $-$   & $-$\\
 {\it reduce}  & $-$   & $+$   & $+$   & $-$\\
 {\it smooth}  & $-$   & $-$   & $+$   & $-$\\
 {\it cause}   & $+$   & $-$   & $-$   & $-$\\
 {\it contain} & $-$   & $-$   & $-$   & $+$\\
 {\it use}     & $-$   & $-$   & $-$   & $+$\\
\hline
\end{tabular}
\caption{\textbf{Some verbs characterizing different kinds of treatments.}}
\label{table:dynCompAnalysis}
\end{center}
\end{table}

Given the similarity between Tables \ref{table:compAnalysis} and \ref{table:dynCompAnalysis},
I propose to analyze the semantics of the relations that hold between the nouns in a noun-noun compound
using a kind of componential analysis, which I call {\it dynamic componential analysis}.
The components of the proposed model are paraphrasing verbs acquired dynamically from the Web
in a principled manner, which addresses the major objection against the classic componential analysis
of being inherently subjective.

\section{Comparison to Abstract Relations in the Literature}

In order to test the paraphrasing approach,
I use noun-noun compound examples from the literature:
I extract corresponding verbal paraphrases for them,
and I manually determine whether these verbs accurately reflect
the expected abstract semantic relations.

\subsection{Comparison to Girju \emph{et al.} (2005)}

First, I study how my paraphrasing verbs relate to
the abstract semantic relations proposed by \namecite{girju:2005:on:the:semantics}
for the semantic classification of noun compounds.
For the purpose, I try to paraphrase the 21 example noun-noun compounds
provided in that article as illustrations of the 21 abstract relations.
I was only able to extract paraphrases for 14 of them,
and I could not find meaningful verbs for the rest:
{\it quality sound} (\texttt{ATTRIBUTE-HOLDER}),
{\it crew investigation} (\texttt{AGENT}),
{\it image team} (\texttt{DEPICTION-DEPICTED}),
{\it girl mouth} (\texttt{PART-WHOLE}),
{\it style performance} (\texttt{MANNER}),
{\it worker fatalities} (\texttt{RECIPIENT}),
and {\it session day} (\texttt{MEASURE}).
For most of these seven cases, there appears either not to be a meaningful predicate for the
particular nouns paired or a nominalization plays the role of the predicate.

Table \ref{table:comparison2girju} shows
the target semantic relation, an example noun-noun compound from that relation,
and the top paraphrasing verbs, optionally followed by prepositions,
that I generate for that example.
The verbs expressing the target relation are in bold,
those referring to a different but valid relation are in italic,
and the erroneous extractions are struck out.
Each verb is followed by a corresponding frequency of extraction.

Overall, the extracted verbs seem to provide a good characterization
of the noun compounds. While in two cases the most frequent verb
is the copula ({\it to be}), the following most frequent verbs
are quite adequate.
In the case of {\it ``malaria mosquito''}, one can argue
that the \texttt{CAUSE} relation, assigned by \namecite{girju:2005:on:the:semantics},
is not exactly correct, in that the disease is only indirectly caused by the mosquitos
(it is rather carried by them), and the proposed most frequent verbs {\it carry}
and {\it spread} actually support a different abstract relation: \texttt{AGENT}.
Still, {\it cause} appears as the third most frequent verb,
indicating that it is common to consider indirect causation as
a causal relation. In the case of {\it combustion gas}, the most frequent verb
{\it support} is a good paraphrase of the noun compound,
but is not directly applicable to the \texttt{RESULT} relation
assigned by \namecite{girju:2005:on:the:semantics};
however, the remaining verbs for that relation do support \texttt{RESULT}.

For the remaining noun-noun compounds, the most frequent verbs accurately capture
the relation assigned by \namecite{girju:2005:on:the:semantics};
in some cases, the less frequent verbs indicate other logical entailments
for the noun combination.

\begin{table}
\begin{center}
\begin{small}
\begin{tabular}{@{}lll@{}}

\multicolumn{1}{c}{\bf Sem. Relation} & \multicolumn{1}{c}{\bf Example} & \multicolumn{1}{c}{\bf Extracted Verbs}\\

\hline
\texttt{POSSESSION} & {\it family estate} &  \textst{be in(29)}, {\bf be held by(9)}, {\bf be owned by(7)}\\
\hline

\texttt{TEMPORAL} & {\it night flight} &  {\bf arrive at(19)}, {\bf leave at(16)}, {\bf be at(6)},\\
 & & {\bf be conducted at(6)}, {\bf occur at(5)}\\
\hline

\texttt{IS-A(HYPERNYMY)} & {\it Dallas city} &  {\bf include(9)}\\
\hline

\texttt{CAUSE} & {\it malaria mosquito} & {\it carry(23)}, {\it spread(16)}, {\bf cause(12)}, {\it transmit(9)},\\
 & &  {\it bring(7)}, {\it have(4)}, {\it be infected with(3)},\\
 & & {\bf be responsible for(3)}, {\it test positive for(3)},\\
 & & {\bf infect many with(3)}, {\it be needed for(3)},\\
 & &  {\bf pass on(2)}, {\bf give(2)}, {\bf give out(2)}\\
\hline

\texttt{MAKE/PRODUCE} & {\it shoe factory} &  {\bf produce(28)}, {\bf make(13)}, {\bf manufacture(11)}\\
\hline

\texttt{INSTRUMENT} & {\it pump drainage} &  {\bf be controlled through(3)}, {\bf use(2)}\\
\hline

\texttt{LOCATION/SPACE} & {\it Texas university} &  \textst{be(5)}, {\bf be in(4)}\\
\hline

\texttt{PURPOSE} & {\it migraine drug} &  {\bf treat(11)}, {\bf be used for(9)}, {\bf prevent(7)},\\
 & & {\bf work for(6)}, {\bf stop(4)}, {\bf help(4)}, \textst{work(4)}\\
 & & {\bf be prescribed for(3)}, {\bf relieve(3)}, {\bf block(3)},\\
 & & {\it be effective for(3)}, {\it be for(3)}, {\bf help ward off(3)},\\
 & & {\it seem effective against(3)}, {\bf end(3)}, {\bf reduce(2)}\\
\hline

\texttt{SOURCE} & {\it olive oil} &  {\bf come from(13)}, {\bf be obtained from(11)},\\
 & & {\bf be extracted from(10)}, {\bf be made from(9)},\\
 & & {\bf be produced from(7)}, {\bf be released from(4)},\\
 & & {\it taste like(4)}, {\bf be beaten from(3)},\\
 & & {\bf be produced with(3)}, {\bf emerge from(3)}\\
\hline

\texttt{TOPIC} & {\it art museum} &  {\bf focus on(29)}, {\it display(16)}, {\bf bring(14)},\\
 & & {\bf highlight(11)}, {\it house(10)}, {\it exhibit(9)}\\
 & & {\bf demonstrate(8)}, {\bf feature(7)}, {\it show(5)},\\
 & & {\bf tell about(4)}, {\bf cover(4)}, {\bf concentrate in(4)}\\
\hline

\texttt{MEANS} & {\it bus service} &  {\bf use(14)}, {\bf operate(6)}, {\it include(6)}\\
\hline

\texttt{EXPERIENCER} & {\it disease victim} &  {\bf spread(12)}, {\bf acquire(12)}, {\bf suffer from(8)},\\
 & & {\bf die of(7)}, {\it develop(7)}, {\bf contract(6)}, {\bf catch(6)},\\
 & & {\bf be diagnosed with(6)}, {\it have(5)}, {\it beat(5)},\\
 & & {\bf be infected by(4)}, {\bf survive(4)}, {\bf die from(4)},\\
 & & {\bf get(4)}, {\bf pass(3)}, {\bf fall by(3)}, {\it transmit(3)}\\
\hline

\texttt{THEME} & {\it car salesman} &  {\bf sell(38)}, \textst{mean inside(13)}, {\bf buy(7)},\\
 & & {\bf travel by(5)}, {\bf pay for(4)}, {\bf deliver(3)},\\
 & & {\bf push(3)}, {\bf demonstrate(3)}, {\bf purr(3)}\\
\hline

\texttt{RESULT} & {\it combustion gas} &  {\it support(22)}, {\bf result from(14)},\\
 & & {\bf be produced during(11)},{\bf be produced by(8)}\\
 & &  {\bf be formed from(8)}, {\bf form during(8)},\\
 & & {\bf be created during(7)}, {\bf originate from(6)},\\
 & &  {\bf be generated by(6)}, {\bf develop with(6)}\\

\hline
\end{tabular}
\caption{\textbf{Top paraphrasing verbs for 14 of the 21 relations in Girju \emph{et al.} (2005).}
    Verbs expressing the target relation are in {\bf bold},
    those referring to a different but semantically valid one are in {\it italic},
    and errors are \textst{struck out}.}
\label{table:comparison2girju}
\end{small}
\end{center}
\end{table}

\begin{table}
\begin{center}
\begin{small}
\begin{tabular}{@{}lll@{}}

\multicolumn{1}{c}{\bf Sem. Relation} & \multicolumn{1}{c}{\bf Example} & \multicolumn{1}{c}{\bf Extracted Verbs}\\
\hline \hline

\texttt{AGENT} & {\it student protest} &  {\bf be led by(6)}, {\bf be sponsored by(6)}, {\bf pit(4)},\\
 & & {\it be(4)}, {\bf be organized by(3)}, {\bf be staged by(3)}, \\
 & &  {\bf be launched by(3)}, {\bf be started by(3)},\\
 & & {\bf be supported by(3)}, {\it involve(3)}, {\it arise from(3)}\\
\hline

\texttt{AGENT} & {\it band concert} &  {\it feature(17)}, {\it capture(10)}, {\it include(6)}, {\bf be given},\\
 & & {\bf by(6)}, {\it play of(4)}, {\it involve(4)}, \textst{be than(4)}\\
 & &  {\bf be organized by(3)}, {\bf be by(3)}, {\it start with(3)}\\
\hline

\texttt{AGENT} & {\it military assault} &  {\bf be initiated by(4)}, {\it shatter(2)}\\
\hline \hline

\texttt{BENEFICIARY} & {\it student price} &  {\it be(14)}, \textst{mean(4)}, \textst{differ from(4)}, {\bf be for(3)},\\
 & & {\bf be discounted for(3)}, {\bf be affordable for(3)},\\
 & & {\bf be unfair for(3)}, {\bf be charged for(3)}\\
\hline \hline

\texttt{CAUSE} & {\it exam anxiety} &  {\it be generated during(3)}\\
\hline \hline

\texttt{CONTAINER} & {\it printer tray} &  {\bf hold(12)}, {\it come with(9)}, {\it be folded(8)}, {\it fit}\\
 & & {\it under(6)}, {\bf be folded into(4)}, {\bf pull from(4)},\\
 & & {\bf be inserted into(4)}, {\it be mounted on(4)},\\
 & & {\it be used by(4)}, {\bf be inside(3)}, {\bf feed into(3)}\\
\hline

\texttt{CONTAINER} & {\it flood water} &  {\it cause(24)}, {\it produce(9)}, {\it remain after(9)},\\
 & & {\it be swept by(6)}, {\it create(5)}, {\it bring(5)}, {\it reinforce(5)}\\
\hline

\texttt{CONTAINER} & {\it film music} &  {\it fit(16)}, {\bf be in(13)}, {\bf be used in(11)}, {\bf be heard},\\
 & & {\bf in(11)}, {\it play throughout(9)}, {\it be written for(9)}\\
\hline

\texttt{CONTAINER} & {\it story idea} &  {\it tell(20)}, {\it make(19)}, {\it drive(15)}, {\it become(13)},\\
 & & {\it turn into(12)}, {\it underlie(12)}, {\bf occur within(8)},\\
 & & {\bf hold(8)}, {\it tie(8)}, {\it be(8)}, {\it spark(8)}, {\it tell(7)}, {\it move(7)}\\
\hline \hline

\texttt{CONTENT} & {\it paper tray} &  {\bf feed(6)}, {\it be lined with(6)}, {\it stand up(6)}, {\bf hold(4)},\\
 & & {\bf contain(4)}, {\it catch(4)}, {\bf overflow with(3)}\\
\hline

\texttt{CONTENT} & {\it eviction notice} &  {\it result in(10)}, {\it precede(3)}, {\it make(2)}\\
\hline \hline

\texttt{DESTINATION} & {\it game bus} &  {\it be in(6)}, {\bf leave for(3)}, {\it be like(3)}, {\it be(3)},\\
 & & {\it make playing(3)}, {\it lose(3)}\\
\hline

\texttt{DESTINATION} & {\it exit route} &  {\it be indicated by(4)}, {\bf reach(2)}, {\it have(1)}, {\it do(1)}\\
\hline

\texttt{DESTINATION} & {\it entrance stairs} &  {\it look like(4)}, {\it stand outside(3)}, {\it have(3)},\\
 & & {\it follow from(3)}, {\it be at(3)}, \textst{be(3)}, {\it descend from(2)}\\
\hline \hline

\texttt{EQUATIVE} & {\it player coach} &  {\it work with(42)}, {\it recruit(28)}, {\bf be(19)}, {\it have(16)},\\
 & & {\it know(16)}, {\it help(12)}, {\it coach(11)}, {\it take(11)}\\
\hline \hline

\texttt{INSTRUMENT} & {\it electron microscope} &  {\bf use(27)}, {\it show(5)}, {\bf work with(4)}, {\bf utilize(4)},\\
 & & {\bf employ(4)}, {\it beam(3)}\\
\hline

\texttt{INSTRUMENT} & {\it diesel engine} &  {\it be(18)}, {\bf operate on(8)}, {\it look like(8)}, {\bf use(7)},\\
 & &  {\it sound like(6)}, {\bf run on(5)}, {\bf be on(5)}\\
\hline

\texttt{INSTRUMENT} & {\it laser printer} &  {\bf use(20)}, {\it consist of(6)}, {\it be(5)}\\
\hline

\end{tabular}
\caption{\textbf{Comparison to Barker \& Szpakowicz (1998):
        top paraphrasing verbs for 8 of the 20 relations.}
    Verbs expressing the target relation are in {\bf bold},
    those referring to a different but semantically valid one are in {\it italic},
    and errors are \textst{struck out}.}
\label{table:comparison2barker}
\end{small}
\end{center}
\end{table}

\subsection{Comparison to Barker \& Szpakowicz (1998)}

Table \ref{table:comparison2barker} compares my paraphrasing verbs
and the first 8 (out of 20) abstract relations from \namecite{Barker:Szpakowicz:1998:nc:sem}:
the paper gives several examples per relation, and I show the results
for each of them, omitting {\it charitable donation} (\texttt{BENEFICIARY})
and {\it overdue fine} (\texttt{CAUSE}) since the modifier in these cases is an
adjective\footnote{\namecite{Barker:Szpakowicz:1998:nc:sem} allow for the modifier to be either a noun or an adjective.},
and {\it composer arranger} (\texttt{EQUATIVE}),
for which I could not extract suitable paraphrases.

I obtain very good results for \texttt{AGENT} and \texttt{INSTRUMENT},
but other relations are problematic, probably due to the varying quality
of the classifications: while {\it printer tray} and {\it film music}
look correctly assigned to \texttt{CONTAINER},
{\it flood water} and {\it story idea} are quite abstract and questionable;
{\it entrance stairs} (\texttt{DESTINATION}) could be equally
well analyzed as \texttt{LOCATION} or \texttt{SOURCE};
and {\it exam anxiety} (\texttt{CAUSE}) could refer to \texttt{TIME}.
Finally, although Table \ref{table:comparison2barker} shows the verb {\it to be}
ranked third for {\it player coach}, in general the \texttt{EQUATIVE} relation
poses a problem since the copula is not very frequent
in the form of paraphrase I am looking for, e.g., `{\it coach who is a player}'.

\begin{table}
\begin{center}
\begin{small}
\begin{tabular}{@{}lll@{}}

 \multicolumn{1}{c}{\bf Categ. Pair} & \multicolumn{1}{c}{\bf Examples} & \multicolumn{1}{c}{\bf Extracted Verbs}\\
 \hline

 \texttt{A01-A07} & ankle artery &  {\it feed}(133), {\it supply}(111), {\it drain}(100), {\it be in}(44),\\
 \texttt{(Body Regions -} & foot vein &  {\it run}(37), {\it appear on}(29), {\it be located in}(22),\\
 \texttt{Cardiovascular} & forearm vein & {\it be found in}(20), {\it run through}(19), {\it be behind}(19),\\
 \texttt{System)} & finger artery & {\it run from}(18), {\it serve}(15), {\it be felt with}(14),\\
  & neck vein & {\it enter}(14), {\it pass through}(12), {\it pass by}(12),\\
  & head vein & {\it show on}(11), {\it be visible on}(11), {\it run along}(11),\\
  & leg artery & {\it nourish}(10), {\it be seen on}(10), {\it occur on}(10),\\
  & thigh vein & {\it occur in}(9), {\it emerge from}(9), {\it go into}(9), $\ldots$\\
 \hline

 \texttt{A01-M01.643} & arm patient &  {\it be}(54), {\it lose}(40), {\it have}(30), {\it be hit in}(11),\\
 \texttt{(Body Regions -} & eye outpatient &  {\it break}(9), {\it gouge out}(9), {\it injure}(8), {\it receive}(7),\\
 \texttt{Disabled Persons)} & abdomen patient & {\it be stabbed in}(7), {\it be shot in}(7), {\it need}(6), $\ldots$\\
 \hline

 \texttt{A01-M01.150} & leg amputee &  {\it lose}(13), {\it grow}(6), {\it have cut off}(4), {\it miss}(2),\\
 \texttt{(Body Regions -} & arm amputee &  {\it need}(1), {\it receive}(1), {\it be born without}(1)\\
 \texttt{Disabled Persons)} & knee amputee & \\
 \hline

 \texttt{A01-M01.898} & eye donor &  {\it give}(4), {\it provide}(3), {\it catch}(1)\\
 \texttt{(Body Regions -} & skin donor & \\
 \texttt{Donors)} & & \\
 \hline

 \texttt{D02-E05.272} & choline diet & {\it be low in}(18), {\it contain}(13), {\it be deficient in}(11),\\
 \texttt{(Organic Chemicals} & methionine diet &  {\it be high in}(7), {\it be rich in}(6), {\it be sufficient in}(6),\\
 \texttt{- Diet)} & carotene diet & {\it include}(4), {\it be supplemented with}(3), {\it be in}(3),\\
 & saccharin diet & {\it be enriched with}(3), {\it contribute}(2), {\it miss}(2), $\ldots$\\
 \hline

\end{tabular}
\caption{\textbf{Top paraphrasing verbs for relations from Rosario \emph{et al.} (2002).}}
\label{table:comparison2rosario}
\end{small}
\end{center}
\end{table}

\subsection{Comparison to Rosario \emph{et al.} (2002)}

\namecite{Rosario:al:2002:descent} characterize noun-noun
compounds based on the semantic category in the MeSH lexical
hierarchy each of the constituent nouns belongs to. For example,
all noun compounds in which the first noun is classified under the
A01 sub-hierarchy ({\it Body Regions}), and the second one falls under A07
({\it Cardiovascular System}), are hypothesized to express the same
relation.  Examples include {\it mesentery artery}, {\it leg vein},
{\it finger capillary}, etc.

Given a category pair of {\it MeSH} labels,
I compare my approach to the descent of hierarchy
by generating paraphrasing verbs on a large scale
for many different compounds belonging to the target category pair.

I collect noun-noun compounds using
the {\it LQL} system and the collection of 1.4 million {\it MEDLINE} abstracts
described in chapter \ref{app:dataset:bio} in the appendix.
I use the heuristic proposed by \namecite{lauer:1995:thesis},
who extracts noun-noun pairs $w_1w_2$ from four-word sequences $w_0w_1w_2w_3$,
where $w_1$ and $w_2$ are nouns and $w_0$, $w_3$ are non-nouns,
with the additional requirement that both $w_1$ and $w_2$ represent
single-word {\it MeSH} terms.
The comparisons against {\it MeSH}
are performed using all inflections and synonyms for a given term
that are listed in {\it MeSH}.
As a result, I obtain 228,702 noun-noun pairs,
40,861 of which are unique, which corresponds to 35,205 unique
{\it MeSH} category pairs of various generalization levels.

Given a category pair, e.g., A01-A07,
I consider all noun-noun compounds whose elements
are in the corresponding {\it MeSH} sub-hierarchies, and I acquire
paraphrasing verbs (+prepositions) for each of them from the Web.
I then aggregate the results in order to obtain a set of characterizing
paraphrasing verbs for the target category pair.

As Table \ref{table:comparison2rosario} shows,
the results are quite good for A01-A07, for which I have a lot of examples,
and for D02-E05.272, which seems relatively unambiguous,
but they are not as good for A01-M01.*, which is both more ambiguous and has fewer examples:
generalizing verbal paraphrases for a category seems to work
best for categories represented by multiple relatively unambiguous examples.

\section{Comparison to Human-Generated Verbs}
\label{sec:sem:human:judgments}

\begin{figure}
   \center
   \includegraphics[width=440pt]{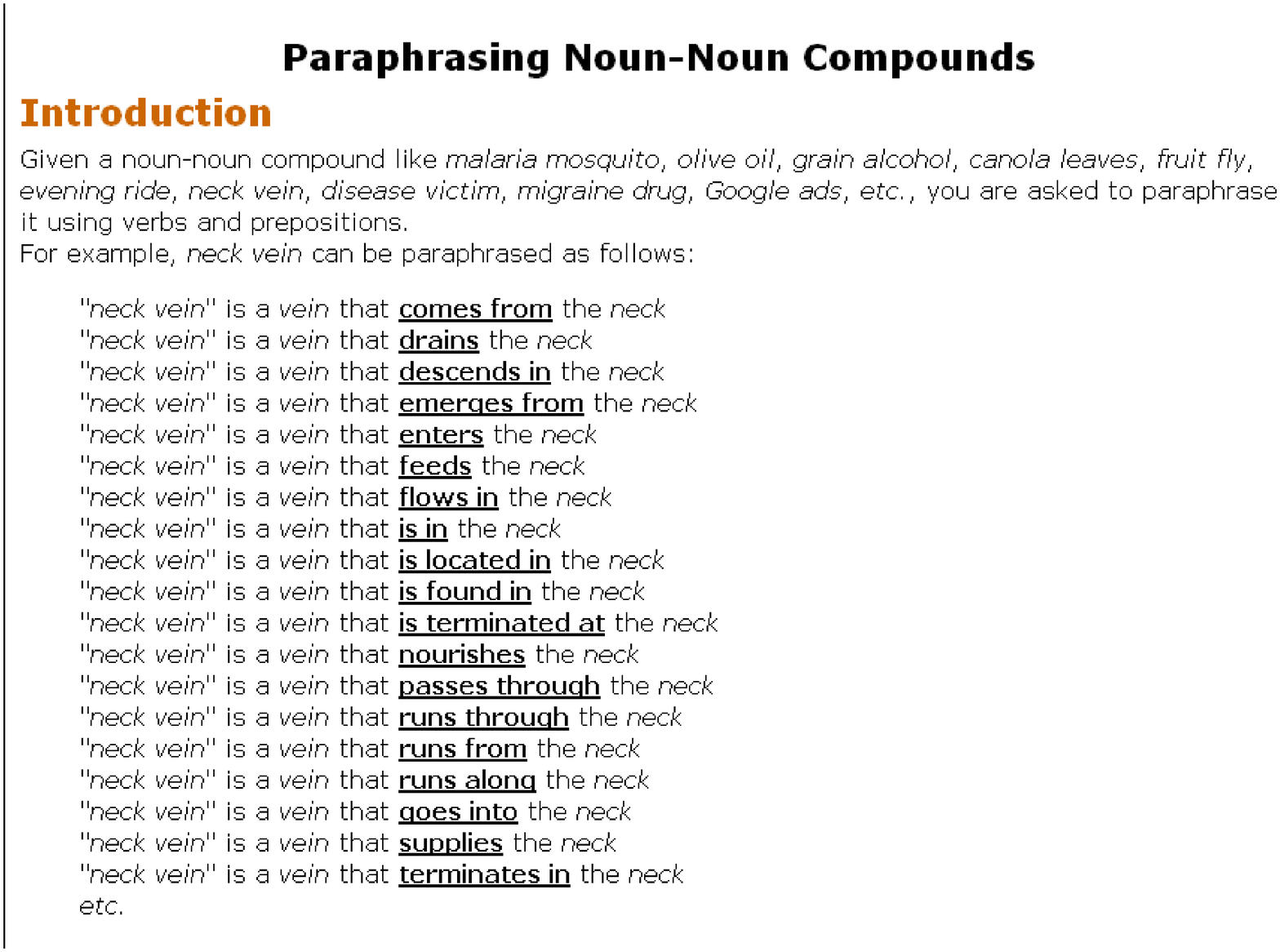}\\
   \caption{\textbf{The noun-noun paraphrasing task in Amazon's Mechanical Turk:} task introduction.}
   \label{Fig:mturk:instr}
 \end{figure}

\begin{figure}
   \center
   \includegraphics[width=440pt]{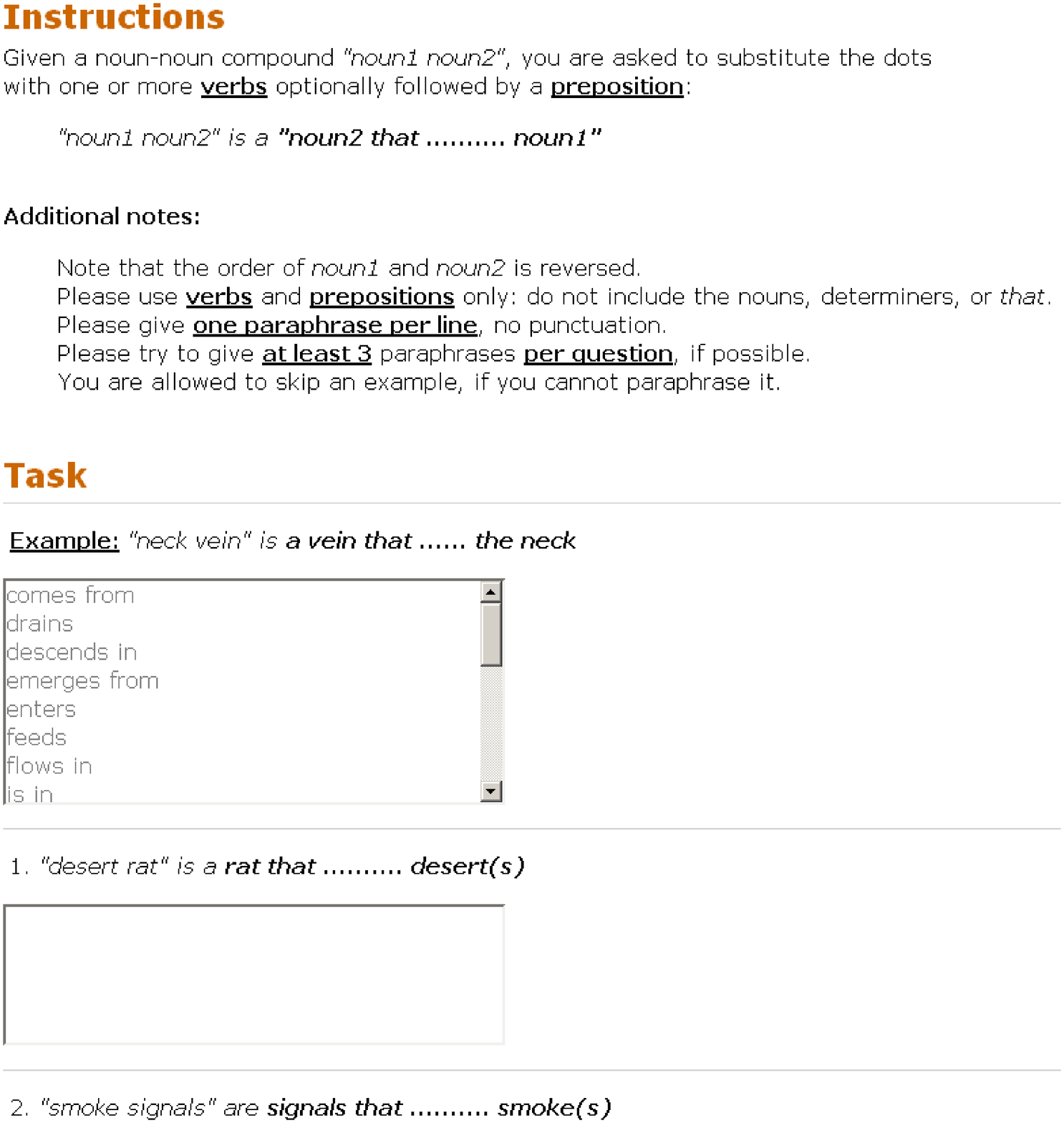}\\
   \caption{\textbf{The noun-noun paraphrasing task in Amazon's Mechanical Turk:}
            instructions, example, and sample questions.}
   \label{Fig:mturk:task}
 \end{figure}

To evaluate the verb-based semantic relations I obtained,
I conducted an experiment in which I gathered paraphrases from human judges.
For the purpose, I defined a special noun-noun compound paraphrasing task
asking human subjects to propose verbal paraphrases of the kind my program generates:
I asked for verbs, possibly followed by prepositions, that could be used in a paraphrase
involving {\it that}. For example, {\it nourish}, {\it run along} and {\it come from}
are good paraphrasing verbs for the noun-noun compound {\it neck vein}
since they can be used in paraphrases like
`{\it a vein that \underline{nourishes} the neck}',
`{\it a vein that \underline{runs along} the neck}'
or `{\it a vein that \underline{comes from} the neck}'.
In an attempt to make the task as clear as possible and to ensure high quality of the results,
I provided detailed instructions, I stated explicit restrictions, and I gave several example paraphrases.
I instructed the participants to propose at least three paraphrasing verbs
per noun-noun compound, if possible.
The instructions I provided and the actual interface the human subjects were seeing
are shown in Figures \ref{Fig:mturk:instr} and \ref{Fig:mturk:task}:
worker's user interface of the {\it Amazon Mechanical Turk} Web service.\footnote{\texttt{http://www.mturk.com}}

%

The service represents a cheap and easy way to recruit subjects for various tasks
that require human intelligence;
it provides an API allowing a computer program
to ask a human to perform a task and return the results,
which {\it Amazon} calls ``{\it artificial artificial intelligence}''.
The idea behind the latter term and behind the origin of service's name
come from the ``mechanical Turk'', a life-sized wooden chess-playing mannequin
the Hungarian nobleman Wolfgang von Kempelen constructed in 1769,
which was able to defeat skilled opponents across Europe, including Benjamin Franklin and Napoleon Bonaparte.
The audience believed the automaton was making decisions
using artificial intelligence, but the actual secret was a chess master hidden inside.
The {\it Amazon Mechanical Turk} Web Service provides a similar solution to computer applications.

Tables \ref{table:mturk:malaria:mosquito}, \ref{table:mturk:olive:oil}
\ref{table:mturk:disease:victim} and \ref{table:mturk:night:flight}
compare human- and program-generated paraphrasing verbs for
{\it malaria mosquito}, {\it olive oil}, {\it disease victim} and {\it night flight}, respectively.
The human-generated paraphrasing verbs, obtained from ten subjects,
are shown on the left sides of the tables, 
while the right sides list the program-generated verbs;
the verbs appearing on both sides are underlined.
We can see in these tables a significant overlap between the human- and the program-generated
paraphrasing verbs for {\it malaria mosquito},
and less overlap for the more ambiguous {\it night flight}, {\it disease victim}, and {\it olive oil}.
For example, the latter can refer to multiple abstract relations, e.g.,
\texttt{CONTAINER} ({\it oil that \underline{is inside} the olive}),
\texttt{SOURCE} or \texttt{ORIGIN} ({\it oil that \underline{comes from} olives}),
\texttt{PRODUCT} ({\it oil that \underline{is produced from} olives}),
\texttt{QUALITY} ({\it oil that \underline{tastes like} olive}), etc.
Still, for all four given examples, there is a general tendency
for the most frequent human-proposed and the top program-generated verbs to overlap.

\begin{table}
\begin{center}
\begin{tabular}{rl|rl}
  {\bf \#} & {\bf Human Judges} & {\bf \#} & {\bf Program}\\
 \hline
  8 & \underline{carries} & 23 & \underline{carries}\\
  4 & \underline{causes}  & 16 & \underline{spreads}\\
  2 & \underline{transmits}  & 12 & \underline{causes}\\
  2 & \underline{is infected with}  & 9 & \underline{transmits}\\
  2 & \underline{infects with}           & 7 & brings\\
  1 & \underline{has}      & 4 & \underline{has}\\
  1 & \underline{gives}   & 3 & \underline{is infected with}\\
  1 & \underline{spreads} & 3 & \underline{infects with}\\
  1 & propagates & 2 & \underline{gives} \\
  1 & supplies & 2 & is needed for \\
 \hline
\end{tabular}
\caption{{\bf Human- and program-generated verbs for \emph{malaria mosquito}.}
    Verbs appearing on both sides of the tables are underlined.}
\label{table:mturk:malaria:mosquito}
\end{center}
\end{table}

\begin{table}
\begin{center}
\begin{tabular}{rl|rl}
  {\bf \#} & {\bf Human Judges} & {\bf \#} & {\bf Program}\\
 \hline
  5 & is pressed from & 13 & \underline{comes from}\\
  4 & \underline{comes from} & 11 & is obtained from\\
  4 & \underline{is made from} & 10 & \underline{is extracted from}\\
  2 & is squeezed from & 9 & \underline{is made from}\\
  2 & is found in & 7 & \underline{is produced} from\\
  1 & \underline{is extracted from} & 4 & is released from\\
  1 & is in & 4 & tastes like\\
  1 & \underline{is produced} out of & 3 & is beaten from\\
  1 & is derived from & 3 & \underline{is produced} with\\
  1 & is created from & 3 & emerges from\\
  1 & contains &  & \\
  1 & is applied to &  & \\
 \hline
\end{tabular}
\caption{{\bf Human- and program-generated verbs for \emph{olive oil}.}
    Verbs appearing on both sides of the tables are underlined;
    both full and partial overlaps are underlined.}
\label{table:mturk:olive:oil}
\end{center}
\end{table}

\begin{table}
\begin{center}
\begin{tabular}{rl|rl}
  {\bf \#} & {\bf Human Judges} & {\bf \#} & {\bf Program}\\
 \hline
    6 & \underline{has} & 12 & spreads\\
    3 & \underline{suffers from} & 12 & acquires \\
    3 & \underline{is infected} with & 8 & \underline{suffers from}\\
    2 & \underline{dies} of & 7 & \underline{dies of}\\
    2 & exhibits & 7 & develops \\
    2 & carries & 6 & \underline{contracts}\\
    1 & \underline{is diagnosed with} & 6 & \underline{is diagnosed with}\\
    1 & \underline{contracts} & 6 & catches\\
    1 & is inflicted with & 5 & \underline{has}\\
    1 & is ill from & 5 & beats\\
    1 & succumbs to & 4 & \underline{is infected} by\\
    1 & is affected by & 4 & survives\\
    1 & presents & 4 & \underline{dies} from\\
    & & 4 & gets\\
    & & 3 & passes\\
    & & 3 & falls by\\
    & & 3 & transmits\\
 \hline
\end{tabular}
\caption{{\bf Human- and program-generated verbs for \emph{disease victim}.}
    Verbs appearing on both sides of the tables are underlined;
    both full and partial overlaps are underlined.}
\label{table:mturk:disease:victim}
\end{center}
\end{table}

\begin{table}
\begin{center}
\begin{tabular}{rl|rl}
  {\bf \#} & {\bf Human Judges} & {\bf \#} & {\bf Program}\\
 \hline
    5 & \underline{occurs at} & 19 & \underline{arrives} at\\
    5 & \underline{is at} & 16 & leaves at\\
    4 & happens at & 6 & \underline{is at}\\
    2 & takes off at & 6 & is conducted at\\
    1 & \underline{arrives} by & 5 & \underline{occurs at}\\
    1 & travels through\\
    1 & runs through \\
    1 & occurs during\\
    1 & is taken at\\
    1 & is performed at\\
    1 & is flown during\\
    1 & departs at \\
    1 & begins at\\
 \hline
\end{tabular}
\caption{{\bf Human- and program-generated verbs for \emph{night flight}.}
    Verbs appearing on both sides of the tables are underlined;
    both full and partial overlaps are underlined.}
\label{table:mturk:night:flight}
\end{center}
\end{table}

I further compared the human and the program-generated paraphrases
in a bigger study using the complex nominals
listed in the appendix of \namecite{levi:1978}.
I had to exclude the examples with an adjectival modifier,
which are allowed by Levi's theory,
as I already explained in section \ref{nc:ling:theory:levi}.
In addition, some of the noun compounds were spelled as a single word,
which, according to my definition of noun compound given in section \ref{sec:compounding:my:def},
represents a single noun.
Therefore, I had to exclude the following concatenated words from Levi's dataset:
{\it whistleberries}, {\it gunboat}, {\it silkworm},
{\it cellblock}, {\it snowball}, {\it meatballs},
{\it windmill}, {\it needlework}, {\it textbook},
{\it doghouse}, and {\it mothballs}.
Some other examples contained a modifier that is
a concatenated noun compound, e.g.,
{\it wastebasket category}, {\it hairpin turn}, {\it headache pills},
{\it basketball season}, {\it testtube baby}.
These examples are noun-noun compounds under my definition, and therefore I retained them.
However, I find them inconsistent with the other examples in the collection
from Levi's theory point of view: the dataset is supposed to contain noun-noun compounds only.
Even more problematic (but not for my definition), is {\it beehive hairdo},
where both the modifier and the head are concatenations; I retained that example as well.
As a result, I ended up with 250 good noun-noun compounds
out of the original 387 complex nominals.

I randomly distributed these 250 noun-noun compounds into groups of 5
as shown in Figure \ref{Fig:mturk:task},
which yielded 50 Mechanical Turk tasks known as HITs (Human Intelligence Tasks),
and I requested 25 different human subjects (workers) per HIT.
I had to reject some of the submissions,
which were empty or were not following the instructions,
in which cases I requested additional workers in order to guarantee
at least 25 good submissions per HIT.
Each human subject was allowed to work on any number of HITs (between 1 and 50),
but was not permitted to do the same HIT twice,
which is controlled by the {\it Amazon Mechanical Turk} Web Service.
A total of 174 different human subjects worked on the 50 HITs, producing 19,018 different verbs.
After removing the empty and the bad submissions, and after normalizing the verbs (see below),
I ended up with a total of 17,821 verbs, which means 71.28 verbs per noun-noun compound on average,
not necessarily distinct.

Since many workers did not strictly follow the instructions,
I performed some automatic cleaning of the results, followed by a manual check and correction,
when it was necessary.
First, some workers included the target nouns, the complementizer {\it that},
or determiners like {\it a} and {\it the}, in addition to the paraphrasing verb,
in which cases I removed this extra material.
For example, {\it star shape} was paraphrased as
{\it shape that looks like a star} or as {\it looks like a}
instead of just {\it looks like}.
Second, the instructions required that a paraphrase be a sequence
of one or more verb forms possibly followed by a preposition
(complex prepositions like {\it because of} were allowed),
but in many cases the proposed paraphrases contained words belonging to other
parts of speech, e.g.,
nouns ({\it is in the \underline{shape} of}, {\it has \underline{responsibilities} of},
{\it has the \underline{role} of}, {\it makes \underline{people} have},
{\it is \underline{part} of}, {\it makes \underline{use} of})
or predicative adjectives
({\it are \underline{local} to}, {\it is \underline{full} of}); I filtered out such paraphrases.
In case a paraphrase contained an adverb, e.g.,
{\it occur \underline{only} in}, {\it will \underline{eventually} bring},
I removed the adverb and kept the paraphrase.
Third, I normalized the verbal paraphrases by removing the leading modals
(e.g., {\it \underline{can} cause} becomes {\it cause}),
perfect tense {\it have} and {\it had} (e.g., {\it \underline{have} joined} becomes {\it joined}),
or continuous tense {\it be} (e.g., {\it \underline{is} donating} becomes {\it donates}).
I converted complex verbal construction of the form `{\it $<$raising verb$>$ to be}'
(e.g., {\it appear to be}, {\it seems to be}, {\it turns to be}, {\it happens to be},
{\it is expected to be}) to just {\it be}.
I further removed present participles introduced by {\it by},
e.g., {\it are caused \underline{by peeling}} becomes {\it are caused}.
I further filtered out any paraphrase that involved {\it to} as part of the infinitive
of a verb different from {\it be}, e.g., {\it is willing \underline{to} donate}
or {\it is painted \underline{to} appear like} are not allowed.
I also added {\it be} when it was missing in passive constructions, e.g., {\it made from}
became {\it be made from}. Finally, I lemmatized the conjugated verb forms using {\it WordNet}, e.g.,
{\it comes from} becomes {\it come from}, and {\it is produced from} becomes {\it be produced from}.
I also fixed some occasional spelling errors that I noticed, e.g.,
{\it b\underline{o}longs to}, {\it happens bec\underline{asue} of},
{\it is \underline{mm}ade from}.

\begin{table}
\begin{center}
\begin{tabular}{|c|c|c|c|}
\hline
{\bf Min \# of} & {\bf Number of} & \multicolumn{2}{c|}{\bf Correlation to Human}\\
{\bf Web Verbs} & {\bf Compounds}  & {\bf All Verbs} & {\bf First Only}\\
\hline
 0 & 250 & 31.81\% & 30.60\%\\
 1 & 236 & 33.70\% & 32.41\%\\
 3 & 216 & 35.39\% & 34.07\%\\
 5 & 203 & 36.85\% & 35.60\%\\
10 & 175 & 37.31\% & 35.53\%\\
\hline
\end{tabular}
\caption{\textbf{Average cosine correlation (in \%s) between the human-
and the program-generated verbs for 250 noun-noun compounds from Levi \emph{et al.} (1978).}
Shown are the results for different limits on the minimum number of program-generated Web verbs.
The last column shows the cosine when only the first verb proposed by each worker is used.}
\label{table:human:levi:comparison}
\end{center}
\end{table}

For each noun-noun compound, I built two frequency vectors $\overrightarrow{h}$ and $\overrightarrow{p}$,
using the human-generated paraphrasing verbs and their frequencies,
and the program-generated ones, respectively.
Each coordinate corresponds to
I then calculated the cosine correlation coefficient between these frequency vectors as follows:
\begin{equation}
    sim_{cos}(\overrightarrow{h},\overrightarrow{p}) = \frac{\sum_{i=1}^n h_i p_i}{\sqrt{\sum_{i=1}^n h_i^2}\sqrt{\sum_{i=1}^n p_i^2}}
\end{equation}

The average cosine correlation (in \%s) for all 250 noun-noun compounds
is shown in Table \ref{table:human:levi:comparison}.
Since the human subjects were instructed to provide at least three paraphrasing verbs
per compound and they tried to comply, this sometimes yielded bad verbs.
In such cases, the very first verb proposed by a worker for a given noun-noun compound
is likely to be the best one. I tested this hypothesis by calculating the cosine using
these first verbs only.
As the last two columns of the table show,
using all verbs produces a consistently better cosine,
which suggests that there are many additional good human-generated verbs
among the ones after the first.
A quick comparison of sections \ref{sec:verbs:all} and \ref{sec:verbs:first} in the Appendix
confirms this. However, the difference is small, about 1-2\%.

A limitation of the Web-based verb-generating method is that it could not
provide paraphrasing verbs for 14 of the noun-noun compounds,
in which cases the cosine was zero.
If the calculation was performed for the remaining 236 compounds only,
the cosine increased by 2\%.
Table \ref{table:human:levi:comparison} shows the results when
the cosine calculations are limited to compounds with at least 1, 3, 5 or 10 different verbs.
We can see that the correlation increases with the minimum number of required verbs,
which means that the extracted verbs are generally good,
and part of the low cosines are due to an insufficient number of extracted verbs.

Overall, all cosines in Table \ref{table:human:levi:comparison}
are in the 30-37\%, which corresponds to a medium correlation \cite{Cohen:1988}.
A detailed comparison for all 250 is shown in Appendix \ref{chap:mturkompare:human}:
see section \ref{sec:verbs:all} for the results when all human-proposed verbs are used,
and section \ref{sec:verbs:first} for only the first verb proposed by each worker is used.

\begin{figure}
   \center
   \includegraphics[width=400pt]{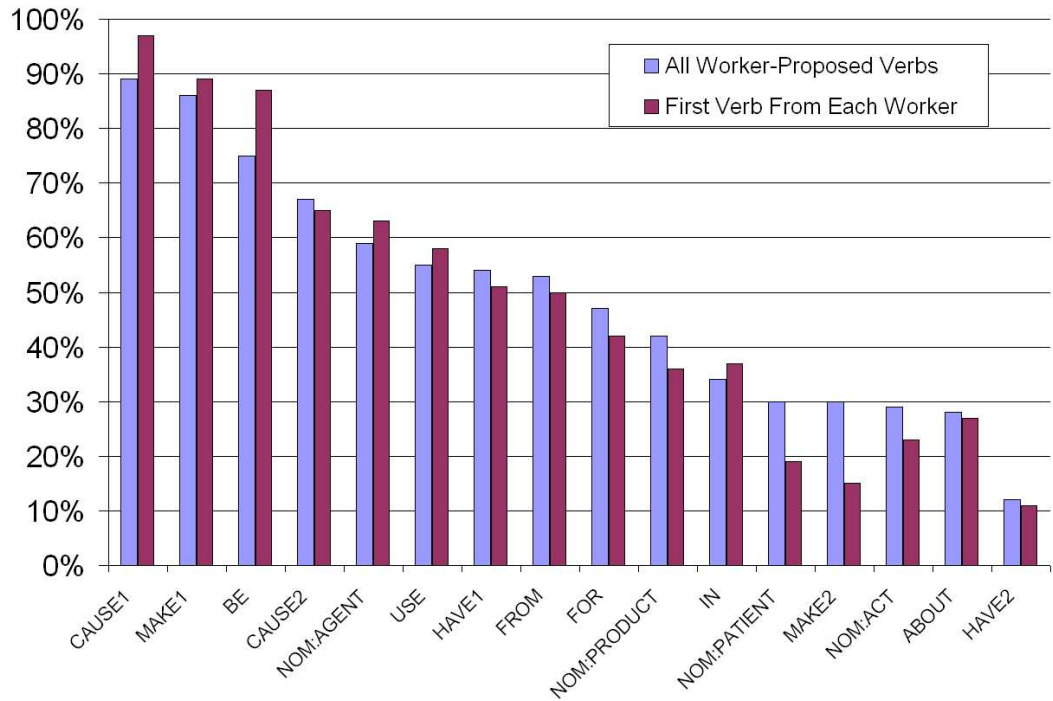}\\
   \caption{\textbf{Cosine correlation (in \%s) between the human- and the program-
        generated verbs aggregated by relation:}
        using all human-proposed verbs vs. the first verb from each worker.}
   \label{Fig:mturk:by:class}
\end{figure}

\begin{figure}
   \center
   \includegraphics[width=300pt]{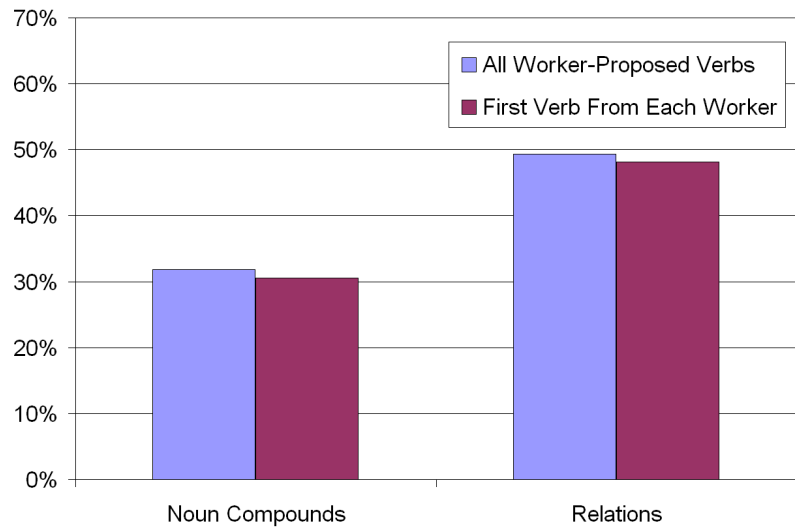}\\
    \caption{\textbf{Average cosine correlation (in \%s) between the human-
        and the program-generated verbs for 250 noun-noun compounds from Levi \emph{et al.} (1978)
        calculated for each noun compound and aggregated by relation:}
        using all human-proposed verbs vs. the first verb from each worker.}
   \label{Fig:mturk:avg:cos}
\end{figure}

I further compare the human- and the program-generated verbs aggregated by relation.
Given a relation, e.g., \texttt{HAVE$_1$}, I collect all verbs belonging
to noun-noun compounds from that relation together with their frequencies.
From a vector-space model point of view, I sum their corresponding frequency vectors.
I do that separately for the human- and the program-generated verbs,
and I then compare them separately for each relation.
A detailed comparison for all 16 relations is shown in Appendix \ref{chap:mturkompare:human}:
see section \ref{sec:verbs:all:class} for the results when all human-proposed verbs are used,
and section \ref{sec:verbs:first:class} for when only the first verb proposed by each worker is used.

Figure \ref{Fig:mturk:by:class} shows the cosine correlations for each of the 16 relations
using all human-proposed verbs and the first verb from each worker.
We can see a very-high correlation (mid 70s to mid 90s) for relations
like \texttt{CAUSE$_1$}, \texttt{MAKE$_1$}, \texttt{BE},
but low correlation 11-30\% for reverse relations like \texttt{HAVE$_2$} and \texttt{MAKE$_2$},
for rare relations like \texttt{ABOUT}, and for most nominalizations (except for \texttt{NOM:AGENT}).
Interestingly, using the first verb only improves the results
for the highly-correlated relations, but damages the low-correlated ones.
This suggests that when a relation is more homogeneous,
the first verbs proposed by the workers are good enough, and the following verbs only introduce noise.
However, when the relation is more heterogeneous, the extra verbs become more likely to be useful.
As Figure \ref{Fig:mturk:avg:cos} shows, overall the average cosine correlation
is slightly higher when all worker-proposed verbs are used vs. the first verb from each worker only:
this is true both when comparing the individual noun-noun compounds
and when the comparison is performed for the 16 relations.
The figure also shows that while the cosine correlation for the individual noun-noun compounds
is in the low 30s, for the relations it is almost 50\%.

\section{Comparison to FrameNet}
\label{sec:sem:framenet}

\begin{table}
\begin{center}
\begin{tabular}{|l|l|c|c|}
\hline
 \textbf{Frames} & \textbf{Verbs} & \textbf{Program} & \textbf{Humans}\\
\hline
\emph{Causation} & (be) because of &        & \checkmark \\
\emph{Causation} & bring       & \checkmark & + \\
\emph{Causation} & bring about & \checkmark & \checkmark \\
\emph{Causation} & bring on    &  +         & \checkmark \\
\emph{Causation} & induce      & \checkmark & \checkmark \\
\emph{Causation} & lead to     & +          & \checkmark \\
\emph{Causation} & make        & \checkmark & \checkmark \\
\emph{Causation} & mean        &            & \\
\emph{Causation} & precipitate &            & \checkmark \\
\emph{Causation} & put         &            & \\
\emph{Causation} & raise       &            & \\
\emph{Causation} & result in   & \checkmark & \checkmark \\
\emph{Causation} & render      &            & \\
\emph{Causation} & send        &            & \\
\emph{Causation} & wreak       &            & \\
\hline
 {\bf Overlap:}  &  & {\bf 5/15 (33.3\%)} & {\bf 8/15 (53.3\%)}\\
\hline
\end{tabular}
\caption{\textbf{\emph{Causation} frame and \texttt{CAUSE$_1$}:}
        comparing \emph{FrameNet} with
        the top 150 program- and the human-generated verbs for \texttt{CAUSE$_1$}.}
\label{table:framenet:cause1}
\end{center}
\end{table}

\begin{table}
\begin{center}
\begin{tabular}{|l|l|c|c|}
\hline
 \textbf{Frame} & \textbf{Verb} & \textbf{Program} & \textbf{Humans}\\
\hline
\emph{Using} & apply         & \checkmark & \checkmark \\
\emph{Using} & avail oneself &            & \\
\emph{Using} & employ        & \checkmark & \checkmark \\
\emph{Using} & operate       & \checkmark & + \\
\emph{Using} & utilise       & \checkmark & \checkmark \\
\emph{Using} & use           & \checkmark & \checkmark \\
\hline
 {\bf Overlap:} & & {\bf 5/6 (83.3\%)} & {\bf 4/6 (66.7\%)}\\
\hline
\end{tabular}
\caption{\textbf{\emph{Using} frame and \texttt{USE}:}
        comparing \emph{FrameNet} with
        the top 150 program- and the human-generated verbs for \texttt{USE}.}
\label{table:framenet:use}
\end{center}
\end{table}

\begin{table}
\begin{center}
\begin{tabular}{|l|l|c|c|}
\hline
 \textbf{Frames} & \textbf{Verbs} & \textbf{Program} & \textbf{Humans}\\
\hline
\emph{Possession} & belong         &  & + \\
\emph{Possession, Have\_associated} & have got       & & \\
\emph{Possession, Have\_associated} & have           & \checkmark & \checkmark \\
\emph{Possession} & lack           & \checkmark & \\
\emph{Possession} & own            & \checkmark & \checkmark \\
\emph{Possession} & possess        & \checkmark & \\
\emph{Possession} & want           & \checkmark & \\
\emph{Inclusion}  & (be) among           & & \\
\emph{Inclusion}  & (be) amongst         & & \\
\emph{Inclusion}  & contain              & \checkmark & \checkmark \\
\emph{Inclusion}  & exclude              & & \\
\emph{Inclusion}  & (be) excluding       & & \\
\emph{Inclusion}  & incorporate          & \checkmark & \checkmark \\
\hline
 {\bf Overlap:} & & {\bf 7/13 (53.8\%)} & {\bf 4/13 (30.8\%)}\\
\hline
\end{tabular}
\caption{\textbf{\emph{Possession}, \emph{Inclusion} and \emph{Have\_associated} frames and \texttt{HAVE$_1$}:}
        comparing \emph{FrameNet} with
        the top 150 program- and the human-generated verbs for \texttt{HAVE$_1$}.}
\label{table:framenet:have1}
\end{center}
\end{table}

\begin{table}
\begin{center}
\begin{tabular}{|l|l|c|c|}
\hline
 \textbf{Frames} & \textbf{Verbs} & \textbf{Program} & \textbf{Humans}\\
\hline
\emph{Intentionally\_create} & create         & \checkmark  & \checkmark \\
\emph{Intentionally\_create} & establish      &  & \\
\emph{Intentionally\_create} & found          &  & \\
\emph{Intentionally\_create} & generate       & \checkmark  & \checkmark \\
\emph{Intentionally\_create, Manufacturing} & make           & \checkmark  & \checkmark \\
\emph{Intentionally\_create, Manufacturing} & produce        & \checkmark  & \checkmark \\
\emph{Intentionally\_create} & setup          &  & \\
\emph{Intentionally\_create} & synthesise     &  & \checkmark\\
\emph{Manufacturing} & fabricate     &  & \\
\emph{Manufacturing} & manufacture     &  & \checkmark\\
\hline
 {\bf Overlap:} & & {\bf 4/10 (40.0\%)} & {\bf 6/10 (60.0\%)}\\
\hline
\end{tabular}
\caption{\textbf{\emph{Intentionally\_create} and \emph{Manufacturing} frames and \texttt{MAKE$_1$}:}
        comparing \emph{FrameNet} with
        the top 150 program- and the human-generated verbs for \texttt{MAKE$_1$}.}
\label{table:framenet:make1}
\end{center}
\end{table}

The idea to approximate noun compounds' semantics by a collection of verbs
is related to the approach taken in
the {\it Berkeley FrameNet project}\footnote{http://www.icsi.berkeley.edu/$\sim$framenet/} \cite{Baker:al:1998},
which builds on the ideas for case frames of \namecite{Fillmore:1968}.
According to \namecite{Fillmore:1982}, frames are
``{\it characterizing a small abstract `scene' or `situation',
so that to understand the semantic structure of the verb it is necessary to
understand the properties of such schematized scenes}''.
In {\it FrameNet}, a lexical item evokes a number of other relations
and concepts representative of the context in which the word applies;
together these make up the speaker's understanding of the word.
For example, the meaning of a sentence such as `{\it Sara faxed Jeremy the invoice.}'
is not derived from the meaning of the verb {\it fax} alone,
but also from speakers' knowledge about situations
where somebody gives something to somebody else \cite{Goldberg:1995,Petruck:1996,Baker:Ruppenhofer:2002}.

In a sense, the above-described program- and human-generated paraphrasing verbs
represent exactly such kind of world knowledge:
a dynamically constructed semantic frame in terms of which the target noun compound is to be understood.
Therefore, it would not be unreasonable to expect similarity of the relations
between the entities in the manually created frames of {\it FrameNet}
and the ones I generate dynamically from the Web:
in particular, there should be some overlap between my automatically generated verbs
and the ones listed in the corresponding {\it FrameNet} frame.
If so, given a sentence, my verbs can help automatically select
the {\it FrameNet} frame that best characterizes the situation described in that sentence.

As a preliminary investigation of the potential of these ideas,
I compared the verbs I generate for four of Levi's relations
\texttt{CAUSE$_1$}, \texttt{USE}, \texttt{HAVE$_1$}, and \texttt{MAKE$_1$},
and the verbs listed in {\it FrameNet} for the frames,
I found to correspond to these relations.
I also tried to compare the {\it FrameNet} verbs to the human-proposed ones.
In both cases, I used the top 150 verbs for the target Levi relation,
as shown in appendix \ref{sec:verbs:all:class}.
The results are shown in Tables \ref{table:framenet:cause1}, \ref{table:framenet:use},
\ref{table:framenet:have1}, and \ref{table:framenet:make1}, respectively.
The results vary per relation, but overall
the performance of the human- and program-proposed verbs is comparable,
and the average overlap is over 50\%, which is quite promising.

\section{Application to Relational Similarity}
\label{sec:relational:similarity}

In this section, I extend the above method which characterizes the semantic relations
that hold between nouns in noun-noun compounds,
to measuring the semantic similarity between pairs of words,
i.e. relational similarity.
The approach leverages the vast size of the Web to building
linguistically-motivated lexical features.
Given a pair of words, it mines the Web for sentences containing these words
and then extracts verbs, prepositions, and coordinating conjunctions that connect them.
These lexical features are then used in a vector-space model to measure semantic similarity
that is needed for building instance-based classifiers.
The evaluation of the approach on several relational similarity problems,
including SAT verbal analogy, head-modifier relations, and relations between complex nominals
shows state-of-the-art performance.

\subsection{Introduction}
\label{sec:rel:sim:intro}

Measuring the semantic similarity between pairs of words,
i.e. relational similarity, is an important but understudied problem.
Despite the tremendous amount of computational
linguistics publications on word similarity (see
\cite{Budanitsky:Hirst:2006} for an overview), there is
surprisingly few work on relational similarity.
Students taking the SAT examination are familiar with verbal analogy
questions, where they need to decide whether, e.g., the relation
between {\it ostrich} and {\it bird} is more similar to the one between
{\it lion} and {\it cat}, or rather between {\it primate} and {\it
monkey}. These kinds of questions are hard; the average test taker
achieves about 56.8\% on the average \cite{Turney:Littman:2005:rel}.

Many NLP applications would benefit from solving the relational similarity problem,
e.g., question answering, information retrieval, machine translation,
word sense disambiguation, information extraction, etc.
While there are few success stories so far,
the ability to measure semantic similarity has already proven
its advantages for textual entailment \cite{Tatu:Moldovan:2005:entailment},
and the importance of the task is being realized:
there was a competition in 2007 on
{\it Classification of Semantic Relations between Nominals} as part of {\it SemEval},
and the journal of {\it Language Resources and Evaluation}
will have a special issue in 2008 on
{\it Computational Semantic Analysis of Language: SemEval-2007 and Beyond}.

Below I introduce a novel Web-based approach, which,
despite its simplicity, rivals the best previous approaches to relational similarity.
Following \namecite{Turney:2006:CL:rel:sim}, I use SAT verbal analogy as a benchmark problem.
I further address two semantic relation classification problems:
head-modifier relations, and relations between complex nominals.


\subsection{Method}

\subsubsection{Feature Extraction}
\label{sec:rel:sim:feat:ext}

Given a pair of nouns $noun_1$ and $noun_2$, I mine the Web for sentences containing them
and then I extract connecting verbs, prepositions, and coordinating conjunctions,
which I will later use as lexical features in a vector-space model
to measure semantic similarity between pairs of nouns.

The extraction process starts with a set of exact phrase queries
generated using the following patterns:

\begin{center}
    ``$infl_1$\texttt{ THAT * }$infl_2$''

    ``$infl_2$\texttt{ THAT * }$infl_1$''

    ``$infl_1$\texttt{ * }$infl_2$''

    ``$infl_2$\texttt{ * }$infl_1$''
\end{center}

\noindent
where:

\begin{center}
\parbox{5.3in}{
$infl_1$ and $infl_2$ are inflected variants of $noun_1$ and $noun_2$; 

\texttt{THAT} can be {\it that}, {\it which}, or {\it who};

and \texttt{*} stands for 0 or more (up to 8) stars, representing the {\it Google} \texttt{*} operator.
}
\end{center}
\vspace{18pt}

For each query, I collect the text snippets (summaries) from the
result set (up to 1,000 per query) and I split them into sentences.
I then filter out the incomplete sentences and the ones
that do not contain the target nouns, I assign POS annotations
using the OpenNLP tagger\footnote{OpenNLP: \texttt{http://opennlp.sourceforge.net}},
and I extract the following features:

{\bf Verb:} I extract a verb,
if the subject NP of that verb is headed by one of the target nouns
(or an inflected form of a target noun), and its direct object NP is
headed by the other target noun (or an inflected form).
For example, the verb {\it include} will be extracted
from ``The {\it committee} \underline{includes} many {\it members}.''
I also extract verbs from relative clauses, e.g.,
``This is a {\it committee} which \underline{includes} many {\it members}.''
Verb particles are also recognized, e.g.,
``The {\it committee} must \underline{rotate off} 1/3 of its {\it members}.''
I ignore modals and auxiliaries, but retain the passive {\it be}.
Finally, I lemmatize the main verb using {\it WordNet}'s morphological analyzer
{\it Morphy} \cite{Fellbaum:1998:wordnet}.

{\bf Verb+Preposition:} If the subject NP of a verb is headed by one
of the target nouns (or an inflected form),
and its indirect object is a PP containing an NP
which is headed by the other target noun (or an inflected form),
I extract the verb and the preposition heading that PP,
e.g., ``The thesis advisory {\it committee} \underline{consists of} three qualified {\it members}.''
As in the verb case, I also extract verb+preposition from relative phrases,
I include particles, I ignore modals and auxiliaries, and I lemmatize the verbs.

{\bf Preposition:} If one of the target nouns is the head of an NP that
contains a PP inside which there is an NP headed by the other target noun
(or an inflected form), I extract the preposition heading that PP,
e.g., ``The {\it members} \underline{of} the {\it committee} held a meeting.''

{\bf Coordinating conjunction:}  If the two target nouns are the heads of
coordinated NPs, I extract the coordinating conjunction.

In addition to the lexical part, for each extracted feature, I keep a direction.
Therefore the preposition {\it of} represents two different features in the following examples
``{\it member} \underline{of} the {\it committee}'' and ``{\it committee}
\underline{of} {\it members}''. See Table \ref{table:committeeMember} for examples.

\begin{table}
\begin{center}
\begin{tabular}{rlcc}
  {\bf Frequency} & {\bf Feature} & {\bf POS} & {\bf Direction}\\
 \hline
  2205 & of & P & $2\rightarrow1$\\
  1923 & be & V & $1\rightarrow2$\\
  771 & include & V & $1\rightarrow2$\\
  382 & serve on & V & $2\rightarrow1$\\
  189 & chair & V & $2\rightarrow1$\\
  189 & have & V & $1\rightarrow2$\\
  169 & consist of & V & $1\rightarrow2$\\
  148 & comprise & V & $1\rightarrow2$\\
  106 & sit on & V & $2\rightarrow1$\\
  81 & be chaired by & V & $1\rightarrow2$\\
  78 & appoint & V & $1\rightarrow2$\\
  77 & on & P & $2\rightarrow1$\\
  66 & and & C & $1\rightarrow2$\\
  66 & be elected & V & $1\rightarrow2$\\
  58 & replace & V & $1\rightarrow2$\\
  48 & lead & V & $2\rightarrow1$\\
  47 & be intended for & V & $1\rightarrow2$\\
  45 & join & V & $2\rightarrow1$\\
  45 & rotate off & V & $2\rightarrow1$\\
  44 & be signed up for & V & $2\rightarrow1$\\
  43 & notify & V & $1\rightarrow2$\\
  40 & provide that & V & $2\rightarrow1$\\
  39 & need & V & $1\rightarrow2$\\
  37 & stand & V & $2\rightarrow1$\\
  36 & be & V & $2\rightarrow1$\\
  36 & vote & V & $1\rightarrow2$\\
  36 & participate in & V & $2\rightarrow1$\\
  35 & allow & V & $1\rightarrow2$\\
  33 & advise & V & $2\rightarrow1$\\
  32 & inform & V & $1\rightarrow2$\\
  31 & form & V & $2\rightarrow1$\\
  $\ldots$ & $\ldots$ &  $\ldots$ & $\ldots$\\
 \hline
\end{tabular}

\caption{{\bf The most frequent Web-derived features for \emph{committee member}.}
Here $V$ stands for verb (possibly +preposition and/or +particle),
$P$ for preposition and $C$ for coordinating conjunction;
$1\rightarrow2$ means {\it committee} precedes the feature and {\it member} follows it;
$2\rightarrow1$ means {\it member} precedes the feature and {\it committee} follows it.}
\label{table:committeeMember}
\end{center}
\end{table}

The proposed method is similar in spirit to previous paraphrase
acquisition approaches which search for similar/fixed endpoints and collect the
intervening material.  \namecite{Lin:Pantel:2001:par} use a dependency parser
and extract paraphrases from dependency tree paths whose ends contain
semantically similar sets of words by generalizing over these ends.
For example, given the target phrase ``{\it X solves Y}'',
they extract paraphrases such as ``{\it X finds a solution to Y}'',
``{\it X tries to solve Y}'', ``{\it X resolves Y}'', ``{\it Y is resolved by X}'', etc.
The approach is extended by \namecite{Shinyama:al:2002:par}, who use named
entity recognizers and look for anchors belonging to a matching semantic class, e.g.,
\texttt{LOCATION}, \texttt{ORGANIZATION}, etc.
This latter idea is further extended by \namecite{Nakov:al:2004:citances},
who apply it in the biomedical domain, imposing the additional restriction
that the sentences from which the paraphrases are to be extracted
represent citation sentences (citances) that cite the same target paper;
this restriction yields higher accuracy.
While the objective of all these approaches is paraphrase acquisition,
here I am interested in extracting linguistic features that can express
the semantic relation between two nouns.

After having extracted the linguistic features,
I use them in a vector-space model
in order to measure semantic similarity between pairs of nouns.
This vector representation is similar to the ones proposed by
\namecite{Alshawi:Carter:1994}, \namecite{Grishman:Sterling:1994},
\namecite{Ruge:1992}, and \namecite{Lin:1998}. For example, the latter
measures {\it word} similarity using triples extracted from a
dependency parser. In particular, given a noun, \namecite{Lin:1998}
finds all verbs that have it as a subject or object, and all adjectives that modify
it, together with the corresponding frequencies.
Unlike this research, whose objective is measuring word similarity,
here I am interested in relational similarity.

The method is also similar to the ideas of \namecite{Devereux:Costello:2006}
that the meaning of a noun-noun compound is characterized by a distribution
over several dimensions, as opposed to be expressed by a single relation.
However, their relations are fixed and abstract,
while mine are dynamic and based on verbs, prepositions and coordinating conjunctions.
It is also similar to \namecite{kim:baldwin:2006:POS},
who characterize the relation using verbs; however they use a fixed set of seed verbs,
and my verbs are dynamically extracted.
My approach is also similar to that of \namecite{OSeaghdha:Copestake:2007}
who use grammatical relations as features to characterize a noun-noun compound;
however I use verbs, prepositions and coordinating conjunctions instead.

\subsubsection{Similarity Measure}

The above-described features are used in the calculation
of the similarity between noun pairs.
I use TF.IDF-weighting in order to downweight very common features like {\it of}:
\begin{equation}
w(x) = TF(x) \times \log{\left(\frac{N}{DF(x)} \right)}
\end{equation}

In the above formula,
$TF(x)$ is the number of times the feature $x$ has been extracted for the target noun pair,
$DF(x)$ is the total number of training noun pairs that have this feature,
and $N$ is the total number of training noun pairs.
%

I use these weights in a variant of the Dice coefficient.
The classic Dice coefficient for two sets $A$ and $B$ is defined as follows:
\begin{equation}
Dice(A,B) = \frac{2 \times |A \cap B|}{|A| + |B|}
\end{equation}

This definition applies to Boolean vectors as well
since they are equivalent to discrete sets,
but it does not apply to numerical vectors in general.
Therefore, I use the following generalized
definition\footnote{Other researchers have proposed different
extensions, e.g., \namecite{Lin:1998}.}:
\begin{equation}
Dice(A,B) = \frac{2 \times \sum_{i=1}^n \min(a_i, b_i)}{\sum_{i=1}^n
a_i + \sum_{i=1}^n b_i}
\end{equation}

A bigger value for the Dice coefficient indicates higher similarity.
Therefore, I take $\min(a_i,b_i)$ in order to avoid giving unbalanced weight to a
feature when the weights are lopsided.  For example, if the noun
pair ({\it committee}, {\it members}) has the feature {\it include} as a verb
in the forward direction 1,000 times,
and the noun pair ({\it ant}, {\it hill}) has it only twice,
I do not want to give a lot of weight for overlapping on that feature.

\subsection{SAT Verbal Analogy}

\begin{table}
\begin{center}
\begin{tabular}{ll|ll}
  & \multicolumn{1}{l}{{\bf ostrich}:{\bf bird}} & & {\bf palatable}:{\bf toothsome}\\
 \hline
 {\it (a)} & {\it lion}:{\it cat} & (a) & rancid:fragrant\\
 (b) & goose:flock & (b) & chewy:textured\\
 (c) & ewe:sheep & {\it (c)} & {\it coarse}:{\it rough}\\
 (d) & cub:bear & (d) & solitude:company\\
 (e) & primate:monkey & (e) & no choice\\
 \hline
\end{tabular}
\caption{{\bf SAT verbal analogy: sample questions.}
The stem is in {\bf bold}, the correct answer is in {\it italic},
and the distractors are in plain text.}
\label{table:SAT:examples}
\end{center}
\end{table}

Following \namecite{Turney:2006:CL:rel:sim}, I use {\it SAT verbal analogy} as a benchmark problem.
I experiment with {\it Turney's dataset}, which consists of 374 SAT questions
from various sources, including 190 from actual SAT tests,
80 from SAT guidebooks, 14 from the ETS Web site, and 90 from other
SAT test preparation Web sites.
Table \ref{table:SAT:examples} shows two example problems:
the top pairs are called {\it stems},
the ones in italic are the {\it solutions},
and the remaining ones are {\it distractors}.
\namecite{Turney:2006:CL:rel:sim} achieves 56\% accuracy,
which matches the 56.8\% average human performance,
and is a significant improvement over the 20\% random-guessing baseline.

Note that the righthand example in Table \ref{table:SAT:examples}
misses one distractor; so do 21 examples in {\it Turney's dataset}.
It also mixes different parts of speech:
while {\it solitude} and {\it company} are nouns, all remaining words are adjectives.
Other examples in {\it Turney's dataset} contain verbs and adverbs,
and even relate pairs of different part of speech.
This is problematic for my approach, which requires that both words be
nouns\footnote{The approach can be extended to handle adjective-noun pairs,
as demonstrated in section \ref{sec:head:modifier:rel} below.}.
After having filtered all examples containing non-nouns, 
I ended up with 184 questions, which I use in the evaluation.

\begin{table}
\begin{center}
\begin{tabular}{lccccc}
\multicolumn{1}{l}{\bf Model} & {\bf Correct} & {\bf Wrong} & \multicolumn{1}{c}{\bf N/A} & {\bf Accuracy} & \multicolumn{1}{c}{\bf Coverage}\\
 \hline
 $v+p+c$ & 129 & 52 & 3 & {\bf 71.27$\pm$6.98} & {\bf 98.37}\\
 $v$     & 122 & 56 & 6 & 68.54$\pm$7.15 & 96.74\\
 $v+p$   & 119 & 61 & 4 & 66.11$\pm$7.19 & 97.83\\
 $v+c$   & 117 & 62 & 5 & 65.36$\pm$7.23 & 97.28\\
 $p+c$   & 90  & 90 & 4 & 50.00$\pm$7.23 & 97.83\\
 $p$     & 84  & 94 & 6 & 47.19$\pm$7.20 & 96.74\\
 \hline
 baseline & 37 & 147 & 0 & 20.00$\pm$5.15 & 100.00\\
 LRA: \cite{Turney:2006:CL:rel:sim} & 122 & 59 & 3 & 67.40$\pm$7.13 & 98.37\\
 \hline
\end{tabular}
\caption{{\bf SAT verbal analogy: evaluation on 184 noun-only questions.}
For each model, the number of correctly classified,
wrongly classified, and non-classified examples is shown,
followed by the accuracy (in \%) and the coverage
(\% of examples for which the model makes prediction).
The letters $v$, $p$ and $c$ indicate the kinds of features used:
$v$ stands for verb (possibly +preposition),
$p$ for preposition, and $c$ for coordinating conjunction.}
\label{table:eval:SAT}
\end{center}
\end{table}

Given a SAT verbal analogy example, I build six feature vectors --
one for each of the six word pairs.
I calculate the similarity between the stem of the analogy
and each of the five candidates using the Dice coefficient with TF.IDF-weighting,
and I choose the pair with the highest score;
I make no prediction if two or more pairs tie for the highest score.


The evaluation results are shown in Table \ref{table:eval:SAT}:
I use leave-one-out cross validation since I need to set the TF.IDF weights.
The last line shows the performance of
Turney's Latent Relational Analysis (LRA) when limited to the 184 noun-only dataset.
My best model $v+p+c$ performs a bit better, 71.27\% vs. 67.40\%,
but the difference is not statistically significant
(Pearson's Chi-square test, see section \ref{sec:stat:significance}).
Note that this ``inferred'' accuracy could be misleading,
and the LRA would have performed better
if it was trained to solve {\it noun-only} analogies, which seem easier,
as demonstrated by the significant increase in accuracy for LRA when limited to nouns:
67.4\% vs. 56.8\% for 184 and 374 questions, respectively.
The learning process for LRA is probably harder on the full dataset,
e.g., its pattern discovery step needs to learn patterns for many POS
as opposed to nouns only, etc.

\subsection{Head-Modifier Relations}
\label{sec:head:modifier:rel}

\begin{table}
\begin{center}
\begin{tabular}{lcccccc}
\multicolumn{1}{l}{\bf Model} & {\bf Correct} & {\bf Wrong} & \multicolumn{1}{c}{\bf N/A} & {\bf Accuracy} & \multicolumn{1}{c}{\bf Coverage}\\
 \hline
 $v+p$      & 240  & 352  & 8  & {\bf 40.54$\pm$3.88} & {\bf 98.67}\\
 $v+p+c$    & 238  & 354  & 8  & 40.20$\pm$3.87 & 98.67\\
 $v$        & 234  & 350  & 16 & 40.07$\pm$3.90 & 97.33\\
 $v+c$      & 230  & 362  & 8  & 38.85$\pm$3.84 & 98.67\\
 $p+c$      & 114  & 471  & 15 & 19.49$\pm$3.01 & 97.50\\
 $p$        & 110  & 475  & 15 & 19.13$\pm$2.98 & 97.50\\
 \hline
 baseline     & 49  & 551 & 0 & 8.17$\pm$1.93  & 100.00\\
 LRA (Turney) & 239 & 361 & 0 & 39.83$\pm$3.84 & 100.00\\
 \hline
\end{tabular}
\caption{{\bf Head-modifier relations, 30 classes:
evaluation on \emph{Barker\&Szpakowicz dataset}.}
For each model, the number of correctly classified,
wrongly classified, and non-classified examples is shown,
followed by the accuracy (in \%) and the coverage
(\% of examples for which the model makes prediction).
Accuracy and coverage are micro-averaged.}
\label{table:eval:Barker:Szpakowicz:30}
\end{center}
\end{table}

Next, I experiment with a head-modifier dataset by \namecite{Barker:Szpakowicz:1998:nc:sem},
which contains head-modifier relations between noun-noun, adjective-noun and adverb-noun pairs.
The dataset contains 600 head-modifier examples,
each annotated with 30 fine-grained relations, grouped into 5 coarse-grained classes
(the covered fine-grained relations are shown in parentheses):
CAUSALITY ({\it cause, effect, purpose, detraction}), TEMPORALITY ({\it
frequency, time\_at, time\_through}), SPATIAL ({\it direction,
location, location\_at, location\_from}), PARTICIPANT ({\it agent,
beneficiary, instrument, object, object\_property, part, possessor,
property, product, source, stative, whole}) and QUALITY ({\it
container, content, equative, material, measure, topic, type}).
For example, {\it exam anxiety} is classified as {\it effect} and therefore as CASUALITY,
and {\it blue book} is {\it property} and therefore also PARTICIPANT.

\begin{table}
\begin{center}
\begin{tabular}{lcccccc}
\multicolumn{1}{l}{\bf Model} & {\bf Correct} & {\bf Wrong} & \multicolumn{1}{c}{\bf N/A} & {\bf Accuracy} & \multicolumn{1}{c}{\bf Coverage}\\
 \hline
 $v+p$    & 328  & 264  & 8  & {\bf 55.41$\pm$4.03} & {\bf 98.67}\\
 $v+p+c$  & 324  & 269  & 7  & 54.64$\pm$4.02 & 98.83\\
 $v$      & 317  & 267  & 16 & 54.28$\pm$4.06 & 97.33\\
 $v+c$    & 310  & 280  & 10 & 52.54$\pm$4.03 & 98.33\\
 $p+c$    & 240  & 345  & 15 & 41.03$\pm$3.91 & 97.50\\
 $p$      & 237  & 341  & 22 & 41.00$\pm$3.94 & 96.33\\
 \hline
 baseline     & 260 & 340 & 0 & 43.33$\pm$3.91 & 100.00\\
 LRA (Turney) & 348 & 252 & 0 & 58.00$\pm$3.99 & 100.00\\
 \hline
\end{tabular}
\caption{{\bf Head-modifier relations, 5 classes:
evaluation on \emph{Barker\&Szpakowicz dataset}.}
For each model, the number of correctly classified,
wrongly classified, and non-classified examples is shown,
followed by the accuracy (in \%) and the coverage
(\% of examples for which the model makes prediction).
Accuracy and coverage are micro-averaged.}
\label{table:eval:Barker:Szpakowicz:5}
\end{center}
\end{table}

There are some problematic examples in this dataset.
First, in three cases, there are two modifiers rather than one,
e.g., {\it \underline{infectious disease} agent}.
In these cases, I ignore the first modifier.
Second, seven examples have an adverb as a modifier, e.g., {\it \underline{daily} exercise},
and 262 examples have an adjective as a modifier, e.g., {\it \underline{tiny} cloud}.
I treat them as if the modifier was a noun, which works in many cases,
since many adjectives and adverbs can be used predicatively,
e.g., `{\it This exercise is performed \underline{daily}.}'
or `{\it This cloud looks very \underline{tiny}.}'

For the evaluation, I create a feature vector for each head-modifier pair,
and I perform a leave-one-out cross-validation:
I leave one example for testing and I train on the remaining 599;
I repeat this procedure 600 times, so that each example gets used for testing.
Following \namecite{Turney:Littman:2005:rel} and \namecite{Barker:Szpakowicz:1998:nc:sem},
I use a 1-nearest-neighbor classifier.
I calculate the similarity between the feature vector of the testing example
and the vectors of the training examples. 
If there is a single highest-scoring training example,
I predict its class for that test example. Otherwise, if
there are ties for first, I assume the class predicted
by the majority of the tied examples, if there is a majority.

The results for the 30-class and 5-class \emph{Barker\&Szpakowicz dataset}
are shown in Tables \ref{table:eval:Barker:Szpakowicz:30}
and \ref{table:eval:Barker:Szpakowicz:5}, respectively.
For the 30-way classification, my best model achieves 40.54\% accuracy,
which is comparable to the accuracy of Turney's LRA: 39.83\%.
For the 5-way classification, I achieve 55.41\% vs. 58.00\% for Turney's LRA.
In either case, the differences are not statistically significant
(tested with Pearson's Chi-square test, see section \ref{sec:stat:significance}).
Given that Turney's algorithm requires substantial resources, synonym expansion,
and costly computation over multiple machines,
I believe that my simple approach is preferable.

Overall, for \emph{Barker\&Szpakowicz dataset},
it is best to use verbs and prepositions ($v+p$); adding coordinating conjunctions
($v+p+c$) lowers the accuracy.
Using prepositions in addition to verbs ($v+p$) is better than using verbs only ($v$),
but combining verbs and coordinating conjunctions ($v+c$) lowers the accuracy.
Coordinating conjunctions only help when combined with prepositions ($p+c$).
Overall, verbs are the most important features, followed by prepositions.


The reason coordinating conjunctions cannot help increase the accuracy is that
head-modifier relations are typically expressed with verbal or prepositional paraphrase;
coordinating conjunctions only help with some infrequent relations, like {\it equative},
e.g., finding {\it player and coach} suggests an equative relation for
{\it player coach} or {\it coach player}.

This is different for SAT verbal analogy, where the best model is $v+p+c$,
as Table \ref{table:eval:SAT} shows.
Verbs are still the most important feature
and also the only one whose presence/absence makes a statistical
difference (as the confidence intervals show). However, this time using $c$ does help:
SAT verbal analogy questions ask for a broader range of
relations, e.g., antonyms, for which coordinating conjunctions like {\it but} can be helpful.

%
%

\subsection{Relations Between Nominals}

\begin{table}
\begin{center}
\begin{tabular}{lllcc}
   \multicolumn{1}{c}{\bf \#} & {\bf Relation Name} & \multicolumn{1}{c}{\bf Examples} & {\bf Train} & {\bf Test}\\
 \hline
   1 & Cause-Effect & hormone-growth, laugh-wrinkle & 140 & 80\\
   2 & Instrument-Agency & laser-printer, ax-murderer & 140 & 78\\
   3 & Product-Producer & honey-bee, philosopher-theory & 140 & 93\\
   4 & Origin-Entity & grain-alcohol, desert-storm & 140 & 81\\
   5 & Theme-Tool & work-force, copyright-law & 140 & 71\\
   6 & Part-Whole & leg-table, door-car & 140 & 72\\
   7 & Content-Container & apple-basket, plane-cargo & 140 & 74\\
 \hline
\end{tabular}
\caption{{\bf SemEval dataset}: relations with examples (context sentences are not shown).
Also shown are the number of training and testing instances for each relation.} \label{table:SemEval:dataset}
\end{center}
\end{table}

The last dataset I experiment with is from {\it SemEval'2007} 
competition Task \#4: {\it Classification of Semantic Relations between Nominals}
\cite{Girju:al:2007:semeval}. 
It contains a total of seven semantic relations,
not exhaustive and possibly overlapping, with 140 training
and about 70 testing examples per relation.
Table \ref{table:SemEval:dataset} lists the seven relations. 
This is a binary classification task and each relation is considered in isolation;
there are approximately 50\% negative and 50\% positive examples (``near misses'') per relation.

Each example consists of a sentence, the target semantic relation,
two nominals to be judged on whether they are in that relation,
manually annotated WordNet 3.0 sense keys for these nominals,
and the Web query used to obtain that example:

\vspace{12pt}
\begin{verbatim}
                "Among the contents of the <e1>vessel</e1>
                were a set of carpenter's <e2>tools</e2>,
                several large storage jars, ceramic
                utensils, ropes and remnants of food, as
                well as a heavy load of ballast stones."
                WordNet(e1) = "vessel%1:06:00::",
                WordNet(e2) = "tool%1:06:00::",
                Content-Container(e2, e1) = "true",
                Query = "contents of the * were a"
\end{verbatim}

Given an entity-annotated example sentence,
I reduce the target entities $e_1$ and $e_2$ to single nouns $noun_1$ and $noun_2$
keeping their last nouns only, which I assume to be the heads.
I then mine the Web for sentences containing both $noun_1$ and $noun_2$,
and I build feature vectors as above.
In addition to the Web-derived features, 
I use the following contextual ones:

{\bf Sentence word:} I use as features the words from the context sentence,
after stop words removal and stemming with the Porter stemmer \cite{Porter:1980:stem}.

{\bf Entity word:} I also use the lemmas of the words that are part of $e_1$ and $e_2$.

{\bf Query word:} Finally, I use the individual words that are part of the query string.
This last feature is used for category $C$ runs only (see below).

Participants in the competition were asked to classify their systems into categories $A$, $B$, $C$ and $D$,
depending on whether they use the manually annotated {\it WordNet} sense keys
and/or the {\it Google} query:

\vspace{6pt}
\begin{center}
\begin{tabular}{lll}
    $A$ & WordNet = NO & Query = NO;\\
    $B$ & WordNet = YES & Query = NO;\\
    $C$ & WordNet = NO & Query = YES;\\
    $D$ & WordNet = YES & Query = YES.\\
\end{tabular}
\end{center}
\vspace{6pt}

I believe that having the target entities annotated
with the correct {\it WordNet} sense keys
is an unrealistic assumption for a real world application.
Therefore, I use no {\it WordNet}-related features
and I only experiment with conditions $A$ and $C$.

As in section \ref{sec:head:modifier:rel} above,
I use a 1-nearest-neighbor classifier with a TF.IDF-weighted Dice coefficient.
If the classifier makes no prediction (due to ties),
I predict the majority class on the training data.
Regardless of classifier's prediction,
if the head nouns of the two entities $e_1$ and $e_2$ have the same lemma,
I classify the example as negative.

In addition to accuracy, I use precision, recall and $F$-measure, defined as follows:

\vspace{12pt}
\begin{center}
\parbox{3.7in}{
$P$ = Pr(manual label = {\scshape true} $|$ system guess = {\scshape true}) \\
$R$ = Pr(system guess = {\scshape true} $|$ manual label = {\scshape true}) \\
$F = (2 \times P \times R) / (P + R)$ \\
$Acc$ = Pr(system guess = manual label)
}
\end{center}
\vspace{12pt}

Tables \ref{table:semeval:eval:A} and \ref{table:semeval:eval:C}
show the results for my type $A$ and $C$ experiments for different amounts of training data:
45 ($A1$, $C1$), 90 ($A2$, $C2$), 105 ($A3$, $C3$), and 140 ($A4$, $C4$) examples.
All results are above the baseline: always propose
the majority label in the test set: `true' or `false'.
My category $C$ results are slightly but consistently better than my category $A$ results
for all four evaluation scores ($P$, $R$, $F$, $Acc$),
which suggests that knowing the query is helpful.
Interestingly, systems $A2$ and $C2$ perform worse than $A1$ and $C1$,
i.e., having more training data does not necessarily help with 1-nearest-neighbor classifiers.
In fact, my category $C$ system is the best-performing 
among the participating systems in {\it SemEval} task \#4,
and I have the third best results for category $A$.

Tables \ref{table:semeval:features1} and \ref{table:semeval:features2}
show additional analysis for models $A4$ and $C4$.
I study the effect of adding extra {\it Google} contexts (using up to 10 stars, rather than 8),
and using different subsets of features. I show the results for the following experimental conditions:
(a) 8 stars, leave-one-out cross-validation on the training data;
(b) 8 stars, testing on the test data; and
(c) 10 stars, testing on the test data.
As the tables show, using 10 stars yields a slight overall improvement in accuracy
for both $A4$ (from 65.4\% to 67.3\%) and $C4$ (from 67.0\% to 68.1\%).
Both results represent improvements over the best $A4$ and $C4$ systems
participating in \emph{SemEval}, which achieved 66\% and 67\% accuracy, respectively.

\begin{table}
\begin{center}
\begin{tabular}{ccclcccc}
{\bf Type} & {\bf Train} & {\bf Test} & {\bf Relation}          & {\bf P}   & {\bf R}  & {\bf F}    & {\bf Accuracy}\\
\hline
$A1$ &  35 & 80 & Cause-Effect      &  58.2 & 78.0 & 66.7 & 60.00$\pm$10.95\\
&  35 & 78 & Instrument-Agency    &  62.5 & 78.9 & 69.8 & 66.67$\pm$11.03\\
&  35 & 93 & Product-Producer     &  77.3 & 54.8 & 64.2 & 59.14$\pm$10.16\\
&  35 & 81 & Origin-Entity        &  67.9 & 52.8 & 59.4 & 67.90$\pm$10.78\\
&  35 & 71 & Theme-Tool           &  50.0 & 31.0 & 38.3 & 59.15$\pm$11.62\\
&  35 & 72 & Part-Whole           &  51.9 & 53.8 & 52.8 & 65.28$\pm$11.52\\
&  35 & 74 & Content-Container    &  62.2 & 60.5 & 61.3 & 60.81$\pm$11.39\\
&   & & {\bf Macro average}  & {\bf 61.4} & {\bf 58.6} & {\bf 58.9} & {\bf 62.7}\\
\hline
$A2$ & 70 & 80 & Cause-Effect      &  58.0 & 70.7 & 63.7 & 58.75$\pm$10.95\\
& 70 & 78 & Instrument-Agency    &  65.9 & 71.1 & 68.4 & 67.95$\pm$10.99\\
& 70 & 93 & Product-Producer     &  80.0 & 77.4 & 78.7 & 72.04$\pm$9.86\\
& 70 & 81 & Origin-Entity        &  60.6 & 55.6 & 58.0 & 64.20$\pm$10.86\\
& 70 & 71 & Theme-Tool           &  45.0 & 31.0 & 36.7 & 56.34$\pm$11.57\\
& 70 & 72 & Part-Whole           &  41.7 & 38.5 & 40.0 & 58.33$\pm$11.53\\
& 70 & 74 & Content-Container    &  56.4 & 57.9 & 57.1 & 55.41$\pm$11.31\\
& & & {\bf Macro average}  &  {\bf 58.2} & {\bf 57.5} & {\bf 57.5} & {\bf 61.9}\\
\hline
$A3$ &  105 & 80 & Cause-Effect      &  62.5 & 73.2 & 67.4 & 63.75$\pm$10.94\\
&  105 & 78 & Instrument-Agency    &  65.9 & 76.3 & 70.7 & 69.23$\pm$10.94\\
&  105 & 93 & Product-Producer     &  75.0 & 67.7 & 71.2 & 63.44$\pm$10.14\\
&  105 & 81 & Origin-Entity        &  48.4 & 41.7 & 44.8 & 54.32$\pm$10.80\\
&  105 & 71 & Theme-Tool           &  62.5 & 51.7 & 56.6 & 67.61$\pm$11.54\\
&  105 & 72 & Part-Whole           &  50.0 & 46.2 & 48.0 & 63.89$\pm$11.54\\
&  105 & 74 & Content-Container    &  64.9 & 63.2 & 64.0 & 63.51$\pm$11.38\\
&  & &  {\bf Macro average}  &  {\bf 61.3} & {\bf 60.0} & {\bf 60.4} & {\bf 63.7}\\
\hline
$A4$ &  140 & 80 & Cause-Effect      & 63.5 & 80.5 & 71.0 & 66.25$\pm$10.89\\
&  140 & 78 & Instrument-Agency    & 70.0 & 73.7 & 71.8 & 71.79$\pm$10.83\\
&  140 & 93 & Product-Producer     & 76.3 & 72.6 & 74.4 & 66.67$\pm$10.07\\
&  140 & 81 & Origin-Entity        & 50.0 & 47.2 & 48.6 & 55.56$\pm$10.83\\
&  140 & 71 & Theme-Tool           & 61.5 & 55.2 & 58.2 & 67.61$\pm$11.54\\
&  140 & 72 & Part-Whole           & 52.2 & 46.2 & 49.0 & 65.28$\pm$11.52\\
&  140 & 74 & Content-Container    & 65.8 & 65.8 & 65.8 & 64.86$\pm$11.36\\
&  & & {\bf Macro average}  & {\bf 62.7} & {\bf 63.0} & {\bf 62.7} & {\bf 65.4}\\
\hline
\multicolumn{4}{l}{\bf Baseline (majority)} &          81.3 & 42.9 & 30.8 &  57.0\\
\end{tabular}
\caption{{\bf Relations between nominals: evaluation results for systems of type $A$}
Shown are the number of training and testing examples used,
and the resulting precision, recall, $F$-measure and accuracy for each relation
and micro-averaged over the 7 relations.}
\label{table:semeval:eval:A}
\end{center}
\end{table}

\begin{table}
\begin{center}
\begin{tabular}{ccclcccc}
{\bf Type} & {\bf Train} & {\bf Test} & {\bf Relation}          & {\bf P}   & {\bf R}  & {\bf F}    & {\bf Accuracy}\\
\hline
$C1$ & 35 & 80 & Cause-Effect         &  58.5 & 75.6 & 66.0 & 60.00$\pm$10.95\\
& 35 & 78 & Instrument-Agency    &  65.2 & 78.9 & 71.4 & 69.23$\pm$10.94\\
& 35 & 93 & Product-Producer     &  81.4 & 56.5 & 66.7 & 62.37$\pm$10.15\\
& 35 & 81 & Origin-Entity        &  67.9 & 52.8 & 59.4 & 67.90$\pm$10.78\\
& 35 & 71 & Theme-Tool           &  50.0 & 31.0 & 38.3 & 59.15$\pm$11.62\\
& 35 & 72 & Part-Whole           &  51.9 & 53.8 & 52.8 & 65.28$\pm$11.52\\
& 35 & 74 & Content-Container    &  62.2 & 60.5 & 61.3 & 60.81$\pm$11.39\\
& & & {\bf Macro average}  &  {\bf 62.4} & {\bf 58.5} & {\bf 59.4} & {\bf 63.5}\\
\hline
$C2$ & 70 & 80 & Cause-Effect         &  58.0 & 70.7 & 63.7 & 58.75$\pm$10.95\\
& 70 & 78 & Instrument-Agency    &  67.5 & 71.1 & 69.2 & 69.23$\pm$10.94\\
& 70 & 93 & Product-Producer     &  80.3 & 79.0 & 79.7 & 73.12$\pm$9.79\\
& 70 & 81 & Origin-Entity        &  60.6 & 55.6 & 58.0 & 64.20$\pm$10.86\\
& 70 & 71 & Theme-Tool           &  50.0 & 37.9 & 43.1 & 59.15$\pm$11.62\\
& 70 & 72 & Part-Whole           &  43.5 & 38.5 & 40.8 & 59.72$\pm$11.54\\
& 70 & 74 & Content-Container    &  56.4 & 57.9 & 57.1 & 55.41$\pm$11.31\\
& & & {\bf Macro average}  &  {\bf 59.5} & {\bf 58.7} & {\bf 58.8} & {\bf 62.8}\\
\hline
$C3$ & 105 & 80 & Cause-Effect         &  62.5 & 73.2 & 67.4 & 63.75$\pm$10.94\\
& 105 & 78 & Instrument-Agency    &  68.2 & 78.9 & 73.2 & 71.79$\pm$10.83\\
& 105 & 93 & Product-Producer     &  74.1 & 69.4 & 71.7 & 63.44$\pm$10.14\\
& 105 & 81 & Origin-Entity        &  56.8 & 58.3 & 57.5 & 61.73$\pm$10.89\\
& 105 & 71 & Theme-Tool           &  62.5 & 51.7 & 56.6 & 67.61$\pm$11.54\\
& 105 & 72 & Part-Whole           &  50.0 & 42.3 & 45.8 & 63.89$\pm$11.54\\
& 105 & 74 & Content-Container    &  64.9 & 63.2 & 64.0 & 63.51$\pm$11.38\\
& & & {\bf Macro average}  &  {\bf 62.7} & {\bf 62.4} & {\bf 62.3} & {\bf 65.1}\\
\hline
$C4$ & 140 & 80 & Cause-Effect         &  63.5 & 80.5 & 71.0 & 66.25$\pm$10.89\\
& 140 & 78 & Instrument-Agency    &  70.7 & 76.3 & 73.4 & 73.08$\pm$10.75\\
& 140 & 93 & Product-Producer     &  76.7 & 74.2 & 75.4 & 67.74$\pm$10.04\\
& 140 & 81 & Origin-Entity        &  59.0 & 63.9 & 61.3 & 64.20$\pm$10.86\\
& 140 & 71 & Theme-Tool           &  63.0 & 58.6 & 60.7 & 69.01$\pm$11.50\\
& 140 & 72 & Part-Whole           &  52.2 & 46.2 & 49.0 & 65.28$\pm$11.52\\
& 140 & 74 & Content-Container    &  64.1 & 65.8 & 64.9 & 63.51$\pm$11.38\\
& & & {\bf Macro average}  &  {\bf 64.2} & {\bf 66.5} & {\bf 65.1} & {\bf 67.0}\\
\hline
\multicolumn{4}{l}{\bf Baseline (majority)} &          81.3 & 42.9 & 30.8 &  57.0\\
\end{tabular}
\caption{{\bf Relations between nominals: evaluation results for systems of type $C$}
Shown are the number of training and testing examples used,
and the resulting precision, recall, $F$-measure and accuracy for each relation
and micro-averaged over the 7 relations.}
\label{table:semeval:eval:C}
\end{center}
\end{table}

\begin{table}
\begin{center}
\begin{tabular}{lccc}
{\bf Features Used} & {\bf Leave-1-out: `8 *'} & {\bf Test: `10 *'} & {\bf Test: `8 *'}\\
\hline
\hline
{\bf Cause-Effect} & & \\
\hline
$sent$                 & 45.7 & 50.0$\pm$10.70 \\
$p$                    & 55.0 & 53.75$\pm$10.85 \\
$v$                    & 59.3 & 68.75$\pm$10.82 \\
$v+p$                  & 57.1 & 63.75$\pm$10.94 \\
$v+p+c$                & 70.5 & 67.50$\pm$10.86 \\
$v+p+c + sent$ ($A4$)        & 58.5 & 66.25$\pm$10.89 & 66.25$\pm$10.89\\
$v+p+c + sent + query$ ($C4$) & 59.3 & 66.25$\pm$10.89 & 66.25$\pm$10.89\\
\hline
\hline
{\bf Instrument-Agency} & & \\
\hline
$sent$                 & 63.6 & 58.97$\pm$11.09 \\
$p$                    & 62.1 & 70.51$\pm$10.89 \\
$v$                    & 71.4 & 69.23$\pm$10.94 \\
$v+p$                  & 70.7 & 70.51$\pm$10.89 \\
$v+p+c$                & 70.0 & 70.51$\pm$10.89 \\
$v+p+c + sent$ ($A4$)        & 68.6 & 71.79$\pm$10.83 & 71.79$\pm$10.83\\
$v+p+c + sent + query$ ($C4$) & 70.0 & 73.08$\pm$10.75 & 73.08$\pm$10.75\\
\hline
\hline
{\bf Product-Producer} & & \\
\hline
$sent$                 & 47.9 & 59.14$\pm$10.16 \\
$p$                    & 55.7 & 58.06$\pm$10.15 \\
$v$                    & 70.0 & 61.29$\pm$10.16 \\
$v+p$                  & 66.4 & 65.59$\pm$10.10 \\
$v+p+c$                & 67.1 & 65.59$\pm$10.10 \\
$v+p+c + sent$ ($A4$)        & 66.4 & 69.89$\pm$9.96 & 66.67$\pm$10.07\\
$v+p+c + sent + query$ ($C4$) & 67.9 & 69.89$\pm$9.96 & 66.67$\pm$10.07\\
\hline
\hline
{\bf Origin-Entity} & & \\
\hline
$sent$                 & 64.3 & 72.84$\pm$10.55 \\
$p$                    & 63.6 & 56.79$\pm$10.85 \\
$v$                    & 69.3 & 71.60$\pm$10.62 \\
$v+p$                  & 67.9 & 69.14$\pm$10.73 \\
$v+p+c$                & 66.4 & 70.37$\pm$10.68 \\
$v+p+c + sent$ ($A4$)        & 68.6 & 72.84$\pm$10.55 & 55.56$\pm$10.83\\
$v+p+c + sent + query$ ($C4$) & 67.9 & 72.84$\pm$10.55 & 64.20$\pm$10.86\\
\hline
\end{tabular}
\caption{{\bf Relations between nominals: accuracy for different features and amount of Web data.}
Shown is the accuracy for the following experimental conditions:
(a) 8 stars, leave-one-out cross-validation on the training data;
(b) 8 stars, testing on the test data; and
(c) 10 stars, testing on the test data. (part 1)}
\label{table:semeval:features1}
\end{center}
\end{table}

\begin{table}
\begin{center}
\begin{tabular}{lccc}
{\bf Features Used} & {\bf Leave-1-out: `8 *'} & {\bf Test: `10 *'} & {\bf Test: `8 *'}\\
\hline
\hline
{\bf Theme-Tool} & & \\
\hline
$sent$                 & 66.4 & 69.01$\pm$11.50 \\
$p$                    & 56.4 & 56.34$\pm$11.57 \\
$v$                    & 61.4 & 70.42$\pm$11.44 \\
$v+p$                  & 56.4 & 67.61$\pm$11.54 \\
$v+p+c$                & 57.1 & 69.01$\pm$11.50 \\
$v+p+c + sent$ ($A4$)        & 52.1 & 61.97$\pm$11.63 & 67.61$\pm$11.54\\
$v+p+c + sent + query$ ($C4$) & 52.9 & 61.97$\pm$11.63 & 69.01$\pm$11.50\\
\hline
\hline
$sent$                 & 47.1 & 51.39$\pm$11.32 \\
$p$                    & 57.1 & 54.17$\pm$11.43 \\
$v$                    & 60.0 & 66.67$\pm$11.49 \\
$v+p$                  & 62.1 & 63.89$\pm$11.54 \\
$v+p+c$                & 61.4 & 63.89$\pm$11.54 \\
$v+p+c + sent$ ($A4$)         & 60.0 & 61.11$\pm$11.55 & 65.28$\pm$11.52\\
$v+p+c + sent + query$ ($C4$) & 60.0 & 61.11$\pm$11.55 & 65.28$\pm$11.52\\
\hline
\hline
{\bf Content-Container} & & \\
\hline
$sent$                 & 56.4 & 54.05$\pm$11.27 \\
$p$                    & 57.9 & 59.46$\pm$11.38 \\
$v$                    & 71.4 & 67.57$\pm$11.30 \\
$v+p$                  & 72.1 & 67.57$\pm$11.30 \\
$v+p+c$                & 72.9 & 67.57$\pm$11.30 \\
$v+p+c + sent$ ($A4$)        & 69.3 & 67.57$\pm$11.30 & 64.86$\pm$11.36\\
$v+p+c + sent + query$ ($C4$) & 71.4 & 71.62$\pm$11.14 & 63.51$\pm$11.38\\
\hline
\hline
{\bf My average A4}                & & {\bf 67.3} & {\bf 65.4}\\
{\bf Best avg. A4 on \emph{SemEval}}    & & {\bf 66.0} & \\
\hline
{\bf My average C4} & & {\bf 68.1} & {\bf 67.0}\\
{\bf Best avg. C4 on \emph{SemEval}}    & & {\bf 67.0} & \\
\hline
\end{tabular}
\caption{{\bf Relations between nominals: accuracy for different features and amount of Web data.}
Shown is the accuracy for the following experimental conditions:
(a) 8 stars, leave-one-out cross-validation on the training data;
(b) 8 stars, testing on the test data; and
(c) 10 stars, testing on the test data. (part 2)}
\label{table:semeval:features2}
\end{center}
\end{table}

\section{Discussion}

The verbal paraphrases I generate can be useful for a number of NLP tasks,
e.g., for noun compound translation (in isolation) \cite{Baldwin:Tanaka:2004},
for paraphrase-augmented machine translation \cite{Callison-Burch:al:2006:mt},
for machine translation evaluation \cite{Russo-Lassner:2005,Kauchak:Barzilay:2006:par},
for summarization evaluation \cite{Zhou:al:2006:mt}, etc.
I will describe in chapter \ref{chapter:MT} an approach that integrates noun compound paraphrasing
using prepositions as part of the machine translation process.

As I have shown above, assuming annotated training data, the paraphrasing verbs
can be used as features in the prediction of abstract relations
like \texttt{TIME} and \texttt{LOCATION},
which can be helpful for other applications.
For example, \namecite{Tatu:Moldovan:2005:entailment} achieve state-of-the-art
results on the PASCAL Recognizing Textual Entailment challenge
by making use of such abstract relations.
See also appendix \ref{chapter:intro:illustration} for a detailed description
of the task and for a discussion of how verbal paraphrases can be utilized directly.



In information retrieval, the paraphrasing verbs can be used for index
normalization \cite{Zhai:1997:nc:parsing}, query expansion, query refinement,
results ranking, etc. For example, when querying for {\it migraine treatment},
pages containing good paraphrasing verbs like {\it relieve} or {\it prevent}
could be preferred.

In data mining, the paraphrasing verbs can be used
to seed a Web search that looks for particular classes
of NPs \cite{Etzioni:al:2005:NER}, such as diseases, drugs, etc.
For example, after having found that {\it prevent} is a good paraphrase for
{\it migraine treatment}, I can use the query
\texttt{"* which prevents migraines"} to obtain different
treatments/drugs for migraine, e.g.
{\it feverfew}, {\it Topamax}, {\it natural treatment},
{\it magnesium}, {\it Botox}, {\it Glucosamine}, etc.
Using a different paraphrasing verb, e.g., using \texttt{"* reduces migraine"}
can produce additional results:
{\it lamotrigine}, {\it PFO closure}, {\it Butterbur Root}, {\it Clopidogrel},
{\it topamax}, {\it anticonvulsant}, {\it valproate},
{\it closure of patent foramen ovale}, {\it Fibromyalgia topamax},
{\it plant root extract}, {\it Petadolex},
{\it Antiepileptic Drug Keppra (Levetiracetam)}, {\it feverfew}, {\it Propranolol}, etc.

This is similar to the idea of a relational Web search of
\namecite{Cafarella:al:2006:relational:web:search},
whose system {\sc TextRunner} serves four types of
relational queries, among which is one asking for all entities that
are in a particular relation with a given entity,
e.g., ``{\it find all $X$ such that $X$ prevents migraines}''.

\section{Conclusion and Future Work}

I have presented a simple unsupervised approach to noun-noun compound
interpretation in terms of the predicates that can be used to paraphrase
the hidden relation between the two nouns,
which could be potentially useful for many NLP tasks.
An important advantage of the approach is that it does not require
knowledge about the meaning of the constituent nouns in order to correctly
assign relations. A potential drawback is that it might not work well for low-frequency words.

I have further extended the method to measuring the semantic similarity
between pairs of words, i.e. relational similarity.
The evaluation of the approach on several relational similarity problems,
including SAT verbal analogy, head-modifier relations, and relations between complex nominals
has shown state-of-the-art performance.

The presented approach can be further extended to other combinations
of parts of speech; not just noun-noun and adjective-noun.
Using a full parser,
or a dependency parser with a richer set of dependency features,
e.g., as proposed by \namecite{Pado:Lapata:2007},
is another promising direction for future work.


\chapter{Improving Machine Translation with Noun Compound Paraphrases}
\label{chapter:MT}

This chapter describes an application
of the methods developed in chapters \ref{chapter:NC:bracketing} and \ref{chapter:NC:semantics}
for noun compound paraphrasing to an important real-world task: {\it machine translation}.
I propose a novel monolingual paraphrasing method
based on syntactic transformations at the NP-level,
which augments the training data with nearly equivalent 
sentence-level syntactic paraphrases of the original corpus,
focused on the noun compounds.
The idea is to recursively generate sentence variants
where noun compounds are paraphrased using suitable prepositions, 
and vice-versa -- NPs with an internal PP-attachment are turned into noun compounds.
The evaluation results show an improvement equivalent to 33\%-–50\% of that
of doubling the amount of training data.
The general idea of the approach was described in \cite{Nakov:Hearst:2007:WMT}.

\section{Introduction}

Most modern Statistical Machine Translation systems rely on
aligned sentences of bilingual corpora for training, from which they
learn how to translate small pieces of text. In many cases, the
learned pieces are equivalent but syntactically different from the
text given at translation time, and the potential for a high-quality
translation is missed.


In this section, I describe a method for expanding the training set
with conceptually similar but syntactically differing paraphrases.
I paraphrase source language side training sentences,
focusing on noun phrases (NPs) and noun compounds (NCs),
which have been previously shown to be very frequent:
\namecite{Koehn:Knight:2003:mt} observe that roughly half of the words in news texts are covered
by NPs/PPs, and \namecite{Baldwin:Tanaka:2004} report that 2.6\% of the tokens
in the {\it British National Corpus} and 3.9\% of the tokens in the {\it Reuters corpus}
are covered by noun compounds.

The proposed approach is novel in that it augments the training corpus with
paraphrases of the original sentences, thus increasing the training
set without increasing the number of training translation pairs needed.
It is also monolingual; other related approaches map from the source
language to other languages in order to obtain paraphrases.  And it is general
enough to be domain independent, although the paraphrasing rules I use
are currently English-specific.

In the experiments below, I show that the proposed monolingual paraphrasing approach
can improve {\it Bleu} scores \cite{Papineni:Roukos:Ward:Zhu:2002}
on small training datasets when translating from English to Spanish.


\section{Related Work}

Recent work in automatic corpus-based paraphrasing includes using
bilingual corpora to find alternative expressions for the same term,
e.g., \cite{Barzilay:McKeown:2001:par} and \cite{Pang:Knight:Marcu:2003:par},
or multiple expressions of the same concept in one language,
as in newswire text \cite{Shinyama:al:2002:par}.

Recently, paraphrases have been used to improve the
{\it evaluation} of machine translation systems.
\namecite{Kauchak:Barzilay:2006:par} argue that
automated evaluation measures like {\it Bleu}
end up comparing $n$-gram overlap
rather than semantic similarity to a reference text.
They performed an experiment asking two human translators to translate 10,000 sentences,
and found that less than 0.2\% of the translations were identical,
and 60\% differed by more than ten words.
Therefore, they proposed an evaluation method that includes paraphrasing
of the machine-produced translations, and demonstrated improved correlation
with human judgements compared to {\it Bleu}.
In a similar spirit, \namecite{Zhou:al:2006:mt} use
a paraphrase table extracted from a bilingual corpus to improve
evaluation of automatic summarization algorithms.

My approach is most closely related to that of
\namecite{Callison-Burch:al:2006:mt}, who
translate English sentences into Spanish and French by trying to
substitute unknown source phrases with suitable paraphrases. The
paraphrases are extracted using the bilingual method of
\namecite{Bannard:Callison-Burch:2005:par}, which finds paraphrases in
one language using a phrase in a second language as a pivot.  For
example, if in a parallel English-German corpus, the English phrases
{\it under control} and {\it in check} happen to be aligned (in
different sentences) to the same German phrase {\it unter controlle},
they would be hypothesized to be paraphrases of each other with some
probability.
\namecite{Callison-Burch:al:2006:mt} extract
English paraphrases using as pivots eight additional languages
from the {\it Europarl} corpus \cite{koehn:2005:europarl}.  These paraphrases are then
incorporated in the machine translation process by adding them as
additional entries in the English-Spanish phrase table and pairing
them with the foreign translation of the original phrase.  Finally,
the system is tuned using minimum error rate training
\cite{Och:2003:mert} with an extra feature penalizing the low
probability paraphrases. The system achieved dramatic increases in
coverage (from 48\% to 90\% of the test word types when using 10,000 training
sentence pairs), and notable increase on {\it Bleu} (up to
1.5 points). However, the method requires large multi-lingual parallel
corpora, which makes it domain-dependent
and most likely limits its source language to Chinese, Arabic, and the
languages of the EU.

Another important related research effort is in translating units of
text smaller than a sentence, e.g., noun-noun compounds
\cite{Grefenstette:1999:web,Tanaka:Baldwin:2003:mt,Baldwin:Tanaka:2004},
noun phrases \cite{Cao:Li:2002:baseNP,Koehn:Knight:2003:mt}, named entities
\cite{Al-Onaizan:Knight:2001:mt}, and technical terms
\cite{Nagata:al:2001:web}.  Although I focus on paraphrasing NPs,
unlike the work above, I paraphrase and translate full sentences,
as opposed to translating NPs as a stand-alone component.

\section{Method}

I propose a novel approach for improving statistical machine
translation using monolingual paraphrases.  Given a
sentence from the source (English) side of the training corpus,
I generate conservative meaning-preserving syntactic paraphrases of the
sentence, and I append these paraphrased sentences to the training
corpus.  Each paraphrased sentence is paired with the foreign
(Spanish) translation that is associated with the original sentence in
the training data.  This augmented training corpus can then be used to
train a statistical machine translation (SMT) system.

I also introduce a variation on this idea that can be used with a
phrase-based SMT.  In this alternative, the source-language phrases
from the phrase table are paraphrased, but again using the source
language phrases for the paraphrasing, as opposed to using foreign
aligned corpora to modify the phrases, as done by
\namecite{Callison-Burch:al:2006:mt}.
I also combine these approaches.



\section{Paraphrasing}
\label{sec:par}

 \begin{figure}
   \center
   \includegraphics[width=440pt]{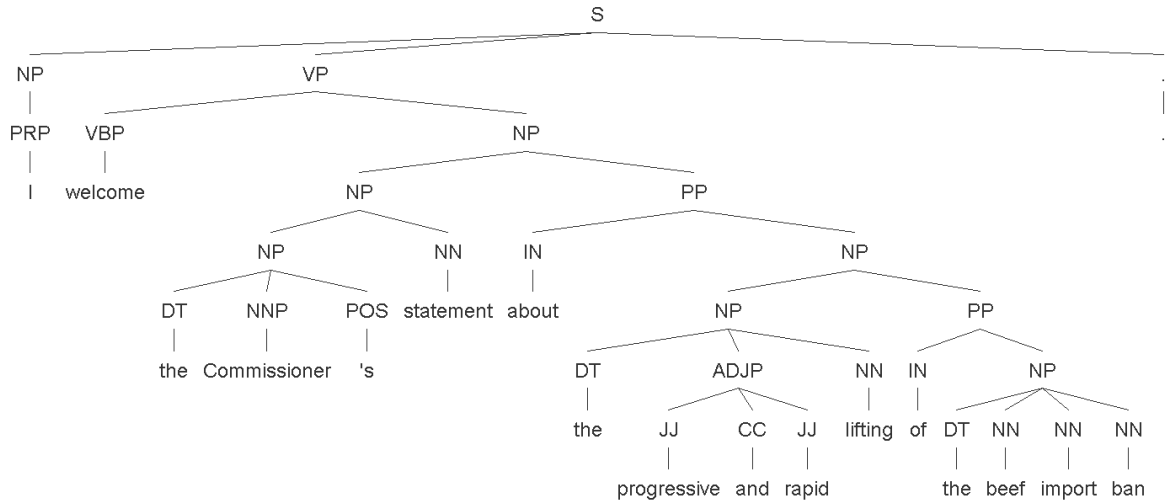}\\
   \caption{\textbf{Sample parse tree generated by the \emph{Stanford parser}.}
            I transform noun compounds into NPs with an internal PP-attachment;
            I turn NPs with an internal PP-attachment into noun compounds,
            or into NPs with an internal possessive marker;
            and I remove possessive markers whenever possible,
            or substitute them with {\it of}.
            All these transformations are applied recursively.}
   \label{Fig:parse}
 \end{figure}

\begin{table}
\begin{center}
\begin{tabular}{|l@{}|}
  \hline
    \textbf{\scriptsize I welcome the Commissioner 's statement about the progressive and rapid beef import ban lifting .}\\
    {\scriptsize I welcome the progressive and rapid beef import ban lifting Commissioner 's statement .}\\
    {\scriptsize I welcome the Commissioner 's statement about the beef import ban 's progressive and rapid lifting .}\\
    {\scriptsize I welcome the beef import ban 's progressive and rapid lifting Commissioner 's statement .}\\
    {\scriptsize I welcome the Commissioner 's statement about the progressive and rapid lifting of the {\it ban on beef imports} .}\\
    {\scriptsize I welcome the Commissioner statement about the progressive and rapid lifting of the beef import ban .}\\
    {\scriptsize I welcome the Commissioner statement about the progressive and rapid beef import ban lifting .}\\
    {\scriptsize I welcome the progressive and rapid beef import ban lifting Commissioner statement .}\\
    {\scriptsize I welcome the Commissioner statement about the beef import ban 's progressive and rapid lifting .}\\
    {\scriptsize I welcome the beef import ban 's progressive and rapid lifting Commissioner statement .}\\
    {\scriptsize I welcome the Commissioner statement about the progressive and rapid lifting of the {\it ban on beef imports} .}\\
    {\scriptsize I welcome the statement of Commissioner about the progressive and rapid lifting of the beef import ban .}\\
    {\scriptsize I welcome the statement of Commissioner about the progressive and rapid beef import ban lifting .}\\
    {\scriptsize I welcome the statement of Commissioner about the beef import ban 's progressive and rapid lifting .}\\
    {\scriptsize I welcome the statement of Commissioner about the progressive and rapid lifting of the {\it ban on beef imports} .}\\
    {\scriptsize I welcome the statement of the Commissioner about the progressive and rapid lifting of the beef import ban .}\\
    {\scriptsize I welcome the statement of the Commissioner about the progressive and rapid beef import ban lifting .}\\
    {\scriptsize I welcome the statement of the Commissioner about the beef import ban 's progressive and rapid lifting .}\\
    {\scriptsize I welcome the statement of the Commissioner about the progressive and rapid lifting of the {\it ban on beef imports} . }\\
\hline
  \textbf{\scriptsize The EU budget , as an instrument of economic policy ,
  amounts to 1.25 \% of European GDP .} \\
  {\scriptsize The EU budget , as an economic policy instrument , amounts to 1.25 \% of European GDP .}\\
  {\scriptsize The EU budget , as an economic policy 's instrument , amounts to 1.25 \% of European GDP .}\\
  {\scriptsize The {\it EU 's budget} , as an instrument of economic policy , amounts to 1.25 \% of European GDP .}\\
  {\scriptsize The {\it EU 's budget} , as an economic policy instrument , amounts to 1.25 \% of European GDP .}\\
  {\scriptsize The {\it EU 's budget} , as an economic policy 's instrument , amounts to 1.25 \% of European GDP .}\\
  {\scriptsize The {\it budget of the EU} , as an instrument of economic policy , amounts to 1.25 \% of European GDP .}\\
  {\scriptsize The {\it budget of the EU} , as an economic policy instrument , amounts to 1.25 \% of European GDP .}\\
  {\scriptsize The {\it budget of the EU} , as an economic policy 's instrument ,
amounts to 1.25 \% of European GDP .}\\
\hline
  \textbf{\scriptsize We must cooperate internationally , and this should
  include UN initiatives .}\\
  {\scriptsize We must cooperate internationally , and this should include {\it initiatives of the UN} .}\\
  {\scriptsize We must cooperate internationally , and this should include {\it initiatives at the UN} .}\\
  {\scriptsize We must cooperate internationally , and this should include {\it initiatives in the UN} .}\\
\hline
  \textbf{\scriptsize Both reports on economic policy confirm the impression
  that environment policy is only a stepchild .}\\
  {\scriptsize Both reports on economic policy confirm the impression that {\it policy on the environment} is only a stepchild .}\\
  {\scriptsize Both reports on economic policy confirm the impression that {\it policy on environment} is only a stepchild .}\\
  {\scriptsize Both reports on economic policy confirm the impression that {\it policy for the environment} is only a stepchild .}\\
  {\scriptsize Both economic policy reports confirm the impression that environment policy is only a stepchild .}\\
  {\scriptsize Both economic policy reports confirm the impression that {\it policy on the environment} is only a stepchild .}\\
  {\scriptsize Both economic policy reports confirm the impression that {\it policy on environment} is only a stepchild .}\\
  {\scriptsize Both economic policy reports confirm the impression that {\it policy for the environment} is only a stepchild .}\\
\hline
\end{tabular}
    \caption{\textbf{Sample sentences and their automatically generated paraphrases.}
            Paraphrased noun compounds are in italics.}
    \label{table:par}
\end{center}
\end{table}

\noindent Consider the sentence:

\begin{quote}
``{\it I welcome the Commissioner's statement about the progressive and
  rapid lifting of the beef import ban.}''.
\end{quote}

The corresponding syntactic parse tree,
produced using the {\it Stanford syntactic parser} \cite{Klein:Manning:2003},
is shown in Figure \ref{Fig:parse}.
From this parse, I generate paraphrases using
the six syntactic transformations shown below,
which are performed recursively.
Table \ref{table:par} shows some sentences and the corresponding
paraphrases for them, including all the paraphrases I generate
from the parse tree in Figure \ref{Fig:parse}.

\begin{enumerate}
\parsep         -2pt
\itemsep        -2pt

  \item {\bf [$_{\mathbf{NP}}$ NP$_1$ P NP$_2$] $\Rightarrow$ [$_{\mathbf{NP}}$ NP$_2$ NP$_1$].}\\
        {\it the lifting of the beef import ban} $\Rightarrow$
        {\it the beef import ban lifting}

  \item {\bf [$_{\mathbf{NP}}$ NP$_1$ \emph{of} NP$_2$] $\Rightarrow$ [$_{\mathbf{NP}}$ NP$_2$ poss NP$_1$].}\\
        {\it the lifting of the beef import ban} $\Rightarrow$
        {\it the beef import ban's lifting}

  \item {\bf NP$_{poss}$ $\Rightarrow$ NP.}\\
        {\it Commissioner's statement} $\Rightarrow$
        {\it Commissioner statement}

  \item {\bf NP$_{poss}$ $\Rightarrow$ NP$_{PP_{of}}$.}\\
        {\it Commissioner's statement} $\Rightarrow$
        {\it statement of (the) Commissioner}

  \item {\bf NP$_{NC}$ $\Rightarrow$ NP$_{poss}$.}\\
        {\it inquiry committee chairman} $\Rightarrow$
         {\it inquiry committee's chairman}

  \item {\bf NP$_{NC}$ $\Rightarrow$ NP$_{PP}$.}\\
        {\it the beef import ban} $\Rightarrow$
        {\it the ban on beef import}
\end{enumerate}

\noindent where:\\
\begin{tabular}{ll}
  {\bf poss}           & possessive marker: ' or 's;\\
  {\bf P}              & preposition;\\
  {\bf NP$_{PP}$}      & NP with an internal PP-attachment;\\
  {\bf NP$_{PP_{of}}$} & NP with an internal PP headed by {\it of};\\
  {\bf NP$_{poss}$}    & NP with an internal possessive marker;\\
  {\bf NP$_{NC}$}      & NP that is a noun compound.\\
\end{tabular}
\\
\\

In order to prevent transformations (1) and (2) from constructing
awkward NPs, I impose certain limitations on NP$_1$ and NP$_2$.  They
cannot span a verb, a preposition or a quotation mark (although they can
contain some kinds of nested phrases, e.g., an ADJP in case of
coordinated adjectives, as in {\it the progressive and controlled
lifting}).  Thus, the phrase {\it reduction in the taxation of labour}
is not transformed into {\it taxation of labour reduction} or {\it
taxation of labour's reduction}.
I further require the head to be a noun and do not allow it to be
an indefinite pronoun like {\it anyone}, {\it everybody}, and {\it someone}.


Transformations (1) and (2) are more complex than they may look.
In order to be able to handle some hard cases,
I apply additional restrictions and transformations.

First, some determiners, pre-determiners and possessive adjectives
must be eliminated in case of conflict between NP$_1$ and NP$_2$,
e.g., {\it \underline{the} lifting of \underline{this} ban}
can be paraphrased as {\it \underline{the} ban lifting},
but not as {\it \underline{this} ban's lifting}.

Second, in case both NP$_1$ and NP$_2$ contain adjectives,
these adjectives have to be put in the right order,
e.g., {\it the \underline{first} statement of the \underline{new} commissioner}
can be paraphrased as {\it the \underline{first} \underline{new} commissioner's statement},
but not {\it the \underline{new} \underline{first} commissioner's statement}.
There is also the option of not re-ordering them, e.g.,
{\it the \underline{new} commissioner's \underline{first} statement}.

Third, further complications are due to scope ambiguities
of modifiers of NP$_1$.  For example, in {\it the first statement of the new
commissioner}, the scope of the adjective {\it first} is not {\it statement} alone,
but {\it statement of the new commissioner}.  This is very different for the NP
{\it the biggest problem of the whole idea}, where the adjective {\it
biggest} applies to {\it problem} only, and therefore it cannot
be transformed to {\it the biggest whole idea's problem}
(although I do allow for {\it the whole idea's biggest problem}).

While the first four transformations are purely syntactic, (5) and (6)
are not.  The algorithm must determine whether a possessive marker is
feasible for (5) and must choose the correct preposition for (6). In
both cases, for noun compounds of length three or more, I also need to
choose the correct position to modify,
e.g., {\it inquiry's committee chairman} vs. {\it inquiry committee's chairman}.

In order to ensure accuracy of the paraphrases,
I use a variation of the Web-based approaches
presented in chapter \ref{chapter:NC:bracketing},
generating and testing the paraphrases
in the context of the preceding
and the following word in the sentence.
First, I split the noun compound into two sub-parts
$N_1$ and $N_2$ in all possible ways, e.g., {\it beef import ban lifting} would be split as:
(a) $N_1$=``{\it beef}'', $N_2$=``{\it import ban lifting}'',
(b) $N_1$=``{\it beef import}'', $N_2$=``{\it ban lifting}'', and
(c) $N_1$=``{\it beef import ban}'', $N_2$=``{\it lifting}''.
For each split, I issue exact phrase queries
to {\it Google} using the following patterns:

\texttt{"lt $N_1$ poss $N_2$ rt"}

\texttt{"lt $N_2$ prep det $N_1'$ rt"}

\texttt{"lt $N_2$ that be det $N_1'$ rt"}

\texttt{"lt $N_2$ that be prep det $N_1'$ rt"}

\noindent where:

\texttt{$N_1'$} can be a singular or a plural form of $N_1$;

\texttt{lt} is the word preceding $N_1$ in the original sentence or empty if none;

\texttt{rt} is the word following $N_2$ in the original sentence or empty if none;

\texttt{poss} is a possessive marker ('s or ');

\texttt{that} is {\it that}, {\it which} or {\it who};

\texttt{be} is {\it is} or {\it are};

\texttt{det} is a determiner ({\it the}, {\it a}, {\it an}, or none);

\texttt{prep} is one of the eight prepositions
used by \namecite{lauer:1995:thesis} for noun compound interpretation: {\it about}, {\it at}, {\it for}, {\it from},
{\it in}, {\it of}, {\it on}, and {\it with}.

Given a split, I collect the number of page hits for each
instantiation of the above paraphrase patterns filtering out the ones whose
page hit counts are less than ten.  I then calculate the total number of
page hits $H$ for all paraphrases (for all splits and all patterns),
and I retain the ones whose page hits counts are at least 10\% of
$H$, which allows for multiple paraphrases (possibly corresponding to different splits)
for a given noun compound.
If no paraphrases are retained, I repeat the above procedure with
\texttt{lt} set to the empty string.  If there are still no good
paraphrases, I set \texttt{rt} to the empty string.  If this does
not help either, I make a final attempt, by setting both \texttt{lt}
and \texttt{rt} to the empty string.

The paraphrased NCs are shown in italics in Table \ref{table:par},
e.g., {\it EU budget} is paraphrased as {\it EU\underline{'s} budget} and
{\it budget \underline{of the} EU};
also {\it environment policy} becomes {\it policy \underline{on} environment},
{\it policy \underline{on the} environment},
and {\it policy \underline{for the} environment};
and {\it UN initiatives} is paraphrased as {\it initiatives \underline{of the} UN},
{\it initiatives \underline{at the} UN}, and {\it initiatives \underline{in the} UN}.

I apply the same algorithm to paraphrasing English {\it phrases} from
the phrase table, but without transformations (5) and (6).
See Table \ref{table:parPhraseTable} for sample paraphrases.

\begin{table}
  \centering
\newcommand\T{\rule{0pt}{2.6ex}}
\newcommand\B{\rule[-1.2ex]{0pt}{0pt}}
\begin{small}
\begin{tabular}{|ll|}
  \hline
{\bf 1} & {\bf \% of members of the irish parliament} \T \\
 & \% of irish parliament members\\
 & \% of irish parliament 's members\\
  \hline
{\bf 2} & {\bf universal service of quality .} \T \\
 & universal quality service .\\
 & quality universal service .\\
 & quality 's universal service .\\
  \hline
{\bf 3} & {\bf action at community level} \T \\
 & community level action\\
  \hline
{\bf 4} & {\bf , and the aptitude for communication and} \T \\
 & , and the communication aptitude and\\
  \hline
{\bf 5} & {\bf to the fall-out from chernobyl .} \T \\
 & to the chernobyl fall-out .\\
 \hline
{\bf 6} & {\bf flexibility in development - and quick} \T \\
 & development flexibility - and quick\\
 \hline
{\bf 7} & {\bf , however , the committee on transport} \T \\
 & , however , the transport committee\\
 \hline
{\bf 8} & {\bf and the danger of infection with aids} \T \\
 & and the danger of aids infection\\
 & and the aids infection danger\\
 & and the aids infection 's danger\\
 \hline
\end{tabular}
\end{small}
  \caption{\textbf{Sample English phrases from the phrase table
                    and corresponding automatically generated paraphrases.}}
  \label{table:parPhraseTable}
\end{table}

\section{Evaluation}

I trained and evaluated several English to Spanish phrase-based SMT
systems using the standard training, development and test sets
of the {\it Europarl} corpus \cite{koehn:2005:europarl}.

First, I generate two directed word-level alignments, English-Spanish and Spanish-English,
using IBM model 4 \cite{Brown:al:1993:mt},
and I combine them using the {\it interect+grow heuristic} described in \namecite{Och:Ney:2003:mt}.
Then I extract phrase-level translation pairs
using the {\it alignment template approach} \cite{Och:Ney:2004:mt}.
The set of pairs of English phrases and their translations into Spanish form a phrase
table where each pair is associated with five parameters:
a forward phrase translation probability, a reverse phrase translation probability,
a forward lexical translation probability, a reverse lexical
translation probability, and a phrase penalty.

I train a log-linear model using the standard set of feature functions
of {\sc Pharaoh} \cite{Koehn:2004:pharaoh}:
language model probability, word penalty, distortion cost, and the five
parameters from the phrase table. I set the feature weights
by optimizing the {\it Bleu} score directly
using {\it minimum error rate training} \cite{Och:2003:mert}
on the first 500 sentences from the development set.
Then, I use the {\sc Pharaoh} beam search decoder \cite{Koehn:2004:pharaoh}
in order to produce the translations for the 2,000 test sentences.
Finally, these translations are compared to the gold standard test set
using {\it Bleu}.

Using the above procedure, I build and evaluate a baseline system
$S$, trained on the original training corpus.
The second system I build, $S_{parW}$, uses a version of the training corpus
which is augmented with syntactic paraphrases of the sentences from the English side
of the training corpus paired with the corresponding Spanish translations
of the original English sentences.
I also build a new system $S_{par}$, which excludes
syntactic transformations (5) and (6) from the paraphrasing process
in order to see the impact of not breaking noun compounds and not using the Web.

System $S^\star$ paraphrases and augments the phrase table
of the baseline system $S$ using syntactic transformations (1)-(4),
similarly to $S_{par}$, i.e. without paraphrasing noun compounds.
Similarly, $S^\star_{parW}$ is obtained by paraphrasing the phrase table of $S_{parW}$.


Finally, I merge the phrase tables for some of the above systems,
which I designate with a ``$+$'', e.g., $S + S_{parW}$ and $S^\star + S^\star_{parW}$.
In these merges, the phrases from the first phrase table
are given priority over those from the second one,
in case a phrase pair is present in both phrase tables.
This is important since the parameters estimated
from the original corpus are more reliable.

Following \namecite{Bannard:Callison-Burch:2005:par}, I also perform an
experiment with an additional feature $F_{parW}$ for each
phrase: Its value is 1 if the phrase is in the phrase table of $S$,
and 0.5 if it comes from the phrase table of $S_{parW}$.  As before,
I optimize the weights using minimum error rate training.  For $S^\star + S_{parW}^\star$,
I also try using two features: in addition to
$F_{parW}$, we I introduce $F_{\star}$, whose value is 0.5 if the
phrase comes from paraphrasing a phrase table entry, and 1 if was in
the original phrase table.

\section{Results and Discussion}

The results of the evaluation are shown in Tables \ref{table:eval10k} and \ref{table:eval}.
I use the notation shown in Table \ref{tab:notation}.

The differences between the baseline
and the remaining systems shown in Table \ref{table:eval10k} are statistically significant:
tested using the bootstrapping method described in
\namecite{Zhang:Vogel:2004:mt:conf} with 2,000 samples.

\begin{table}
  \centering
\begin{tabular}{|l|c|cccc|cc|cc|}
  \hline
                             &  & \multicolumn{4}{|c|}{\bf $n$-gram precision} &  \multicolumn{2}{|c|}{\bf Bleu} & \multicolumn{2}{|c|}{\bf \# of phrases}\\
  {\bf System } & {\bf Bleu} & {\bf 1-gr.} & {\bf 2-gr.} & {\bf 3-gr.} & {\bf 4-gr.} & {\bf BP} & {\bf ration} & {\bf gener.} & {\bf used}\\
  \hline
  $S$ (baseline)             & 22.38 & 55.4 & 27.9 & 16.6 & 10.0 & 0.995 & 0.995 & 180,996 & 40,598\\
  $S_{par}$                  & 21.89 & 55.7 & 27.8 & 16.5 & 10.0 & 0.973 & 0.973 & 192,613 & 42,169\\
  $S_{parW}$                 & 22.57 & 55.1 & 27.8 & 16.7 & 10.2 & 1.000 & 1.000 & 202,018 & 42,523\\
  $S^\star$                  & 22.58 & 55.4 & 28.0 & 16.7 & 10.1 & 1.000 & 1.001 & 206,877 & 40,697\\
  $S + S_{par}$              & 22.73 & 55.8 & 28.3 & 16.9 & 10.3 & 0.994 & 0.994 & 262,013 & 54,486\\
  $S + S_{parW}$             & {\bf 23.05} & 55.8 & 28.5 & 17.1 & 10.6 & 0.995 & 0.995 & 280,146 & 56,189\\
  $S + S_{parW} \dag$    & {\bf 23.13} & 55.8 & 28.5 & 17.1 & 10.5 & 1.000 & 1.002 & 280,146 & 56,189\\
  $S^\star + S_{parW}^\star$ & {\bf 23.09} & 56.1 & 28.7 & 17.2 & 10.6 & 0.993 & 0.993 & 327,085 & 56,417\\
  $S^\star + S_{parW}^\star \ddagger$ & {\bf 23.09} & 55.8 & 28.4 & 17.1 & 10.5 & 1.000 & 1.001 & 327,085 & 56,417\\
  \hline
\end{tabular}
  \caption{\textbf{\emph{Bleu} scores and $n$-gram precisions (in \%s) for 10k training sentences}.
           The last two columns show the total number of phrase pairs in the phrase table and
           the number usable on the test set, respectively. See the text for a description of the systems.}
  \label{table:eval10k}
\end{table}

\begin{table}
\centering
\begin{tabular}{l@{ }l}
  $S$ & baseline, trained on the original corpus;\\
  $S_{par}$ & original corpus, augmented with sentence-level paraphrases, no transformations\\
            & (5) and (6) (i.e. without using the Web);\\
  $S_{parW}$ & original corpus, augmented with sentence-level paraphrases, all transformations;\\
  $\star$ & means paraphrasing the phrase table;\\
  $+$ & means merging the phrase tables;\\
  $\dag$ & using an extra feature: $F_{parW}$;\\
  $\ddagger$ & using two extra features: $F_{\star}$, $F_{parW}$.\\
\end{tabular}
\caption{\textbf{Notation for the experimental runs.}}
\label{tab:notation}
\end{table}

\begin{table}
  \centering
\begin{tabular}{|l|cccc|}
  \hline
                    & \multicolumn{4}{|c|}{\bf \# of training sentences}\\
  \multicolumn{1}{|c|}{\bf System}          & 10k   & 20k   & 40k & 80k\\
  \hline
  $S$ (baseline)   & 22.38 & 24.33 & 26.48 & 27.05\\
  \cline{5-5}
  $S_{parW}$   & 22.57 & 24.41 & 25.96 & \multicolumn{1}{|c}{}\\
  $S^\star$  & 22.58 & 25.00   & 26.48 & \multicolumn{1}{|c}{}\\
  $S + S_{parW}$    & {\bf 23.05} & {\bf 25.01} & {\bf 26.75} & \multicolumn{1}{|c}{}\\
  \cline{1-4}
\end{tabular}
  \caption{\textbf{\emph{Bleu} scores for different number of training sentences.}}
  \label{table:eval}
\end{table}

{\bf Gain of 33\%--50\% compared to doubling the training data.}
As Table \ref{table:eval} shows,
neither paraphrasing the training sentences, $S_{parW}$, nor
paraphrasing the phrase table, $S^\star$, lead to notable
improvements. For 10k training sentences the two systems are
comparable and improve {\it Bleu} by 0.3, for 40k sentences, $S^\star$
performs as the baseline and $S_{parW}$ even drops below it.
However, when merging the phrase tables of $S$ and $S_{parW}$,
I get an improvement of almost 0.7 for 10k and 20k sentences, and
about 0.3 for 40k sentences. While this improvement might look
small, it is comparable to that of
\namecite{Bannard:Callison-Burch:2005:par}, who obtain 0.7 improvement
for 10k sentences and 1.0 for 20k sentences (but translating in the
reverse direction: from Spanish into English). Note also, that the
0.7 {\it Bleu} improvement for 10k and 20k sentences is about 1/3 of the 2
{\it Bleu} points obtained by the baseline system by doubling the
training size.  Note also that the 0.3 gain on {\it Bleu} for 40k sentences
is equal to half of what will be gained if I train on 80k sentences.

{\bf Improved precision for all $n$-grams.}
Table \ref{table:eval10k}
provides a comparison of different systems trained on 10k sentences.
In addition to the {\it Bleu} score, I also give its elements:
$n$-gram precisions, BP (brevity penalty) and ration. Comparing the baseline
with the last four systems, we can see that unigram, bigram, trigram
and fourgram precisions are all improved by between 0.4\% and 0.7\%.

{\bf Importance of noun compound splitting.}
System $S_{par}$ is trained on the training corpus augmented with paraphrased
sentences, where the noun compound splitting transformations
(5) and (6) are not used, i.e. the paraphrases are purely syntactic and use no Web counts.
We can see that omitting these rules
causes the results to go below the baseline: while
there is a 0.3\% gain on unigram precision,
bigram and trigram precision go down by about 0.1\%.
More importantly, BP goes down as well: since
the sentence-level paraphrases (except for the possessives which are
infrequent) mainly convert NPs into noun compounds,
the resulting sentences are shorter, which causes the translation model
to learn to generate shorter sentences.
Note that this effect is not observed for $S_{parW}$, where transformations (5) and (6)
make the sentences longer, and therefore balance the BP to be
exactly 1.0.  A somewhat different kind of argument applies to
$S + S_{par}$, which is worse than $S + S_{parW}$, but not
because of BP.  In this case, there is no improvement for unigrams,
but a consistent 0.2-0.3\% drop for bigrams, trigrams and fourgrams.
The reason for this is shown in the last column of Table \ref{table:eval} --
omitting the noun compound splitting transformations (5) and (6)
results in fewer training sentences,
which means fewer phrases in the phrase table and
consequently fewer phrases compatible with the test set.

{\bf More usable phrases.}
The last two columns of Table
\ref{table:eval10k} show that more phrases in the phrase table
mean an increased number of usable phrases as well.
A notable exception is $S^\star$:
its phrase table is bigger than those of $S_{par}$ and $S_{parW}$,
but it contains fewer phrases compatible with the test set then them
(but still more than the baseline).
This suggests that additional phrases extracted from paraphrased sentences
are more likely to be usable at test time
than additional phrases generated by paraphrasing the phrase table.

{\bf Paraphrasing the phrase table vs. paraphrasing the training corpus.}
As Tables \ref{table:eval} and \ref{table:eval10k} show,
paraphrasing the phrase table $S^\star$ ({\it Bleu} score 22.58\%)
does not compete against paraphrasing the training corpus
followed by merging the resulting phrase table with the phrase table
for the original corpus\footnote{Note that $S^\star$
does not use syntactic transformations (5) and (6).
However, as the results for $S+S_{par}$ show, the claim still holds even
if transformations (5) and (6) are also excluded when paraphrasing the sentences:
the \emph{Bleu} score for $S+S_{par}$ is 22.73\% vs. 22.58\% for $S^\star$.},
as in $S+S_{parW}$ ({\it Bleu} score 23.05\%). I also try to paraphrase the phrase table of
$S+S_{parW}$, but the resulting system $S^\star+S^\star_{parW}$ yields
little improvement: 23.09\% {\it Bleu} score. Adding the two
extra features, $F_\star$ and $F_{parW}$, does not
yield improvements as well: $S^\star + S_{parW}^\star \ddagger$
achieves the same {\it Bleu} score as $S^\star+S^\star_{parW}$. This shows
that extracting additional phrases from the augmented corpus is
a better idea than paraphrasing the phrase table,
which can result in erroneous splitting of noun phrases.
Paraphrasing whole sentences as opposed to paraphrasing
the phrase table could potentially improve
the approach of \namecite{Callison-Burch:al:2006:mt}: while low probability
and context dependency could be problematic,
a language model could help filtering the bad sentences out.
Such filtering could potentially improve my results as well.
Finally, note that different paraphrasing strategies could be used
when paraphrasing phrases vs. sentences.
For example, paraphrasing the phrase table can be done more aggressively:
if an ungrammatical phrase is generated in the phrase table,
it will probably have no negative effect on translation quality
since it will be unlikely to occur in the test data.

{\bf Quality of the paraphrases.}
An important difference between my syntactic paraphrasing method
and the multi-lingual approach of \namecite{Callison-Burch:al:2006:mt} is that
their paraphrases are only contextually synonymous
and often depart significantly from the original meaning.
As a result, they could not achieve improvements by simply augmenting the phrase table:
this introduces too much noise and yields an accuracy significantly
below the baseline -- by 3-4\% on {\it Bleu}.
In order to achieve an improvement, they had to introduce an extra feature
penalizing the low probability paraphrases and promoting the
original phrases in the phrase table. In contrast,
my paraphrases are meaning preserving and context-independent:
introducing the feature $F_{parW}$,
which penalizes phrases coming from the paraphrased corpus,
in system $S + S_{parW} \dag$ yields a tiny improvement
on {\it Bleu} score (23.13\% vs. 23.05\%), i.e., the phrases
extracted from my augmented corpus
are almost as good as the ones from the original corpus.
Finally, note that my paraphrasing method is
complementary to that of \namecite{Callison-Burch:al:2006:mt} and therefore
the two can be combined: the strength of my approach is at
improving the coverage of longer phrases using syntactic paraphrases,
while the strength of theirs is at improving the vocabulary coverage
with extra words extracted from additional corpora
(although they do get some gain from using longer phrases as well).

{\bf Translating into English.}
I tried translating
in the reverse direction, paraphrasing the target (English)
language side, which resulted in decreased performance.
This is not surprising: the set of available source phrases remains the same,
and a possible improvement could potentially come from producing
a more fluent translation only, e.g., from turning an NP with an internal PP
into a noun compound. However, unlike the original translations,
the extra ones are less likely to be judged correct
since they were not observed in training.

{\bf Problems and limitations.}
Error analysis suggests that the major problems
for the proposed paraphrasing method are caused by incorrect PP-attachments
in the parse tree. Somewhat less frequent and therefore less
important are errors of POS tagging. At present I use a parser,
which limits the applicability of the method to languages for which
syntactic parsers are available. In fact, the kinds of paraphrases
I use are simple and can be approximated by using a shallow parser or
just a POS tagger; this would lead to errors of PP-attachment, but
these attachments are often assigned incorrectly by parsers anyway.

The central target of my paraphrases are noun compounds --
I turn an NP with an internal PP into a noun compound and vice versa --
which limits its applicability of the approach to languages
where noun compounds are a frequent phenomenon,
e.g., Germanic, but not Romance or Slavic languages.

From a practical viewpoint, a limitation of the method is that
it increases the size of the phrase table and/or of the training corpus,
which slows down the processes of both training and translation,
and limits its applicability to relatively small corpora for computational reasons.

Finally, as Table \ref{table:eval} shows,
the improvements get smaller for bigger training corpora,
which suggests that it gets harder to generate useful paraphrases
that are not already present in the corpus.

\section{Conclusion and Future Work}

I have presented a novel domain-independent approach for improving
statistical machine translation by augmenting the training corpus
with monolingual paraphrases of the source language side sentences,
thus increasing the training data ``for free'',
by creating it from data that is already available
rather than having to create more aligned data.

While in my experiments I use phrase-based SMT,
any other MT approach that learns from parallel corpora could
potentially benefit from the proposed syntactic corpus augmentation idea.
At present, my paraphrasing rules are English-specific,
but they could be easily adapted to other Germanic languages,
which make heavy use of noun compounds; the general idea of automatically
generating nearly equivalent source side syntactic paraphrases
can in principle be applied to any language.
The current version of the method can be considered preliminary,
as it is limited to NPs; still, the results are already encouraging, and
the approach is worth considering when building MT systems from small corpora,
e.g., in case of resource-poor language pairs, in specific domains, etc.

Better use of the Web could be made for paraphrasing noun compounds
(e.g., using verbal paraphrases as in chapters \ref{chapter:NC:bracketing} and \ref{chapter:NC:semantics}),
other syntactic transformations could be tried
(e.g., adding/removing complementizers like {\it that},
which are not obligatory in English, or
adding/removing commas from non-obligatory positions), and
a language model could be used to filter out the bad paraphrases.

Even more promising, but not that simple, seems using a tree-to-tree syntax-based SMT system
and learning transformations that can make the source language trees
structurally closer to the target-language ones,
e.g., the English sentence
``{\it Remember the guy who you are \underline{with}!}'' would be transformed to
``{\it Remember the guy \underline{with} whom you are!}'',
whose word order is closer to Spanish
``{\it !`Recuerda al individuo \underline{con} quien est\'{a}s!}'',
which might facilitate the translation process.

Finally, the process could be made
part of the decoding, which would eliminate the need of paraphrasing
the training corpus and might allow the dynamic generation
of paraphrases both for the phrase table and for the test sentence.


\chapter{Other Applications}
\label{chapter:appl}

This chapter describes applications
of the methods developed in chapter \ref{chapter:NC:bracketing}:
to two important structural ambiguity problems a syntactic parser faces --
{\it prepositional phrase attachment} and {\it NP coordination}.
Using word-association scores, Web-derived surface features, and paraphrases,
I achieve 84\% accuracy for the former task and 80\% for the latter one,
which is on par with the state-of-the-art.
A shorter version of this work appeared in \cite{nakov-hearst:2005:HLTEMNLP}.

\section{Prepositional Phrase Attachment}
\label{sec:PP:attach}

\subsection{Introduction}

A long-standing challenge for syntactic parsers is the attachment decision for
prepositional phrases.  In a configuration where a verb takes a noun complement
that is followed by a prepositional phrase (PP), the problem arises of whether
the prepositional phrase attaches to the noun or to the verb.
Consider for example the following contrastive
pair of sentences:

\vspace{6pt}
\begin{tabular}{lll}
(1) & {\it Peter spent millions of dollars.}  & (noun)\\
(2) & {\it Peter spent time with his family.} & (verb)
\end{tabular}
\vspace{6pt}

In the first sentence, the prepositional phrase {\it millions of dollars} attaches to the noun
{\it millions}, while in the second sentence the prepositional phrase {\it with his family}
attaches to the verb {\it spent}.

Below I will address a formulation of the PP-attachment problem
that casts these associations out of context,
as the quadruple $(v,n_1,p,n_2)$, where $v$ is the verb, $n_1$ is
the head of the direct object, $p$ is the preposition (the head
of the PP), and $n_2$ is the head of the noun phrase inside the prepositional phrase.
For example, the quadruple for (2) is ({\it spent}, {\it time}, {\it with}, {\it family}).

\subsection{Related Work}

While most early work on PP-attachment ambiguity resolution relied
primarily on deterministic pragmatic
and syntactic considerations (e.g., {\it minimal attachment}, {\it right association}, etc.),
recent research on the problem is dominated by probabilistic
machine learning approaches: supervised and unsupervised.

\subsubsection{Supervised Approaches}


\namecite{Ratnaparkhi:Reynar:Roukos:1994} created a dataset of 27,937
(20,801 training; 4,039 development; 3,097 testing)
quadruples $(v,n_1,p,n_2)$ from the {\it Wall Street Journal};
I will refer to it as {\it Ratnaparkhi's dataset}.
They achieved 81.6\% accuracy with a maximum entropy classifier
and a binary hierarchy of word classes derived using mutual information.
They also found the human performance for the task to be
88\% without context, and 93\% when sentence context is provided.

{\it Ratnaparkhi's dataset} has been used as the benchmark dataset for this task.
\namecite{Collins:Brooks:1995} propose a supervised back-off model, which
achieves 84.5\% accuracy.
\namecite{Zavrel:al:1997} use memory-based learning
and various similarity metrics yielding 84.4\% accuracy.
\namecite{Stetina:Nagao:1997} use a supervised method with
decision trees and {\it WordNet} classes which achieves 88.1\% accuracy.
\namecite{Abney:al:1999:pp} report 84.6\% accuracy using boosting.
Using SVMs with weighted polynomial information gain kernels,
\namecite{Vanschoenwinkel:Manderick:2003:pp} achieve 84.8\% accuracy.
\namecite{Alegre:al:1999:pp} report 86\% with neural networks.
Using a nearest-neighbor classifier, \namecite{Zhao:Lin:2004:pp} achieve 86.5\% accuracy.
\namecite{Toutanova:al:2004} report 87.5\% accuracy using {\it WordNet} and morphological and syntactic analysis.

On their own dataset of 500 test examples,
\namecite{Brill:Resnik:1994} use supervised transformation-based
learning and conceptual classes derived from {\it WordNet},
achieving 82\% accuracy.
On a different dataset, \namecite{Li:2002} achieves 88.2\% accuracy
with word clustering based on co-occurrence data.
\namecite{Olteanu:Moldovan:2005:PPattach} use SVM with a large number
of complex syntactic and semantic features, and $n$-gram frequencies from the Web.
They could not use {\it Ratnaparkhi's dataset},
since their formulation of the problem is a bit different:
they use features extracted from the whole sentence, as opposed to using the quadruple only.
They achieve 93.62\% on a dataset extracted from the {\it Penn Treebank};
on another dataset derived from {\it FrameNet} \cite{Baker:al:1998}, they achieve 91.79\% and 92.85\%,
depending on whether manually annotated semantic information is made available to the model.

\subsubsection{Unsupervised Approaches}
\label{sec:pp:unsup}

Unsupervised approaches make attachment decisions
using co-occurrence statistics drawn from text collections.
The idea was pioneered by \namecite{Hindle:Rooth:1993},
who use a partially parsed corpus to
calculate lexical associations over subsets of the
tuple $(v,n_1,p)$ (e.g., comparing
$\mathrm{Pr}(p|n_1)$ to $\mathrm{Pr}(p|v)$),
achieving 80\% accuracy at 80\% coverage.

\namecite{Ratnaparkhi:1998} develops an
unsupervised method that collects statistics from text annotated
with POS tags and morphological base forms.  An
extraction heuristic is used to identify unambiguous attachment
decisions, for example, the algorithm can assume a
{\it noun attachment} if there is no verb within $k$ words to the left of
the preposition in a given sentence, among other conditions.
In his experiments, this heuristic uncovers 910K unique tuples of the
form $(v,p,n_2)$ and $(n,p,n_2)$, which suggest
the correct attachment only about 69\% of the time.
The tuples are used by classifiers,
the best of which achieved 81.9\% accuracy on {\it Ratnaparkhi's dataset}.
Note that these classifiers are not trained on real training data,
but on the output of an unsupervised method, i.e. the method stays unsupervised.

\namecite{Pantel:Lin:2000} describe an unsupervised method,
which uses a collocation
database, a thesaurus, a dependency parser, and a large corpus
(125M words), achieving 84.31\% accuracy on the same dataset.

Recently, there has been a growing interest in using $n$-gram frequencies
derived from the Web, as opposed to estimating them from a static corpus.
\namecite{Volk:2000:www:PPattach} used Web-derived $n$-gram frequencies
to solve the PP-attachment problem for German.
He compared $\mathrm{Pr}(p|n_1)$ to $\mathrm{Pr}(p|v)$,
estimated as $\mathrm{Pr}(p|x) = \#(x,p) / \#(x)$, where $x$ can be
$n_1$ or $v$, and the frequencies are obtained from {\it Altavista}
using the \texttt{NEAR} operator.
The approach was able to make a decision for 58\% of his examples
with a accuracy of 75\% (baseline 63\%).
\namecite{Volk:2001:www:PPattach} later improved on these results by comparing
$\mathrm{Pr}(p,n_2|n_1)$ to $\mathrm{Pr}(p,n_2|v)$.
These conditional probabilities were estimated as $\mathrm{Pr}(p,n_2|x) = \#(x,p,n_2) / \#(x)$,
where $x$ can be $n_1$ or $v$.
Using inflected forms, he was able to classify 85\% of his examples
with 75\% accuracy.

\namecite{Calvo:Gelbukh:2003:PPattach} used exact phrases
instead of the \texttt{NEAR} operator,
and compared the frequencies ``$v$ $p$ $n_2$'' and ``$n_1$ $p$ $n_2$''.
For example, to disambiguate {\it Veo al gato con un telescopio},
they compared the $n$-gram frequencies for phrases ``{\it veo con telescopio}''
and ``{\it gato con telescopio}''.
They tested this idea on 181 randomly chosen Spanish examples,
and were able to classify 89.5\% of them with a accuracy of 91.97\%.

\namecite{Lapata:Keller:05:Web:based:Models} used exact-phrase queries in  {\it AltaVista}
and the models of \namecite{Volk:2000:www:PPattach},
\namecite{Volk:2001:www:PPattach} and \namecite{Calvo:Gelbukh:2003:PPattach}.
On a random subset of 1,000 examples from {\it Ratnaparkhi's dataset},
they achieved 69.40\% accuracy (baseline 56.80\%),
which they found to be worse than using {\it BNC} with the same models: 74.40\%.

\subsection{Models and Features}

As in Chapter \ref{chapter:NC:bracketing},
I use three kinds of features, all derived from the Web:
(a) word association scores;
(b) paraphrases; and
(c) surface features.
I used the training part of {\it Ratnaparkhi's dataset}
when developing the features in the last two groups.

\subsubsection{$n$-gram Models}

\noindent I use the following co-occurrence models:

\vspace{12pt}
\begin{tabular}{ll}
(1) & $\#(n_1,p)$ vs. $\#(v,p)$\\
(2) & $\mathrm{Pr}(p|n_1)$ vs. $\mathrm{Pr}(p|v)$\\
(3) & $\#(n_1,p,n_2)$ vs. $\#(v,p,n_2)$\\
(4) & $\mathrm{Pr}(p,n_2|n_1)$ vs. $\mathrm{Pr}(p,n_2|v)$\\
\end{tabular}
\vspace{12pt}

In the above models, the left score is associated with a {\it noun attachment},
and the right score with a {\it verb attachment}.
I estimate the $n$-gram counts using inflected exact phrase queries
against {\it MSN Search};
the conditional probabilities are estimated as ratios
of $n$-gram frequencies as described in section \ref{sec:pp:unsup} above.
I allow for determiners where appropriate, e.g.,
between the preposition and the noun when querying for
$\#(v,p,n_2)$, and I add up the frequencies for all possible
variations of the query.
I tried using the class-based smoothing
technique described by \cite{Hindle:Rooth:1993}, which led to
better coverage for some of the models.
I also tried backing off from (3) to (1), and from (4) to (2),
as well as backing off plus smoothing, but there was no improvement
over smoothing alone.  I found $n$-gram counts to be
unreliable when pronouns appear in the test set as $n_1$ or $n_2$,
and therefore I disabled those models in these cases. Happily, such examples can still
be handled by paraphrases or surface features (see below).

%


\subsubsection{Web-Derived Surface Features}

%
\noindent Consider for example, the sentence

\begin{center}
{\it John opened the door with his key.}
\end{center}

\noindent which contains a difficult verb attachment example since
{\it doors}, {\it keys}, and {\it opening} are all semantically related.
To determine if this should be a verb or a
noun attachment, I look for surface cues in other contexts
that could indicate which of these terms tend to associate most closely.
For example, if I can find an instance with parentheses like

\begin{center}
  {\it ``opened the door (with his key)''}
\end{center}

\noindent it would suggest a verb attachment, since the parentheses
signal that ``{\it with a key}'' acts as a unit that is independent of {\it door},
and therefore has to attach to the verb.

Hyphens, colons, capitalization, and other punctuation can also help
making disambiguation decisions. For example, when disambiguating the sentence

\begin{center}
    {\it John eats spaghetti with sauce.}
\end{center}

\noindent finding the following example on the Web

\begin{center}
  {\it ``eat: spaghetti with sauce''}
\end{center}

\noindent would suggest a noun attachment.

Table \ref{table:surfacePP} illustrates a variety of
surface features, along with the attachment decisions they are
assumed to suggest (singleton events are ignored).
The table shows that such surface features have low coverage.

Since search engines ignore punctuation characters,
I issue exact phrase inflected queries against {\it Google}
and then post-process the top 1,000 resulting
summaries,
looking for the surface features of interest.

\begin{table}
\begin{center}
\begin{small}
\begin{tabular}{|lccc|}
\multicolumn{1}{l}{\bf Surface Feature} & {\bf Prediction} & {\bf Accuracy (\%)} & \multicolumn{1}{c}{\bf Coverage (\%)} \\
\hline
open Door with a key & noun & 100.00 & 0.13\\
(open) door with a key & noun & 66.67 & 0.28 \\
open (door with a key) & noun & 71.43 & 0.97 \\
open - door with a key & noun & 69.70 & 1.52 \\
open / door with a key & noun & 60.00 & 0.46 \\
open, door with a key & noun & 65.77 & 5.11 \\
open: door with a key & noun & 64.71 & 1.57 \\
open; door with a key & noun & 60.00 & 0.23 \\
open. door with a key & noun & 64.13 & 4.24 \\
open? door with a key & noun & 83.33 & 0.55 \\
open! door with a key & noun & 66.67 & 0.14 \\
\hline
open door With a Key & verb & 0.00 & 0.00 \\
(open door) with a key & verb & 50.00 & 0.09 \\
open door (with a key) & verb & 73.58 & 2.44\\
open door - with a key & verb & 68.18 & 2.03 \\
open door / with a key & verb & 100.00 & 0.14 \\
open door, with a key & verb & 58.44 & 7.09 \\
open door: with a key & verb & 70.59 & 0.78 \\
open door; with a key & verb & 75.00 & 0.18 \\
open door. with a key & verb & 60.77 & 5.99 \\
open door! with a key & verb & 100.00 & 0.18 \\
\hline
{\bf Surface features (sum)} &  & {\bf 73.13$\pm$5.41} & {\bf 9.26} \\
\hline
\end{tabular}
\end{small} \caption{{\bf Surface features for PP-attachment 2,171.}
Accuracy and coverage (in \%s) are shown across all examples,
and are illustrated by the quadruple (\emph{open}, \emph{door}, \emph{with}, \emph{key}).}
\label{table:surfacePP}
\end{center}
\end{table}

\subsubsection{Paraphrases}

The second way I extend the use of Web counts
is by paraphrasing the relation
in an alternative form that could suggest the correct attachment.
Given a quadruple $(v,n_1,p,n_2)$,
I look on the Web for instances of the following patterns:

\vspace{12pt}
\begin{tabular}{lll}
(1) & $v$ $n_2$ $n_1$ & (noun)\\
(2) & $v$ $p$ $n_2$ $n_1$ & (verb)\\
(3) & $p$ $n_2$ * $v$ $n_1$ & (verb)\\
(4) & $n_1$ $p$ $n_2$ $v$ & (noun)\\
(5) & $v$ {\it pronoun} $p$ $n_2$ & (verb)\\
(6) & {\it be} $n_1$ $p$ $n_2$ & (noun)
\end{tabular}
\vspace{12pt}

These patterns are linguistically motivated
and each one supports either a noun or a verb attachment,
as indicated above.

{\bf Pattern (1)} predicts a noun attachment
if ``$n_1$ $p$ $n_2$'' can be expressed as a noun compound ``$n_2$ $n_1$'',
which would mean that the verb has a single object.
For example, pattern (1) would paraphrase

\begin{center}
    {\it meet/$v$ demands/$n_1$ \underline{from/$p$ customers/$n_2$}}
\end{center}

\noindent as

\begin{center}
    {\it meet/$v$ the \underline{customer/$n_2$} demands/$n_1$}
\end{center}

Note that in case of ditransitive verbs,
the pattern could make an incorrect prediction.
For example

\begin{center}
    {\it give/$v$ an apple/$n_1$ \underline{to/$p$ the boy/$n_2$}} $\rightarrow$
    {\it give/$v$ \underline{the boy/$n_2$} an apple/$n_1$}
\end{center}

\noindent where ``{\it boy an apple}'' clearly is not a noun compound,
and the verb still has two objects: only the order
of the direct and the indirect object has changed.

In order to prevent this, in the paraphrase,
I do not allow a determiner before $n_1$,
and I do require one before $n_2$. In addition, I disallow the
pattern if the preposition is {\it to} and I require both $n_1$
and $n_2$ to be nouns (as opposed to numbers, percents,
pronouns, determiners, etc.).

{\bf Pattern (2)} predicts a verb attachment. It presupposes that
``$p$ $n_2$'' is an indirect object of the verb $v$ and tries to
switch it with the direct object $n_1$, which is possible in some cases,
e.g.,

\begin{center}
    {\it had/$v$ a program/$n_1$ \underline{in/$p$ place/$n_2$}}
\end{center}

\noindent would be transformed into

\begin{center}
    {\it had/$v$ \underline{in/$p$ place/$n_2$} a program/$n_1$}
\end{center}

I require that $n_1$ be preceded by a determiner
in order to prevent ``$n_2$ $n_1$'' from forming a noun compound
in the paraphrase.

{\bf Pattern (3)} predicts a verb attachment.
It looks for appositions, where the prepositional phrase ``$p$ $n_2$'' has moved in
front of the verb, e.g.,

\begin{center}
    {\it I gave/$v$ an apple/$n_1$ \underline{to/$p$ the boy/$n_2$}}.
\end{center}

\noindent can be paraphrased as

\begin{center}
    {\it It was \underline{to the boy/$n_2$} that I gave/$v$ an apple/$n_1$}.
\end{center}

I allow zero or more (up to three) intervening words at the position of the asterisk
in pattern (3): ``$p$ $n_2$ * $v$ $n_1$''.

{\bf Pattern (4)} predict a noun attachment.
It looks for appositions, where the whole complex NP ``$n_1$ $p$ $n_2$''
has moved in front of the verb $v$.
For example, it would transform

\begin{center}
    {\it shaken/$v$ \underline{confidence/$n_1$ in/$p$ markets/$n_2$}}
\end{center}

\noindent into

\begin{center}
    {\it \underline{confidence/$n_1$ in/$p$ markets/$n_2$} shaken/$v$}
\end{center}


{\bf Pattern (5)} is motivated by the observation that
prepositional phrases do not like attaching to pronouns \cite{Hindle:Rooth:1993}.
Therefore, if $n_1$ is a pronoun,
the probability of a verb attachment is very high.
(I have a separate model that checks whether $n_1$ is a pronoun.)
Pattern (5) substitutes $n_1$ with a dative pronoun: {\it him} or {\it her}.
For example, it would paraphrase

\begin{center}
    {\it put/$v$ \underline{a client/$n_1$} at/$p$ odds/$n_2$}
\end{center}

\noindent as

\begin{center}
    {\it put/$v$ \underline{him} at/$p$ odds/$n_2$}
\end{center}

{\bf Pattern (6)} is motivated by the observation that the verb {\it to be} is
typically used with a noun attachment. (I have a separate model that checks whether
$v$ is a form of the verb {\it to be}.)  The pattern substitutes $v$ with {\it
is} and {\it are}, e.g., it will turn

\begin{center}
    {\it eat/$v$ spaghetti/$n_1$ with/$p$ sauce/$n_2$}
\end{center}

\noindent into

\begin{center}
    {\it is spaghetti/$n_1$ with/$p$ sauce/$n_2$}.
\end{center}

All six patterns allow for determiners where appropriate,
unless otherwise stated. A prediction is made
if at least one instance of the pattern has been found.

\subsection{Experiments and Evaluation}

In order to make my results directly comparable
to those of other researchers,
I perform the evaluation on the test part of {\it Ratnaparkhi's dataset},
consisting of 3,097 quadruples $(v,n_1,p,n_2)$.
However, there are some problems with that dataset,
due primarily to extraction errors.
First, the dataset contains 149 examples in which a bare determiner
like {\it the}, {\it a}, {\it an}, etc.,
is labeled as $n_1$ or $n_2$, instead of the actual head
noun, e.g.,
(\emph{is}, \emph{\underline{the}}, \emph{of}, \emph{kind}),
(\emph{left}, \emph{chairmanship}, \emph{of}, \emph{\underline{the}}),
(\emph{acquire}, \emph{securities}, \emph{for}, \emph{\underline{an}}), etc.
While supervised algorithms can compensate for the problem
by learning from the training set
that {\it the} can be a ``noun'',
unsupervised algorithms cannot do so.
Second, there are 230 examples in which the nouns contain
symbols like \%, /, \&, ',
which cannot be used in queries against a search engine,
e.g.,
(\emph{buy}, \emph{\underline{\%}}, \emph{for}, \emph{10}),
(\emph{beat}, \emph{\underline{S\&P-down}}, \emph{from}, \emph{\underline{\%}}),
(\emph{is}, \emph{\underline{43\%-owned}}, \emph{by}, \emph{firm}), etc.
While being problematic for my models,
such quadruples do not necessarily represent errors of extraction
when the dataset was created.

Following \namecite{Ratnaparkhi:1998} and \namecite{Pantel:Lin:2000},
I predict a noun attachment for all 926 test examples
whose preposition is {\it of} (which is very accurate: 99.14\%),
and I only try to classify the remaining
2,171 test examples, which I will call {\it Ratnaparkhi's dataset}-2,171.
The performance of the individual models on that smaller dataset
is shown in Table \ref{table:PP:models}.
The table also shows 95\%-level confidence intervals for the accuracy,
calculated as described in section \ref{sec:conf:interval}.
The baseline accuracy of always assigning verb attachment is 58.18\%,
which is close to the accuracy for three of the models:
``$\#(v,p):\#(n_1,p)$'' with 58.91\%,
``$v$ $n_2$ $n_1$'' with 59.29\%, and
``$p$ $n_2$ $v$ $n_1$'' with 57.79\%.

\begin{table}
\begin{center}
\begin{small}
\begin{tabular}{|l|c|c|}
\multicolumn{1}{c}{\bf Model} & \multicolumn{1}{c}{\bf Accuracy (\%)} & \multicolumn{1}{c}{\bf Coverage (\%)} \\
\hline
    Baseline-2,171 (verb attachment) & 58.18$\pm$2.09 & 100.00 \\
\hline
$\#(v,p) : \#(n_1,p)$ & 58.91$\pm$2.27 & 83.97\\
$\mathrm{Pr}(p|v) : \mathrm{Pr}(p|n_1)$ & 66.81$\pm$2.19 & 83.97\\
$\mathrm{Pr}(p|v) : \mathrm{Pr}(p|n_1)$ smoothed & {\bf 66.81$\pm$2.19} & 83.97\\
\hline
$\#(v,p,n_2) : \#(n_1,p,n_2)$ & 65.78$\pm$2.25 & 81.02\\
$\mathrm{Pr}(p,n_2|v) : \mathrm{Pr}(p,n_2|n_1)$ & 68.34$\pm$2.20 & 81.62\\
$\mathrm{Pr}(p,n_2|v) : \mathrm{Pr}(p,n_2|n_1)$ smoothed & {\bf 68.46$\pm$2.17} & 83.97\\
\hline
(1) ``$v$ $n_2$ $n_1$'' & {\bf 59.29$\pm$4.46} & 22.06\\
(2) ``$p$ $n_2$ $v$ $n_1$'' & {\bf 57.79$\pm$2.47} & 71.58\\
(3) ``$n_1$ * $p$ $n_2$ $v$'' & {\bf 65.78$\pm$4.50} & 20.73\\
(4) ``$v$ $p$ $n_2$ $n_1$'' & {\bf 81.05$\pm$6.17} & 8.75\\
(5) ``$v$ {\it pronoun} $p$ $n_2$'' & {\bf 75.30$\pm$3.43} & 30.40\\
(6) ``{\it be} $n_1$ $p$ $n_2$'' & {\bf 63.65$\pm$3.73} & 30.54\\
\hline
$n_1$ is a {\it pronoun} & {\bf 98.48$\pm$6.58} & 3.04\\
$v$ is the verb {\it to be} & {\bf 79.23$\pm$6.03} & 9.53\\
\hline
Surface features & {\bf 73.13$\pm$6.52} & 9.26\\
\hline
\hline
\emph{of} $\rightarrow$ noun, majority vote & 85.01$\pm$1.36 & 91.77\\
\emph{of} $\rightarrow$ noun, majority vote, verb & 83.63$\pm$1.34 & 100.00\\
\hline
\emph{of} $\rightarrow$ noun, bootstrapping, maj. vote & 84.91$\pm$1.33 & 96.93\\
\emph{of} $\rightarrow$ noun, bootstrapping, maj. vote, verb & 84.37$\pm$1.32 & 100.00\\
\hline
\end{tabular}
\end{small}
\caption{{\bf PP-attachment: evaluation results.}
The results for the baseline and for the individual models are calculated on \emph{Ratnaparkhi's dataset}-2,171;
the last four lines show the performance of majority vote combinations on the full \emph{Ratnaparkhi's dataset}.}
\label{table:PP:models}
\end{center}
\end{table}

\begin{table}
\begin{center}
\begin{small}
\begin{tabular}{|l|c|c|}
\multicolumn{1}{c}{\bf Model} & \multicolumn{1}{c}{\bf Accuracy (\%)}\\
\hline
    Baseline 1: {\it noun attachment} & 58.96$\pm$1.74\\
    Baseline 2: \emph{of} $\rightarrow$ noun; the rest $\rightarrow$ {\it verb attachment} & 70.42$\pm$1.63\\ Baseline 3: {\it most likely for each $p$} & 72.20$\pm$1.60\\
\hline
    \cite{Lapata:Keller:05:Web:based:Models} -- {\it AltaVista} & 69.40$\pm$2.93\\
    \cite{Lapata:Keller:05:Web:based:Models} -- {\it BNC} & 74.40$\pm$2.79\\
    \cite{Ratnaparkhi:1998} & 81.60$\pm$1.40\\
    \cite{Pantel:Lin:2000}  & 84.31$\pm$1.32\\
    {\bf My bootstrapped majority vote classifier} & {\bf 84.37$\pm$1.32}\\
\hline
    Average human: {\it quadruple} & 88.20 \\
    Average human: {\it whole sentence} & 93.20\\
\hline
\end{tabular}
\end{small}
\caption{{\bf PP-attachment: comparisons on \emph{Ratnaparkhi's dataset}.}
        Shown are human performance, different baselines, and previous unsupervised results.
        Note that the results of Lapata \& Keller (2005) are calculated
        for 1,000 random examples from \emph{Ratnaparkhi's dataset} (total: 3,097 testing examples).}
\label{table:PP:all}
\end{center}
\end{table}

The poor performance of these models probably stems to an extent from the removal of
the quadruples whose preposition is {\it of}.
For example, they are likely to be paraphrasable as noun compounds
(e.g., {\it includes refinancing of debt} $\rightarrow$
{\it includes debt refinancing}), and therefore many of them would have been
classified correctly by paraphrasing pattern (1).
Most of the $n$-gram based word association models cover
about 81-83\% of the examples with about 66-68\% accuracy.
They are outperformed by two of the paraphrasing patterns,
which show significantly better accuracy, but much lower coverage:
(4) ``$v$ $p$ $n_2$ $n_1$'' (81.05 accuracy, 8.75\% coverage) and
(5) ``$v$ {\it pronoun} $p$ $n_2$'' (75.30\% accuracy, 30.40\% coverage).
Checking if $n_1$ is a pronoun yields 98.48\% accuracy, but has a very low coverage: 3.04\%
Checking if $v$ is a form of the verb {\it to be} has a better coverage, 9.53\%,
and a very good accuracy: 79.23\%.
The surface features are represented in the table by a single model:
for a given example, I add together the number of
noun attachment predicting matches and I compare that number
to the total number of verb attachment predicting matches.
This yields 73.13\% accuracy and 9.26\% coverage.
The performance for the individual surface features is shown in Table \ref{table:surfacePP}.
Some of them have very good accuracy, but also very low coverage;
this is why I add them together.

The last four lines of Table \ref{table:PP:models} show
the performance of some majority vote combinations on the full \emph{Ratnaparkhi's dataset}.
I first assign the 926 {\it of}-examples to {\it noun attachment},
and then, for the remaining 2,171 examples, I use a majority vote
which combines the bold rows in Table \ref{table:PP:models}.
This yields 85.01\% accuracy, and 91.77\% coverage.
In order to achieve 100\% coverage,
I further assign all undecided examples to {\it verb attachment},
which gives 83.63\% accuracy.

The last two lines in Table \ref{table:PP:models}
show the results of bootstrapping the above-described majority voting algorithm,
which made attachment predictions for 1,992 of the 2,171 examples.
I use these examples and the voting predictions
as training data for an $n$-gram based back-off classifier,
similar to the one used by \namecite{Collins:Brooks:1995}, but limited to words and bigrams only.

First, following \namecite{Collins:Brooks:1995},
I normalize the examples by substituting all 4-digit numbers
(possibly followed by `{\it s}', e.g., {\it 1990s}) with \texttt{YEAR},
all other numbers and \%s with \texttt{NUM},
all pronouns with \texttt{PRO},
the articles {\it a}, {\it an} and {\it the} with \texttt{ART},
and all other determiners (e.g., {\it this}, {\it one}) with \texttt{DET}.
I also lemmatize all nouns and verbs using {\it WordNet}.

Then I use the following ratio
\begin{equation}
    R_1 = \frac{\#(v,p|noun) + \#(n1,p|noun) + \#(p,n_2|noun)}{\#(v,p) + \#(n1,p) + \#(p,n_2)}
\end{equation}

\noindent where the counts are estimated from the 1,992 training examples
(not from the Web, which makes the higher-order $n$-grams unreliable);
conditioning on $noun$ means that the count
is calculated over the examples that had a {\it noun attachment} predicted.

I choose a {\it noun attachment} if $R_1 > 0.5$,
and a {\it verb attachment} if $R_1 < 0.5$.
I only make a decision if the denominator is greater than 3.
If no decision can be made, I back-off to the following ratio:
\begin{equation}
    R_2 = \frac{\#(p|N)}{\#(p)}
\end{equation}

Finally, I add this new model as an additional voter in the majority vote,
and I obtain 84.91\% accuracy, and 96.93\% coverage.
The last line in Table \ref{table:PP:models}
shows the results when I further assign all undecided examples
to {\it verb attachment}, which yields 84.37\% accuracy.

As Table \ref{table:PP:all} shows, my latest result (84.37\% accuracy) is as strong as
that of the best unsupervised approach on this collection:
\namecite{Pantel:Lin:2000} achieved 84.31\%.
Unlike their work, I do not need a collocation database,
a thesaurus, a dependency parser, nor a large domain-dependent
text corpus, which makes my approach easier to implement
and to extend to other languages.

Table \ref{table:PP:all} shows the results of other
previous researchers on \emph{Ratnaparkhi's dataset}.
It also shows the accuracy for three different baselines:

\begin{itemize}
  \item Always predicting a {\it noun attachment};
  \item Predicting {\it noun attachment} if the preposition is {\it of}, and a {\it verb attachment} otherwise;
  \item Make the most likely prediction for each preposition.
\end{itemize}

The last two baselines are pretty high, accuracy in the low seventies,
which is higher than the {\it AltaVista} results of \namecite{Lapata:Keller:05:Web:based:Models}.
However, my results and those of \namecite{Pantel:Lin:2000}
are statistically better than the all three baselines
and than all other previously published results.

\subsection{Conclusion and Future Work}

I have shown that simple unsupervised models that make use of $n$-grams,
surface features and paraphrases extracted from the largest existing corpus, the Web,
are effective for solving the problem of prepositional phrase attachment,
yielding 84.37\% accuracy, which matches the best previously published unsupervised results.
The approach does not require labeled training data, lexicons, or ontologies,
thesaurus, parsers, which makes it promising for a wide range of other NLP tasks.


There are many ways in which the presented approach can be extended.
First, there should be many more linguistically-motivated
paraphrases and surface features that are worth exploring.
For example, \namecite{Hindle:Rooth:1993} mention some interesting heuristics.
They predict a {\it verb attachment} if the verb is passive
(unless the preposition is {\it by}). They also mention that
if the NP object includes a superlative adjective as a pre-modifier
then the noun attachment is certain, which I could use as a paraphrase pattern.
In addition, when annotating the testing data, they always attach {\it light verbs} to the noun,
and {\it small clauses} to the verb.
Many other interesting features proposed and used
by \namecite{Olteanu:Moldovan:2005:PPattach} are worth trying as well.


\newpage
\section{Noun Phrase Coordination}
\label{sec:coord}

\subsection{Introduction}

Coordinating conjunctions, such as {\it and}, {\it or}, {\it but}, etc.,
pose major challenges to parsers; their proper handling
is also essential for understanding the semantics of the sentence.
Consider the following ``cooked'' example:

\vspace{6pt}
\begin{quotation}
``\emph{The Department of Chronic Diseases \textbf{and} Health
Promotion leads \textbf{and} strengthens global efforts to prevent \textbf{and}
control chronic diseases \textbf{or} disabilities \textbf{and} to promote health
\textbf{and} quality of life.}''
\end{quotation}
\vspace{-6pt}

Coordinating conjunctions can link two words,
two constituents (e.g., NPs), two clauses, two sentences, etc.
Therefore, the first challenge is to identify the
boundaries of the conjuncts of each coordination. The next
problem comes from the interaction of the coordinations with
other constituents that attach to its conjuncts (most often
prepositional phrases). In the example above, we need to decide
between the following two bracketings: {\it $[$ health and $[$quality of life$]$ $]$} and
{\it $[$ $[$health and quality$]$ of life $]$}.
From a semantic point of view, we need to determine whether
the {\it or} in {\it chronic diseases or disabilities} really means {\it
or} or is used as an {\it and} \cite{Agarwal:Boggess:1992:conj}. Finally,
there is an ambiguity between {\it NP-coordination} and {\it noun-coordination},
i.e. between {\it $[$ $[$chronic diseases$]$ or $[$disabilities$]$ $]$}
and {\it $[$ chronic $[$diseases or disabilities$]$ $]$}.

Below I focus on a special case of the latter problem.
Consider the noun phrase {\it car and truck production}.
Its actual meaning is {\it car production and truck production}.
However, for reasons of economy of expression,
the first instance of {\it production} has been left out.
By contrast, in {\it president and chief executive}, {\it president}
is coordinated with {\it chief executive}, and nothing has been compressed out
(it does not mean {\it president executive and chief executive}).
There is also a third option, an all-way coordination,
where the coordinated parts are inseparable from the whole,
as in {\it Securities and Exchange Commission}.

More formally, I consider quadruples $(n_1,c,n_2,h)$,
where $n_1$ and $n_2$ are nouns, $c$ is a coordinating conjunction,
and $h$ is the head noun\footnote{Quadruples of the kind $(n,h_1,c,h_2)$,
e.g. {\it company/$n$ cars/$h_1$ and/$c$ trucks/$h_2$},
can be handled in a similar way.}. I further limit $c$ to be {\it and} or {\it or} only.
The task is to decide between an NP-coordination and a noun-coordination,
given the quadruple only and independently of the local context.
Syntactically, the distinction can be expressed by the following bracketings:

\begin{center}
\begin{tabular}{ll}
$[$ $n_1$ $c$ $n_2$ $]$ $h$ & (noun coordination)\\
$[$ $n_1$ $]$ $c$ $[$ $n_2$ $h$ $]$ & (NP coordination)
\end{tabular}
\end{center}

In order to make the task more realistic
(from a parser's perspective), I ignore the option of all-way
coordination and I try to predict the bracketing in {\it Penn Treebank}
\cite{Marcus:al:1994} for configurations of this kind 
The {\it Penn Treebank} has a flat NP in case of noun-coordination,
e.g.,

\vspace{12pt}
\begin{verbatim}
    (NP car/NN and/CC truck/NN production/NN)
\end{verbatim}

In case of NP-coordination, there is an NP that contains two internal NPs:

\begin{verbatim}
    (NP
        (NP president/NN)
        and/CC
        (NP chief/NN executive/NN))
\end{verbatim}

All-way coordinations can appear bracketed either way and make the task harder.
\vspace{6pt}

\subsection{Related Work}

Coordination ambiguity is under-explored, despite being
one of the major sources of structural ambiguity
(e.g., {\it and} and {\it or} account for about 3\% of the word tokens in {\it BNC}),
and despite being the hardest type of dependency to parse
(e.g., \namecite{Collins:2003:CL:parsing} reports only 61.47\% recall and 62.20\% precision
when parsing dependencies involving coordination).


\namecite{Rus:al:2002:bracketing} present a deterministic rule-based approach
for bracketing {\it in context} of coordinated NPs of the kind
``$n_1$ $c$ $n_2$ $h$'' as a necessary step towards logical form
derivation. Their algorithm uses POS tagging, syntactic parsing,
manually annotated semantic senses, lookups in a semantic network ({\it WordNet}),
and the type of the coordination conjunction (e.g., {\it and}, {\it or}),
in order to make a three-way decision:
NP-coordination, noun-coordination, and all-way coordination.
Using a back-off sequence of three different heuristics,
they achieve 83.52\% accuracy (baseline 61.52\%) on a set of 298 examples.
When three additional context-dependent heuristics and 224 additional examples
with local contexts were added, the accuracy jumped to 87.42\%
(baseline 52.35\%), with 71.05\% coverage.

\namecite{Resnik:1999:taxonomy} tries to bracket the following kinds
of three- and four-noun configurations: 
``$n_1$ $and$ $n_2$ $n_3$'' and ``$n_1$ $n_2$ $and$ $n_3$ $n_4$''.
While there are two bracketing options for the former,
five valid bracketings exist for the latter.
Following \namecite{Kurohashi:Nagao:1992:conj}, Resnik makes decisions based
on similarity of form (i.e., number agreement: 53\% accuracy,
90.6\% coverage), similarity of meaning (66\% accuracy, 71.2\% coverage), and
conceptual association (75\% accuracy, 69.3\% coverage). Using a decision tree ,
he achieves 80\% accuracy (baseline 66\%) at 100\% coverage for the three-noun
coordinations. For the four-noun coordinations the accuracy is
81.6\% (baseline 44.9\%), 85.4\% coverage.

\namecite{Chantree:2005:coord} cover a large set of bracketing ambiguities, not limited to nouns.
Using distributional information from the {\it BNC},
they calculate similarities between words, similarly to \namecite{Resnik:1999:taxonomy},
and achieve an F-measure below 50\%.

\namecite{Goldberg:1999:pp} brackets phrases of the kind ``$n_1$ $p$ $n_2$ $c$ $n_3$'', e.g.,
{\it box/$n_1$ of/$p$ chocolates/$n_2$ and/$c$ roses/$n_3$}.
Using an adaptation of the maximum entropy algorithm
used by \namecite{Ratnaparkhi:1998} for PP-attachment,
she achieves 72\% accuracy (baseline 64\%). 

\namecite{Agarwal:Boggess:1992:conj} focus on
determining the boundaries of the conjuncts of coordinate conjunctions.
Using POS and case labels in a deterministic
algorithm, they achieve 81.6\% accuracy. 
\namecite{Kurohashi:Nagao:1992:conj} work on the same problem for Japanese. Their algorithm
looks for similar word sequences among with sentence simplification and
achieves 81.3\% accuracy. 
\namecite{Okumura:Muraki:1994:coord} address that problem heuristically
using orthographical, syntactic, and semantic information.

In recent work, \namecite{hogan:2007:coord} presents a method
for improving NP coordination disambiguation within the framework of a
lexicalised history-based parsing model. Using two main information sources,
symmetry in conjunct structure and dependency between conjunct's lexical heads,
he achieves 73.8\% F-measure (baseline: 69.9\%).
This work is related to that of
\namecite{Ratnaparkhi:al:1994:parsing} and \namecite{Charniak:Johnson:2005:parsing},
who use specialized features targeting coordination in discriminative re-rankers for parsing.

\namecite{Buyko:al:2007:coord} resolve coordination ellipses for biological named entities.
Using conditional random fields (CRFs), they achieve 93\% $F$-measure on the GENIA corpus.
On the same corpus, \namecite{shimbo:hara:2007:coord} apply a sequence-alignment
model for detecting and disambiguating coordinate conjunctions,
achieving F-measure of 70.5\% and 57.2\%, respectively.

\subsection{Models and Features}

\subsubsection{$n$-gram Models}


I use the following $n$-gram models:

({\it i}) $\#(n_1,h)$ vs. $\#(n_2,h)$

({\it ii}) $\#(n_1,h)$ vs. $\#(n_1,c,n_2)$
\vspace{6pt}

Model ({\it i}) compares how likely it is for $n_1$ to modify
$h$, as opposed to $n_2$ modifying $h$. Model ({\it ii}) checks
which association is stronger: between $n_1$ and $h$, or between
$n_1$ and $n_2$. When calculating the frequency for $\#(n_1,c,n_2)$,
I query for both {\it or} and {\it and} and I add up the counts,
regardless of whether the original coordination $c$ was {\it or} or {\it and}.

\subsubsection{Paraphrasing Patterns}

I use the following paraphrasing patterns:

\vspace{12pt}
\begin{tabular}{lll}
(1) & $n_2$ $c$ $n_1$ $h$ & (noun-coordination)\\
(2) & $n_2$ $h$ $c$ $n_1$ & (NP-coordination)\\
(3) & $n_1$ $h$ $c$ $n_2$ $h$ & (noun-coordination)\\
(4) & $n_2$ $h$ $c$ $n_1$ $h$ & (noun-coordination)
\end{tabular}
\vspace{12pt}

If matched frequently enough, the patterns predict
the coordination decision indicated in parentheses. If not found
or found infrequently, the opposite decision is made.

{\bf Pattern (1)} predicts a noun-coordination.
It switches the places of $n_1$ and $n_2$ in the coordinated NP, e.g.,
\begin{center}
    {\it \underline{bar/$n_1$} and/$c$ \textbf{pie/$n_2$} graph/$h$}
\end{center}

\noindent is paraphrased as
\begin{center}
    {\it \textbf{pie/$n_2$} and/$c$ \underline{bar/$n_1$} graph/$h$}
\end{center}

{\bf Pattern (2)} predicts an NP-coordination.
It moves $n_2$ and $h$ together to the left of the coordination
conjunction, and places $n_1$ to the right, e.g.,
\begin{center}
    {\it \underline{president/$n_1$} and/$c$ \textbf{chief/$n_2$ executive/$h$}}
\end{center}

\noindent is paraphrased as
\begin{center}
    {\it \textbf{chief/$n_2$ executive/$h$} and/$c$ \underline{president/$n_1$}}
\end{center}

{\bf Pattern (3)} predicts a noun-coordination.
It inserts the elided head $h$ after $n_1$ with the hope
that if there is ellipsis of the head $h$,
a coordination involving the full phrase ``$n_1$ $h$''
will be likely to be found elsewhere on the Web, e.g.,
\begin{center}
    {\it bar/$n_1$ and/$c$ pie/$n_2$ \underline{graph/$h$}}
\end{center}

\noindent is paraphrased as
\begin{center}
    {\it bar/$n_1$ \underline{graph/$h$} and/$c$ pie/$n_2$ \underline{graph/$h$}}
\end{center}

{\bf Pattern (4)} predicts a noun-coordination.
It combines pattern (1) and pattern (3)
by not only inserting $h$ after $n_1$,
but also switching the places of $n_1$ and $n_2$.
For example,
\begin{center}
    {\it \textbf{bar/$n_1$} and/$c$ \underline{pie/$n_2$ graph/$h$}}
\end{center}

\noindent is paraphrased as
\begin{center}
    {\it \underline{pie/$n_2$ graph/$h$} and/$c$ \textbf{bar/$n_2$ graph/$h$}}
\end{center}

\subsubsection{Heuristics}

I also included some of the heuristics proposed by \namecite{Rus:al:2002:bracketing},
as shown in Table \ref{table:CC}.
Heuristic 1 predicts an NP-coordination when $n_1$ and $n_2$ are the same word,
e.g., {\it milk/$n_1$ and/$c$ milk/$n_2$ products/$h$}. Heuristics 2 and 3 perform a
lookup in {\it WordNet} and I decided not not to use them.
Heuristics 4, 5 and 6 exploit the local context,
namely the adjectives modifying $n_1$ and/or $n_2$.
Heuristic 4 predicts an NP-coordination if both $n_1$
and $n_2$ are modified by adjectives. Heuristic 5 predicts
a noun-coordination if $c$ is {\it or} and $n_1$ is modified
by an adjective, but $n_2$ is not. Heuristic 6 predicts NP-coordination
if $n_1$ is not modified by an adjective, but $n_2$ is.

Since I target coordination disambiguation independent of context,
and since search engines lack POS annotations,
I could not use the original heuristics 4, 5 and 6.
Therefore, in my experiments, I used adapted versions
of these heuristics that look for a determiner
(e.g., {\it a, an, the}) rather than an adjective.

\subsubsection{Number Agreement}

I also include the number agreement model proposed by
\namecite{Resnik:1993:thesis}, which makes prediction as follows:

(a) if $n_1$ and $n_2$ match in number, but $n_1$ and $h$ do not,
predict noun-coordination;

(b) if $n_1$ and $n_2$ do not match in number, but $n_1$
and $h$ do, predict NP-coordination;

(c) otherwise leave undecided.

\subsubsection{Web-Derived Surface Features}

The set of surface features is similar to the one I used for
PP-attachment in section \ref{sec:PP:attach},
and include brackets, slash, comma, colon, semicolon, dot,
question mark, exclamation mark, and any special character.
There are two additional NP-coordination predicting features: a dash after
$n_1$ and a slash after $n_2$, see Table \ref{table:surfaceCC}.

\begin{table}
\begin{center}
\begin{small}
\begin{tabular}{|l@{ }ccc|}
\multicolumn{1}{l}{\bf Surface Feature} & {\bf Prediction} & {\bf Accuracy (\%)} & \multicolumn{1}{c}{\bf Coverage (\%)} \\
\hline
(buy) and sell orders & NP-coordination & 33.33 & 1.40\\
buy (and sell orders) & NP-coordination & 70.00 & 4.67 \\
buy: and sell orders & NP-coordination & 0.00 & 0.00 \\
buy; and sell orders & NP-coordination & 66.67 & 2.80 \\
buy. and sell orders & NP-coordination & 68.57 & 8.18 \\
buy[...] and sell orders & NP-coordination & 49.00 & 46.73 \\
\hline
buy- and sell orders & noun-coordination & 77.27 & 5.14\\
buy and sell / orders & noun-coordination & 50.54 & 21.73\\
(buy and sell) orders & noun-coordination & 92.31 & 3.04 \\
buy and sell (orders) & noun-coordination & 90.91 & 2.57 \\
buy and sell, orders & noun-coordination & 92.86 & 13.08 \\
buy and sell: orders & noun-coordination & 93.75 & 3.74 \\
buy and sell; orders & noun-coordination & 100.00 & 1.87 \\
buy and sell. orders & noun-coordination & 93.33 & 7.01 \\
buy and sell[...] orders & noun-coordination & 85.19 & 18.93 \\
\hline
{\bf Surface features (sum)} &  & {\bf 82.80$\pm$8.93} & {\bf 21.73} \\
\hline
\end{tabular}
\end{small} \caption{{\bf NP coordination surface features.}
Accuracy and coverage shown are across all examples, not just the
{\it buy and sell orders} shown.} \label{table:surfaceCC}
\end{center}
\end{table}

\begin{table}
\begin{center}
\begin{small}
\begin{tabular}{|l|c|c|}
  \multicolumn{1}{c}{\bf Model} & \multicolumn{1}{c}{\bf Accuracy (\%)} & \multicolumn{1}{c}{\bf Coverage (\%)} \\
  \hline
  Baseline: {\it noun-coordination} & {\small 56.54$\pm$4.73} & {\small 100.00}\\
  \hline
  $(n_1,h)$ vs. $(n_2,h)$ & {\small \textbf{80.33$\pm$7.93}} & {\small 28.50}\\
  $(n_1,h)$ vs. $(n_1,c,n_2)$ & {\small 61.14$\pm$7.03} & {\small 45.09}\\
  \hline
  $(n_2,c,n_1,h)$ & {\small \textbf{88.33$\pm$10.51}} & {\small 14.02}\\
  $(n_2,h,c,n_1)$ & {\small \textbf{76.60$\pm$9.50}} & {\small 21.96}\\
  $(n_1,h,c,n_2,h)$ & {\small \textbf{75.00$\pm$18.36}} & {\small 6.54}\\
  $(n_2,h,c,n_1,h)$ & {\small \textbf{78.67$\pm$10.55}} & {\small 17.52}\\
  \hline
  Heuristic 1 & {\small \textbf{75.00$\pm$44.94}} & {\small 0.93}\\
  Heuristic 4 & {\small 64.29$\pm$18.46} & {\small 6.54}\\
  Heuristic 5 & {\small 61.54$\pm$13.58} & {\small 12.15}\\
  Heuristic 6 & {\small \textbf{87.09$\pm$15.95}} & {\small 7.24}\\
  \hline
  Number agreement & {\small \textbf{72.22$\pm$6.62}} & {\small 46.26}\\
  \hline
  Surface features (sum) & {\small \textbf{82.80$\pm$8.93}} & {\small 21.73}\\
  \hline
  {\it Majority vote} & {\small 83.82$\pm$4.25} & {\small 80.84}\\
  {\it Majority vote, N/A $\rightarrow$ no ellipsis} & {\small \emph{\textbf{80.61$\pm$4.01}}} & {\small \emph{\textbf{100.00}}}\\
\hline
\end{tabular}
\end{small}

\caption{\bf NP Coordination: evaluation results.}
\label{table:CC}
\end{center}
\end{table}

\subsection{Evaluation}

I evaluate the above-described models on a collection of 428 quadruples
of the form ($n_1$, $c$, $n_2$, $h$), extracted from the {\it Penn Treebank}.
I look for NPs of the following kinds:

\vspace{-3pt}
\begin{center}
    (NP $\ldots$ $n_1$/N $c$/CC $n_2$/N $h$/N)
\end{center}
\vspace{-3pt}
\noindent and
\vspace{-3pt}
\begin{center}
    (NP
        (NP  $\ldots$ $n_1$/N)
        $c$/CC
        (NP  $\ldots$ $n_2$/N $h$/N))
\end{center}
\vspace{-3pt}

The nouns $n_1$, $n_2$ and $h$ can be POS tagged as NN, NNS, NNP, or NNPS,
and the NPs can contain determiners and non-noun modifiers preceding $n_1$ and/or $n_2$,
but no other nouns.

The full dataset is shown in Appendix \ref{appendix:coord:set}.
It contains some repetitions of quadruples, e.g.,
the NP-coordinated (\emph{test}, \emph{and}, \emph{Learning}, \emph{Materials});
in some cases, the different instances of the same quadruple can have different
labels, e.g., (\emph{FTC}, \emph{and}, \emph{Justice}, \emph{Department})
appears both noun-coordinated and NP-coordinated.
Such inconsistencies stem from either context dependence or from inconsistencies
in the {\it Penn Treebank} annotations.

As Table \ref{table:CC} shows, the $n$-gram model ({\it i})
performs very well (80.33\% accuracy, 28.50\% coverage),
but the $n$-gram model ({\it ii}) is not that accurate (61.14\% accuracy, 7.03\% coverage).
This is probably because the ``$n_1$ $c$ $n_2$'' is a trigram,
which makes it less likely to be observed
than the alternative ``$n_1$ $h$'', which is a bigram.

Interestingly, the number agreement feature yields 72.22\% accuracy and
46.26\% coverage, while in the experiments of \namecite{Resnik:1993:thesis}
it was the best feature 90\% accuracy and 53\% coverage.
This might indicate that my dataset is harder.

The surface features are represented in Table \ref{table:CC} by a single model:
for a given example, I add up together the number of
noun-coordination predicting matches and I compare that number
to the total number of NP-coordination predicting matches,
which yields 82.80\% accuracy and 21.73\% coverage.
The performance for the individual surface features is shown in Table \ref{table:surfaceCC}.
We can see that they are very good predictors of noun-coordination,
but are less reliable when predicting NP-coordination.

I combine the bold rows of Table \ref{table:CC} in a majority vote,
obtaining 83.82\% accuracy and 80.84\% coverage.
Assigning all undecided cases to NP-coordination,
yields 80.61\% accuracy (and 100\% coverage).

\subsection{Conclusion and Future Work}

I have shown that the simple Web-based unsupervised models
similar to the ones I proposed for {\it noun compound bracketing} in chapter \ref{chapter:NC:bracketing} --
$n$-grams, surface features and linguistically-motivated paraphrases --
are effective for solving the problem of {\it NP coordination}:
I have achieved 80.61\% accuracy, which is on par with other approaches,
whose best scores fall into the low 80's for accuracy;
direct comparison is not possible though,
as the tasks and the datasets all differ.

Other kinds of coordination-related attachment ambiguity problems
might be addressable in a similar way, e.g.,
identification of the boundaries of the coordination conjuncts, or
interactions between coordinations and prepositional phrases,
with the ultimate goal of extending and applying the proposed approach to
real NLP tasks, e.g., named entity recognition and syntactic parsing,
as done in \namecite{Buyko:al:2007:coord} and \namecite{hogan:2007:coord}, respectively.

\chapter{Quality and Stability of Page Hit Estimates}
\label{chapter:instability}

Web search engines provide an easy access for NLP researchers to
world's biggest corpus, but not without drawbacks.
In this chapter, I point to some problems and limitations of using
search engine page hits as a proxy for $n$-gram frequency estimates.
I further describe a study on the quality and stability of such estimates
across search engines and over time,
as well as on the impact of using word inflections and of limiting the queries to English pages.
Using the task of noun compound bracketing and 14 different $n$-gram based models,
I illustrate that while sometimes causing sizable fluctuations,
variability's impact generally is not statistically significant.

An initial version of this study appeared in \cite{nakov:hearst:2005:RANLP:web:hits:proxy}.

\section{Using Web Page Hits: Problems and Limitations}

\subsection{Lack of Linguistic Restrictions on the Query}

Many problems with page hit estimates stem from
Web search engines not allowing for linguistic restrictions on the query.
While some linguistic annotations might be supported and used by Web search engines internally,
they are not accessible to the end user, which can limit
search engine's value as a tool for corpus linguistics research.

For example, when bracketing the three-word noun compound {\it home health care},
the probabilistic association models described in section \ref{sec:assoc:scores}
would need to calculate the probability that {\it health} modifies {\it care}, i.e.
$\mathrm{Pr}(health \rightarrow care|care) = \frac{\#(\mathrm{''}health\ care\mathrm{''})}{\#(care)}$.
This calculation requires estimates for the frequencies of {\it ``health care''} and {\it care},
ideally where both words are used as nouns.
While the exact phrase query ``{\it health care}'' would almost
guarantee that both words are used nouns, a query for {\it care}
would return many pages where that word is used as a verb,
which would lead to an overestimated value for the denominator,
and thus to an underestimated probability.
On the other hand, since the word {\it health} can only be a noun, the estimate for
$\mathrm{Pr}(home \rightarrow health|health) = \frac{\#(\mathrm{''}home\ health\mathrm{''})}{\#(health)}$
would not be affected. Therefore, the {\it adjacency} model (see section \ref{sec:adj})
would compare one correct probability and one underestimated probability,
which can cause potential problems when their values are close.

Further, as I have already mentioned in section \ref{sec:paraphrases},
search engines do not support queries containing place-holders
asking for a particular part-of-speech, e.g.,

\begin{center}
{\it stem cells VERB PREP DET brain}
\end{center}

\noindent where the uppercase typed placeholders stand for a verb,
a preposition and a determiner, respectively.

This makes typed placeholder queries prohibitively expensive:
for example, in the above pattern, one would need to substitute every possible
verb, every possible preposition and every possible determiner,
and then to issue a separate exact-phrase query for each combination.

\subsection{No Indexing for Punctuation}

A further problem is caused by the fact that search engines ignore punctuation at indexing time.
Consider for example the bigram {\it health care}. An exact-phrase query for
``{\it health care}'' does not guarantee that the words will be really adjacent:
they might belong to different NPs, different sentences, even different paragraphs.
For example, a search engine would find a ``legitimate'' exact-phrase bigram match
in the following sentence:

\vspace{12pt}
\begin{quotation}
``\emph{This section is for you if you care about \underline{health}, \underline{care}
about fitness, like sports and above all, want to help people.}''\footnote{From  \texttt{http://www.jobmonkey.com/sports/html/health\_fitness\_careers\_overvie.html}}
\end{quotation}
\vspace{6pt}

As I have already mentioned in section \ref{sec:bracketing:web:derived:features},
the lack of punctuation also makes impossible the direct execution
of exact-phrase queries containing
plural genitive markers (e.g., ``{\it protein synthesis' inhibition}''),
hyphens (e.g., ``{\it law-enforcement officer}''),
parentheses (e.g., ``{\it bronchoalveolar lavage (BAL) fluid}''), etc.
In the last example, a further potential problem is caused by the lack of distinction
between letters in uppercase vs. in lowercase.

\subsection{Page Hits are Not n-grams}

There are other reasons why using page hits as a proxy for $n$-gram frequencies can
yield some counter-intuitive results. Consider the bigrams
$w_1w_4$, $w_2w_4$ and $w_3w_4$ and a page that contains each
of them exactly once. A search engine will contribute a page
count of 1 for $w_4$ instead of a frequency of 3; thus the
number of page hits for $w_4$ can be smaller than that for the
sum of the bigrams that contain it.
\namecite{Keller:Lapata:03:web:unseen:bigrams} describe more potential problems with page hits.

\subsection{Rounding and Extrapolation of Page Hit Estimates}
\label{sec:pagehits:rounding}

Another potentially undesirable aspect of using page hits for linguistic research
is that two of the major Web search engines, {\it Google} and {\it Yahoo!}
provide rounded estimates rather than exact numbers.
For example, in June 2005, {\it MSN Search}
returns 60,\underline{219,609} page hits for {\it cell},
while {\it Google} and {\it Yahoo!} return 397,\underline{000,000} and 637,\underline{000,000}, respectively.
This rounding can be problematic, especially when comparing two ratios of page hits,
e.g., $\frac{\#(stem,cell)}{\#(cell)}$ and $\frac{\#(brain,stem)}{\#(stem)}$.

The rounding is probably done since, once the numbers get somewhat large,
exact counts are not necessary for the ordinary user,
who hardly expects them from a search engine that only indexes part
of the constantly changing Web anyway.
From a technical viewpoint, calculating the exact page hits
from multiple distributed and continually changing indexes is computationally expensive.
Therefore, commercial search engines, which are optimized for quickly returning results
to the user under high load, may sample from their indexes, rather than performing exact computations
\cite{Veronis:blogGoogle:extrapolate}.
They would stop scanning their indexes once they are ready to return
$N$ results (typically 10, but no more than 100 at a time,
and no more than 1,000 in total for a given query),
and they would calculate the expected total number of pages by extrapolation,
taking into account how many results have been produced so far,
and what part of the index has been consumed in order to produce them.
Another possible cause for obtaining inflated numbers
is that page hit estimates come in part from anchor text of unexplored pages.
In other words, search engines find links to pages that contain
the indexed words, without necessarily crawling all those pages.

It is possible to observe page hits inflation by issuing
an exact-phrase query that matches less than 1,000 actual pages.
For example, at the time of writing, a {\it Google} query for ``{\it searches engines}''
returns 10,200 page hits, but if one actually asks to see them all,
it turns out that there are only 340 actual pages.

\subsection{Inconsistencies of Page Hit Estimates and Boolean Logic}

The extrapolation process has caused various speculations and allegations.
For example, in a series of publications on his popular blog\footnote{\texttt{http://aixtal.blogspot.com/}},
Jean V\'{e}ronis suggested that many search engines inflate their page hits for marketing reasons.
Since they never provide access to more than 1,000 result pages per query,
the user cannot actually verify whether the listed total number of pages is correct.
In 2005, after having observed over a period of
time the frequencies of fifty English words drawn randomly from
mid-range frequencies in a one million word corpus of English
text, V\'{e}ronis estimated that the true index sizes of {\it Google} and
{\it MSN Search} are 60\% and 75\%, respectively, of the officially announced
numbers \cite{Veronis:blogGoogle,Veronis:blogMSN}.
While he was unable to find similar inconsistencies for {\it Yahoo!},
V\'{e}ronis claims it suspiciously doubled its index in March 2005 \cite{Veronis:blogYahoo}.

V\'{e}ronis also reported problems with {\it Google}'s Boolean
logic (which he called a ``{\it Googlean logic}''). For example, in
February 2005, a query for {\it Chirac and Sarkozy} produced about
half of the page hits for {\it Chirac} alone. Even stranger
things happened for queries like {\it x AND x}, {\it x OR x},
or for ones repeating the same word multiple times \cite{Veronis:blogGoogle}.
Apparently, these problems have been fixed \cite{Veronis:blogGoogleFixing}.

The actual implications of these allegations remain unclear.
If the page hit estimates are consistent across different queries,
this should not impact {\it ratios} of page hit estimates,
and while some researchers do use operations like OR and NEAR,
e.g., \namecite{Volk:2001:www:PPattach}, \namecite{Mihalcea:Moldovan:1999:wsd}, \namecite{Lapata:Keller:04},
this is not strictly necessary: it is possible to use multiple queries instead of OR,
and NEAR could be emulated to some extent using the `*' operator in {\it Google}.

\subsection{Instability of Page Hit Estimates}

A major problem, from a research perspective, is caused
by the instability of query results.
Nowadays, Web search engines are too complex to run on a single
machine, and the queries are served by multiple servers,
which collaborate to produce the final result. In addition, the Web is dynamic,
many pages change frequently, others disappear or are created for the first time,
and therefore search engines need to update their indexes frequently;
commercial search engines often compete on how ``fresh'' their indexes are.
Finally, search engines constantly improve their algorithms
for sampling from the index, for page hit extrapolation, for result ranking, etc.
As a result, the number of page hits for a given query,
as well as the query results in general,
could change over time in unpredictable ways.

The indexes themselves are too big to be stored on a single
machine and therefore are spread across multiple servers
\cite{Brin:Page:1998:anatomy}. For availability and
efficiency reasons, there are multiple copies of the same
part of the index, which are not always synchronized with
one another since the different copies are updated at different
times. As a result, if one issues the same query multiple times
in rapid succession, connections to different physical machines
could be opened, which could yield different results.
This effect is known as search engine ``dancing''.
With a little bit of luck, one can observe {\it Google} ``dancing'' by
comparing the results of different data centers, e.g.,
\texttt{www.google.com}, \texttt{www2.google.com},
\texttt{www3.google.com}. Alternatively, one can try the {\it
Google dance tool} at:
\texttt{http://www.seochat.com/googledance}. More on the
phenomenon can be found on the Web, e.g., at
\texttt{http://dance.efactory.de} or simply by asking a search
engine about ``{\it Google dancing}''.

From a research perspective, dynamics over time is highly undesirable,
as it precludes the exact replicability of the results obtained using search engines.
At best, one could reproduce the same initial conditions, and expect similar results.
While this kind of variability is uncommon in NLP,
where one typically experiments with a static frozen corpus, it is common in
natural sciences, like physics, chemistry etc., where the same experiment
is unlikely to always yield exactly the same outcome. In fact, even in computer science,
for a fixed corpus, the exact replicability of the results may not be guaranteed,
e.g., when the algorithm has elements of randomization,
when the explanation in the paper is not detailed enough,
or the authors had a programming bug, etc.

\section{Noun Compound Bracketing Experiments}

I believe that the best way to test the impact of rounding,
possible inconsistencies and dynamics over time
is to design a set of suitable experiments organized
around a real NLP task. I chose {\it noun compound bracketing},
which can be solved using several different methods
that make use of $n$-grams of different lengths.
Below I perform series of experiments on {\it Lauer's dataset}
(see section \ref{dataset:lauer})
comparing the accuracy of 14 different $n$-gram models
from Chapter \ref{chapter:NC:bracketing}, across the following four dimensions:
(1) {\it search engine}; (2) {\it time}; (3) {\it language filter};
and (4) {\it word inflection}.


For each model, I issue exact phrase queries within a
single day. Unless otherwise stated, the queries are not
inflected and no language filter is applied. I use a
threshold on the module of the difference between the left- and
the right-predicting $n$-gram frequencies:
I only make a bracketing decision if the difference is at least five.

\subsection{Variability over Time}

I study the variability over time for {\it Google} and for {\it MSN Search}.
I chose time snapshots at varying time intervals in
order to lower the potential impact of major search engines' index changes,
in case they are scheduled at fixed time intervals.

\begin{figure}
\begin{center}
  \includegraphics[width=15cm,height=8cm]{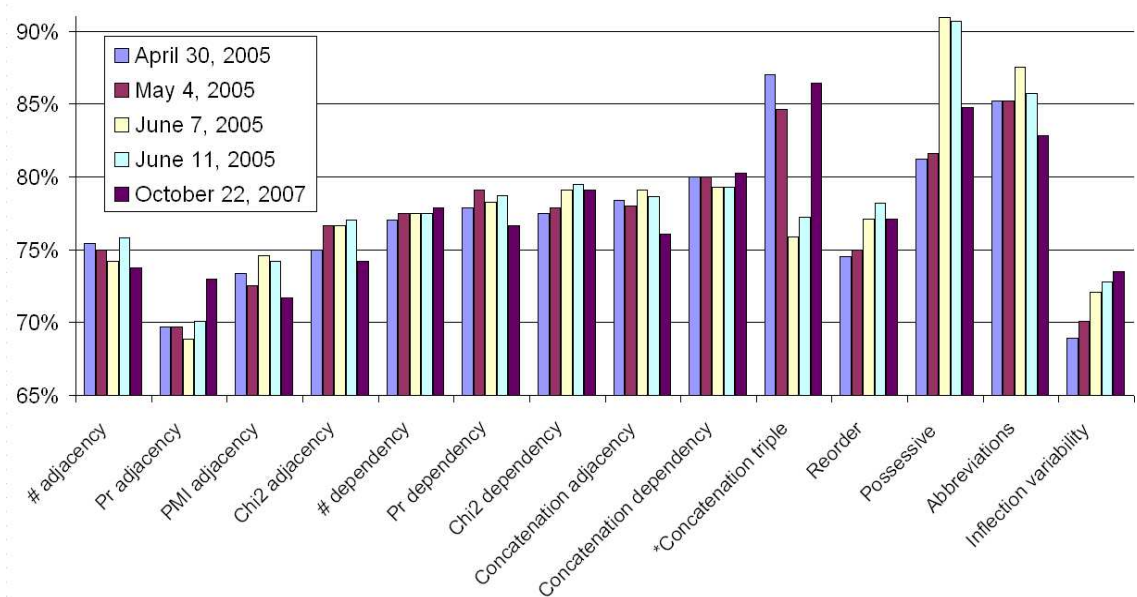}\\
  \caption{{\bf Accuracy over time for \emph{Google}:} pages in any language, no word inflections.
            The models with statistically significant variations are marked with an asterisk.}
  \label{figure:byTimeGoogle:noinfl:anylang}
\end{center}
\end{figure}

\begin{figure}
\begin{center}
  \includegraphics[width=15cm,height=8cm]{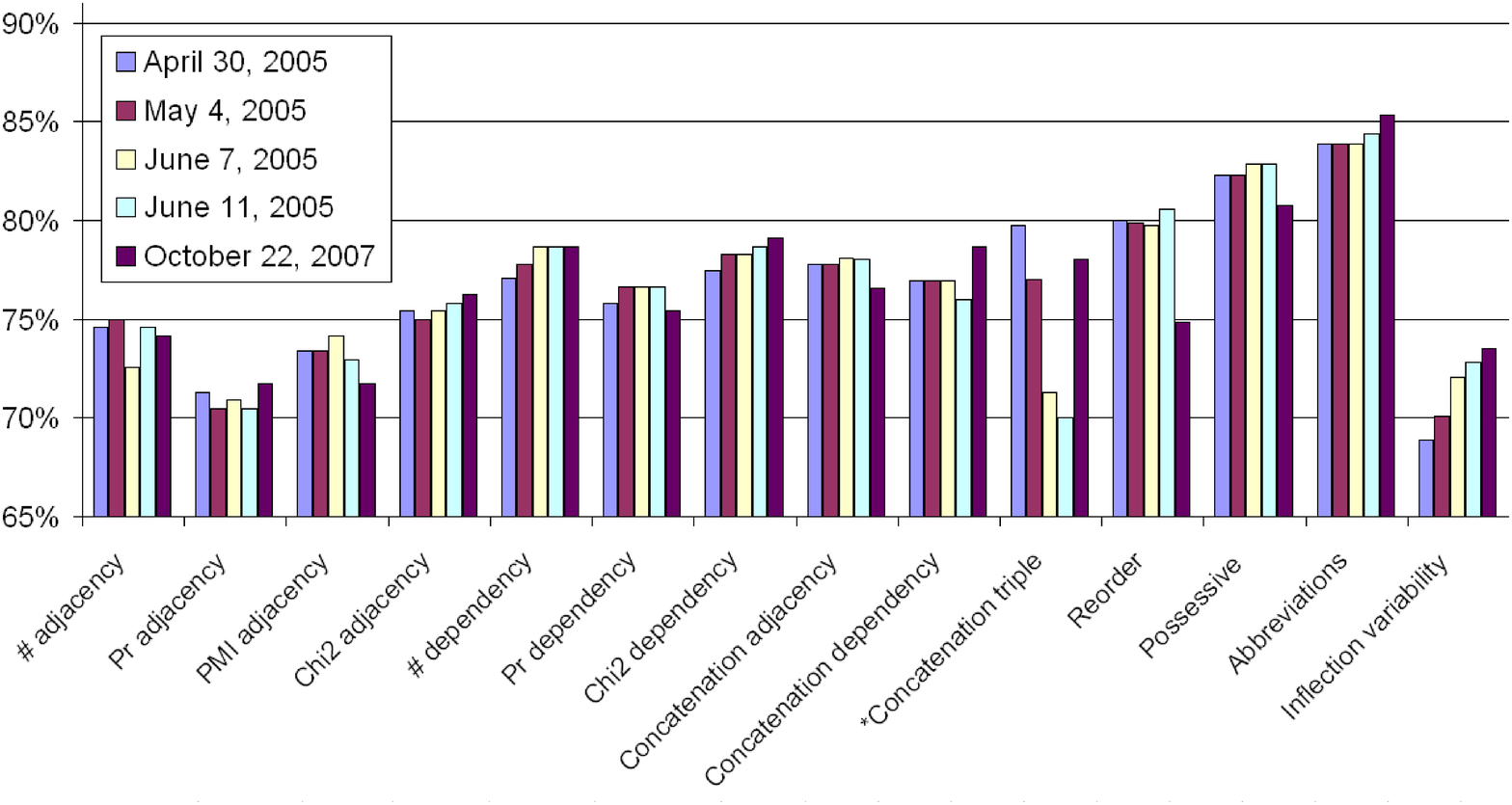}\\
  \caption{{\bf Accuracy over time for \emph{Google}:} pages in any language, with word inflections.
            The models with statistically significant variations are marked with an asterisk.}
  \label{figure:byTimeGoogle:withinfl:anylang}
\end{center}
\end{figure}

\begin{figure}
\begin{center}
  \includegraphics[width=15cm,height=8.5cm]{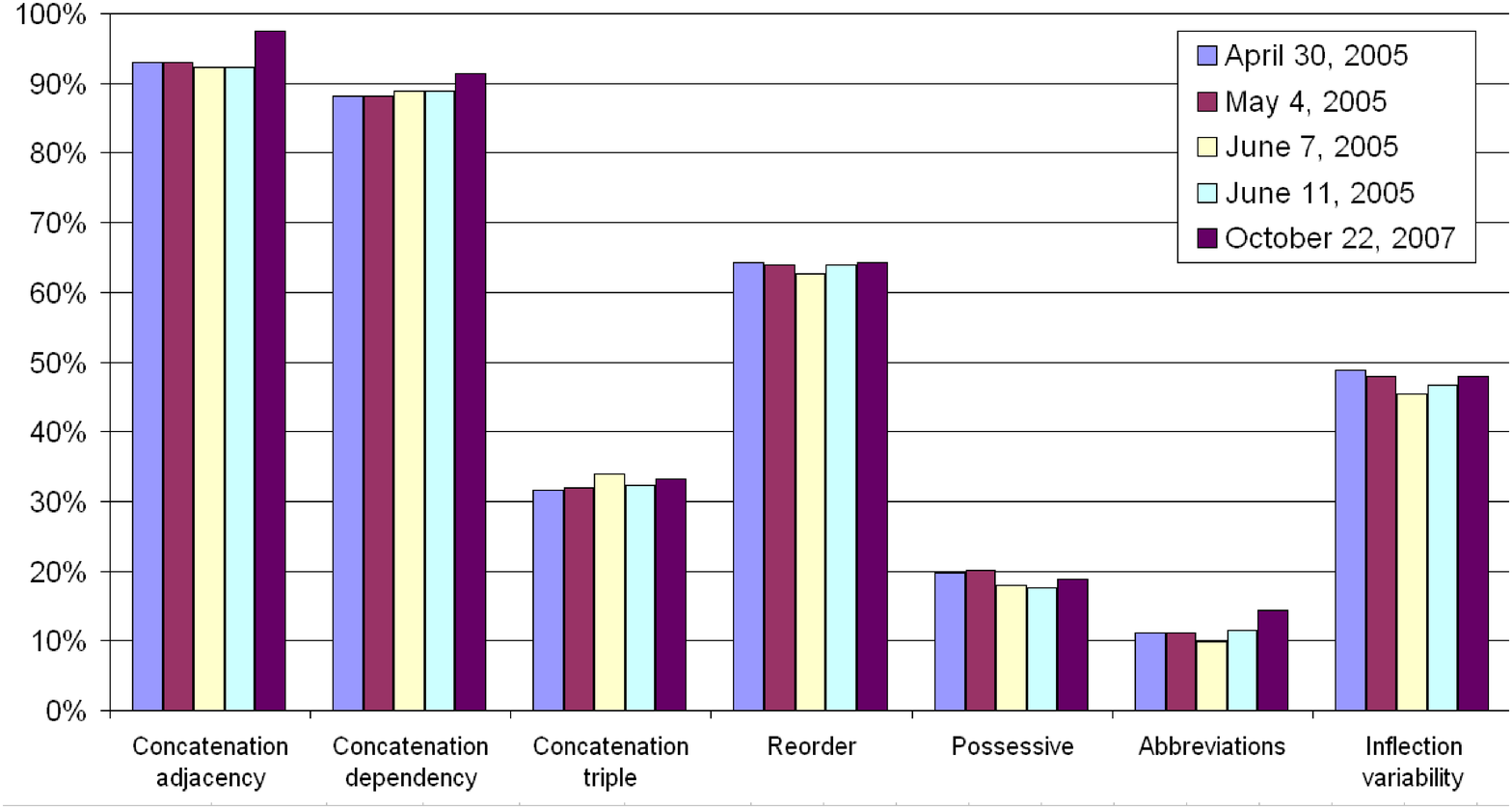}\\
  \caption{{\bf Coverage over time for \emph{Google}:} pages in any language, no word inflections.}
  \label{figure:byTimeGoogle:noinfl:anylang:R}
\end{center}
\end{figure}

\begin{figure}
\begin{center}
  \includegraphics[width=15cm,height=8.5cm]{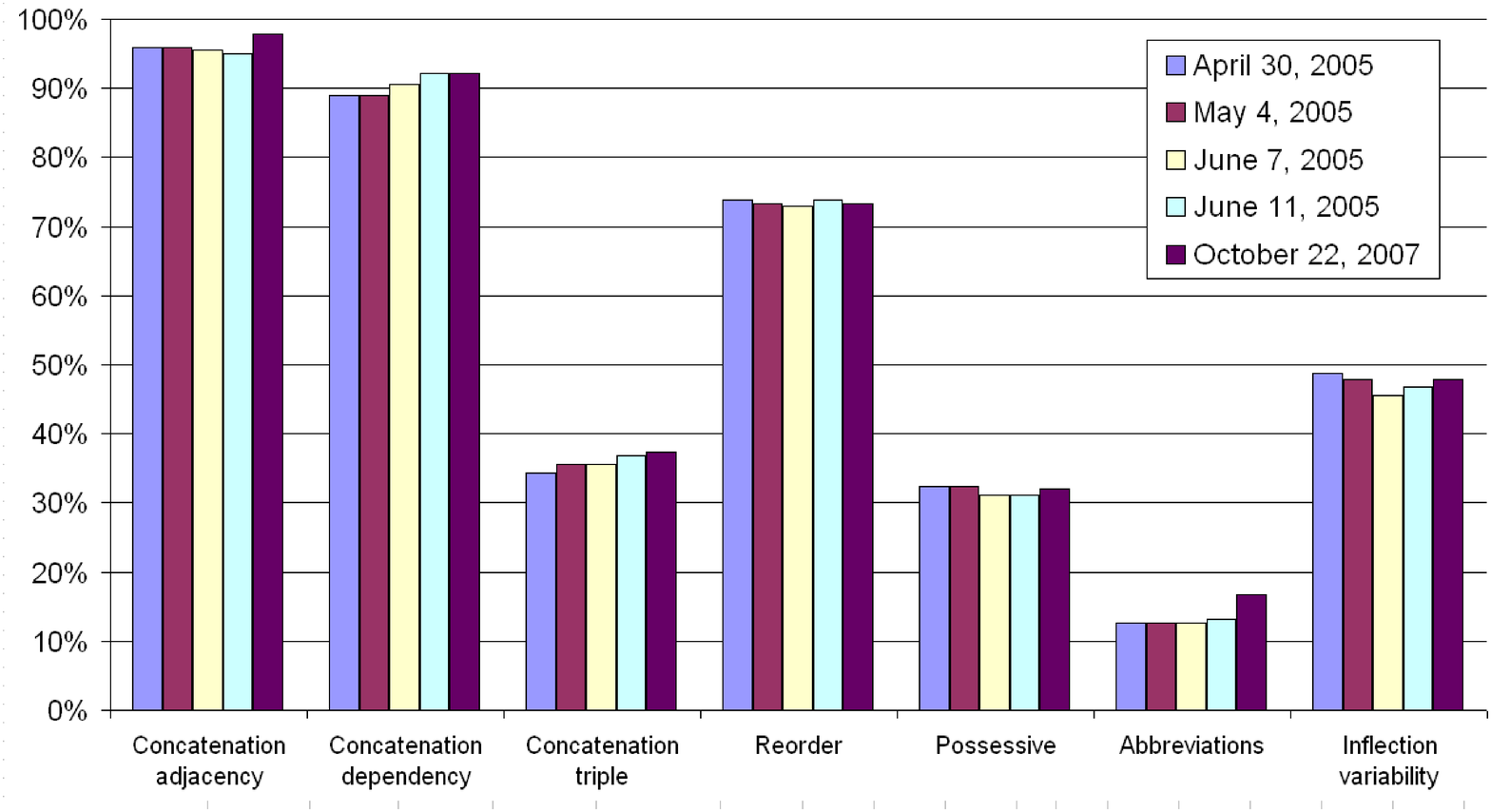}\\
  \caption{{\bf Coverage over time for \emph{Google}:} pages in any language, with word inflections.}
  \label{figure:byTimeGoogle:withinfl:anylang:R}
\end{center}
\end{figure}

\begin{figure}
\begin{center}
  \includegraphics[width=15cm,height=8cm]{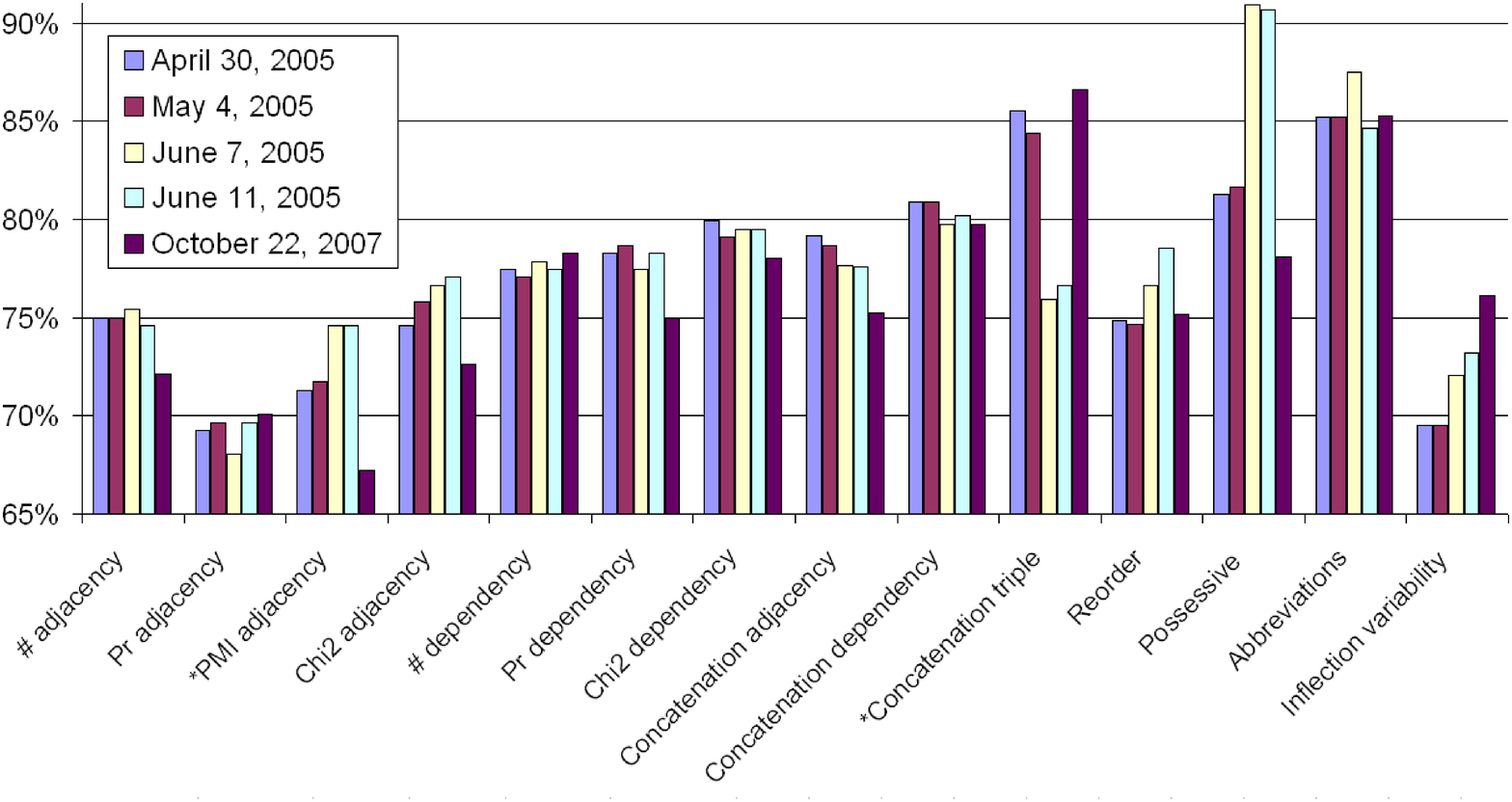}\\
  \caption{{\bf Accuracy over time for \emph{Google}:} English pages only, no word inflections.
            The models with statistically significant variations are marked with an asterisk.}
  \label{figure:byTimeGoogle:noinfl:english}
\end{center}
\end{figure}

\begin{figure}
\begin{center}
  \includegraphics[width=15cm,height=8cm]{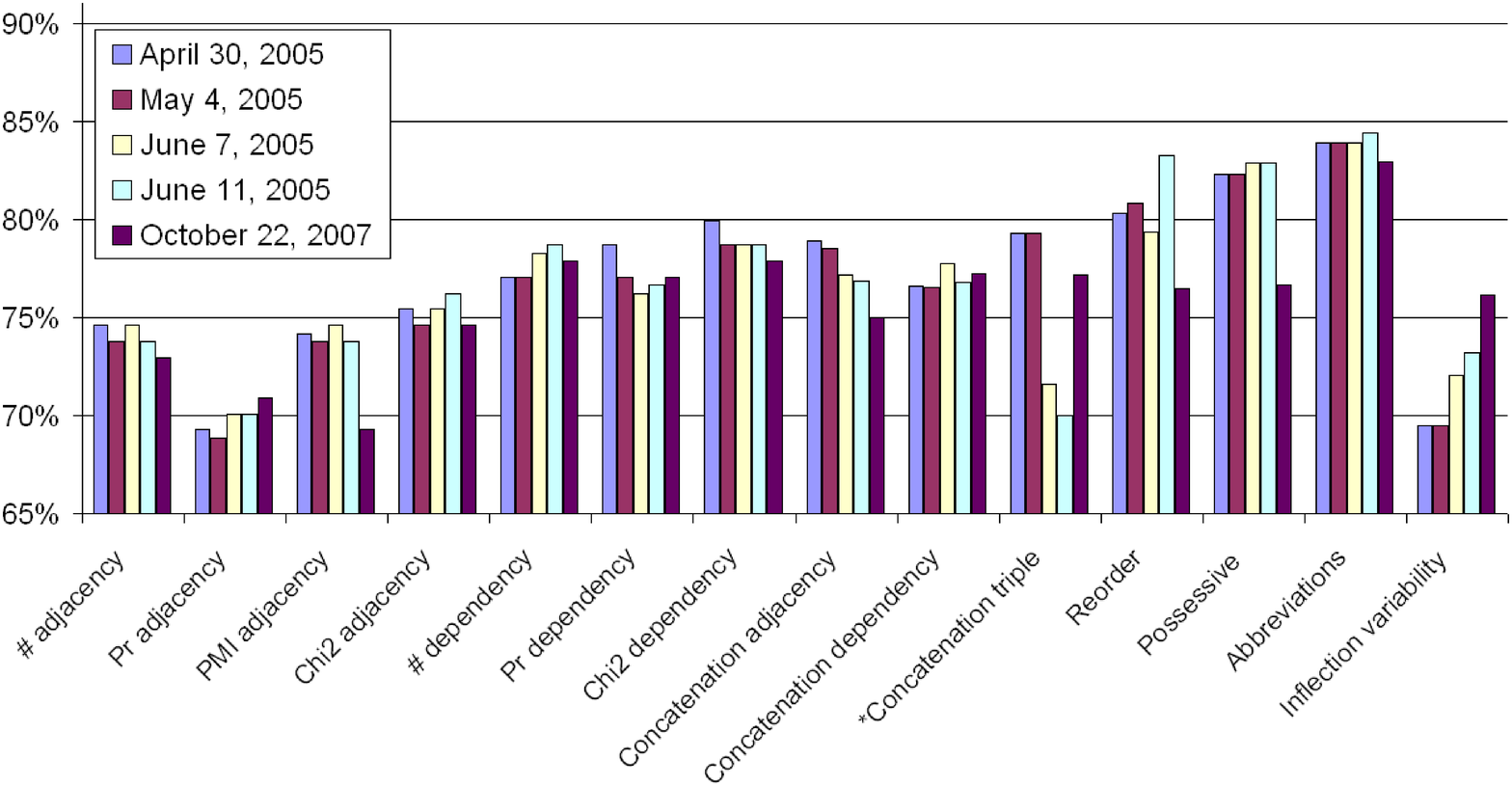}\\
  \caption{{\bf Accuracy over time for \emph{Google}:} English pages only, with word inflections.
            The models with statistically significant variations are marked with an asterisk.}
  \label{figure:byTimeGoogle:withinfl:english}
\end{center}
\end{figure}

\begin{figure}
\begin{center}
  \includegraphics[width=15cm,height=8.5cm]{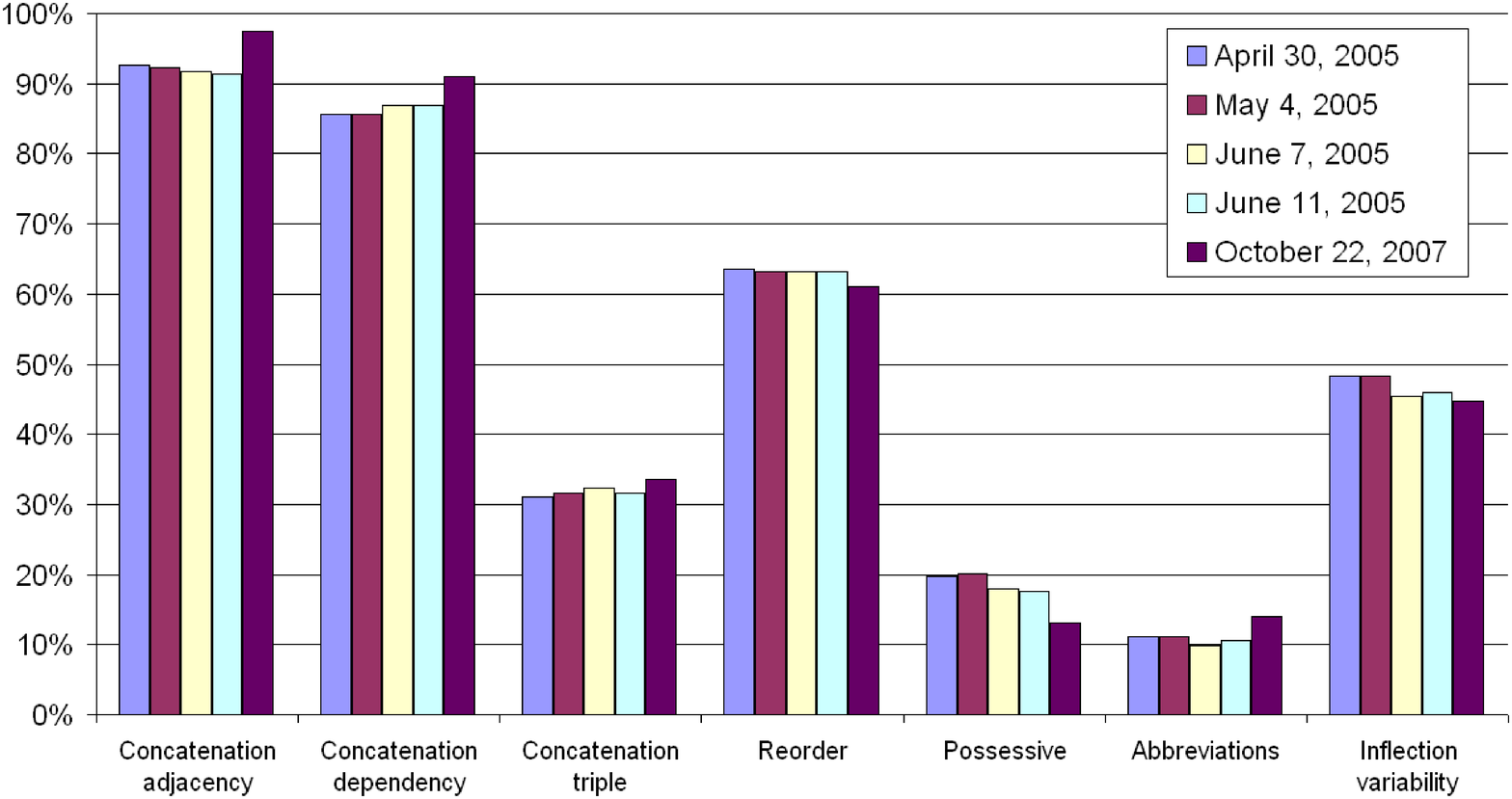}\\
  \caption{{\bf Coverage over time for \emph{Google}:} English pages only, no word inflections.}
  \label{figure:byTimeGoogle:noinfl:english:R}
\end{center}
\end{figure}

\begin{figure}
\begin{center}
  \includegraphics[width=15cm,height=8.5cm]{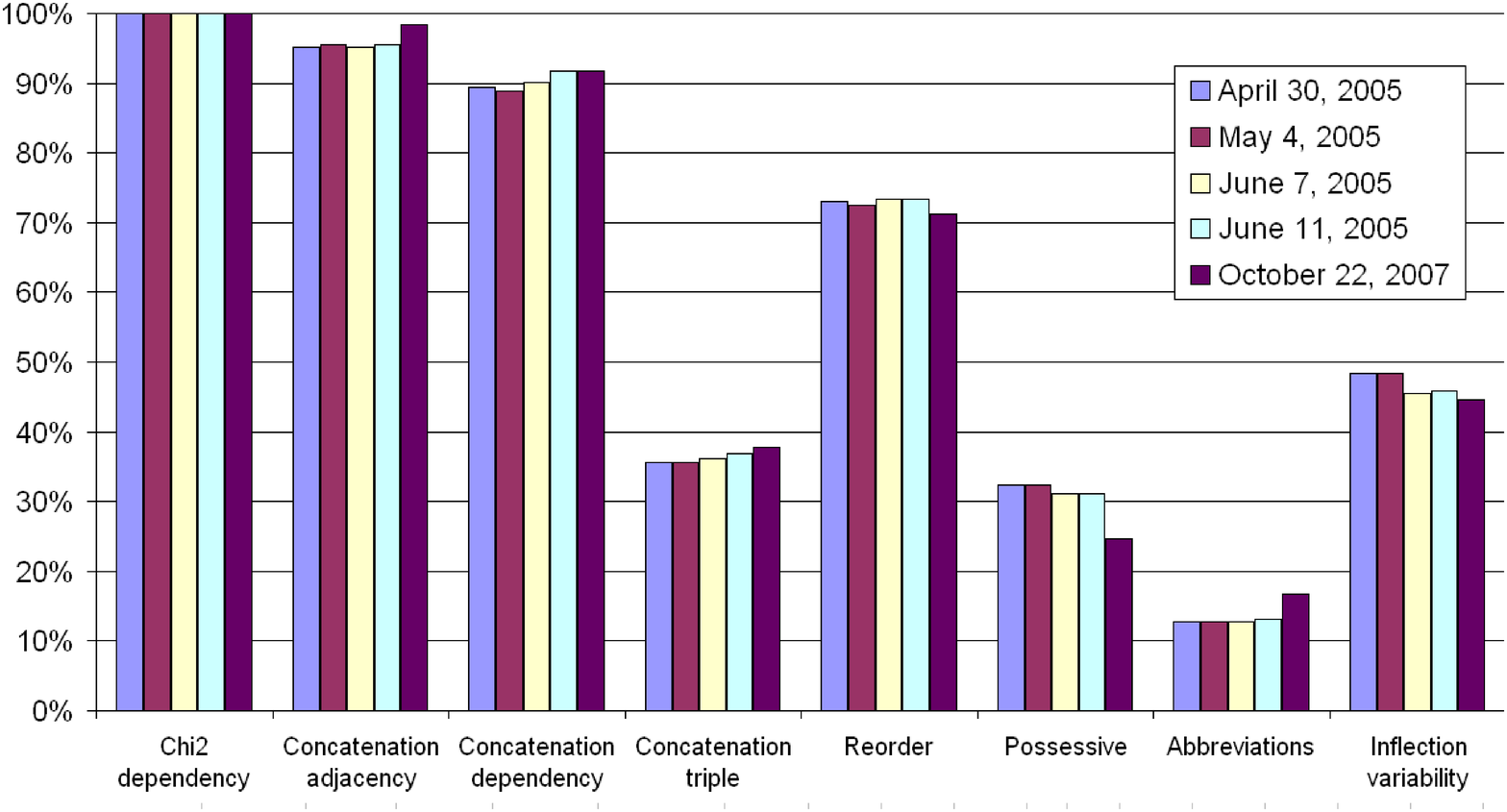}\\
  \caption{{\bf Coverage over time for \emph{Google}:} English pages only, with word inflections.}
  \label{figure:byTimeGoogle:withinfl:english:R}
\end{center}
\end{figure}

\begin{figure}
\begin{center}
  \includegraphics[width=15cm,height=8.5cm]{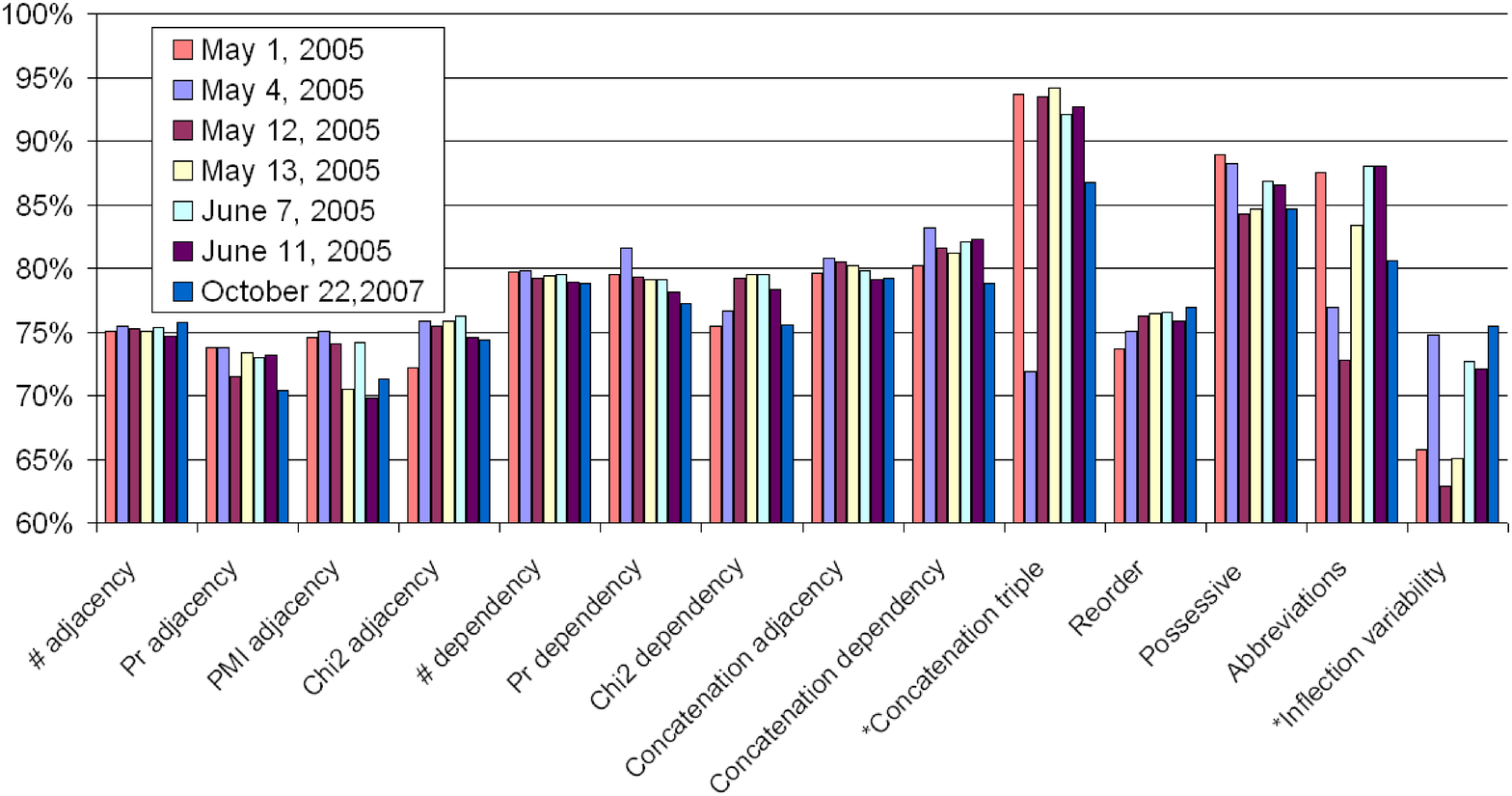}\\
  \caption{{\bf Accuracy over time for \emph{MSN}:} pages in any language, no word inflections.
            The models with statistically significant variations are marked with an asterisk.}
  \label{figure:byTimeMSN:noinfl:anylang}
\end{center}
\end{figure}

\begin{figure}
\begin{center}
  \includegraphics[width=15cm,height=8.5cm]{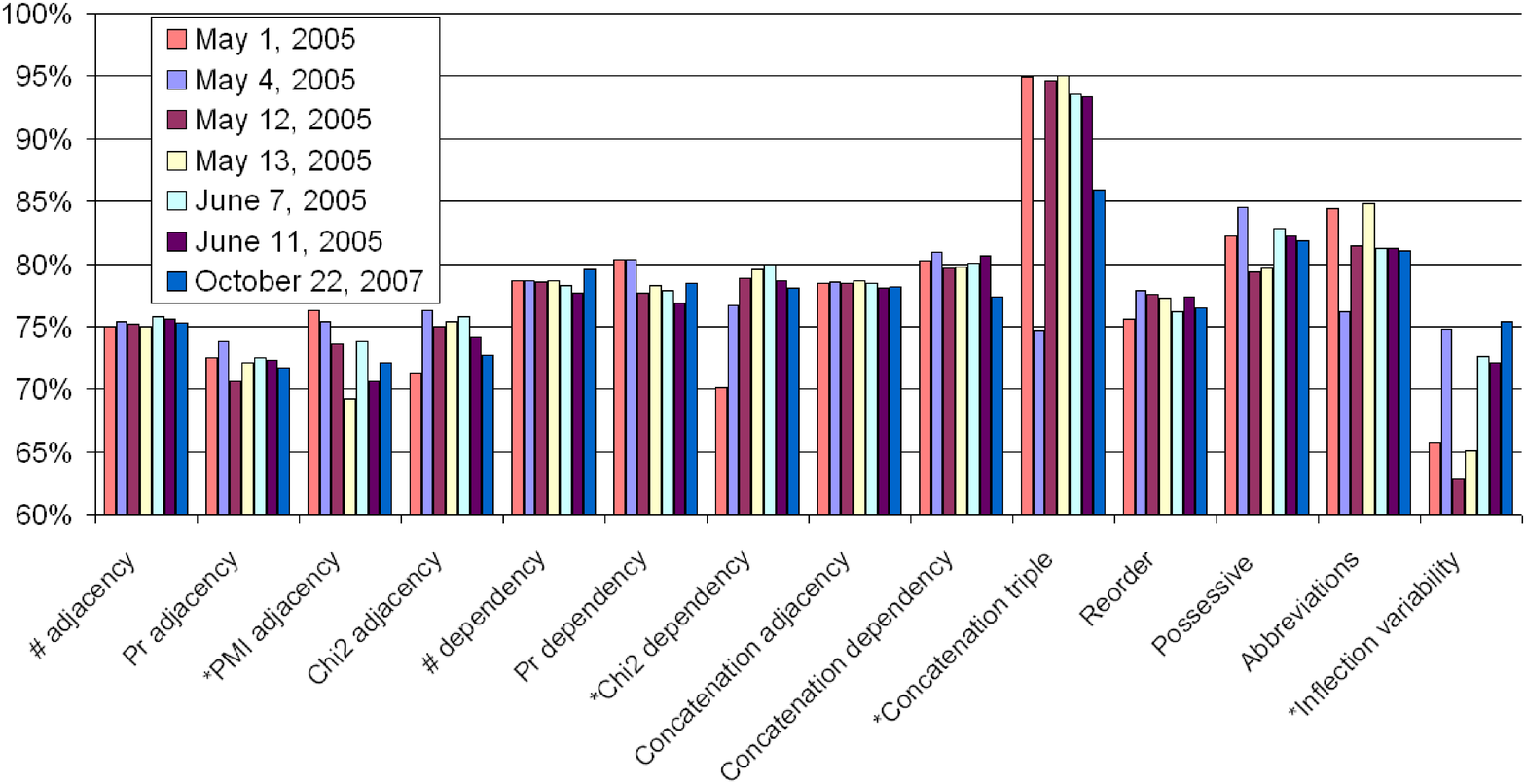}\\
  \caption{{\bf Accuracy over time for \emph{MSN}:} pages in any language, with word inflections.
            The models with statistically significant variations are marked with an asterisk.}
  \label{figure:byTimeMSN:withinfl:anylang}
\end{center}
\end{figure}

\begin{figure}
\begin{center}
  \includegraphics[width=15cm,height=8.5cm]{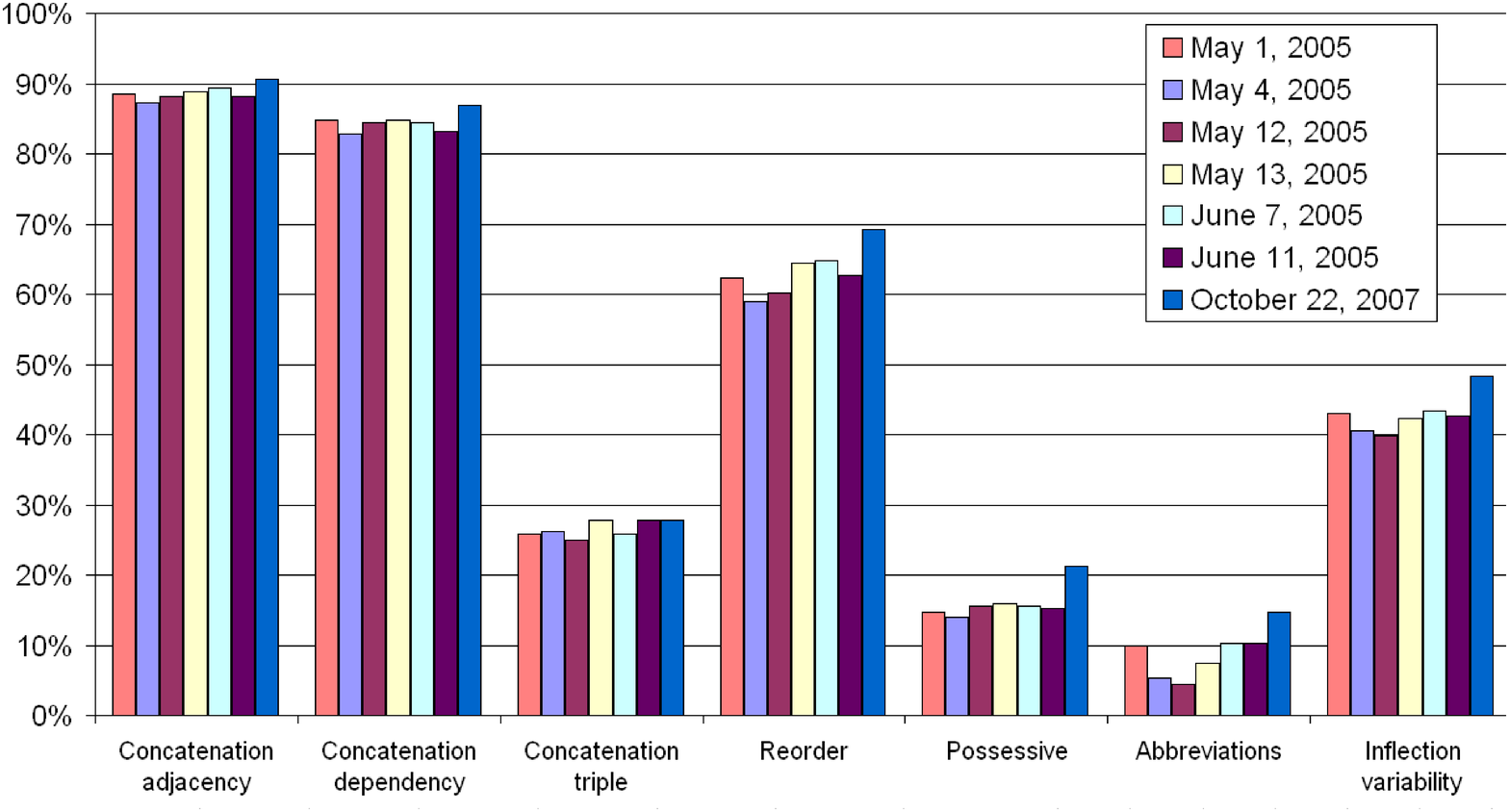}\\
  \caption{{\bf Coverage over time for \emph{MSN}:} pages in any language, no word inflections.}
  \label{figure:byTimeMSN:noinfl:anylang:R}
\end{center}
\end{figure}

\begin{figure}
\begin{center}
  \includegraphics[width=15cm,height=8.5cm]{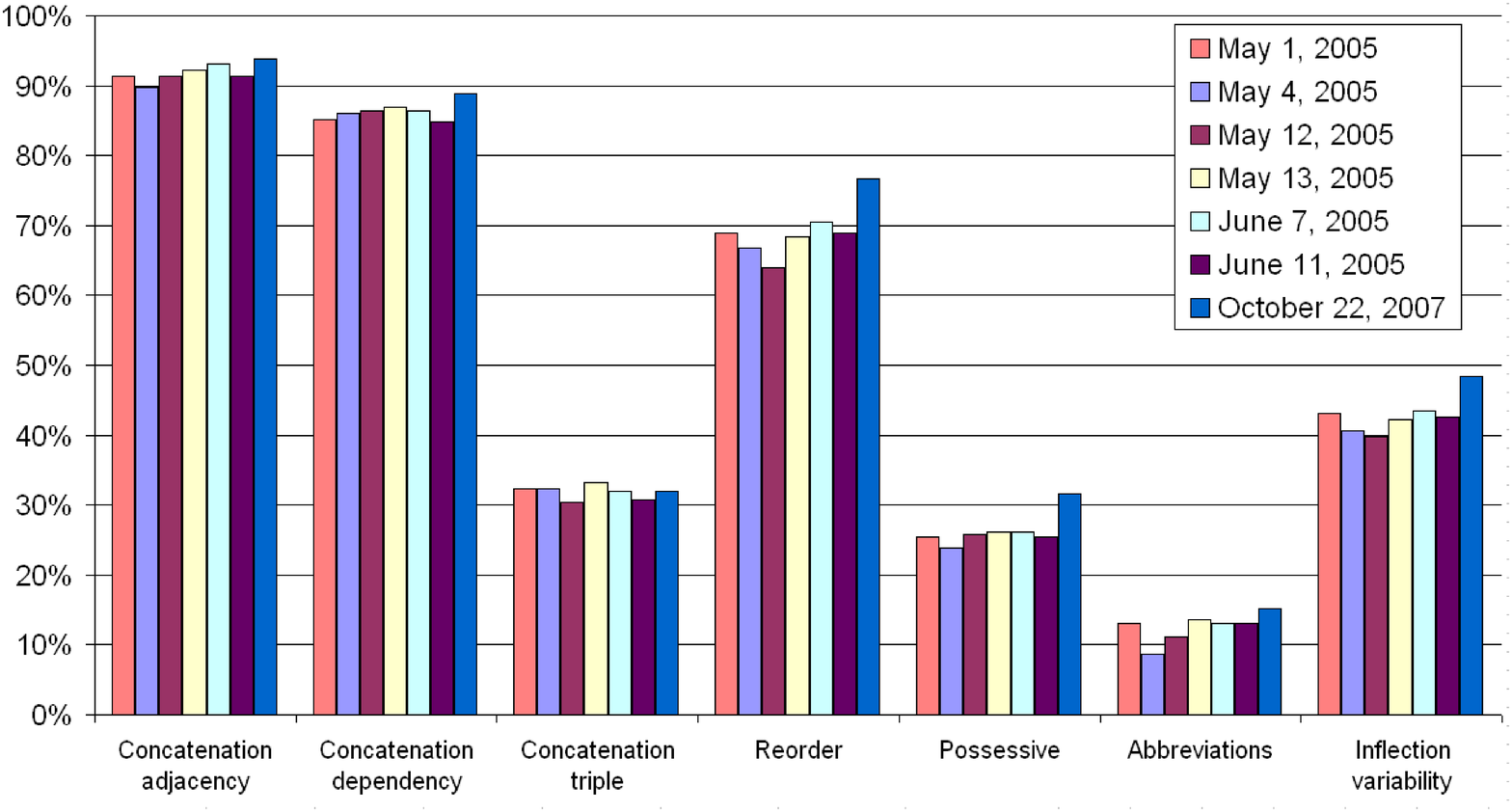}\\
  \caption{{\bf Coverage over time for \emph{MSN}:} pages in any language, with word inflections.}
  \label{figure:byTimeMSN:withinfl:anylang:R}
\end{center}
\end{figure}

\begin{figure}
\begin{center}
  \includegraphics[width=15cm,height=8.5cm]{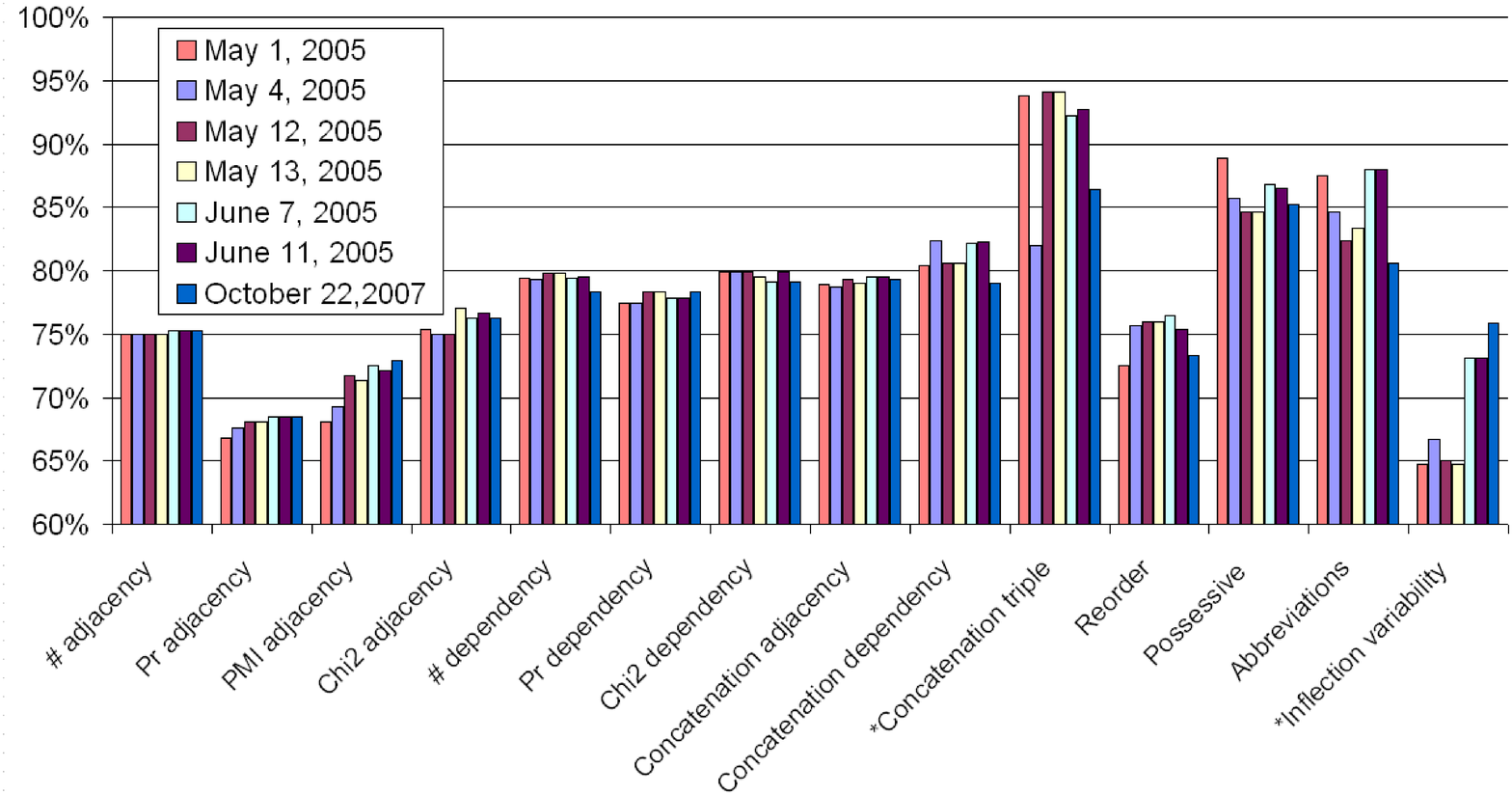}\\
  \caption{{\bf Accuracy over time for \emph{MSN}:} English pages only, no word inflections.
            The models with statistically significant variations are marked with an asterisk.}
  \label{figure:byTimeMSN:noinfl:english}
\end{center}
\end{figure}

\begin{figure}
\begin{center}
  \includegraphics[width=15cm,height=8.5cm]{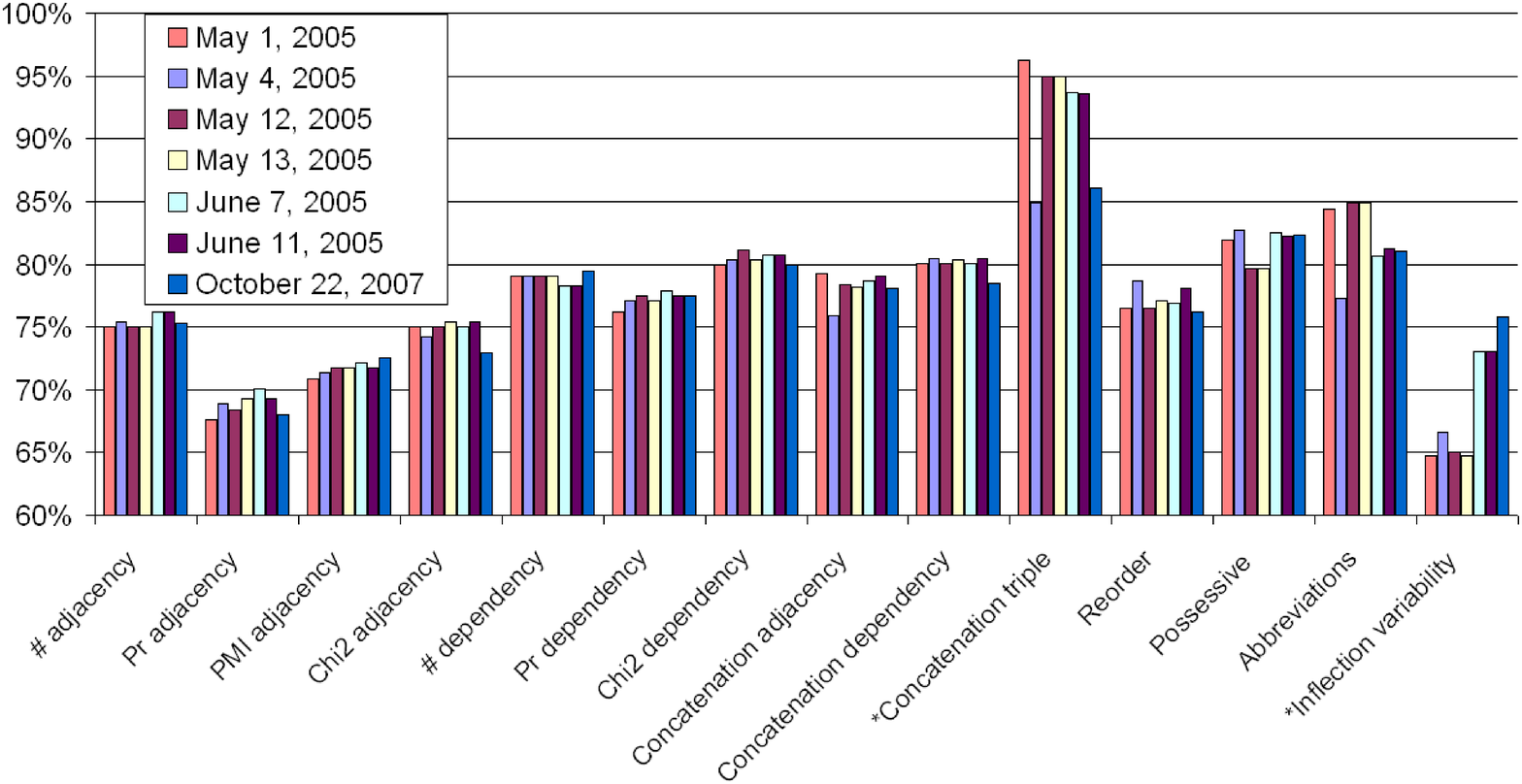}\\
  \caption{{\bf Accuracy over time for \emph{MSN}:} English pages only, with word inflections.
            The models with statistically significant variations are marked with an asterisk.}
  \label{figure:byTimeMSN:withinfl:english}
\end{center}
\end{figure}

\begin{figure}
\begin{center}
  \includegraphics[width=15cm,height=8.5cm]{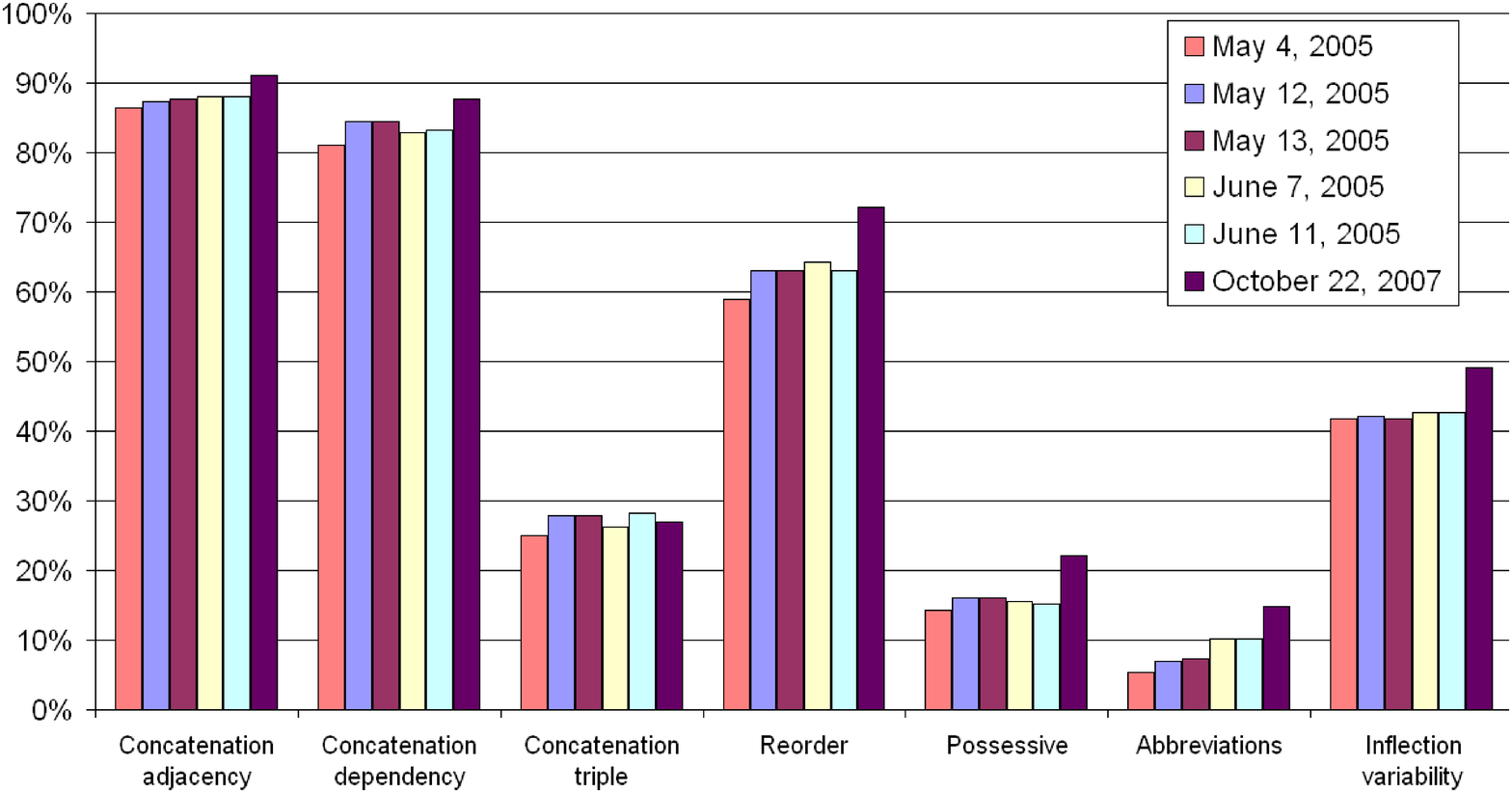}\\
  \caption{{\bf Coverage over time for \emph{MSN}:} English pages only, no word inflections.}
  \label{figure:byTimeMSN:noinfl:english:R}
\end{center}
\end{figure}

\begin{figure}
\begin{center}
  \includegraphics[width=15cm,height=8.5cm]{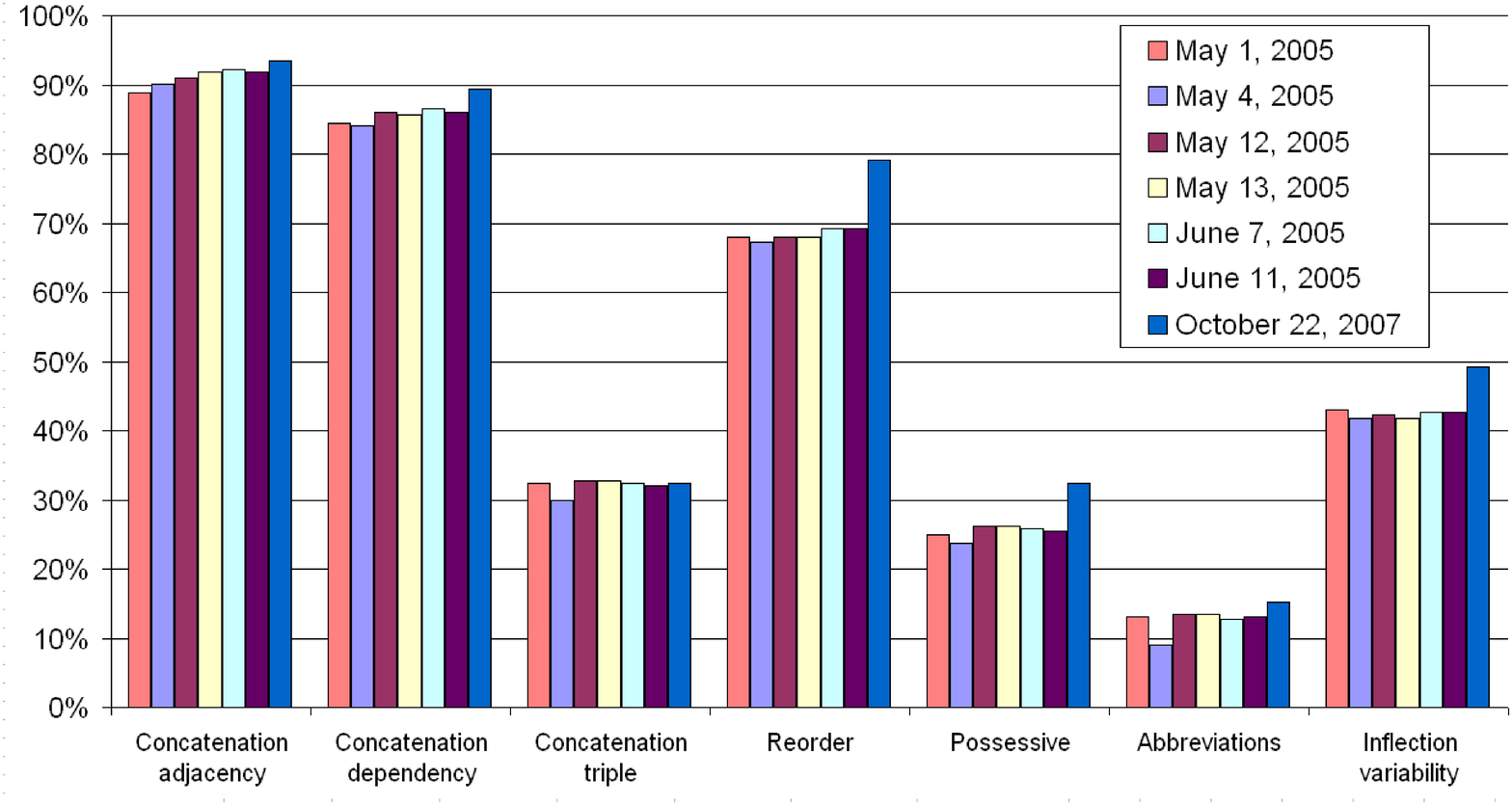}\\
  \caption{{\bf Coverage over time for \emph{MSN}:} English pages only, with word inflections.}
  \label{figure:byTimeMSN:withinfl:english:R}
\end{center}
\end{figure}

For {\it Google}, I collected $n$-gram statistics for October 22, 2007,
and for four different dates in 2005: April 30, May 4, June 7, and June 11.
The results for the accuracy are shown in
Figure \ref{figure:byTimeGoogle:noinfl:anylang} (no language filter, no word inflections),
Figure \ref{figure:byTimeGoogle:withinfl:anylang} (no language filter, using word inflections),
Figure \ref{figure:byTimeGoogle:noinfl:english} (English pages only, no word inflections)
and Figure \ref{figure:byTimeGoogle:withinfl:english} (English pages only, using word inflections).
The variability in accuracy is low for the first nine models (about 1-4\%),
and more sizable for the last five ones.
{\it Concatenation triple} is marked with an asterisk,
which indicates statistically significant differences for it (see section \ref{sec:stat:significance}).
In Figure \ref{figure:byTimeGoogle:withinfl:english},
there is a statistically significant difference
for {\it PMI adjacency} and for {\it possessive} as well.
While in the first two time snapshots the accuracy for
the {\it possessive} is much lower than in the last two, it is the
reverse for {\it concatenation triple};
{\it Google} may have changed how it handles possessive markers
and word concatenations in the interrum.

The coverage (\% of examples for which a model makes predictions) for the last seven models is shown in
Figure \ref{figure:byTimeGoogle:noinfl:anylang:R} (no language filter, no word inflections),
Figure \ref{figure:byTimeGoogle:withinfl:anylang:R} (no language filter, using word inflections),
Figure \ref{figure:byTimeGoogle:noinfl:english:R} (English pages only, no word inflections)
and Figure \ref{figure:byTimeGoogle:withinfl:english:R} (English pages only, using word inflections).
The maximum difference in coverage for these models is rather small and does not exceed 7\%.
The coverage for the remaining seven models is not shown since it is 100\%
(or very close to 100\%) for all models and under all observed conditions.

For {\it MSN Search}, I collected $n$-gram statistics for October 22, 2007
and for six dates in 2005: May 1, May 4, May 12, May 13, June 7, and June 11.
The results for the accuracy are shown in
Figure \ref{figure:byTimeMSN:noinfl:anylang} (no language filter, no word inflections),
Figure \ref{figure:byTimeMSN:withinfl:anylang} (no language filter, using word inflections),
Figure \ref{figure:byTimeMSN:noinfl:english} (English pages only, no word inflections)
and Figure \ref{figure:byTimeMSN:withinfl:english} (English pages only, using word inflections).
The corresponding results for the coverage are shown in Figures
\ref{figure:byTimeMSN:noinfl:anylang:R}, \ref{figure:byTimeMSN:withinfl:anylang:R},
\ref{figure:byTimeMSN:noinfl:english:R}, \ref{figure:byTimeMSN:withinfl:english:R}.
As these figures show, {\it MSN Search} exhibits much higher variability
in accuracy compared to {\it Google}.
There is almost 10\% difference for the adjacency- and dependency-based models:
for example, for {\it $\chi^2$ dependency}, Figure \ref{figure:byTimeMSN:withinfl:anylang} shows
an accuracy of 70.08\% for May 1, 2005 and 79.92\% for June 7, 2005.
This difference is statistically significant (and so is the difference for
{\it PMI adjacency}: 76.23\% on May 1, 2005 vs. 69.26\% on May 13, 2005).
Even bigger differences can be observed for the last seven models, e.g.,
Figure \ref{figure:byTimeMSN:noinfl:anylang} shows about 14\%
absolute variation for {\it inflection variability} (75.42\% on October 22, 2005 vs. 62.89\% on May 12, 2007),
and over 20\% for the {\it concatenation triple} (74.68\% on May 4, 2005 vs. 95.06\% on May 13, 2005);
both differences are statistically significant.
The higher variability in accuracy for {\it MSN Search} is probably due to differences in how rounding is computed:
since {\it Google}'s page hits are rounded, they change less over time.
By contrast, these counts are exact for {\it MSN Search},
which makes them more sensitive to variation.
This hypothesis is supported by the fact that overall
both engines exhibit a higher variability for the last seven models,
which use smaller counts that are more likely to be represented by exact
numbers in {\it Google} as well.

Finally, the variability in coverage for {\it MSN Search}
is slightly bigger than for {\it MSN Search} and goes up to 11\%
(as opposed to up to 7\% for {\it Google}).

\subsection{Variability by Search Engine}
\label{sec:byu:search:engine}

\begin{figure}
\begin{center}
  \includegraphics[width=15cm,height=8.5cm]{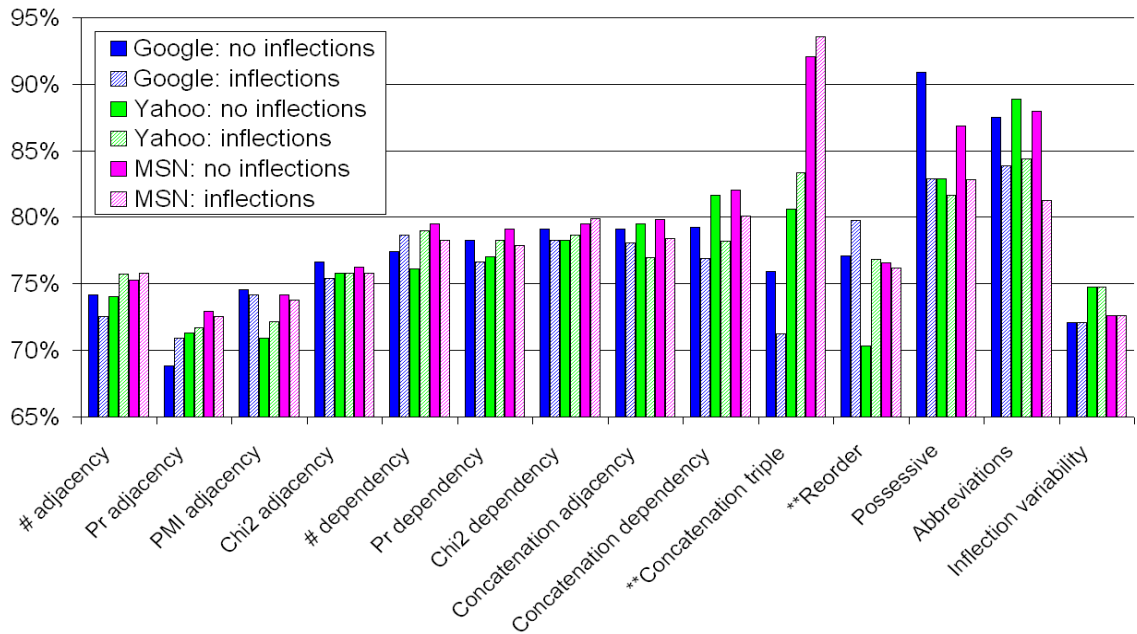}\\
  \caption{{\bf Accuracy by search engine and inflection, June 6, 2005:} any language.
            Significant differences between different engines are marked with a double asterisk.}
  \label{figure:byEngineAndInfl:anylang}
\end{center}
\end{figure}

\begin{figure}
\begin{center}
  \includegraphics[width=15cm,height=8.5cm]{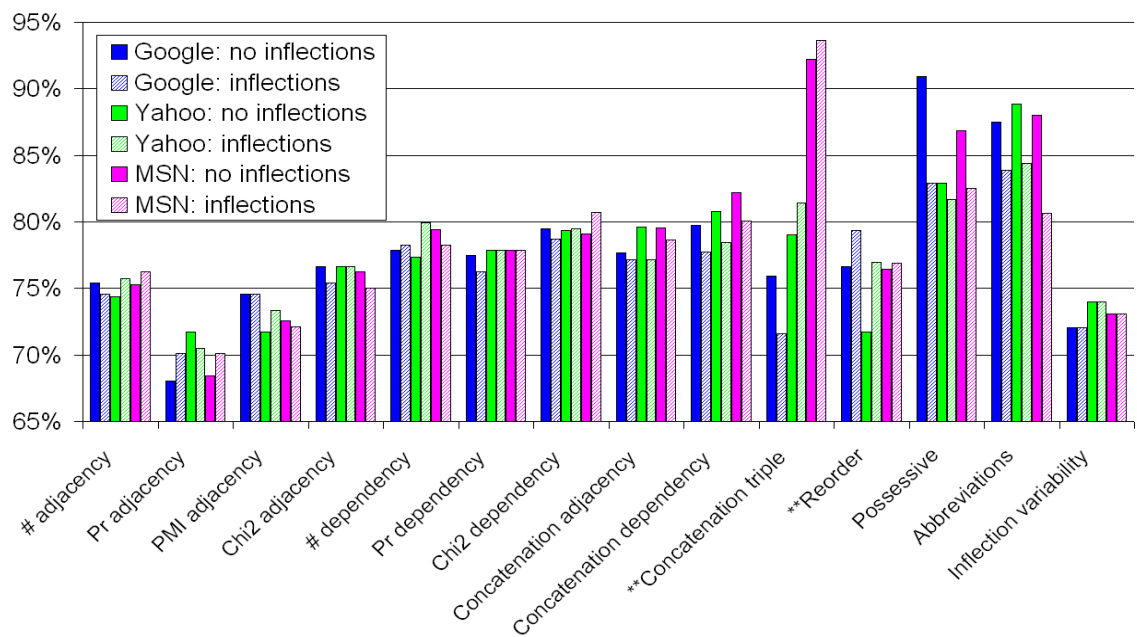}\\
  \caption{{\bf Accuracy by search engine and inflection, June 6, 2005:} English only.
            Significant differences between different engines are marked with a double asterisk.}
  \label{figure:byEngineAndInfl:english}
\end{center}
\end{figure}

\begin{figure}
\begin{center}
  \includegraphics[width=15cm,height=8.5cm]{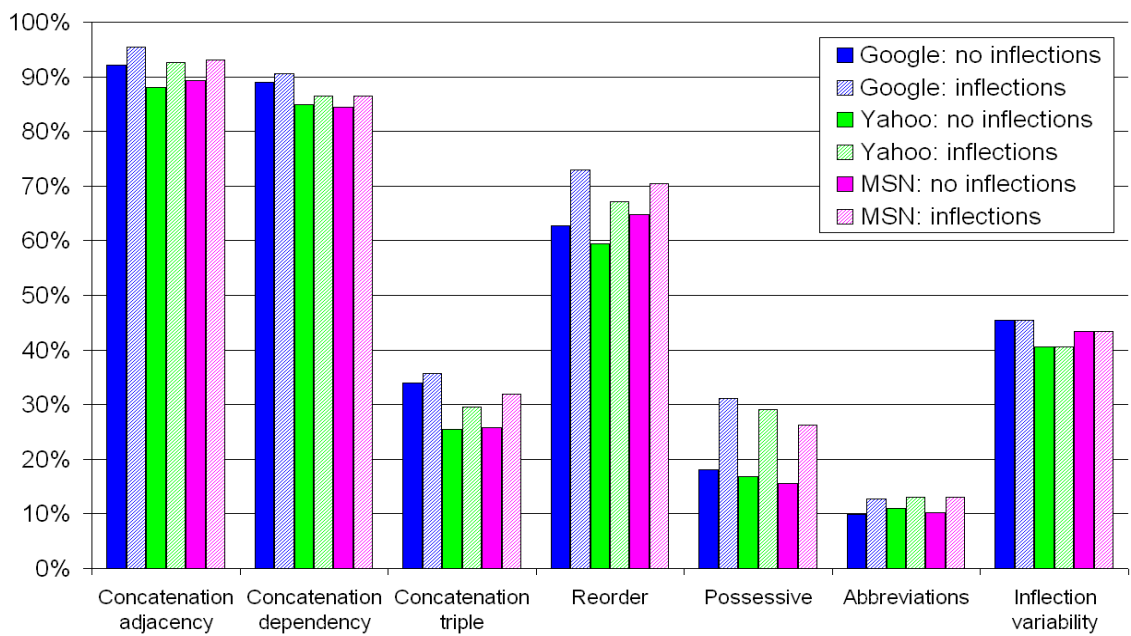}\\
  \caption{{\bf Coverage by search engine and inflection, June 6, 2005:} any language.}
  \label{figure:byEngineAndInfl:anylang:R}
\end{center}
\end{figure}

\begin{figure}
\begin{center}
  \includegraphics[width=15cm,height=8.5cm]{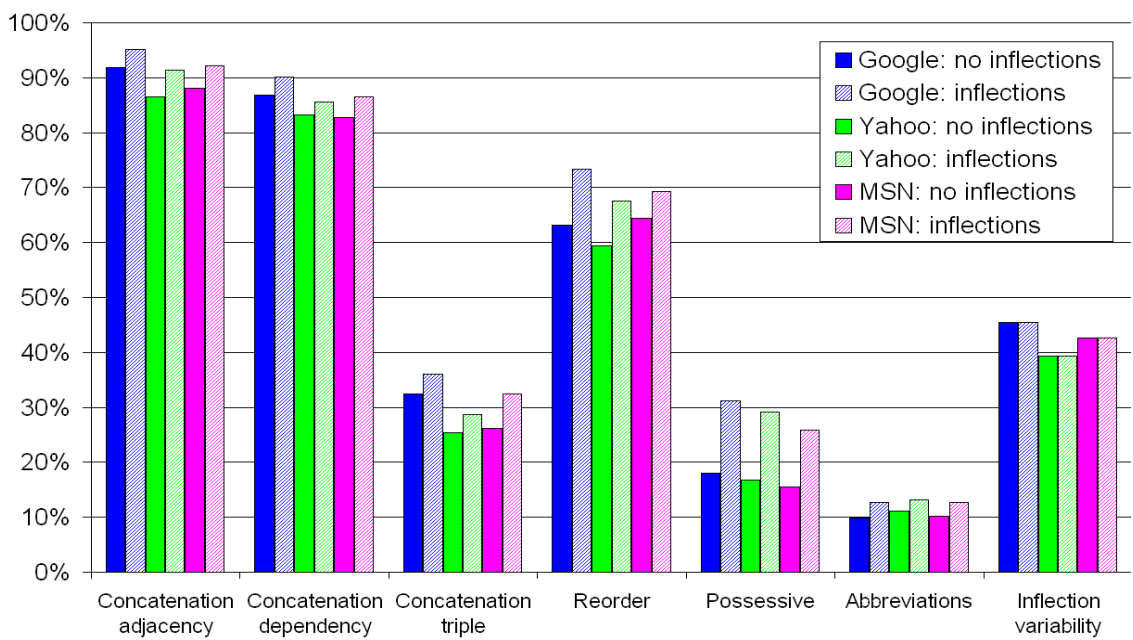}\\
  \caption{{\bf Coverage by search engine and inflection, June 6, 2005:} English only.}
  \label{figure:byEngineAndInfl:english:R}
\end{center}
\end{figure}

I study the variability by search engine
by comparing {\it Google}, {\it MSN Search} and {\it Yahoo!}
on the same day, June 6, 2005, in order to minimize
the impact of index changes over time.
I show the variability in accuracy by search engine and word inflection in
Figure \ref{figure:byEngineAndInfl:anylang} (without language filters) and
Figure \ref{figure:byEngineAndInfl:english} (English pages only).
The corresponding results for the coverage are shown in
Figures \ref{figure:byEngineAndInfl:anylang:R} and \ref{figure:byEngineAndInfl:english:R}.
I also show the variability in accuracy by search engine and language in
Figure \ref{figure:byEngineAndLang:noinfl} (without word inflections) and
Figure \ref{figure:byEngineAndLang:withinfl} (using word inflections);
the coverage is shown in
Figures \ref{figure:byEngineAndLang:noinfl:R} and \ref{figure:byEngineAndLang:withinfl:R}.

As Figure \ref{figure:byEngineAndLang:withinfl} shows,
the biggest difference in accuracy is exhibited
by {\it concatenation triple}: using inflected queries and English pages only,
{\it MSN Search} achieves an accuracy of 93.67\%,
while {\it Google} under the same conditions only achieves 71.59\%,
which is a statistically significant difference.
The results without the language filter are similar: 93.59\% vs. 71.26\%.
The differences between {\it MSN Search} and {\it Google}
are also statistically significant without the language filter.
The differences for {\it reorder} between {\it Google} and {\it Yahoo!}
are statistically significant as well.
Other large variations (not statistically significant) can be seen
for {\it possessive}.
Overall, {\it MSN Search} performs best, which I attribute to it not rounding
its page hit estimates. However, as Figure \ref{figure:byEngineAndLang:noinfl} shows,
{\it Google} is almost 5\% ahead on {\it possessive} for non-inflected queries,
while {\it Yahoo!} leads on {\it abbreviations} and {\it inflection variability}.
The fact that different search engines exhibit strength on different
kinds of queries and models suggests that combining them could offer potential benefits.
The results for the coverage in
Figures \ref{figure:byEngineAndLang:noinfl:R} and \ref{figure:byEngineAndLang:withinfl:R}
are slightly better for {\it Google}, which
suggests it might have had a bigger index on June 6, 2005,
compared to {\it Yahoo!} and {\it MSN Search}.

\subsection{Impact of Language Filtering}

The impact of language filtering, meaning requiring only documents
in English as opposed to having no restrictions on the language,
can be observed on several graphs.

\begin{figure}
\begin{center}
  \includegraphics[width=15cm,height=8cm]{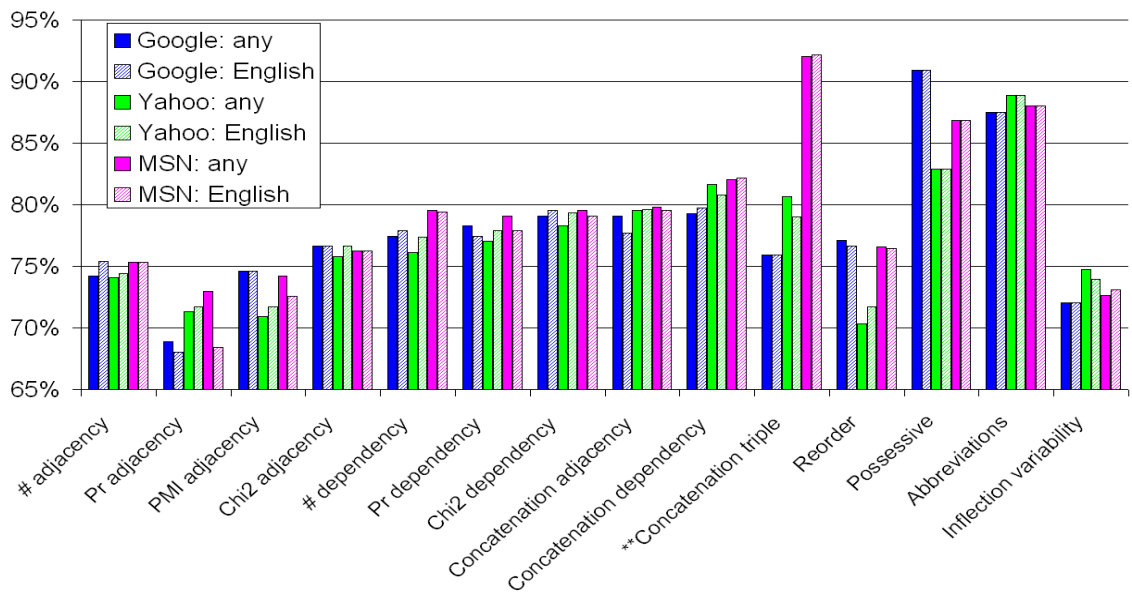}\\
  \caption{{\bf Accuracy by search engine and language filtering, June 6, 2005:} no inflections.
            Significant differences between different engines are marked with a double asterisk.}
  \label{figure:byEngineAndLang:noinfl}
\end{center}
\end{figure}

\begin{figure}
\begin{center}
  \includegraphics[width=15cm,height=8cm]{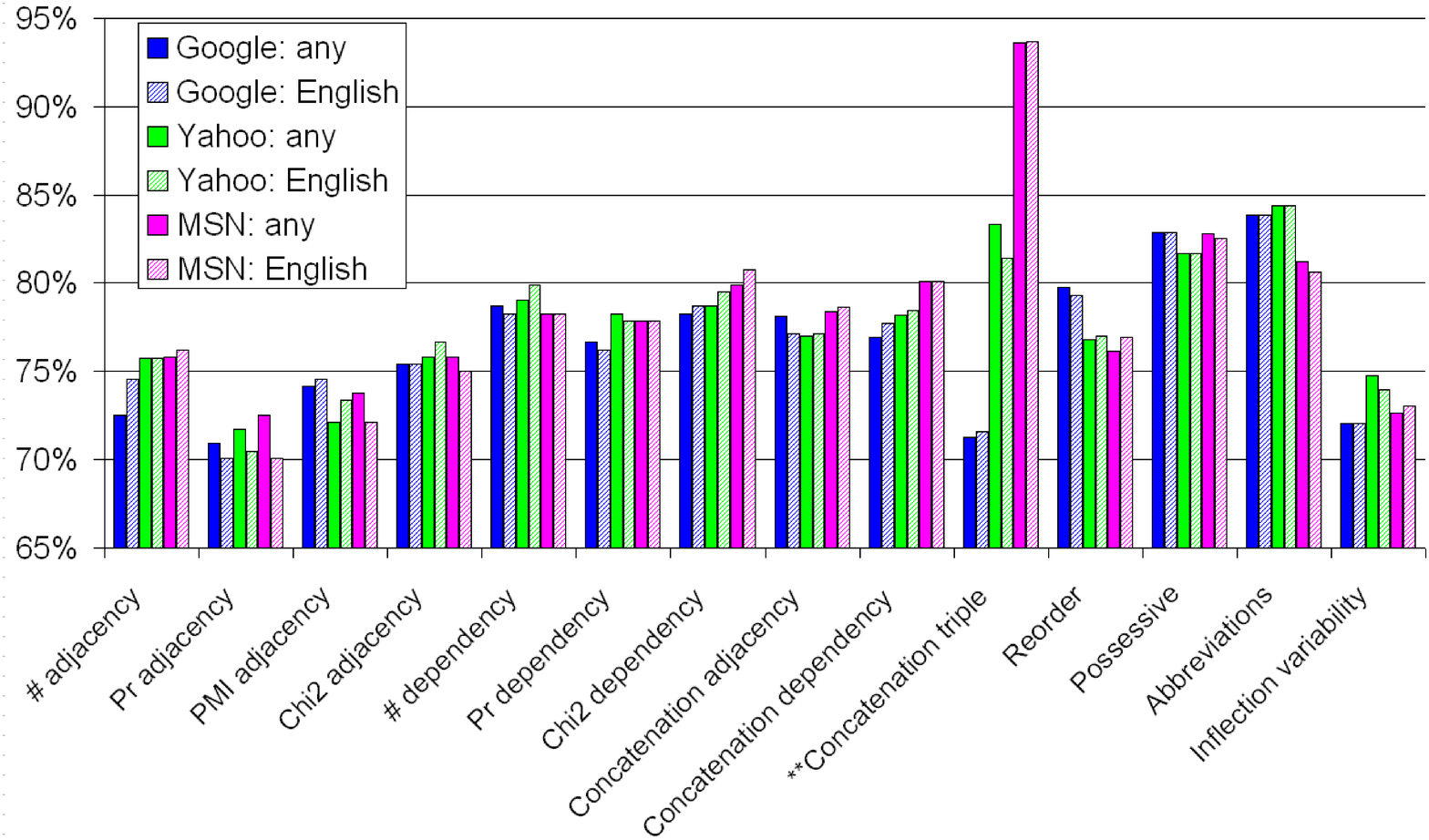}\\
  \caption{{\bf Accuracy by search engine and language filtering, June 6, 2005:} with inflections.
            Significant differences between different engines are marked with a double asterisk.}
  \label{figure:byEngineAndLang:withinfl}
\end{center}
\end{figure}

\begin{figure}
\begin{center}
  \includegraphics[width=15cm,height=8.5cm]{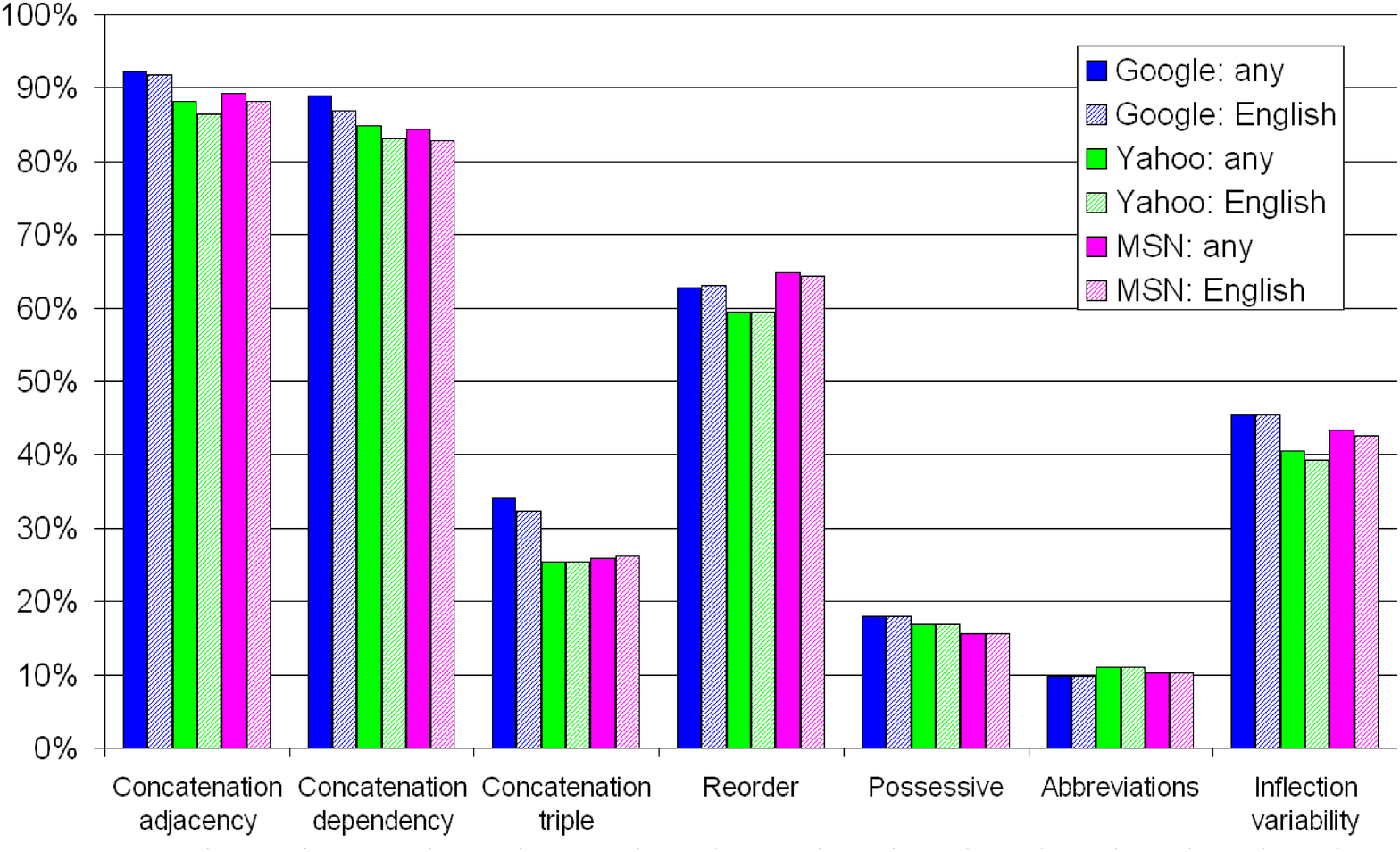}\\
  \caption{{\bf Coverage by search engine and language filtering, June 6, 2005:} no inflections.}
  \label{figure:byEngineAndLang:noinfl:R}
\end{center}
\end{figure}

\begin{figure}
\begin{center}
  \includegraphics[width=15cm,height=8.5cm]{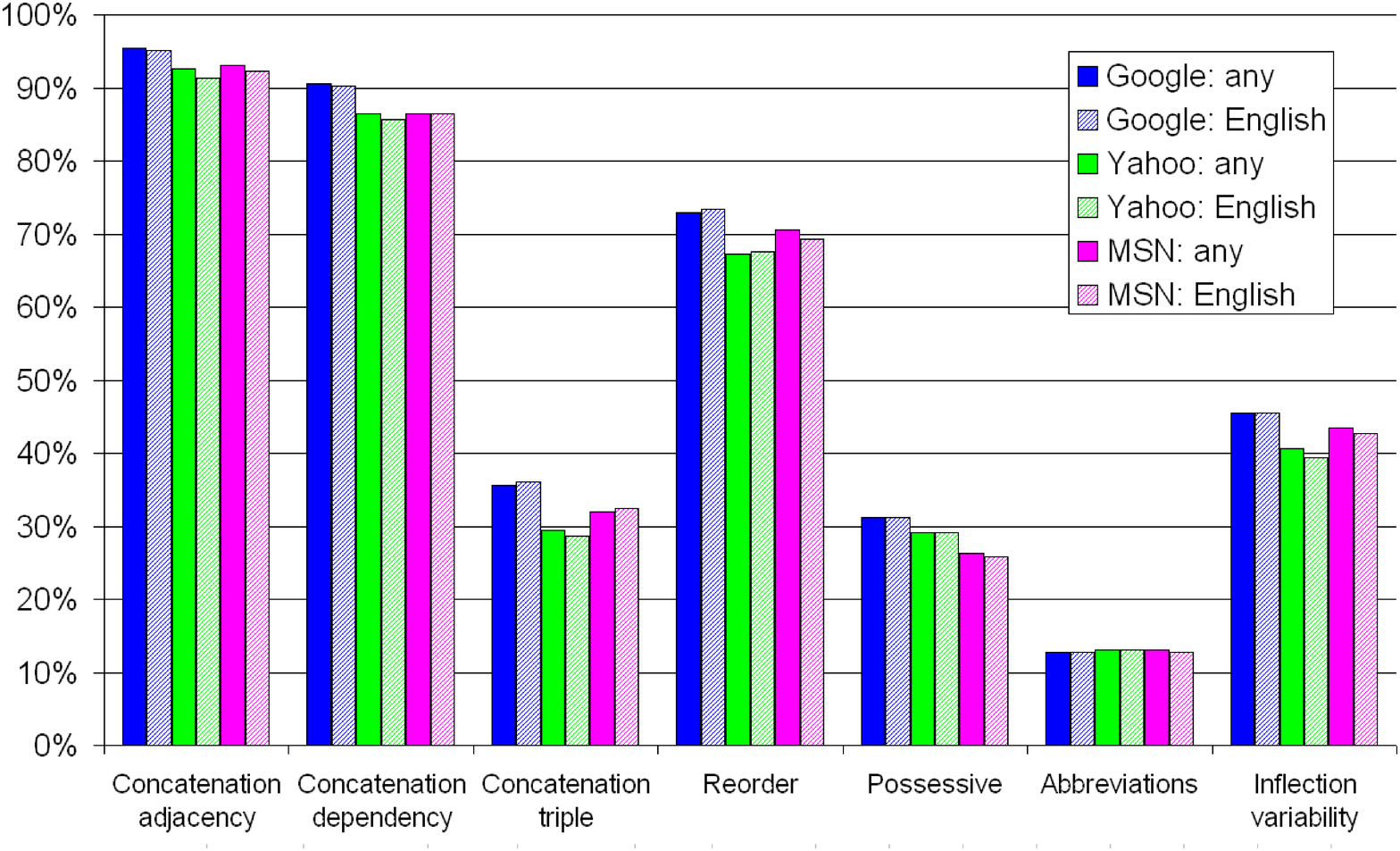}\\
  \caption{{\bf Coverage by search engine and language filtering, June 6, 2005:} with inflections.}
  \label{figure:byEngineAndLang:withinfl:R}
\end{center}
\end{figure}

First, the variation in accuracy can be observed across search engines for June 6, 2005, as shown
in Figure \ref{figure:byEngineAndLang:noinfl} (without word inflections)
and Figure \ref{figure:byEngineAndLang:withinfl} (using word inflections).
The impact of language filtering on accuracy is minor and inconsistent
for all three search engines: sometimes the results are improved
slightly and sometimes they are negatively impacted,
with the differences varying up to 3\% (usually 0-1\%).
Figures \ref{figure:byEngineAndLang:noinfl:R} and \ref{figure:byEngineAndLang:withinfl:R},
compare the impact of language filtering on coverage for the last seven models.
As we can see, using English only leads to a drop in coverage, as one could
expect, but this drop is quite small: less than 1\%.

Second, the impact of language filtering can also be observed across time
for {\it Google} and for {\it MSN Search}.
For {\it Google},
we can study the impact without inflections by comparing
Figure \ref{figure:byTimeGoogle:noinfl:anylang} (no language filter)
and
Figure \ref{figure:byTimeGoogle:noinfl:english} (English pages only).
We can also study the impact with inflections by comparing
Figure \ref{figure:byTimeGoogle:withinfl:anylang} (no language filter)
and
Figure \ref{figure:byTimeGoogle:withinfl:english} (English pages only).
Similarly,
for {\it MSN Search}, the impact without inflections would compare
Figure \ref{figure:byTimeMSN:noinfl:anylang} (no language filter)
and
Figure \ref{figure:byTimeMSN:noinfl:english} (English pages only).
We can also study the impact with inflections by comparing
Figure \ref{figure:byTimeMSN:withinfl:anylang} (no language filter)
and
Figure \ref{figure:byTimeMSN:withinfl:english} (English pages only).
In all cases, the differences in accuracy are very small: less than 3\%.
The corresponding differences in coverage are within 0-1\%.

\subsection{Impact of Word Inflections}

Finally, I study the impact of using different word inflections and adding up the frequencies.
I generate the inflections (e.g., {\it tumor} and {\it tumors})
using {\it Carroll's morphological tools} \cite{Minnen:Carroll:Pearce:2001}
and {\it WordNet} \cite{Fellbaum:1998:wordnet}, as described in section \ref{sec:bracketing:eval}.
See sections \ref{sec:assoc:scores} and \ref{sec:bracketing:web:derived:features}
for a detailed description of what kinds of inflections are used for each model
and how the page hits are summed up.

The impact of word inflections can be studied by looking at several graphs.
First, the variation in accuracy can be observed across search engines for June 6, 2005, as shown
in Figure \ref{figure:byEngineAndInfl:anylang} (no language filter)
and Figure \ref{figure:byEngineAndInfl:english} (English only).
While the results are mixed, the impact on accuracy here is bigger
compared to language filtering (and, of course, there is no impact on {\it inflection variability}).
Interestingly, for all search engines and regardless of whether language filtering has been used,
inflected queries positively impact the accuracy
for {\it possessive} (by up to 8\%), {\it abbreviations} (by up to 8\%),
{\it concatenation adjacency} (by up to 3-4\%) and {\it concatenation dependency} (by up to 4\%).
Using word inflections for {\it concatenation triple}
improves accuracy by 2-4\% for {\it Yahoo!} and {\it MSN Search},
but negatively affects {\it Google} (by about 5\%).
Overall, word inflections improve accuracy for {\it reorder}
by about 7\% for {\it Yahoo!}, and by about 3\% for {\it Google},
but have an insignificant impact for {\it MSN Search}.
The impact of using word inflections on coverage can be seen in
Figure \ref{figure:byEngineAndInfl:anylang:R} (no language filter)
and Figure \ref{figure:byEngineAndInfl:english:R} (English only).
As one would expect, the coverage improves consistently (by up to 7\%),
especially for {\it possessive}, {\it reorder} and {\it concatenation triple}.
However, none of these differences in accuracy or coverage
is statistically significant.\footnote{Figures
\ref{figure:byEngineAndInfl:anylang} and \ref{figure:byEngineAndInfl:english}
only contain double asterisks,
which indicate statistically significant difference between search engines,
not between inflected and non-inflected queries.}

Second, the impact of using word inflections can be observed across time
for {\it Google} and for {\it MSN Search}.
For {\it Google},
the impact of word inflections, in case of no language filtering, can be seen by comparing
Figure \ref{figure:byTimeGoogle:noinfl:anylang} (without word inflections)
and
Figure \ref{figure:byTimeGoogle:withinfl:anylang} (using word inflections),
while in case of filtering out the non-English pages we can compare
Figure \ref{figure:byTimeGoogle:noinfl:english} (without word inflections)
and
Figure \ref{figure:byTimeGoogle:withinfl:english} (using word inflections).
Similarly,
for {\it MSN Search}, the impact of word inflections,
without language filtering, can be seen by comparing
Figure \ref{figure:byTimeMSN:noinfl:anylang} (without word inflections)
and
Figure \ref{figure:byTimeMSN:withinfl:anylang} (using word inflections),
while in case of filtering out the non-English pages we can compare
Figure \ref{figure:byTimeMSN:noinfl:english} (without word inflections)
and
Figure \ref{figure:byTimeMSN:withinfl:english} (using word inflections).
In either case, the differences in accuracy are up to 8\%,
and the corresponding differences in coverage are up to 7\%;
none of these variations is statistically significant.

\subsection{Variability by Search Engine: 28 Months Later}

I have studied the interesting question of how
the variability by search engine changes over a long period of time
for different search engines. For the purpose,
I repeated the experiments described in section \ref{sec:byu:search:engine}
comparing the performance of {\it Google}, {\it Yahoo!} and {\it MSN}
on June 6, 2005 and on October 22, 2007.

Many things have changed over this period of time:
the size of search engines' indexes has increased
query analysis has been updated (e.g., {\it Google} seems
to handle automatically inflectional variation for English,
which has an impact on inflected queries), and
page hit estimation formulas employed by search engines have been altered.
The most important change occurred to {\it MSN Search},
now {\it Live Search}, which started rounding its page hit estimates,
just like {\it Google} and {\it Yahoo!} do,
rather than providing exact estimates as in 2005.

\begin{figure}
\begin{center}
  \includegraphics[width=15cm,height=8cm]{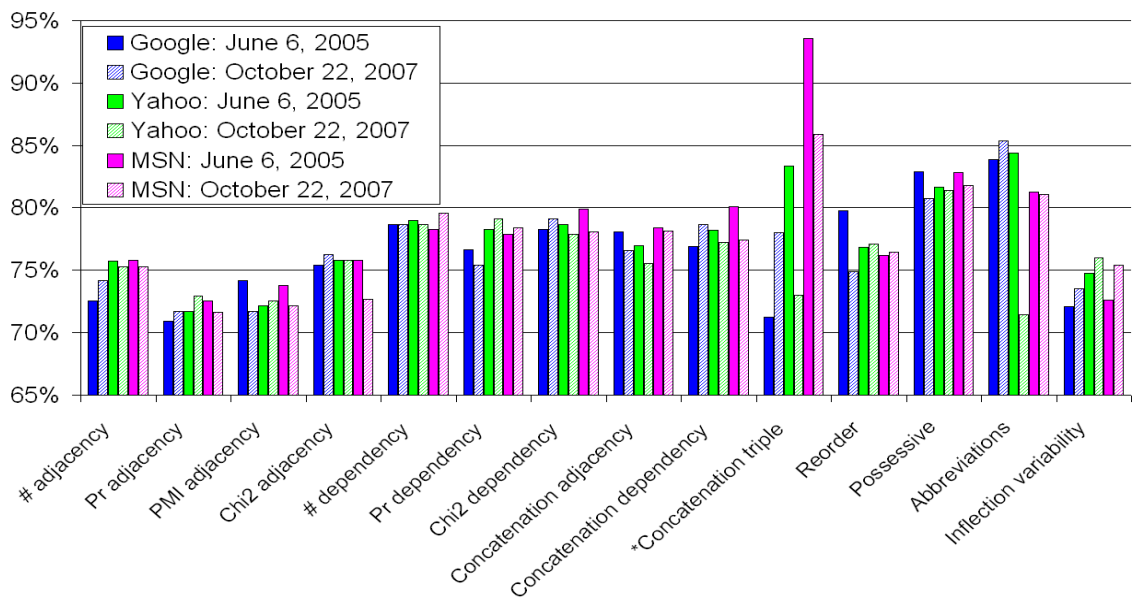}\\
  \caption{{\bf Accuracy by search engine for June 6, 2005 and October 22, 2007:} using inflections, any language.
            Statistically significant differences between the same search engine are marked with an asterisk.}
  \label{figure:byEngineTwoDates:withinfl:anylang}
\end{center}
\end{figure}

\begin{figure}
\begin{center}
  \includegraphics[width=15cm,height=8cm]{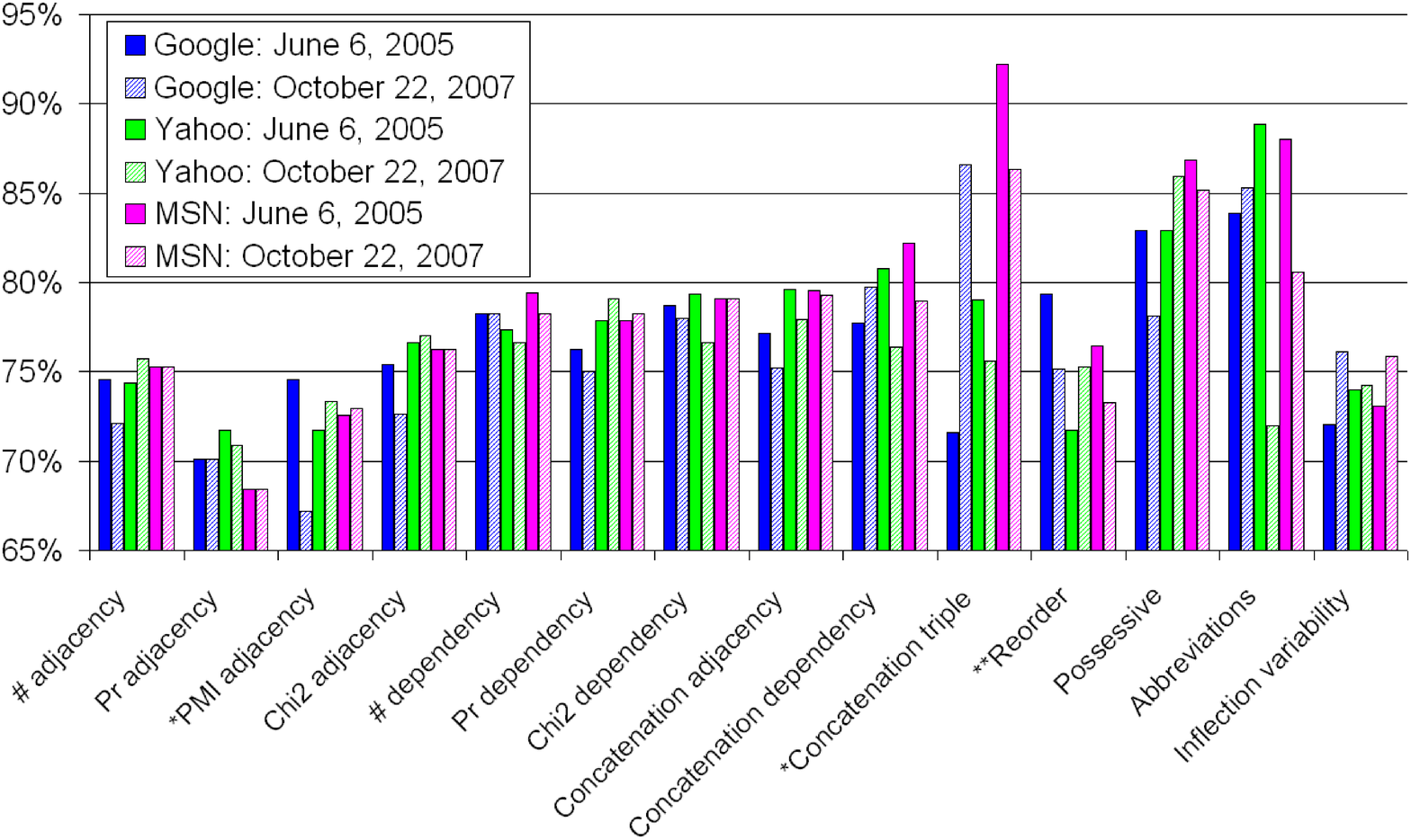}\\
  \caption{{\bf Accuracy by search engine for June 6, 2005 and October 22, 2007:} no inflections, English pages only.
            Statistically significant differences between the same/different search engines is marked with a single/double asterisk.}
  \label{figure:byEngineTwoDates:noinfl:english}
\end{center}
\end{figure}

\begin{figure}
\begin{center}
  \includegraphics[width=15cm,height=8cm]{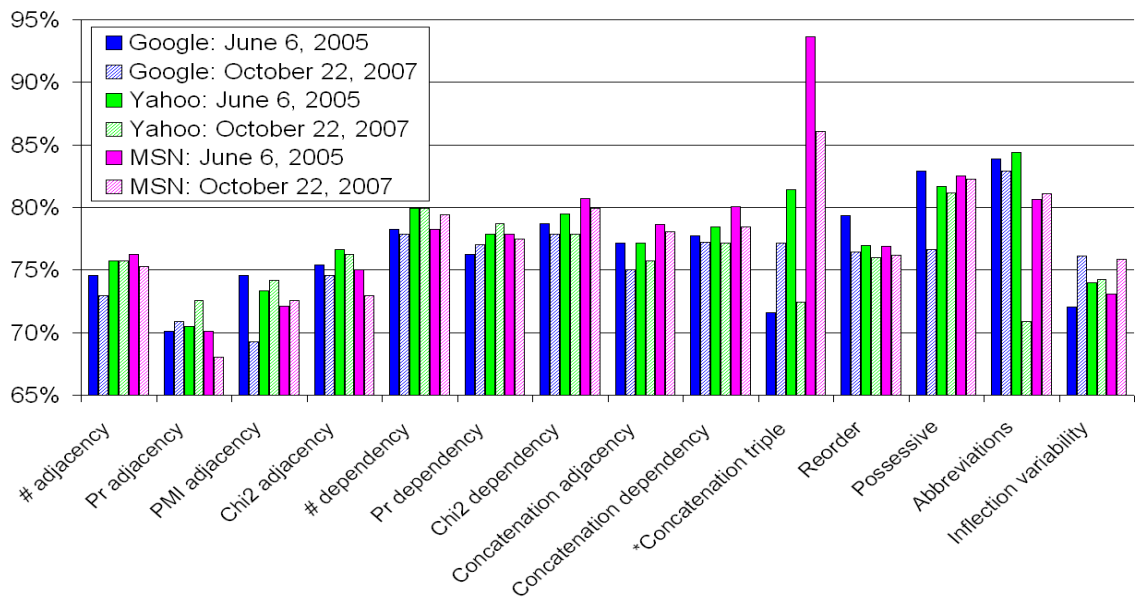}\\
  \caption{{\bf Accuracy by search engine for June 6, 2005 and October 22, 2007:} using inflections,
            English pages only.
            Statistically significant differences between the same search engine are marked with an asterisk.}
  \label{figure:byEngineTwoDates:withinfl:english}
\end{center}
\end{figure}

The evaluation results are shown on the following figures:
Figure \ref{figure:byEngineTwoDates:noinfl:anylang} (no language filter, no word inflections),
Figure \ref{figure:byEngineTwoDates:withinfl:anylang} (no language filter, using word inflections),
Figure \ref{figure:byEngineTwoDates:noinfl:english} (English pages only, no word inflections)
and Figure \ref{figure:byEngineTwoDates:withinfl:english} (English pages only, using word inflections).
We can see that, while the differences in performance are inconsistent,
the accuracies for {\it Live Search} on October 22, 2007
are generally worse compared to those for {\it MSN Search} on June 6, 2005,
especially for the last six models,
which I attribute to {\it Live Search} rounding its page hit estimates.

\begin{figure}
\begin{center}
  \includegraphics[width=15cm,height=8.5cm]{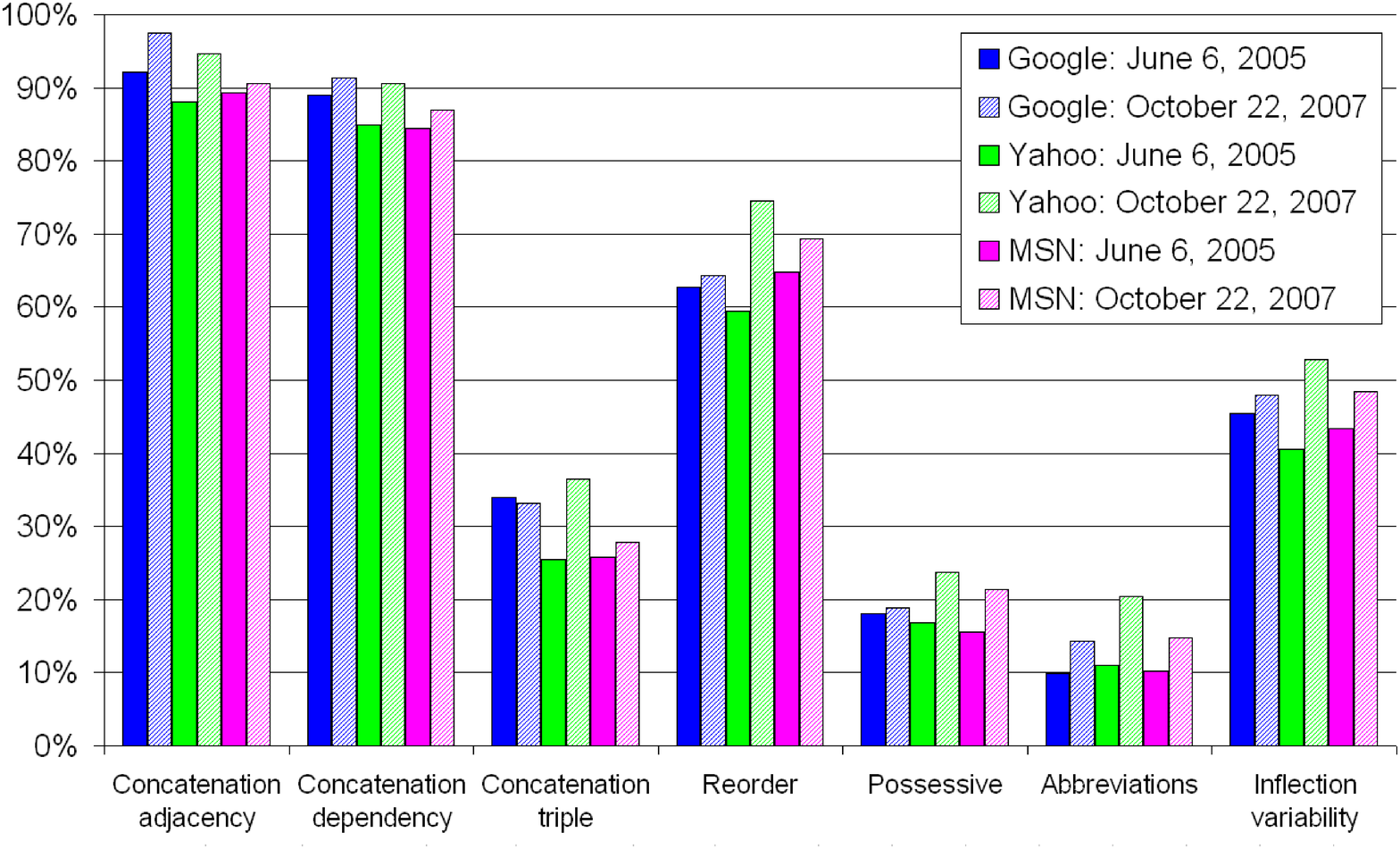}\\
  \caption{{\bf Coverage by search engine for June 6, 2005 and October 22, 2007:} no inflections, any language.}
  \label{figure:byEngineTwoDates:noinfl:anylang:R}
\end{center}
\end{figure}

\begin{figure}
\begin{center}
  \includegraphics[width=15cm,height=8.5cm]{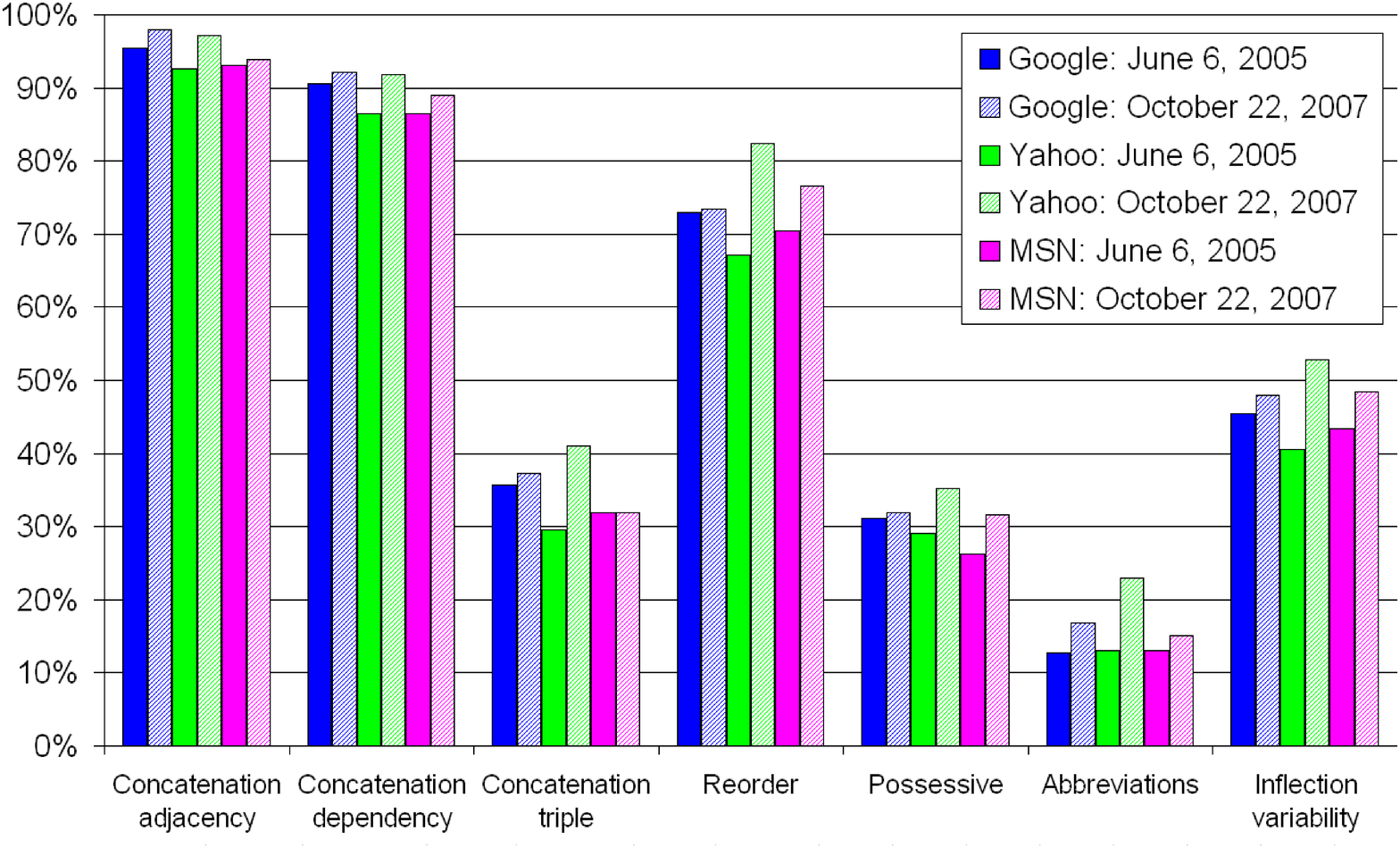}\\
  \caption{{\bf Coverage by search engine for June 6, 2005 and October 22, 2007:} using inflections, any language.}
  \label{figure:byEngineTwoDates:withinfl:anylang:R}
\end{center}
\end{figure}

\begin{figure}
\begin{center}
  \includegraphics[width=15cm,height=8.5cm]{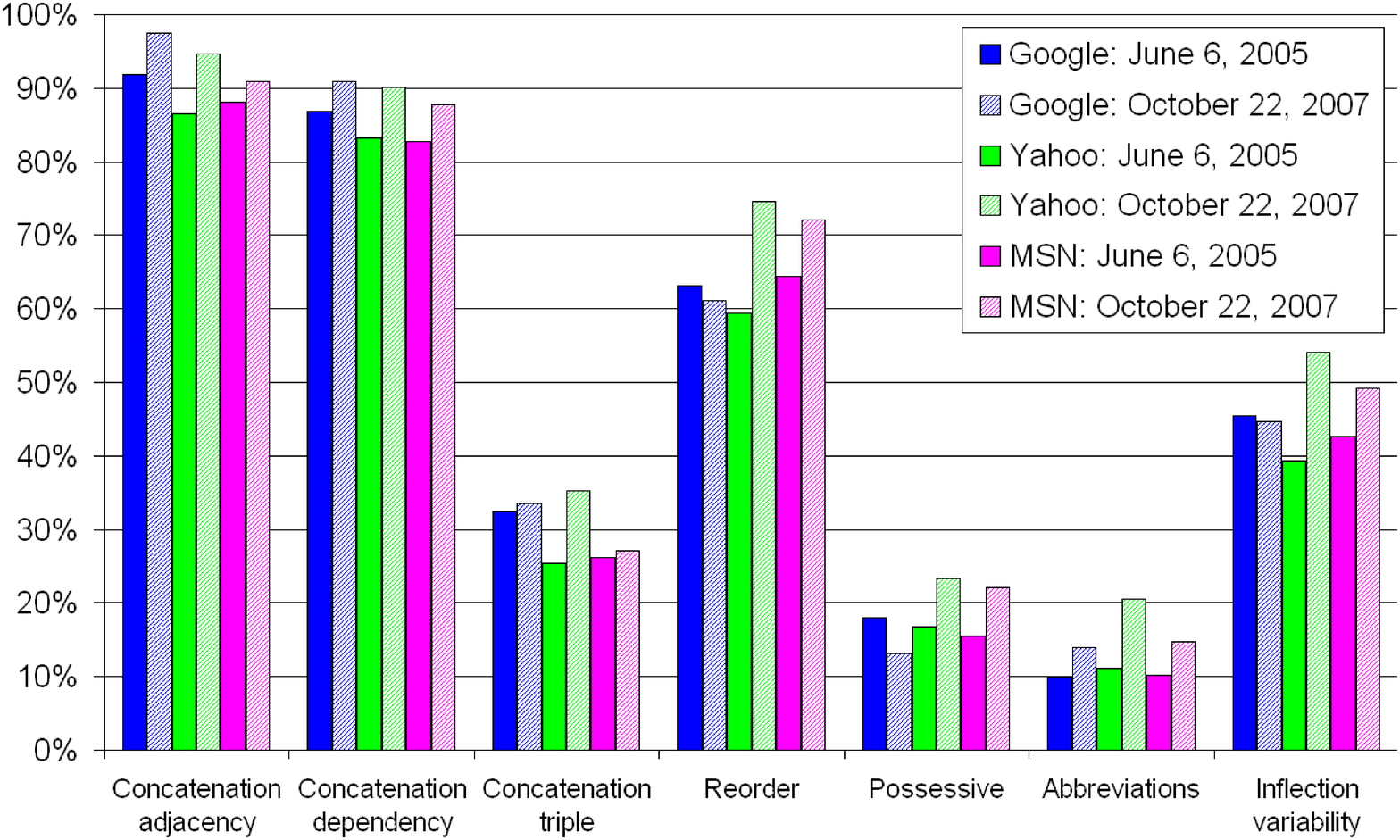}\\
  \caption{{\bf Accuracy by search engine for June 6, 2005 and October 22, 2007:} no inflections, English pages only.}
  \label{figure:byEngineTwoDates:noinfl:english:R}
\end{center}
\end{figure}

\begin{figure}
\begin{center}
  \includegraphics[width=15cm,height=8.5cm]{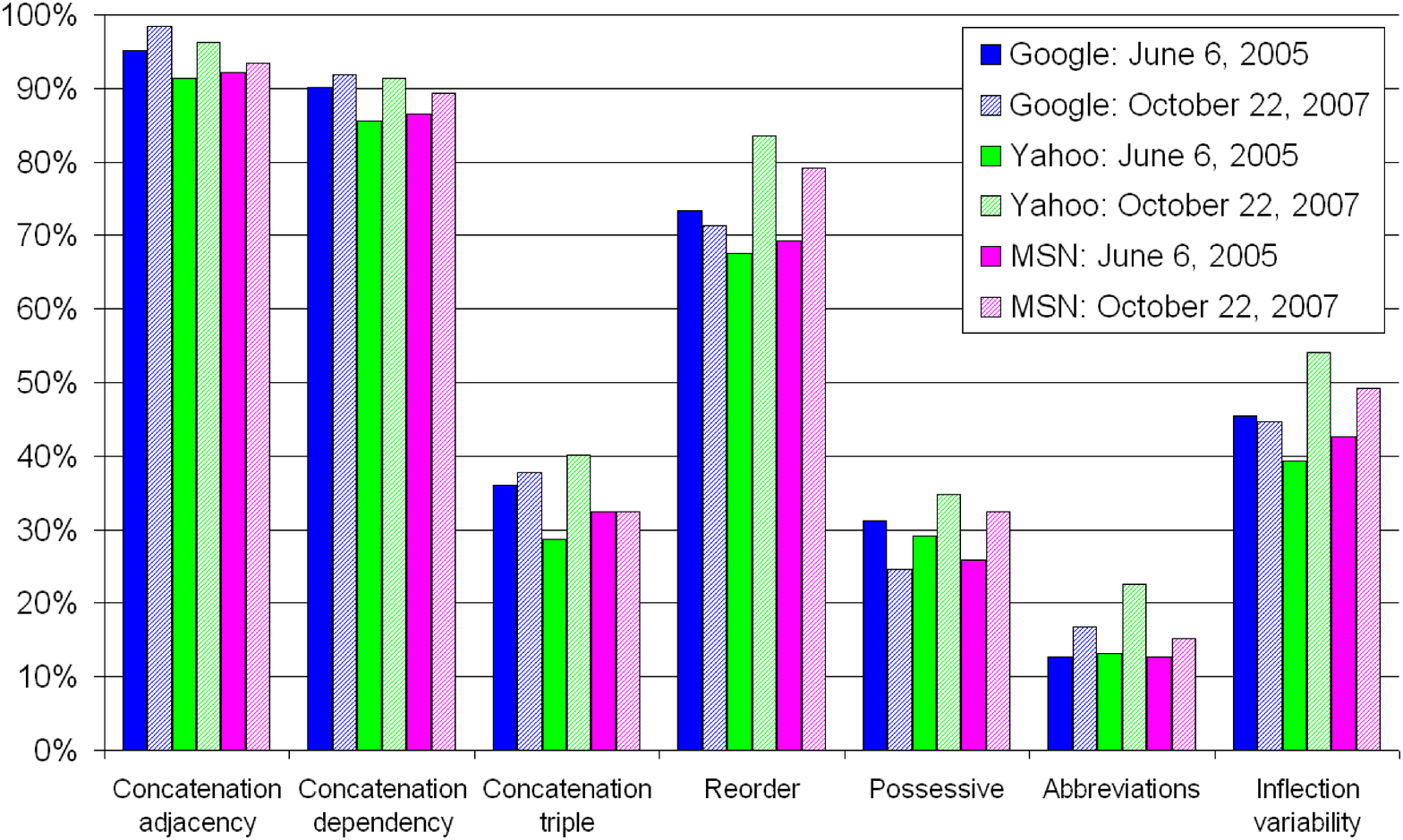}\\
  \caption{{\bf Accuracy by search engine for June 6, 2005 and October 22, 2007:} using inflections,
            English pages only.}
  \label{figure:byEngineTwoDates:withinfl:english:R}
\end{center}
\end{figure}

The corresponding results for the coverage are shown in Figures
\ref{figure:byEngineTwoDates:noinfl:anylang:R}, \ref{figure:byEngineTwoDates:withinfl:anylang:R},
\ref{figure:byEngineTwoDates:noinfl:english:R}, \ref{figure:byEngineTwoDates:withinfl:english:R}.
As we can see, the coverage has improved over time for all three search engines,
which can be expected, given the increase in the number of indexed Web pages.

\subsection{Page Hit Estimations: Top 10 vs. Top 1,000}

As I mentioned in section \ref{sec:pagehits:rounding},
search engines may sample from their indexes,
rather than performing exact computations,
in which case index scanning might stop once the requested number of pages is produced:
10 by default, and up to 1,000 at most.
The more pages the user requests, the larger the part of the indexes scanned,
and the more accurate the estimates would be.

\begin{figure}
\begin{center}
  \includegraphics[width=15cm,height=8cm]{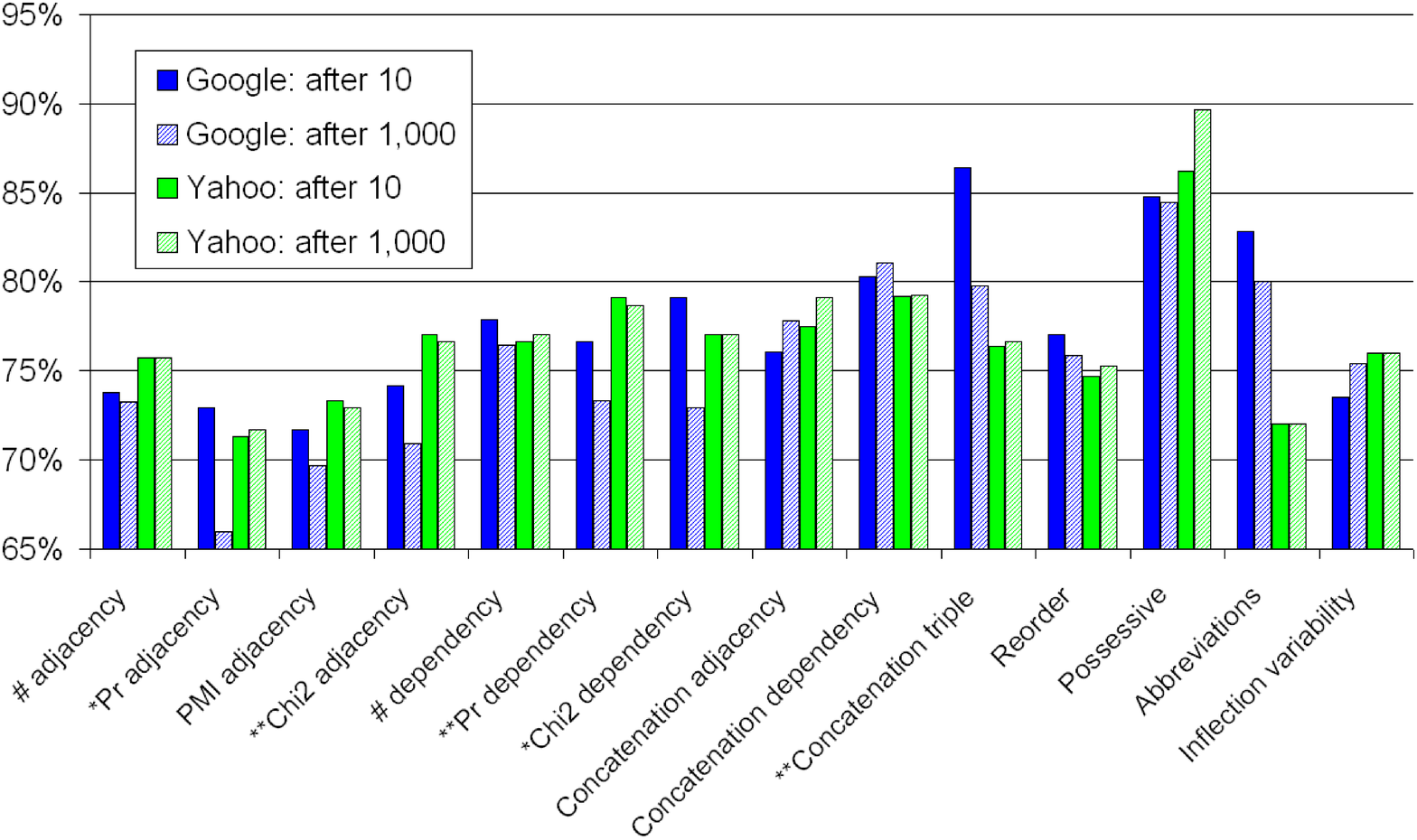}\\
  \caption{{\bf Accuracy by search engine for October 22, 2007, 10 vs. 1,000 results:} no inflections, any language.
            Statistically significant differences between the same/different search engines is marked with a single/double asterisk.}
  \label{figure:byEngineTwoDates:noinfl:anylang:990}
\end{center}
\end{figure}

\begin{figure}
\begin{center}
  \includegraphics[width=15cm,height=8cm]{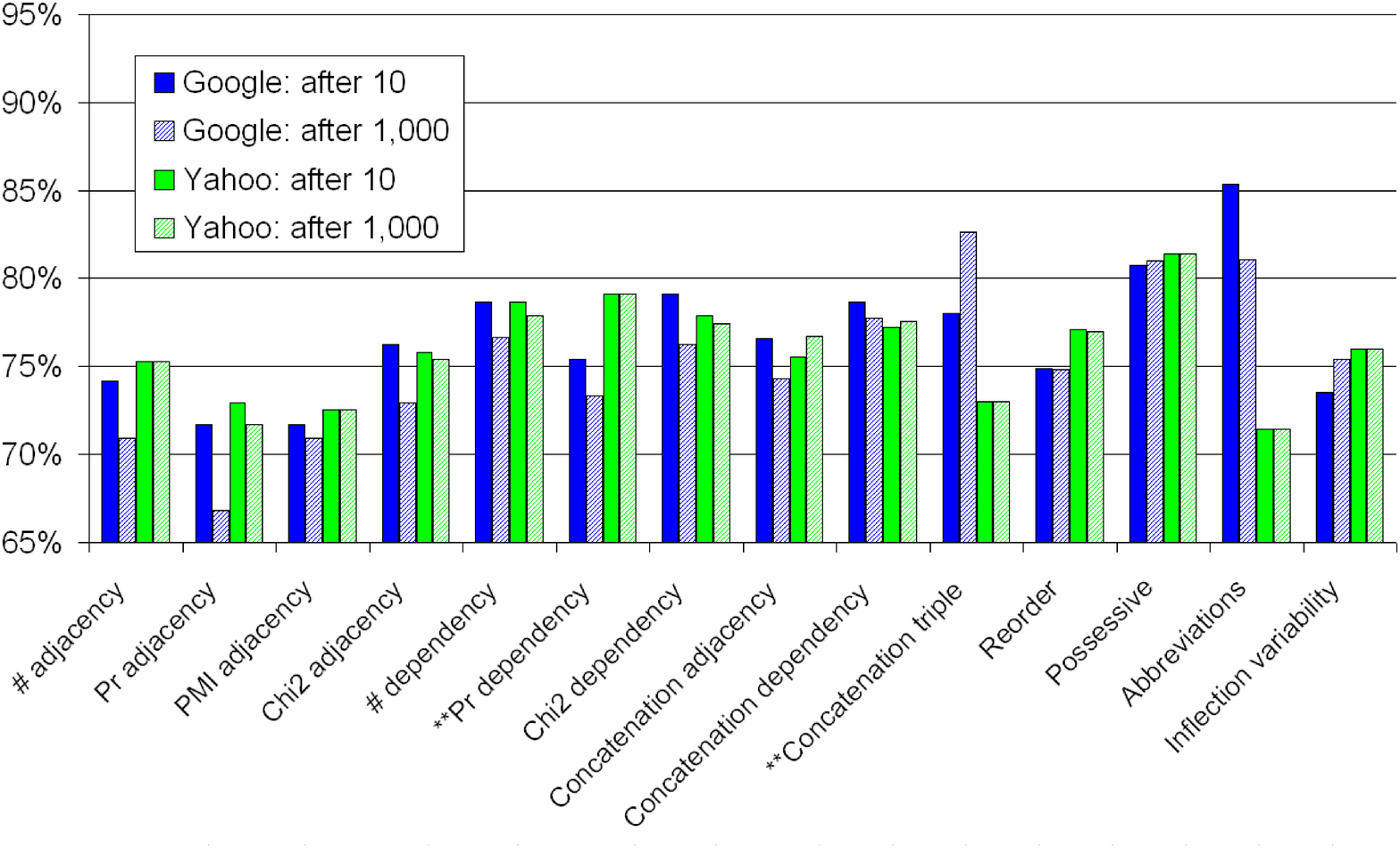}\\
  \caption{{\bf Accuracy by search engine for October 22, 2007, 10 vs. 1,000 results:} with inflections, any language.
            Statistically significant differences between the same/different search engines is marked with a single/double asterisk.}
  \label{figure:byEngineTwoDates:withinfl:anylang:990}
\end{center}
\end{figure}

To see the impact of the more accurate estimates, I extract the page hit estimates
after having asked the search engine to produce 10 and 1,000 results.
The corresponding accuracies for {\it Google} and {\it Yahoo!} on October 22, 2007
are shown in
Figure \ref{figure:byEngineTwoDates:noinfl:anylang:990} (no inflections, any language)
and Figure \ref{figure:byEngineTwoDates:withinfl:anylang:990} (with inflections, any language).
Interestingly, using estimates after having requested 1,000 results has a negative impact on accuracy,
i.e. the initial page hit estimates were actually better.
Note also that {\it Yahoo!} shows less variability in accuracy than {\it Google},
which suggests {\it Yahoo!}'s page hit estimates are more stable.

\section{Conclusion and Future Work}

\begin{figure}
\begin{center}
  \includegraphics[width=15cm,height=8cm]{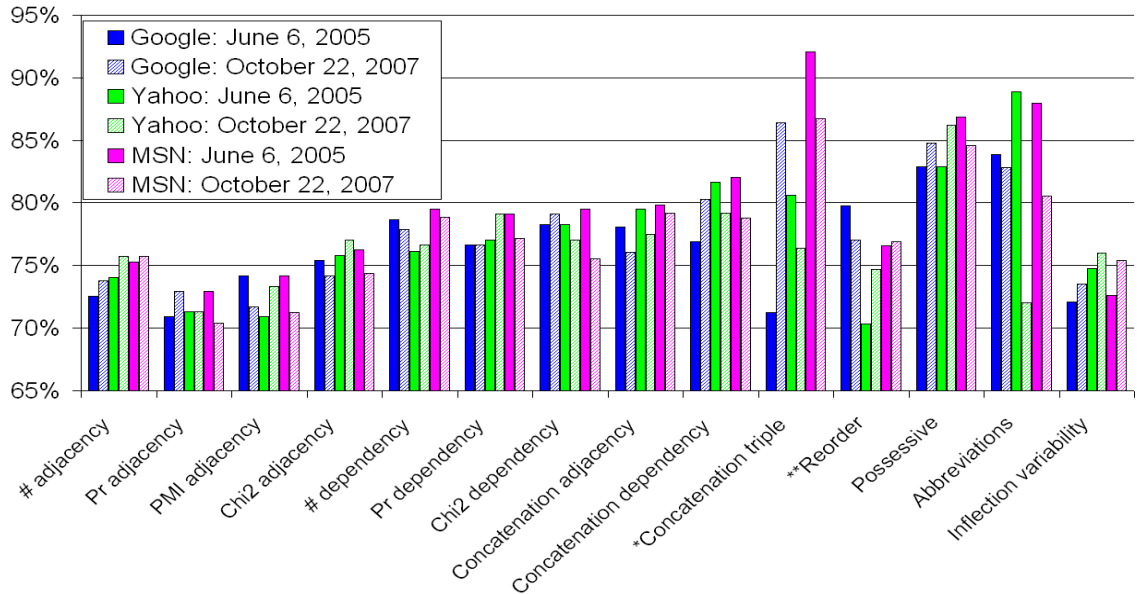}\\
  \caption{{\bf Accuracy by search engine for June 6, 2005 and October 22, 2007:} no inflections, any language.
            Statistically significant differences between the same/different search engines is marked with a single/double asterisk.}
  \label{figure:byEngineTwoDates:noinfl:anylang}
\end{center}
\end{figure}

In the present chapter, I have described some problems
with using search engine page hits as a proxy for $n$-gram frequency estimates.
Then, using a real NLP task, {\it noun compound bracketing},
and 14 different $n$-gram based models, I have shown that variability
over time and across search engines,
as well as using language filters and word inflections,
while sometimes causing sizable fluctuations,
generally have no statistically significant impact
on accuracy and coverage for that particular task
(except for ``exotic'' models like
{\it concatenation triple}, {\it inflection variability}, etc.
under some of the experimental conditions).
These results suggest that, despite the fluctuations,
overall experiments with models using search engines statistics
can be considered relatively stable and reproducible.
This is good news which should reassure researchers using Web as a data source that,
despite the recent scepticism of \namecite{Kilgarriff:2007},
{\it Googleology} is not ``bad science'' after all.

In order to further bolster these results,
more experiments with other NLP tasks which
make use of Web-derived $n$-gram estimates would be needed.
It would be good to perform a similar study
for {\it surface features} (see chapter \ref{sec:bracketing:web:derived:features})
and for {\it paraphrases} (see chapter \ref{sec:paraphrases}) as well.
It would be also interesting to try languages other than English, where the
language filter could be much more important, and where the
impact of the inflection variability may differ, especially in
case of a morphologically rich language like Bulgarian or Russian.
Finally, since different experimental conditions, e.g., different search engines,
exhibit strength for different kinds of models,
it looks promising to combine them, e.g., in a log-linear model,
similarly to the way \namecite{Lapata:Keller:05:Web:based:Models}
combined $n$-gram statistics from the Web and from the {\it British National Corpus}.


\chapter{Conclusions and Future Work}
\label{chapter:conclusion}

\section{Contributions of the Thesis}

In the present thesis, 
I have introduced novel surface features and paraphrases which go beyond simple $n$-gram counts,
which have been demonstrated highly effective for a wide variety of NLP tasks.
Based on them, I have built several novel, highly accurate
Web-based approaches to solving problems
related to the syntax and semantics of noun compounds,
and to other related problems like relational similarity,
machine translation, PP-attachment and NP coordination.

In chapter \ref{chapter:NC:bracketing},
I have extended and improved upon the state-of-the-art approaches
to noun compound bracketing, achieving 89.34\% accuracy
on the benchmark {\it Lauer's dataset},
which is a statistically significant improvement over the best previously published results.

In chapter \ref{chapter:NC:semantics},
I have presented a novel, simple unsupervised approach to noun compound
interpretation in terms of predicates characterizing the hidden
relation between the head and the modifier, which could be useful for many NLP tasks.
Using these verbs as features in a classifier,
I have demonstrated state-of-the-art results for several relational similarity problems,
including: mapping noun-modifier pairs to abstract relations like \texttt{TIME} and \texttt{LOCATION},
classifying relation between nominals, and solving SAT verbal analogy questions.

In chapter \ref{chapter:MT}, I have described an application
of the methods developed in chapters \ref{chapter:NC:bracketing} and \ref{chapter:NC:semantics}
for noun compound paraphrasing to an important real-world task: {\it machine translation}.
I have proposed a novel monolingual paraphrasing method
based on syntactic transformations at the NP-level,
used to increase the training data
with nearly equivalent sentence-level syntactic paraphrases of the original corpus.
The evaluation has shown an improvement equivalent to 33\%-–50\% of that
of doubling the amount of training data.

In chapter \ref{chapter:instability},
I have addressed the important question of the stability and reproducibility
of the results obtained using search engines.
Using a real NLP task, {\it noun compound bracketing},
I have shown that variability
over time and across search engines,
as well as using language filters and word inflections,
generally has no statistically significant impact.
These results suggest that, despite the fluctuations,
overall experiments with models using search engines statistics
can be considered stable and reproducible.

Finally, I have created three datasets which might be useful to other researchers:

(1) a domain-specific {\it Biomedical dataset} for noun compound bracketing
is given in Appendix \ref{appendix:bio:set};

(2) an {\it NP-coordination dataset} is listed in Appendix \ref{appendix:coord:set};

(3) a distribution of the top 10 human-generated verbs paraphrasing the hidden relation
between the nouns in 250 noun-noun compounds used in the theory of \namecite{levi:1978}
is given in Appendix \ref{chap:mturkompare:human}.

\section{Directions for Future Research}


In my opinion, the most exciting and probably the most promising direction 
for future research is on using the generated verbs and paraphrases of noun-noun compounds 
for various related NLP tasks, e.g.,
paraphrase-augmented machine translation \cite{Callison-Burch:al:2006:mt},
noun compound translation \cite{Baldwin:Tanaka:2004},
machine translation evaluation \cite{Russo-Lassner:2005,Kauchak:Barzilay:2006:par},
summarization evaluation \cite{Zhou:al:2006:mt},
textual entailment (see Appendix \ref{chapter:sem:entailment:nc}),
information retrieval, e.g. index normalization \cite{Zhai:1997:nc:parsing}, 
query segmentation \cite{bergsma:wang:2007:query:segm}.
relational Web search \cite{Cafarella:al:2006:relational:web:search}, 
lexical acquisition, e.g., extracting names of diseases and drugs
from the Web \cite{Etzioni:al:2005:NER}, etc.

My future plans for the noun compound bracketing problem include
extending the approach to noun compounds consisting of more than three nouns,
bracketing NPs in general, and recognizing structurally ambiguous noun compounds.
As the evaluation results in chapter \ref{chapter:NC:bracketing} show,
there may be potential benefits in combining Web and corpus-based statistics, e.g.,
similarly to the way \namecite{Lapata:Keller:05:Web:based:Models}
combined $n$-gram estimates from the Web and from the {\it British National Corpus}.
Another information source that seems promising for the syntax and semantics of noun compounds
is cross-linguistic evidence, e.g., extracted from parallel multi-lingual corpora like {\it Europarl}
as described by \namecite{girju:2007:ACLMain}.

It would be interesting to extend the page hit instability study from chapter \ref{chapter:instability},
and to apply it to {\it surface features} (see section \ref{sec:bracketing:web:derived:features})
and {\it paraphrases} (see section \ref{sec:paraphrases}).
Trying other NLP tasks would be interesting as well,
as will be experimenting with languages other than English, where the
language filter might be much more important, and where the
impact of the inflection variability may be different, especially in
case of a morphologically rich language like Bulgarian and Russian.

Last but not least, special efforts should be made to reduce the number of queries to the search engines.
{\it Google}'s {\it Web 1T 5-gram} dataset,
which is available from the Linguistic Data Consortium,
while somewhat limited, might still be quite useful in this respect.


\bibliography{bibliography}
\bibliographystyle{lsalike}

\appendix

\chapter{Semantic Entailment}
\label{chapter:sem:entailment:nc}

For a typical right-headed non-lexicalized noun-noun compound $n_1n_2$,
it can be expected that the assertion ``$n_1n_2$ is a type/kind of $n_2$'' 
(e.g., ``{\it lung \underline{cancer}} is a kind/type of {\it cancer}'') would be true, 
i.e. that $n_1n_2$ semantically entails $n_2$.
Note that this is not true for all noun-noun compounds, e.g., 
{\it bird\underline{brain}} is not a kind/type of {\it brain}; it is a kind of person.
Similarly, {\it vitamin \underline{D}} is not a kind/type of {\it D}; it is a kind of {\it vitamin}.

For a three-word right-headed non-lexicalized noun compound $n_1n_2n_3$,
it can be expected that the assertion ``$n_1n_2n_3$ is a type/kind of $n_3$'' would be true,
e.g., ``{\it lung cancer \underline{drug}} is a kind/type of {\it drug}''.
If the compound is left-bracketed $(n_1n_2)n_3$, 
it can be further expected that the assertion ``$n_1n_2n_3$ is a type/kind of $n_2n_3$'' 
would be true as well, e.g., ``{\it lung \underline{cancer} \underline{drug}} is a kind/type of {\it cancer drug}''.
If the compound is right-bracketed $n_1(n_2n_3)$,
then it can be expected that the following two assertions are true
``$n_1n_2n_3$ is a type/kind of $n_2n_3$'' and
``$n_1n_2n_3$ is a type/kind of $n_1n_3$'',
e.g., ``{\it oxydant \underline{wrinkle} \underline{treatment}} is a kind/type of {\it wrinkle treatment}''
and ``{\it \underline{oxydant} wrinkle \underline{treatment}} is a kind/type of {\it oxydant treatment}''.

In other words, given a three-word right-headed non-lexicalized noun compound $(n_1n_2)n_3$,
it is expected to entail $n_3$ and $n_2n_3$, but not $n_1n_3$, if left-bracketed,
and all three $n_3$, $n_2n_3$, and $n_1n_3$, if right-bracketed.
Therefore, noun compound bracketing can be used to suggest semantic entailment.
Unfortunately, there is no easy way to distinguish left-headed from right-headed compounds,
and it is even harder to distinguish between lexicalized and non-lexicalized compounds, 
the boundary between which is not well-established.
The matter is further complicated due to nominalization, 
metaphor, metonymy, contextual dependency, world knowledge, and pragmatics.

\begin{table}
\begin{center}
\begin{tabular}{|l|lc|lc|c|}
 \hline
    $\mathbf{n_1n_2n_3}$ & $\mathbf{n_1n_3}$ & {\bf ``$\mathbf{\Rightarrow}$''} & $\mathbf{n_2n_3}$ & {\bf ``$\mathbf{\Rightarrow}$''} & {\bf Brack.}\\
 \hline
   \small{\it army ant behavior} & \small{\it army behavior} & no & \small{\it ant behavior} & yes & left\\
   \small{\it health care reform} & \small{\it health reform} & no & \small{\it care reform} & {\bf NO} & left\\
   \small{\it heart rate variability} & \small{\it heart variability} & {\bf YES} & \small{\it rate variability} & yes & left\\
   \small{\it lung cancer doctor} & \small{\it lung doctor} & {\bf YES} & \small{\it cancer doctor} & yes & left\\
   \small{\it lung cancer physician} & \small{\it lung physician} & {\bf YES} & \small{\it cancer physician} & yes & left\\
   \small{\it lung cancer patient} & \small{\it lung patient} & {\bf YES} & \small{\it cancer patient} & yes & left\\
   \small{\it lung cancer survivor} & \small{\it lung survivor} & no & \small{\it cancer survivor} & yes & left\\
   \small{\it lung cancer treatment} & \small{\it lung treatment} & {\bf YES} & \small{\it cancer treatment} & yes & left\\
   \small{\it lung cancer drug} & \small{\it lung drug} & no & \small{\it cancer drug} & yes & left\\
   \small{\it science fiction satire} & \small{\it science satire} & no & \small{\it fiction satire} & {\bf NO} & left\\
   \small{\it science fiction writer} & \small{\it science writer} & no & \small{\it fiction writer} & {\bf NO} & left\\
   \small{\it science fiction novels} & \small{\it science novels} & no & \small{\it fiction novels} & {\bf NO} & left\\
   \small{\it US army forces} & \small{\it US forces} & {\bf YES} & \small{\it army forces} & yes & left\\
   \small{\it vitamin D deficiency} & \small{\it vitamin deficiency} & {\bf YES} & \small{\it D deficiency} & {\bf NO} & left\\
 \hline
   \small{\it oxydant wrinkle treatment} & \small{\it oxydant treatment} & yes & \small{\it wrinkle treatment} & yes & right\\
   \small{\it space science fiction} & \small{\it space fiction} & {\bf NO} & \small{\it science fiction} & yes & right\\
 \hline
   \small{\it brain stem cell} & \small{\it brain cell} & yes & \small{\it stem cell} & yes & both\\
 \hline
\end{tabular}
\caption{\textbf{Semantic entailment and noun compound bracketing.} 
        The entailments that are inconsistent with the bracketing are shown in bold.}
\label{table:HypernymDatasetExample}
\end{center}
\end{table}

Table \ref{table:HypernymDatasetExample} shows sample three-word noun compounds,
the corresponding bracketing and whether $n_1n_3$ and $n_2n_3$ are entailed or not.
The entailments that are inconsistent with the bracketing are shown in bold.
For some of them, it is easy to come up an explanation:

\begin{enumerate}
    \item {\bf Lexicalization/morphology}.
    As I mentioned in section \ref{sec:compounding:notion},
    noun compounds lie on the boundary between
    lexicalization/morphology and syntax. While some are completely
    lexicalized (e.g. {\it birdbrain}), and other are purely compositional
    (e.g., {\it ant behavior}), the majority occupy the continuum in between.
    For example, {\it science fiction} looks lexicalized to a higher
    degree compared to {\it lung cancer} and {\it heart rate}, which in
    turn are more lexicalized compared to {\it cancer drug}. In the process
    of lexicalization, noun-noun compounds become less and less
    compositional, which could explain the {\it science fiction}
    examples: {\it science fiction} is not a kind of {\it fiction}. The same explanation
    can be applied to the {\it health care} example, which is probably even more
    lexicalized as it is often written as a single word
    {\it healthcare}: while its meaning does not depart too far from a pure
    compositional interpretation, it is non-compositional enough to prevent
    {\it health care reform} from entailing {\it care reform}.

    \item {\bf Individual Nouns' Semantics}. The fact
    that {\it lung cancer physician} implies {\it lung physician} cannot
    be explained by the relatively low degree of lexicalization of {\it lung cancer} alone. 
    The noun compound is left bracketed and therefore this entailment
    should not be true. It looks like the semantics of the individual nouns
    and world knowledge come into play, e.g., {\it physician} is likely to be modified by organs.

    \item {\bf Nominalization}. The really problematic noun compounds are
    {\it lung cancer patient} and {\it lung cancer survivor}.
    While {\it patient} and {\it survivor} have a very similar meaning in
    the context of these compounds, {\it lung patient} is
    implied, but {\it lung survivor} is not. One possible
    explanation is that {\it survivor} is a nominalization of the
    verb {\it survive}, which likes {\it cancer}, and therefore {\it lung cancer}, 
    as a direct object, but does not like {\it lung}.

    \item {\bf Left Head}. The entailments are completely reversed for {\it vitamin D deficiency}, 
    which contains the left-headed noun compound {\it vitamin D} as a modifier:
    the interpretation of the noun compound changes completely when the head precedes the modifier.

    \item {\bf Metonymy}. Due to metonymy, 
    in the case of {\it US army forces}, all possible subsequences are entailed:
    not only {\it US forces}, {\it army forces} and {\it forces}, 
    but also {\it US army}, {\it army} and, in some contexts, even just {\it US}.
    
\end{enumerate}

\chapter{Phrase Variability in Textual Entailment}
\label{chapter:intro:illustration}

Here I present a short study of some examples from the
development dataset of the Second Pascal Recognizing Textual
Entailment (RTE2)
Challenge.\footnote{\texttt{http://www.pascal-network.org/Challenges/RTE2}}
The challenge addresses a generic semantic inference task needed by
many natural language processing applications, including Question
Answering (QA), Information Retrieval (IR), Information Extraction
(IE), and (multi-)document summarization (SUM). Given two textual
fragments, a text $T$ and a hypothesys $H$, the goal is
to recognize whether the meaning of $H$ is entailed (can be inferred)
from $T$: 

\vspace{12pt}
\begin{quote}
``{\it We say that $T$ entails $H$ if, typically, a human
reading $T$ would infer that $H$ is most likely true. This somewhat
informal definition is based on (and assumes) common human
understanding of language as well as common background knowledge.}''
(RTE2 task definition)
\end{quote}

Below I list some examples where $T$ does entail $H$
in order to illustrate different kinds of phrase variability problems
that a successful inference procedure might need to solve.
The target phrases are underlined.

\section{Prepositional Paraphrase}

\noindent Here a noun compound has to be matched to a prepositional paraphrase.

\begin{center}
``{\it paper costs}'' $\Rightarrow$ ``{\it cost of paper}''
\end{center}
\vspace{6pt}

\begin{center}
\framebox{
\parbox{4.83in}{
\begin{small}
\texttt{<pair id="503" entailment="YES" task="IR">}

\texttt{<t>Newspapers choke on rising \underline{paper costs} and falling revenue.</t>}

\texttt{<h>The \underline{cost of paper} is rising.</h>}

\texttt{</pair>}
\end{small}
}}
\end{center}

\section{Prepositional Paraphrase \& Genitive Noun Compound}

\noindent Here a genitive noun compound needs to be
matched to a prepositional paraphrase.

\begin{center}
 ``{\it Basra's governor}'' $\Longrightarrow$ ``{\it
governor of Basra}''
\end{center}
\vspace{6pt}

\begin{center}
\framebox{
\parbox{5.36in}{
\begin{small}
 \texttt{<pair id="337" entailment="YES" task="SUM">}

 \texttt{<t>\underline{Basra's governor} said he would
not cooperate with British troops until}

 \indent \texttt{there was an apology for a raid to
free two UK soldiers.</t>}

 \texttt{<h>The \underline{governor of Basra} will not
work with British troops until}

 \indent \texttt{there is an apology for a raid to
free two UK soldiers.</h>}

 \texttt{</pair>}
\end{small}
}}
\end{center}
\vspace{6pt}

\section{Verbal \& Prepositional Paraphrases}

\noindent Here we have two different paraphrases, one verbal
and one prepositional, of the same noun compound {\it WTO Geneva
headquarters}.

\begin{center}
 ``{\it Geneva headquarters of the WTO}''
$\Rightarrow$ ``{\it WTO headquarters are located in Geneva}''
\end{center}
\vspace{6pt}

\begin{center}
\framebox{
\parbox{5.3in}{
\begin{small}
 \texttt{<pair id="284" entailment="YES" task="QA">}

 \texttt{<t>While preliminary work goes on at the
\underline{Geneva headquarters of the WTO},}

 \texttt{with members providing input, key
decisions are taken at the ministerial}

 \texttt{meetings.</t>}

 \texttt{<h>The \underline{WTO headquarters are located in Geneva}.</h>}

 \texttt{</pair>}
\end{small}
}}
\end{center}
\vspace{6pt}

\section{Nominalization}

\noindent Here we have two nouns representing different
nominalizations referring to the same process.

\begin{center}
``{\it legalizing drugs}'' $\Rightarrow$ ``{\it drug legalization}''
\end{center}
\vspace{6pt}

\begin{center}
\framebox{
\parbox{5.35in}{
\begin{small}
 \texttt{<pair id="103" entailment="YES" task="IR">}

 \texttt{<t>This paper describes American alcohol
use, the temperance movement,}

 \indent \texttt{Prohibition, and the War on Drugs
and explains how \underline{legalizing drugs} would}

 \indent \texttt{reduce crime and public health
problems.</t>}

 \texttt{<h>\underline{Drug legalization} has
benefits.</h>}

 \texttt{</pair>}
\end{small}
}}
\end{center}
\vspace{6pt}

\section{Bracketing}

\noindent Here we have an entailment which extracts a
component consistent with a left bracketing.

\begin{center}
``{\it $[$breast cancer$]$ patients}'' $\Rightarrow$ ``{\it $[$breast cancer$]$}''
\end{center}
\vspace{6pt}

\begin{center}
\framebox{
\parbox{4.43in}{
\begin{small}
 \texttt{<pair id="177" entailment="YES" task="SUM">}

 \texttt{<t>Herceptin was already approved to treat
the sickest}

 \indent \texttt{\underline{breast cancer patients}, and the company said, Monday,}

 \indent \texttt{it will discuss with federal regulators the possibility}

 \indent \texttt{of prescribing the drug for more \underline{breast cancer patients}.</t>}

 \texttt{<h>Herceptin can be used to treat
\underline{breast cancer}.</h>}

 \texttt{</pair>}
\end{small}
}}
\end{center}
\vspace{6pt}

\vspace{12pt}

\noindent Here is a longer example where the extracted component is consistent with a left bracketing.

\begin{center}
``{\it $[$ocean $[$remote sensing$]]$ science}'' $\Rightarrow$
``{\it $[$ocean $[$remote sensing$]]$}''
\end{center}
\vspace{6pt}

\begin{center}
\framebox{
\parbox{4.43in}{
\begin{small}
 \texttt{<pair id="155" entailment="YES" task="IR">}

 \texttt{<t>Scientists and engineers in the APL Space
Department have}

 \indent \texttt{contributed to \underline{ocean remote sensing science} and technology}

 \indent \texttt{for more than a quarter century.</t>}

 \texttt{<h>\underline{Ocean remote sensing} is developed.</h>}

 \texttt{</pair>}
\end{small}
}}
\end{center}
\vspace{6pt}

\section{Bracketing \& Paraphrase}

\noindent Here we have a noun compound which is paraphrased
using a preposition in a way that is consistent with left
bracketing.

\begin{center}
``{\it $[$ivory trade$]$ ban}'' $\Rightarrow$ ``{\it ban on $[$ivory
trade$]$}''
\end{center}
\vspace{6pt}

\begin{center}
\framebox{
\parbox{5.35in}{
\begin{small}
 \texttt{<pair id="117" entailment="NO" task="IR">}

 \texttt{<t>The release of its report led to calls
for a complete \underline{ivory trade ban},}

 \indent \texttt{and at the seventh conference in 1989, the
African Elephant was moved to}

 \indent \texttt{appendix one of the treaty.</t>}

 \texttt{<h>The \underline{ban on ivory trade} has been
effective in protecting the elephant}

 \indent \texttt{from extinction.</h>}

 \texttt{</pair>}
\end{small}
}}
\end{center}
\vspace{6pt}

\section{Synonymy}

\noindent Here we have two noun compounds with the same modifiers and
synonymous heads.

\begin{center}
``{\it marine plants}'' $\Rightarrow$ ``{\it marine vegetation}''
\end{center}
\vspace{6pt}

\begin{center}
\framebox{
\parbox{5.5in}{
\begin{small}
 \texttt{<pair id="65" entailment="YES" task="IR">}

 \texttt{<t>A number of \underline{marine plants} are
harvested commercially}

 \texttt{in Nova Scotia.</t>}

 \texttt{<h>\underline{Marine vegetation} is
harvested.</h>}

 \texttt{</pair>}
\end{small}
}}
\end{center}
\vspace{6pt}

\section{Hyponymy \& Prepositional Paraphrase}

\noindent Here we have a prepositional paraphrase and a noun compound
with the same modifiers and heads that are in a hyponymy
relation\footnote{Alternatively, we have paraphrases for the
longer noun compounds {\it marijuana legalization benefits} and {\it drug
legalization benefits}, i.e.,
``{\it benefits in the legalization of marijuana}'' $\Rightarrow$
``{\it drug legalization has benefits}''.}: {\it marijuana} is a kind of {\it drug}.

\begin{center}
``{\it legalization of marijuana}'' $\Rightarrow$ ``{\it drug
legalization}''
\end{center}
\vspace{6pt}

\begin{center}
\framebox{
\parbox{5.85in}{
\begin{small}
 \texttt{<pair id="363" entailment="YES" task="IR">}

 \texttt{<t>One economic study will not be the basis
of Canada's public policy decisions,}

 \indent \texttt{but Easton's research does
conclusively show that there are economic benefits}

 \indent \texttt{in the \underline{legalization of
marijuana}.</t>}

 \texttt{<h>\underline{Drug legalization} has
benefits.</h>}

 \texttt{</pair>}
\end{small}
}}
\end{center}
\vspace{6pt}

\section{Prepositional Paraphrase, Nominalization \& Synonymy}

\noindent Here the verb {\it enrich} has been substitued by
its synonym {\it enhance}, which in turn has been nominalized. On
the other hand, the verbs {\it enrich} and {\it feed} are good
paraphrases for the noun compound {\it soil enhancers}, e.g., ``{\it enhancers
that enrich the soil}'', and {\it fertilizer} is a contextual
synonym of {\it enhancer}.

\begin{center}
``{\it enriches and feeds the soil}'' $\Rightarrow$ ``{\it soil
enhancers}''
\end{center}
\vspace{6pt}

\begin{center}
\framebox{
\parbox{5.85in}{
\begin{small}
 \texttt{<pair id="192" entailment="YES" task="IR">}

 \texttt{<t>Organic fertilizer slowly
\underline{enriches and feeds the soil}. Fast acting synthetic
fertilizers harm soil life.</t>}

 \texttt{<h>Organic fertilizers are used as
\underline{soil enhancers}.</h>}

 \texttt{</pair>}
\end{small}
}}
\end{center}
\vspace{6pt}

\section{Nominalization, Bracketing \& Synonymy}

\noindent The example this time is
Here the modificational past participle
{\it damaged} becomes the new noun compound head: {\it injury}. One possible
interpretation is that {\it damaged}, which is a form
of the verb {\it to damage}, has been substituted by the synonymous
verb {\it to injure}, which has been nominalized and
has become the new noun compound head.
Note however that the verb {\it to damage} does
not have to be analyzed as a synonym of {\it to injure}; it is also
a good candidate for a verbal paraphrase of
{\it spinal cord injury}, e.g., as ``{\it injury which damaged the spinal cord''}.
Note that while {\it damaged spinal cord} does entail {\it spinal cord injury},
the entailment does not hold for the overall example.

\begin{center}
``{\it damaged $[$spinal cord$]$}'' $\Rightarrow$ ``{\it $[$spinal
cord$]$ injury}''
\end{center}

\begin{center}
\framebox{
\parbox{5.9in}{
\begin{small}
 \texttt{<pair id="261" entailment="NO" task="SUM">}

 \texttt{<t>The new work went an extra step,
suggesting that the connections that the stem}

 \indent \texttt{cells form to help bridge the
\underline{damaged spinal cord}, are key to recovery.</t>}

 \texttt{<h>The experiment, reported Monday, isn't
the first to show that stem cells offer}

 \indent \texttt{tantalizing hope for \underline{spinal cord injury}.</h>}

 \texttt{</pair>}
\end{small}
}}
\end{center}
\vspace{6pt}

\section{Discussion}

As I have shown, solving many, if not all, of the above textual entailments might
require understanding the syntax and/or the semantics of the involved noun compounds.
Some of the examples require checking whether a particular
verb or preposition is acceptable as a paraphrase for a given noun compound,
other call for judging on semantic entailment after word dropping
from a noun compound. Additional factors come into play as well, e.g.,
genitives,
inflectional and derivational morphological alternations,
lexical relations like synonymy and hyponymy. 

An important observation to make is that
noun compound interpretation techniques can naturally extend
to cases where noun compounds are formally
not involved. For example, checking whether ``{\it Geneva headquarters
of the WTO}'' could textually entail ``{\it WTO headquarters are located in
Geneva}'' can be performed by understanding the semantics of the
noun compound ``{\it WTO Geneva headquarters}'' in terms of possible
prepositional and verbal paraphrases.

\chapter{\emph{Biomedical Dataset} for Noun Compound Bracketing}
\label{app:dataset:bio}

\section{Extracting the Most Frequent Noun Compounds from MEDLINE}
\label{sec:dataset:bio:extract}

The three-word noun compounds for the {\it Biomedical dataset}
for noun compound bracketing are extracted from a collection of 1.4 million
MEDLINE abstracts (citations between 1994 and 2003),
which constitutes about 8.23\% of the total 17 million abstracts in MEDLINE.\footnote{\texttt{http://www.ncbi.nlm.nih.gov/sites/entrez}}

The abstracts have been sentence split, tokenized, POS tagged and shallow parsed
using the Open NLP tools.\footnote{\texttt{http://opennlp.sourceforge.net/}}
The resulting linguistic annotations -- sentences, tokens, POS, and shallow parses --
were indexed by a specialized system architecture developed as part of
the {\it Biotext project}\footnote{\texttt{http://biotext.berkeley.edu}},
which supports queries over layers of annotation on natural language text.
The system allows for both hierarchical and
overlapping layers and for querying at multiple levels of
description. The implementation is built on top of a standard
relational database management system (RDBMS),
and, by using carefully constructed indexes, it can execute complex
queries efficiently.
More information about the
system\footnote{The LQL query language and the architecture have been
developed by Ariel Schwartz, Brian Wolf, Gaurav Bhalotia, and myself
under the supervision of Prof. Marti Hearst.
Rowena Luk and Archana Ganapathi contributed to the design.} and its query language,
the {\it Layered Query Language} ({\it LQL}), can be found in
\cite{nakov:2005:lql:biolink,nakov:2005:lql:layers}.
A detailed description of the language syntax, a broader example-based introduction
and an online demo can be found online at \texttt{http://biotext.berkeley.edu/lql/}.

Using the LQL query language, I extracted all sequences
of three nouns falling in the last three positions of a noun phrase (NP)
found in the shallow parse. If the NP contained other nouns, the sequence was discarded.
This allows for noun compounds that are modified by adjectives, determiners, and so on,
but prevents extracting three-word noun compounds that are part of longer noun compounds.
Details follow below.

In order to simplify the explanations, I start with a simple LQL query
and I refine it in two additional steps in order to obtain the final version that I use
to extract the three-word noun compounds for the {\it Biomedical dataset}.
Consider the following query:

The new query contains an assertion that a layer does not exist, via
the negation operator (``\verb\!\''). The outer brackets
indicate that gaps (intermediate words) are allowed between any
layers, thus disallowing the non-existing layer anywhere before the
sequence of three nouns (not just when it immediately precedes the
first noun). One more pair of brackets counteracts
\texttt{ALLOW GAPS} and keeps the three internal nouns adjacent. 

\vspace{6pt}

\noindent {\bf \underline{Query (1)}:}
\vspace{6pt}

\begin{scriptsize}
\begin{verbatim}
FROM
   [layer='shallow_parse' && tag_name='NP'
     ^ [layer='pos' && tag_name="noun"] AS n1
       [layer='pos' && tag_name="noun"] AS n2
       [layer='pos' && tag_name="noun"] AS n3 $
   ]
SELECT n1.content, n2.content, n3.content
\end{verbatim}
\end{scriptsize}
\vspace{6pt}

The above LQL query looks for a noun phrase from the shallow-parse layer,
containing exactly three nouns from the POS layer. Each layer in an
LQL query is enclosed in brackets and is optionally followed by a
binding statement. Layers have names, e.g.,
\texttt{pos} and \texttt{shallow\_parse}, which determine a possible set of
types such as \texttt{NP} for \texttt{shallow\_parse} and \texttt{NN}
for \texttt{pos}. There are macros for groups of tags, such as
\texttt{noun}, which refers to all parts of speech which are
nouns in the Penn Treebank. Single quotes are used for exact
matches, and double quotes for case insensitive matches and macros.

Enclosure of one layer within another indicates that the outer spans
the inner. By default, layers' contents are adjacent, but the keywords
\texttt{ALLOW GAPS} can change this default.
LQL uses the UNIX-style delimiters \texttt{\^} and \texttt{\$}
in order to constrain the
results so that no other words can precede or follow the three nouns
within the NP. Finally, the \texttt{select}
statement specifies that the contents of the {\it compound}
(i.e. the text span of the compound) are to be returned.

{\it Query (1)} is case sensitive. While this is useful in the
biomedical domain, where the capitalization can distinguish between
different genes or proteins, to get frequency counts we want
normalization. {\it Query (2)} below converts the results to lowercase by
using the corresponding SQL function.

{\it Query (2)} encloses the LQL inside an SQL block and then uses SQL
aggregation functions to produce a sorted list (in descending order)
of the noun compounds and their corresponding frequencies.
Note the two \texttt{SELECT} statements -- the first one belongs to SQL,
and the second one to LQL.
This query will extract three-word noun compounds, provided that the POS tagging and the
shallow-parse annotations are correct. However, since it
requires that the NP consist of exactly three nouns (without preceding
adjectives, determiners etc.), it will miss some three-word noun compounds that are
preceded by modifiers and determiners (such as {\it the}, {\it a},
{\it this}).

\vspace{12pt}
\noindent {\bf \underline{Query (2)}:}
\vspace{6pt}

\begin{scriptsize}
\begin{verbatim}
SELECT
    LOWER(n1.content) lc1,
    LOWER(n2.content) lc2,
    LOWER(n3.content) lc3,
    COUNT(*) AS freq
FROM
  BEGIN_LQL
     FROM
        [layer='shallow_parse' && tag_name='NP'
          ^ [layer='pos' && tag_name="noun"] AS n1
            [layer='pos' && tag_name="noun"] AS n2
            [layer='pos' && tag_name="noun"] AS n3 $
        ]
     SELECT n1.content, n2.content, n3.content
  END_LQL
GROUP BY lc1, lc2, lc3
ORDER BY freq DESC
\end{verbatim}
\end{scriptsize}

{\it Query (3)} below asserts that the nouns should occupy the three last
positions in the NP and disallows other nouns within the same NP,
but allows for other parts of speech.  Note the trade-off between query complexity
and amount of control afforded by the language.

This new query contains an assertion that a layer does not exist, via
the negation operator (``\verb\!\''). The outer brackets
indicate that gaps (intermediate words) are allowed between any
layers, thus disallowing the non-existing layer anywhere before the
sequence of three nouns (not just when it immediately precedes the
first noun). One more pair of brackets counteracts
\texttt{ALLOW GAPS} and keeps the three internal nouns adjacent. 

\vspace{6pt}
\noindent {\bf \underline{Query (3)}:}
\vspace{6pt}

\begin{scriptsize}
\begin{verbatim}
SELECT LOWER(n1.content) lc1,
       LOWER(n2.content) lc2,
       LOWER(n3.content) lc3,
       COUNT(*) AS freq
FROM
  BEGIN_LQL
    FROM
      [layer='shallow_parse' && tag_name='NP'
        ^ ( { ALLOW GAPS }
            ![layer='pos' && tag_name="noun"]
             ( [layer='pos' && tag_name="noun"] AS n1
               [layer='pos' && tag_name="noun"] AS n2
               [layer='pos' && tag_name="noun"] AS n3 ) $
          ) $
      ]
    SELECT n1.content, n2.content, n3.content
  END_LQL
GROUP BY lc1, lc2, lc3
ORDER BY freq DESC
\end{verbatim}
\end{scriptsize}

{\it Query (3)} returns a total of 418,678 distinct noun compounds
with corresponding frequencies.
I manually investigate the most frequent ones, removing those with errors
in tokenization (e.g., containing words like {\it transplan} or {\it
tation}), POS tagging (e.g., {\it acute lung injury}, where {\it
acute} is wrongly tagged as a noun) or shallow parsing (e.g., {\it
situ hybridization}, that misses {\it in}). I had to consider the
first 843 examples in order to obtain 500 good ones, which
suggests an extraction accuracy of 59\%.  This number is low for various reasons.
First, the tokenizer handles dash-connected words as a single token (e.g.,
{\it factor-alpha}). Second, many tokens contain other special
characters (e.g., {\it cd4+}), which I exclude since
they cannot be used in a query against a search engine.
Note that while the extraction process does not allow for exact repetitions,
some examples are similar due to various kinds of variability:

\begin{itemize}
  \item {\bf spelling variation}
      \begin{itemize}
        \item ``{\it iron deficiency \underline{anaemia}}'' vs. ``{\it iron deficiency \underline{anemia}}'';
        \item ``{\it motor \underline{neuron} disease}'' vs. ``{\it motor \underline{neurone} disease}'', etc.
      \end{itemize}
  \item {\bf inflectional variation}
      \begin{itemize}
        \item ``{\it blood flow \underline{velocity}}'' vs. ``{\it blood flow \underline{velocities}}'';
        \item ``{\it group b \underline{streptococcus}}'' vs. ``{\it group b \underline{streptococci}}'', etc.
      \end{itemize}
  \item {\bf derivational variation}
      \begin{itemize}
        \item ``{\it bone marrow \underline{transplantation}}'' vs. ``{\it bone marrow \underline{transplants}}'';
        \item ``{\it dna strand \underline{breakage}}'' vs. ``{\it dna strand \underline{breaks}}'', etc.
      \end{itemize}
  \item {\bf disease/gene/protein/chemical/etc. name variation}
      \begin{itemize}
        \item ``{\it hepatitis \underline{c} virus}'' vs. ``{\it hepatitis \underline{e} virus}'';
        \item ``{\it rna polymerase \underline{ii}}'' vs. ``{\it rna polymerase \underline{iii}}'';
      \end{itemize}
  \item {\bf hyponymy}
      \begin{itemize}
        \item ``{\it colon \underline{cancer} cells}'' vs. ``{\it colon \underline{carcinoma} cells}'', etc.
      \end{itemize}
\end{itemize}

\section{Extracting Statistics from MEDLINE instead of the Web}
\label{app:dataset:bio:lql}

Here I describe the process of acquisition of $n$-gram and paraphrase frequency statistics
for the {\it Biomedical dataset} from 1.4 million MEDLINE abstracts with suitable linguistic annotations
using LQL.

\subsection{Extracting $n$-gram Frequencies}

For a three-word noun compound $w_1w_2w_3$, the {\it adjacency} and
the {\it dependency} models need the corpus frequencies for $w_1w_2$,
$w_1w_3$ and $w_2w_3$. For the purpose, I use a case insensitive LQL query that
allows for inflections of the second word, which respects the sentence boundaries
and requires both words to be nouns and to be the last two words in a noun phrase.
The LQL query below shows an example instantiation for the bigram ``{\it immunodeficiency virus}'':

\vspace{12pt}
\begin{scriptsize}
\begin{verbatim}
SELECT
    COUNT(*) AS freq
FROM
  BEGIN_LQL
    FROM
      [layer='shallow_parse' && tag_name='NP'
        [layer='pos' && tag_name="noun"
           && content="immunodeficiency"] AS word1
        [layer='pos' && tag_name="noun"
           && (content="virus" || content="viruses")] $
      ]
    SELECT word1.content, word2.content
  END_LQL
\end{verbatim}
\end{scriptsize}

The query for the word frequencies is much simpler.
It only requires the word to be a noun, allowing for inflections.
Only the total number of occurrences is selected.
The LQL query below shows an example instantiation for the word ``{\it human}'':

\vspace{12pt}
\begin{scriptsize}
\begin{verbatim}
SELECT COUNT(*) AS freq
FROM
  BEGIN_LQL
    FROM
      [layer='pos' && tag_name="noun" &&
        content IN ("human", "humans")] AS word
    SELECT word.content
  END_LQL
\end{verbatim}
\end{scriptsize}

\subsection{Extracting Paraphrase Frequencies}

%
%
%

As I have already explained in section \ref{sec:paraphrases},
paraphrases are an important feature type for noun compound bracketing.
The example LQL query below follows the general pattern
``{\it $[_{\mathrm{NP}}$ ... $w_2$ $w_3$$]$ PREP $[_{\mathrm{NP}}$ ... $w_1$$]$}'',
looking for right-predicting {\bf prepositional paraphrases}
for the noun compound {\it human immunodeficiency virus}:

%
%

\vspace{12pt}
\begin{scriptsize}
\begin{verbatim}
SELECT LOWER(prep.content) lp,
       COUNT(*) AS freq
FROM
  BEGIN_LQL
    FROM
      [layer='sentence'
        [layer='shallow_parse' && tag_name='NP'
           [layer='pos' && tag_name="noun" &&
               content = "immunodeficiency"]
           [layer='pos' && tag_name="noun" &&
               content IN ("virus", "viruses")] $
        ]
        [layer='pos' && tag_name='IN'] AS prep
        [layer='shallow_parse' && tag_name='NP'
           [layer='pos' && tag_name="noun" &&
               content IN ("human", "humans") ] $
        ]
      ]
    SELECT prep.content
  END_LQL
GROUP BY lp
ORDER BY freq DESC
\end{verbatim}
\end{scriptsize}
\vspace{12pt}

The query above requires that the three words be nouns
and allows for inflections of $w_1$ and $w_3$,
but not of $w_2$ since it modifies $w_3$ in the paraphrase.
It also required $w_1$ and ``$w_2$ $w_3$'' be the last words
in their corresponding NPs, and accepts any preposition
between these NPs.

{\bf Verbal paraphrases} and {\bf copula paraphrases} can be handled
in a similar manner, but have been excluded from my experiments
since they are very infrequent because of the relatively small size of my subset of MEDLINE.
As already mentioned in section \ref{sec:dataset:bio},
my corpus consists of about 1.4 million MEDLINE abstracts, each one being about 300 words long
on the average, which means about 420 million indexed words in total.
For comparison, {\it Google} indexes about eight billion pages;
assuming each one contains 500 words on the average,
this means about four trillion indexed words, which is about a million times
bigger than my corpus.
Since the {\it prepositional} paraphrases are much more frequent
and since I add up the total number of left-predicting vs. right-predicting
paraphrase occurrences, the {\it verbal} and the {\it copula} paraphrases
are unlikely to change the overall bracketing prediction of the paraphrase model:
the total sums would still be dominated by the {\it prepositional} paraphrases.
In addition, in the Web experiments above,
the {\it verbal} paraphrases used a limited set of verbs which act as prepositions
(see section \ref{sec:paraphrases} for a full list and for additional details).
Since the LQL query above allows for any preposition,
it would return some of these verbal paraphrases anyway.
Therefore, I chose not to make use any {\it copula} or {\it verbal} paraphrases,
thus limiting myself to {\it prepositional} paraphrases only
which makes my experiments more directly comparable to the above described Web-based
ones.
\vspace{12pt}

\newpage
\section{The Annotated \emph{Biomedical Dataset}}
\label{appendix:bio:set}

Below I list the 430 test examples (sorted by decreasing frequency)
from the {\it Biomedical dataset} for noun compound bracketing,
described in section \ref{sec:dataset:bio}.

\vspace{18pt}

\begin{center}
\begin{small}
\begin{tabular}{|lll|c|r|}
  \multicolumn{3}{c}{\bf Noun Compound} & \multicolumn{1}{c}{\bf Bracketing} & \multicolumn{1}{c}{\bf Freq.} \\
  \hline
    polymerase & chain & reaction & right & 5208 \\
    body & mass & index & left & 2096 \\
    nerve & growth & factor & right & 1918 \\
    tumor & necrosis & factor & right & 1888 \\
    positron & emission & tomography & left & 1447 \\
    protein & kinase & c & left & 1399 \\
    transmission & electron & microscopy & left & 1331 \\
    bone & mineral & density & right & 1257 \\
    hepatitis & c & virus & left & 1211 \\
    bone & marrow & transplantation & left & 979 \\
    bone & marrow & cells & left & 918 \\
    t & cell & activation & left & 912 \\
    heart & rate & variability & left & 845 \\
    health & care & reform & left & 842 \\
    lymph & node & metastasis & left & 755 \\
    cell & cycle & progression & left & 749 \\
    calcium & channel & blockers & right & 739 \\
    health & care & providers & left & 700 \\
    lymph & node & metastases & left & 670 \\
    health & care & workers & left & 634 \\
    dna & sequence & analysis & left & 615 \\
    phorbol & myristate & acetate & left & 615 \\
    growth & factor & beta & left & 613 \\
    rna & polymerase & ii & left & 608 \\
    stem & cell & factor & left & 589 \\
    male & wistar & rats & right & 570 \\
    hepatitis & b & virus & left & 560 \\
    type & iv & collagen & left & 560 \\
    pneumocystis & carinii & pneumonia & left & 546 \\
    tissue & plasminogen & activator & right & 515 \\
    tumor & suppressor & genes & left & 510 \\
    breast & cancer & cells & left & 503 \\
    breast & cancer & patients & left & 486 \\
    dna & strand & breaks & right & 481 \\
    health & care & professionals & left & 456 \\
  \hline
\end{tabular}
\end{small}
\end{center}

\begin{center}
\begin{small}
\begin{tabular}{|lll|c|r|}
  \multicolumn{3}{c}{\bf Noun Compound} & \multicolumn{1}{c}{\bf Bracketing} & \multicolumn{1}{c}{\bf Freq.} \\
  \hline
    signal & transduction & pathways & left & 450 \\
    hepatocyte & growth & factor & right & 450 \\
    t & cell & proliferation & left & 444 \\
    protein & tyrosine & kinases & right & 410 \\
    class & ii & molecules & left & 397 \\
    xenopus & laevis & oocytes & left & 394 \\
    body & weight & gain & left & 392 \\
\hline
\end{tabular}
\end{small}
\end{center}

\begin{center}
\begin{small}
\begin{tabular}{|lll|c|r|}
  \multicolumn{3}{c}{\bf Noun Compound} & \multicolumn{1}{c}{\bf Bracketing} & \multicolumn{1}{c}{\bf Freq.} \\
  \hline
    amino & acid & substitutions & left & 259 \\
    skin & blood & flow & right & 259 \\
    borderline & personality & disorder & right & 246 \\
    receptor & tyrosine & kinases & right & 246 \\
    growth & factor & receptors & left & 243 \\
    cell & cycle & control & left & 229 \\
    wheat & germ & agglutinin & left & 229 \\
\hline
\end{tabular}
\end{small}
\end{center}

\begin{center}
\begin{small}
\begin{tabular}{|lll|c|r|}
  \multicolumn{3}{c}{\bf Noun Compound} & \multicolumn{1}{c}{\bf Bracketing} & \multicolumn{1}{c}{\bf Freq.} \\
  \hline
    monte & carlo & simulations & left & 181\\
    b & cell & development & left & 179\\
    dna & adduct & formation & left & 178\\
    primer & extension & analysis & left & 178\\
    skin & prick & tests & left & 177\\
    t & cell & recognition & left & 177\\
    gel & permeation & chromatography & left & 173\\
\hline
\end{tabular}
\end{small}
\end{center}

\begin{center}
\begin{small}
\begin{tabular}{|lll|c|r|}
  \multicolumn{3}{c}{\bf Noun Compound} & \multicolumn{1}{c}{\bf Bracketing} & \multicolumn{1}{c}{\bf Freq.} \\
  \hline
    tumor & cell & growth & left & 151\\
    type & ii & pneumocytes & left & 151\\
    guinea & pig & ileum & left & 150\\
    erythrocyte & sedimentation & rate & left & 150\\
    blood & pressure & variability & left & 148\\
    rna & blot & analysis & left & 148\\
    anion & exchange & chromatography & left & 147\\
\hline
\end{tabular}
\end{small}
\end{center}

\begin{center}
\begin{small}
\begin{tabular}{|lll|c|r|}
  \multicolumn{3}{c}{\bf Noun Compound} & \multicolumn{1}{c}{\bf Bracketing} & \multicolumn{1}{c}{\bf Freq.} \\
  \hline
    health & maintenance & organizations & left & 125\\
    mast & cell & activation & left & 125\\
    restriction & enzyme & digestion & left & 125\\
    t & cell & function & left & 125\\
    chlamydia & trachomatis & infection & left & 124\\
    health & services & research & left & 123\\
    world & health & organization & right & 122\\
\hline
\end{tabular}
\end{small}
\end{center}

\begin{center}
\begin{small}
\begin{tabular}{|lll|c|r|}
  \multicolumn{3}{c}{\bf Noun Compound} & \multicolumn{1}{c}{\bf Bracketing} & \multicolumn{1}{c}{\bf Freq.} \\
  \hline
    b & cell & lines & left & 108\\
    blood & pressure & changes & left & 108\\
    contact & lens & wear & left & 108\\
    restriction & enzyme & analysis & left & 108\\
    somatostatin & receptor & scintigraphy & left & 108\\
    bone & marrow & aspirates & left & 107\\
    heparan & sulfate & proteoglycan & left & 107\\
\hline
\end{tabular}
\end{small}
\end{center}

\begin{center}
\begin{small}
\begin{tabular}{|lll|c|r|}
  \multicolumn{3}{c}{\bf Noun Compound} & \multicolumn{1}{c}{\bf Bracketing} & \multicolumn{1}{c}{\bf Freq.} \\
  \hline
    bone & marrow & samples & left & 100\\
    heart & transplant & patients & left & 100\\
    sister & chromatid & exchange & left & 100\\
    substance & abuse & treatment & left & 100\\
    type & iii & procollagen & left & 100\\
    rat & mast & cells & right & 100\\
    life & table & analysis & left & 99\\
\hline
\end{tabular}
\end{small}
\end{center}

\begin{center}
\begin{small}
\begin{tabular}{|lll|c|r|}
  \multicolumn{3}{c}{\bf Noun Compound} & \multicolumn{1}{c}{\bf Bracketing} & \multicolumn{1}{c}{\bf Freq.} \\
  \hline
    transient & transfection & experiments & left & 92\\
    colon & carcinoma & cells & left & 91\\
    t & cell & receptors & left & 91\\
    blood & pressure & values & left & 90\\
    bone & marrow & biopsy & left & 90\\
    g & protein & activation & left & 90\\
    neutron & activation & analysis & left & 90\\
\hline
\end{tabular}
\end{small}
\end{center}

\begin{center}
\begin{small}
\begin{tabular}{|lll|c|r|}
  \multicolumn{3}{c}{\bf Noun Compound} & \multicolumn{1}{c}{\bf Bracketing} & \multicolumn{1}{c}{\bf Freq.} \\
  \hline
    borna & disease & virus & left & 81\\
    carbon & dioxide & production & left & 81\\
    dinucleotide & repeat & polymorphism & left & 81\\
    messenger & rna & expression & left & 81\\
    nadph & diaphorase & activity & left & 81\\
    nmda & receptor & function & left & 81\\
    rabbit & reticulocyte & lysate & left & 81\\
\hline
\end{tabular}
\end{small}
\end{center}

\begin{center}
\begin{small}
\begin{tabular}{|lll|c|r|}
  \multicolumn{3}{c}{\bf Noun Compound} & \multicolumn{1}{c}{\bf Bracketing} & \multicolumn{1}{c}{\bf Freq.} \\
  \hline
    phase & contrast & microscopy & left & 74\\
    tyrosine & kinase & activation & left & 74\\
    gc & b & cells & right & 74\\
    memory & b & cells & right & 74\\
    glutamate & receptor & antagonists & left & 73\\
    motor & nerve & terminals & left & 73\\
    t & cell & help & left & 73\\
    time & series & analysis & left & 73\\
    adult & wistar & rats & right & 73\\
    color & doppler & flow & right & 73\\
    protein & tyrosine & phosphatase & right & 73\\
    dna & strand & breakage & left & 72\\
    ground & reaction & forces & left & 72\\
    kidney & transplant & patients & left & 72\\
    integration & host & factor & right & 72\\
    murine & t & cells & right & 72\\
  \hline
\end{tabular}
\end{small}
\end{center}

\chapter{Noun Phrase Coordination Dataset}
\label{appendix:coord:set}

Below I list the 428 test examples from the {\it Noun Phrase Coordination Dataset},
which were used in the experiments described in section \ref{sec:coord}.

\vspace{6pt}
\begin{center}
\begin{small}
\begin{tabular}{|llll|c|r|}
  \multicolumn{4}{c}{\bf Coordinated Noun Phrase} & \multicolumn{1}{c}{\bf Coord. Type} \\
  \hline
    cotton & and & acetate & fibers & noun\\
    parts & and & marketing & operations & noun\\
    sales & and & marketing & executive & noun\\
    royalty & or & rock & stars & NP\\
    Trade & and & Industry & Ministry & noun\\
    Newsweek & and & U.S. & News & NP\\
    chairman & and & chief & designer & NP\\
    Securities & and & Exchange & Commission & noun\\
    Taiwan & and & Saudi & Arabia & NP\\
    investigations & and & trade & sanctions & NP\\
    movie & and & book & pirates & noun\\
    patent & and & copyright & owners & noun\\
    Taiwan & and & Saudi & Arabia & NP\\
    Washington & and & New & York & NP\\
    Perch & and & Dolphin & fields & noun\\
    Seahorse & and & Tarwhine & fields & noun\\
    U.S. & and & Virgin & Islands & NP\\
    capacity & and & debt & load & NP\\
    president & and & chief & executive & NP\\
    France & and & Hong & Kong & NP\\
    U.S. & and & Soviet & Union & noun\\
  \hline
\end{tabular}
\end{small}
\end{center}

\begin{center}
\begin{small}
\begin{tabular}{|llll|c|r|}
  \multicolumn{4}{c}{\bf Coordinated Noun Phrase} & \multicolumn{1}{c}{\bf Coord. Type} \\
  \hline
    milk & and & milk & powder & NP\\
    practice & and & team & image & NP\\
    business & and & computer & science & NP\\
    analysts & and & business & people & NP\\
    Taiwan & and & South & Korea & NP\\
    business & and & research & program & noun\\
    Public & and & Internatonal & Affairs & noun\\
\hline
\end{tabular}
\end{small}
\end{center}

\begin{center}
\begin{small}
\begin{tabular}{|llll|c|r|}
  \multicolumn{4}{c}{\bf Coordinated Noun Phrase} & \multicolumn{1}{c}{\bf Coord. Type} \\
  \hline
    Triton & and & Mr. & Chase & NP\\
    firm & and & Mr. & Whelen & NP\\
    pulp & and & paper & segment & noun\\
    credit & and & loan & guarantees & noun\\
    cash & and & development & assistance & noun\\
    Justice & and & Commerce & departments & noun\\
    health & and & safety & violations & noun\\
\hline
\end{tabular}
\end{small}
\end{center}

\begin{center}
\begin{small}
\begin{tabular}{|llll|c|r|}
  \multicolumn{4}{c}{\bf Coordinated Noun Phrase} & \multicolumn{1}{c}{\bf Coord. Type} \\
  \hline
    California & and & New & York & NP\\
    oil & and & mining & shares & noun\\
    Securities & and & Exchange & Commission & noun\\
    felony & and & misdemeanor & counts & noun\\
    Philadelphia & and & Cleveland & districts & noun\\
    estate & and & mortgage & loans & noun\\
    Monopolies & and & Mergers & Commission & noun\\
\hline
\end{tabular}
\end{small}
\end{center}

\begin{center}
\begin{small}
\begin{tabular}{|llll|c|r|}
  \multicolumn{4}{c}{\bf Coordinated Noun Phrase} & \multicolumn{1}{c}{\bf Coord. Type} \\
  \hline
    servants & or & government & functionaries & NP\\
    chemical & and & biotechnology & companies & noun\\
    tennis & and & golf & tours & noun\\
    accountants & and & estate & developers & NP\\
    diesel & and & gas & turbines & noun\\
    president & and & executive & officer & NP\\
    economy & and & currency & issues & NP\\
\hline
\end{tabular}
\end{small}
\end{center}

\begin{center}
\begin{small}
\begin{tabular}{|llll|c|r|}
  \multicolumn{4}{c}{\bf Coordinated Noun Phrase} & \multicolumn{1}{c}{\bf Coord. Type} \\
  \hline
    pulp & and & newsprint & mill & noun\\
    Oregon & and & South & Carolina & NP\\
    Senate & and & House & bills & noun\\
    treatment & and & tax & relief & NP\\
    chairman & and & executive & officer & NP\\
    Hyundai & and & J.P. & Morgan & NP\\
    delver & and & detail & guy & NP\\
\hline
\end{tabular}
\end{small}
\end{center}

\begin{center}
\begin{small}
\begin{tabular}{|llll|c|r|}
  \multicolumn{4}{c}{\bf Coordinated Noun Phrase} & \multicolumn{1}{c}{\bf Coord. Type} \\
  \hline
    oil & and & gas & properties & noun\\
    oil & and & gas & acquisitions & noun\\
    oil & and & gas & revenue & noun\\
    oil & and & gas & properties & noun\\
    pipeline & and & oil & assets & noun\\
    mining & and & energy & assets & noun\\
    clothing & and & food & retailer & noun\\
\hline
\end{tabular}
\end{small}
\end{center}

\begin{center}
\begin{small}
\begin{tabular}{|llll|c|r|}
  \multicolumn{4}{c}{\bf Coordinated Noun Phrase} & \multicolumn{1}{c}{\bf Coord. Type} \\
  \hline
    business & and & labor & leaders & noun\\
    aid & and & food & grants & NP\\
    Warner & and & Mr. & Azoff & NP\\
    compensation & and & business & plans & NP\\
    Ed & and & Lorraine & Warren & noun\\
    crane & and & bulldozer & operators & noun\\
    road & and & bridge & design & noun\\
\hline
\end{tabular}
\end{small}
\end{center}

\begin{center}
\begin{small}
\begin{tabular}{|llll|c|r|}
  \multicolumn{4}{c}{\bf Coordinated Noun Phrase} & \multicolumn{1}{c}{\bf Coord. Type} \\
  \hline
    terms & and & syndicate & manager & NP\\
    Remics & and & mortgage & securities & NP\\
    Paribas & and & Navigation & Mixte & NP\\
    telecommunications & and & defense & equipment & noun\\
    wire & and & cable & businesses & noun\\
    oil & and & gas & acquisition & noun\\
    exploration & and & production & unit & noun\\
\hline
\end{tabular}
\end{small}
\end{center}

\begin{center}
\begin{small}
\begin{tabular}{|llll|c|r|}
  \multicolumn{4}{c}{\bf Coordinated Noun Phrase} & \multicolumn{1}{c}{\bf Coord. Type} \\
  \hline
    polyester & and & rayon & markets & noun\\
    policy & and & exchange & rates & NP\\
    viability & and & ownership & changes & NP\\
    volume & and & profit & impact & noun\\
    food & and & tobacco & giant & noun\\
    steam & and & combustion & turbines & noun\\
    Friday & and & Saturday & nights & noun\\
\hline
\end{tabular}
\end{small}
\end{center}

\begin{center}
\begin{small}
\begin{tabular}{|llll|c|r|}
  \multicolumn{4}{c}{\bf Coordinated Noun Phrase} & \multicolumn{1}{c}{\bf Coord. Type} \\
  \hline
    competition & and & client & needs & NP\\
    Portugal & and & Puerto & Rico & NP\\
    sales & and & dealer & orders & NP\\
    tax & and & acccounting & practices & noun\\
    wildlife & and & fishing & industry & NP\\
    debt & and & debt & service & NP\\
    production & and & pricing & differences & noun\\
    hurricane & and & earthquake & damage & noun\\
    Securities & and & Exchange & Board & noun\\
    bonds & or & equity & shares & NP\\
    petrochemical & and & agrochemical & company & noun\\
    regulation & and & disclosure & requirements & noun\\
    Securities & and & Exchange & Board & noun\\
    media & and & resources & company & noun\\
    Bond & and & Bell & Resources & noun\\
    government & and & Jaguar & holders & NP\\
    manufacturing & and & marketing & ventures & noun\\
    banks & and & securities & companies & NP\\
    beer & and & food & concern & noun\\
    West & and & South & Carolina & NP\\
    Austin & and & Fort & Worth & NP\\
    Merger & and & acquisition & activity & noun\\
    adviser & and & investment & banker & NP\\
    film & and & television & operations & noun\\
    cosmetology & and & business & schools & noun\\
    Patterson & or & Sugarman & groups & noun\\
    spending & and & capital & investment & NP\\
  \hline
\end{tabular}
\end{small}
\end{center}

\chapter{Comparing Human- and Web-Generated Paraphrasing Verbs}
\label{chap:mturkompare:human}

Below I list the Web- and the human-derived paraphrasing verbs
for 250 noun-noun compounds from \namecite{levi:1978}.
The process of extraction is described in sections
\ref{sec:that:verbs:method} and \ref{sec:sem:human:judgments}, respectively.
For each noun-noun compound, I show the cosine correlation,
the semantic class\footnote{One of the recoverably deletable predicates proposed by \namecite{levi:1978},
as described in section \ref{nc:ling:theory:levi}.},
and the paraphrasing verbs of each type,
each followed by the corresponding frequency; the overlapping verbs are underlined.
I show the results (top 10 verbs) by noun-noun compound when all human-proposed verbs are used and
when only the first verb proposed by each worker is used
in sections \ref{sec:verbs:all} and \ref{sec:verbs:first}, respectively.
I further show a comparison aggregated by semantic class (top 150 verbs)
in sections \ref{sec:verbs:all:class} and \ref{sec:verbs:first:class}, respectively.

\section{Comparison by Noun Compound: Using All Human-Proposed Verbs}
\label{sec:verbs:all}
\vspace{6pt}

\begin{small}\begin{center}
\begin{tabular}{p{15cm}}
{\bf 0.96 ``blood donor'' NOMINALIZATION:AGENT}\\
{\scriptsize \textbf{MTurk:} \underline{give(30)}, \underline{donate(16)}, \underline{supply(8)}, \underline{provide(6)}, \underline{share(2)}, \underline{contribute(1)}, \underline{volunteer(1)}, \underline{offer(1)}, choose(1), hand over(1), $\ldots$}\\
{\scriptsize \textbf{Web:} \underline{give(653)}, \underline{donate(395)}, receive(74), sell(41), \underline{provide(39)}, \underline{supply(17)}, be(13), match(11), \underline{contribute(10)}, \underline{offer(9)}, $\ldots$}\\
\hline
\end{tabular}\end{center}\end{small}
\begin{small}\begin{center}\begin{tabular}{p{15cm}}
{\bf 0.95 ``women professors'' BE}\\
{\scriptsize \textbf{MTurk:} \underline{be(22)}, \underline{teach(2)}, look like(2), be born(2), \underline{research(1)}, \underline{study(1)}, be comprised of(1), behave like(1), include(1), be gendered(1), $\ldots$}\\
{\scriptsize \textbf{Web:} \underline{be(251)}, \underline{teach(46)}, \underline{study(38)}, specialize in(26), appear as(13), think(10), take(10), \underline{research(9)}, work with(9), think that(9), $\ldots$}\\
\hline
\end{tabular}\end{center}\end{small}
\begin{small}\begin{center}\begin{tabular}{p{15cm}}
{\bf 0.94 ``student friends'' BE}\\
{\scriptsize \textbf{MTurk:} \underline{be(18)}, \underline{come from(2)}, \underline{help(2)}, be had by(2), be made by(2), \underline{include(1)}, \underline{involve(1)}, \underline{act like(1)}, \underline{live as(1)}, \underline{work as(1)}, $\ldots$}\\
{\scriptsize \textbf{Web:} \underline{be(1250)}, have(34), \underline{support(33)}, know(31), teach(22), give(21), \underline{meet as(19)}, \underline{help(16)}, guide(15), pose as(12), $\ldots$}\\
\hline
\end{tabular}\end{center}\end{small}
\begin{small}\begin{center}\begin{tabular}{p{15cm}}
{\bf 0.93 ``city wall'' HAVE$_2$}\\
{\scriptsize \textbf{MTurk:} \underline{surround(24)}, \underline{protect(10)}, \underline{enclose(8)}, \underline{encircle(7)}, \underline{encompass(3)}, be in(3), \underline{contain(2)}, \underline{snake around(1)}, \underline{border(1)}, \underline{go around(1)}, $\ldots$}\\
{\scriptsize \textbf{Web:} \underline{surround(708)}, \underline{encircle(203)}, \underline{protect(191)}, \underline{divide(176)}, \underline{enclose(72)}, \underline{separate(49)}, ring(41), be(34), \underline{encompass(25)}, \underline{defend(25)}, $\ldots$}\\
\hline
\end{tabular}\end{center}\end{small}
\begin{small}\begin{center}\begin{tabular}{p{15cm}}
{\bf 0.91 ``citizen soldier'' BE}\\
{\scriptsize \textbf{MTurk:} \underline{be(20)}, represent(3), come from(3), act as(2), fight for(2), \underline{protect(1)}, \underline{defend(1)}, care for(1), start out as(1), guard(1), $\ldots$}\\
{\scriptsize \textbf{Web:} \underline{be(185)}, become(44), be treated as(14), save(14), dehumanize(11), shoot(10), rob(6), murder(5), fire upon(5), be drafted from(5), $\ldots$}\\
\hline
\end{tabular}\end{center}\end{small}
\begin{small}\begin{center}\begin{tabular}{p{15cm}}
{\bf 0.91 ``accident weather'' CAUSE$_1$}\\
{\scriptsize \textbf{MTurk:} \underline{cause(21)}, promote(4), lead to(4), \underline{provoke(2)}, create(2), occur by(2), result in(2), contribute to(2), be condusive to(1), occasion(1), $\ldots$}\\
{\scriptsize \textbf{Web:} \underline{cause(25)}, \underline{provoke(3)}, be(3), include(2), end in(2), exist during(2), be proven with(1).}\\
\hline
\end{tabular}\end{center}\end{small}
\begin{small}\begin{center}\begin{tabular}{p{15cm}}
{\bf 0.91 ``disease germ'' CAUSE$_1$}\\
{\scriptsize \textbf{MTurk:} \underline{cause(20)}, \underline{spread(5)}, \underline{carry(4)}, \underline{create(4)}, \underline{produce(3)}, generate(3), \underline{start(2)}, promote(2), lead to(2), result in(2), $\ldots$}\\
{\scriptsize \textbf{Web:} \underline{cause(919)}, \underline{produce(63)}, \underline{spread(37)}, \underline{carry(20)}, \underline{propagate(9)}, \underline{create(7)}, transmit(7), be(7), bring(5), give(4), $\ldots$}\\
\hline
\end{tabular}\end{center}\end{small}
\begin{small}\begin{center}\begin{tabular}{p{15cm}}
{\bf 0.89 ``flu virus'' CAUSE$_1$}\\
{\scriptsize \textbf{MTurk:} \underline{cause(19)}, \underline{spread(4)}, \underline{give(4)}, \underline{result in(3)}, create(3), infect with(3), contain(3), \underline{be(2)}, \underline{carry(2)}, \underline{induce(1)}, $\ldots$}\\
{\scriptsize \textbf{Web:} \underline{cause(906)}, \underline{produce(21)}, \underline{give(20)}, differentiate(17), \underline{be(16)}, have(13), include(11), \underline{spread(7)}, mimic(7), trigger(6), $\ldots$}\\
\hline
\end{tabular}\end{center}\end{small}
\begin{small}\begin{center}\begin{tabular}{p{15cm}}
{\bf 0.89 ``collie dog'' BE}\\
{\scriptsize \textbf{MTurk:} \underline{be(12)}, \underline{look like(8)}, \underline{resemble(2)}, \underline{come from(2)}, belong to(2), be related to(2), be called(2), be classified as(2), be made from(1), be named(1), $\ldots$}\\
{\scriptsize \textbf{Web:} \underline{be(24)}, \underline{look like(14)}, \underline{resemble(8)}, be border(5), feature(3), \underline{come from(2)}, tend(2), be bearded(1), include(1), betoken(1), $\ldots$}\\
\hline
\end{tabular}\end{center}\end{small}
\begin{small}\begin{center}\begin{tabular}{p{15cm}}
{\bf 0.89 ``gas stove'' USE}\\
{\scriptsize \textbf{MTurk:} \underline{use(20)}, \underline{run on(9)}, \underline{burn(8)}, \underline{cook with(6)}, utilize(4), \underline{emit(3)}, \underline{be heated by(2)}, \underline{need(2)}, \underline{consume(2)}, \underline{work with(2)}, $\ldots$}\\
{\scriptsize \textbf{Web:} \underline{use(98)}, \underline{run on(36)}, \underline{burn(33)}, be(25), \underline{be heated by(10)}, \underline{work with(7)}, be used with(7), \underline{leak(6)}, \underline{need(6)}, \underline{consume(6)}, $\ldots$}\\
\hline
\end{tabular}\end{center}\end{small}
\begin{small}\begin{center}\begin{tabular}{p{15cm}}
{\bf 0.87 ``music box'' MAKE$_1$}\\
{\scriptsize \textbf{MTurk:} \underline{play(19)}, \underline{make(12)}, \underline{produce(10)}, emit(5), \underline{create(4)}, \underline{contain(4)}, \underline{provide(2)}, \underline{generate(2)}, give off(2), \underline{include(1)}, $\ldots$}\\
{\scriptsize \textbf{Web:} \underline{play(104)}, \underline{make(34)}, \underline{produce(18)}, \underline{have(16)}, \underline{provide(14)}, be(13), \underline{contain(9)}, access(8), say(7), \underline{store(6)}, $\ldots$}\\
\hline
\end{tabular}\end{center}\end{small}
\begin{small}\begin{center}\begin{tabular}{p{15cm}}
{\bf 0.87 ``cooking utensils'' FOR}\\
{\scriptsize \textbf{MTurk:} \underline{be used for(17)}, \underline{be used in(9)}, \underline{facilitate(4)}, help(3), aid(3), \underline{be required for(2)}, \underline{be used during(2)}, be found in(2), be utilized in(2), involve(2), $\ldots$}\\
{\scriptsize \textbf{Web:} \underline{be used for(43)}, \underline{be used in(11)}, make(6), be suited for(5), replace(3), \underline{be used during(2)}, \underline{facilitate(2)}, turn(2), keep(2), \underline{be for(1)}, $\ldots$}\\
\hline
\end{tabular}\end{center}\end{small}
\begin{small}\begin{center}\begin{tabular}{p{15cm}}
{\bf 0.87 ``honey bee'' MAKE$_1$}\\
{\scriptsize \textbf{MTurk:} \underline{make(25)}, \underline{produce(16)}, \underline{create(9)}, \underline{manufacture(5)}, \underline{store(4)}, \underline{eat(2)}, \underline{provide(2)}, secrete(2), \underline{generate(1)}, \underline{gather(1)}, $\ldots$}\\
{\scriptsize \textbf{Web:} \underline{make(292)}, \underline{produce(189)}, \underline{gather(104)}, have(69), collect(57), suck(39), extract(19), drink(16), bring(15), carry(13), $\ldots$}\\
\hline
\end{tabular}\end{center}\end{small}
\begin{small}\begin{center}\begin{tabular}{p{15cm}}
{\bf 0.85 ``song bird'' MAKE$_1$}\\
{\scriptsize \textbf{MTurk:} \underline{sing(22)}, \underline{chirp(4)}, \underline{warble(4)}, \underline{make(4)}, \underline{create(3)}, \underline{produce(3)}, \underline{be known for(2)}, generate(2), \underline{have(1)}, \underline{perform(1)}, $\ldots$}\\
{\scriptsize \textbf{Web:} \underline{sing(264)}, learn(80), \underline{have(69)}, hear(41), be(27), lose(18), continue in(18), know(15), develop(11), give(11), $\ldots$}\\
\hline
\end{tabular}\end{center}\end{small}
\begin{small}\begin{center}\begin{tabular}{p{15cm}}
{\bf 0.85 ``fruit tree'' HAVE$_1$}\\
{\scriptsize \textbf{MTurk:} \underline{bear(20)}, \underline{produce(16)}, \underline{grow(15)}, \underline{have(6)}, \underline{give(4)}, \underline{provide(3)}, \underline{develop(2)}, \underline{supply(2)}, \underline{make(2)}, \underline{hold(1)}, $\ldots$}\\
{\scriptsize \textbf{Web:} \underline{bear(2113)}, \underline{produce(1215)}, \underline{have(337)}, bore(269), yield(201), bring(157), \underline{provide(155)}, bringeth(140), \underline{give(135)}, \underline{grow(74)}, $\ldots$}\\
\hline
\end{tabular}\end{center}\end{small}
\begin{small}\begin{center}\begin{tabular}{p{15cm}}
{\bf 0.83 ``concussion force'' CAUSE$_1$}\\
{\scriptsize \textbf{MTurk:} \underline{cause(23)}, create(9), result in(5), \underline{produce(4)}, \underline{make(3)}, trigger(2), lead to(2), inflict(2), induce(2), form(2), $\ldots$}\\
{\scriptsize \textbf{Web:} \underline{cause(25)}, \underline{produce(15)}, \underline{make(2)}, suffer(2), be termed(1).}\\
\hline
\end{tabular}\end{center}\end{small}
\begin{small}\begin{center}\begin{tabular}{p{15cm}}
{\bf 0.83 ``servant girl'' BE}\\
{\scriptsize \textbf{MTurk:} \underline{work as(17)}, \underline{be(14)}, \underline{act as(6)}, \underline{be employed as(5)}, \underline{serve as(3)}, \underline{be used as(2)}, be paid as(2), act like(2), be considered(2), \underline{have(1)}, $\ldots$}\\
{\scriptsize \textbf{Web:} \underline{be(193)}, \underline{work as(119)}, \underline{become(78)}, live as(20), require(16), live(10), \underline{be employed as(9)}, grow up as(9), go as(9), toil as(8), $\ldots$}\\
\hline
\end{tabular}\end{center}\end{small}
\begin{small}\begin{center}\begin{tabular}{p{15cm}}
{\bf 0.82 ``coke machine'' FOR}\\
{\scriptsize \textbf{MTurk:} \underline{dispense(21)}, \underline{sell(19)}, \underline{contain(9)}, \underline{vend(7)}, distribute(6), hold(4), store(4), \underline{give(3)}, \underline{serve(2)}, \underline{supply(2)}, $\ldots$}\\
{\scriptsize \textbf{Web:} \underline{sell(25)}, \underline{dispense(16)}, \underline{give(9)}, \underline{offer(7)}, draw(7), \underline{supply(6)}, \underline{have(4)}, be camouflaged as(4), \underline{serve(3)}, carry(3), $\ldots$}\\
\hline
\end{tabular}\end{center}\end{small}
\begin{small}\begin{center}\begin{tabular}{p{15cm}}
{\bf 0.80 ``snow blindness'' CAUSE$_2$}\\
{\scriptsize \textbf{MTurk:} \underline{be caused by(24)}, come from(7), result from(7), be created by(3), be derived from(2), emerge from(2), be induced by(2), be related to(2), come about from(1), be spawned by(1), $\ldots$}\\
{\scriptsize \textbf{Web:} \underline{be caused by(2)}, be like(1).}\\
\hline
\end{tabular}\end{center}\end{small}
\begin{small}\begin{center}\begin{tabular}{p{15cm}}
{\bf 0.80 ``desert rat'' IN}\\
{\scriptsize \textbf{MTurk:} \underline{live in(22)}, reside in(10), be found in(6), come from(6), \underline{inhabit(5)}, \underline{survive in(3)}, exist in(3), thrive in(2), feed in(2), dwell in(2), $\ldots$}\\
{\scriptsize \textbf{Web:} \underline{live in(16)}, occur in(4), do(3), \underline{inhabit(2)}, \underline{love(1)}, \underline{survive in(1)}, inhibit(1).}\\
\hline
\end{tabular}\end{center}\end{small}
\begin{small}\begin{center}\begin{tabular}{p{15cm}}
{\bf 0.79 ``deficiency disease'' CAUSE$_2$}\\
{\scriptsize \textbf{MTurk:} \underline{be caused by(18)}, \underline{come from(6)}, \underline{result from(5)}, be due to(3), \underline{be associated with(2)}, \underline{cause(2)}, start with(2), be created by(2), happen because of(2), \underline{stem from(1)}, $\ldots$}\\
{\scriptsize \textbf{Web:} \underline{be caused by(253)}, \underline{result from(203)}, be characterized by(81), \underline{have(69)}, \underline{cause(59)}, \underline{be associated with(56)}, \underline{involve(45)}, \underline{result in(40)}, be(39), \underline{arise from(34)}, $\ldots$}\\
\hline
\end{tabular}\end{center}\end{small}
\begin{small}\begin{center}\begin{tabular}{p{15cm}}
{\bf 0.79 ``mail sorter'' NOMINALIZATION:AGENT}\\
{\scriptsize \textbf{MTurk:} \underline{sort(18)}, organize(9), handle(5), \underline{separate(4)}, work with(4), process(3), classify(3), divide(3), categorize(2), go through(2), $\ldots$}\\
{\scriptsize \textbf{Web:} \underline{sort(25)}, sweep(4), \underline{separate(1)}, deliver(1), result in(1).}\\
\hline
\end{tabular}\end{center}\end{small}
\begin{small}\begin{center}\begin{tabular}{p{15cm}}
{\bf 0.78 ``cigarette burn'' CAUSE$_2$}\\
{\scriptsize \textbf{MTurk:} \underline{be caused by(21)}, \underline{come from(9)}, \underline{be made by(7)}, result from(6), be due to(4), originate from(2), be done with(2), emerge from(2), be left by(2), be created by(2), $\ldots$}\\
{\scriptsize \textbf{Web:} \underline{be caused by(6)}, \underline{come from(5)}, \underline{be from(4)}, consume(2), \underline{be inflicted by(1)}, \underline{be made by(1)}, \underline{resemble(1)}, suggest(1).}\\
\hline
\end{tabular}\end{center}\end{small}
\begin{small}\begin{center}\begin{tabular}{p{15cm}}
{\bf 0.77 ``novelty item'' BE}\\
{\scriptsize \textbf{MTurk:} \underline{be(13)}, be called(4), be classified as(3), \underline{be considered(2)}, represent(2), have(2), resemble(2), be seen as(2), be sold as(2), inspire(2), $\ldots$}\\
{\scriptsize \textbf{Web:} \underline{be(14)}, include(4), \underline{display(3)}, range from(3), possess(2), appear as(2), \underline{be considered(1)}, be imroduced as(1), seem like(1).}\\
\hline
\end{tabular}\end{center}\end{small}
\begin{small}\begin{center}\begin{tabular}{p{15cm}}
{\bf 0.77 ``sap tree'' MAKE$_1$}\\
{\scriptsize \textbf{MTurk:} \underline{produce(17)}, \underline{have(7)}, \underline{exude(5)}, \underline{contain(5)}, \underline{ooze(3)}, \underline{give(3)}, \underline{make(3)}, \underline{supply(2)}, \underline{create(2)}, \underline{secrete(2)}, $\ldots$}\\
{\scriptsize \textbf{Web:} \underline{produce(78)}, force(52), \underline{have(42)}, \underline{drip(33)}, \underline{exude(32)}, \underline{ooze(20)}, cause(18), get(17), \underline{secrete(16)}, yield(15), $\ldots$}\\
\hline
\end{tabular}\end{center}\end{small}
\begin{small}\begin{center}\begin{tabular}{p{15cm}}
{\bf 0.77 ``city folk'' IN}\\
{\scriptsize \textbf{MTurk:} \underline{live in(21)}, reside in(13), \underline{come from(8)}, \underline{inhabit(5)}, \underline{stay in(2)}, \underline{be from(2)}, emerge from(2), be located in(2), be found in(2), occupy(2), $\ldots$}\\
{\scriptsize \textbf{Web:} \underline{live in(245)}, run(26), be in(22), leave(22), move from(22), \underline{come from(19)}, \underline{work in(17)}, flee(17), make(16), populate(13), $\ldots$}\\
\hline
\end{tabular}\end{center}\end{small}
\begin{small}\begin{center}\begin{tabular}{p{15cm}}
{\bf 0.75 ``farm boy'' FROM}\\
{\scriptsize \textbf{MTurk:} \underline{work on(23)}, \underline{live on(22)}, \underline{come from(9)}, \underline{be born on(3)}, \underline{reside on(3)}, \underline{work in(3)}, \underline{grow up on(2)}, \underline{be raised on(2)}, toil on(2), labor on(2), $\ldots$}\\
{\scriptsize \textbf{Web:} \underline{live on(175)}, \underline{grow up on(103)}, \underline{work on(72)}, leave(67), remain on(13), \underline{be raised on(12)}, go from(11), be reared on(9), \underline{come from(8)}, live in(8), $\ldots$}\\
\hline
\end{tabular}\end{center}\end{small}
\begin{small}\begin{center}\begin{tabular}{p{15cm}}
{\bf 0.75 ``target structure'' BE}\\
{\scriptsize \textbf{MTurk:} \underline{be(10)}, \underline{have(4)}, \underline{include(3)}, \underline{contain(3)}, involve(3), \underline{represent(2)}, \underline{resemble(2)}, \underline{provide(2)}, \underline{become(2)}, \underline{be used as(1)}, $\ldots$}\\
{\scriptsize \textbf{Web:} \underline{be(513)}, serve as(68), make(45), \underline{become(34)}, \underline{represent(21)}, \underline{provide(18)}, form(18), constitute(13), host(13), mimic(12), $\ldots$}\\
\hline
\end{tabular}\end{center}\end{small}
\begin{small}\begin{center}\begin{tabular}{p{15cm}}
{\bf 0.73 ``peer judgments'' NOMINALIZATION:PRODUCT}\\
{\scriptsize \textbf{MTurk:} \underline{be made by(16)}, come from(11), be given by(4), involve(2), be rendered by(2), be from(2), originate from(1), be processed by(1), flow through(1), be handed down from(1), $\ldots$}\\
{\scriptsize \textbf{Web:} \underline{be made by(3)}, be had against(1).}\\
\hline
\end{tabular}\end{center}\end{small}
\begin{small}\begin{center}\begin{tabular}{p{15cm}}
{\bf 0.73 ``field mouse'' IN}\\
{\scriptsize \textbf{MTurk:} \underline{live in(22)}, come from(9), be found in(7), inhabit(7), reside in(4), dwell in(3), \underline{breed in(2)}, habitates(2), hide in(2), \underline{burrow in(1)}, $\ldots$}\\
{\scriptsize \textbf{Web:} \underline{live in(38)}, ravage(10), be(8), suggest that(8), have(6), \underline{burrow in(5)}, infest(5), occur in(4), extend(4), be deprived of(3), $\ldots$}\\
\hline
\end{tabular}\end{center}\end{small}
\begin{small}\begin{center}\begin{tabular}{p{15cm}}
{\bf 0.72 ``beehive hairdo'' BE}\\
{\scriptsize \textbf{MTurk:} \underline{look like(21)}, \underline{resemble(16)}, be shaped like(5), remind of(3), appear(2), emulate(2), \underline{be described as(1)}, be formed like(1), be referred to as(1), be modeled after(1), $\ldots$}\\
{\scriptsize \textbf{Web:} \underline{look like(4)}, \underline{be described as(4)}, \underline{resemble(2)}, be(1).}\\
\hline
\end{tabular}\end{center}\end{small}
\begin{small}\begin{center}\begin{tabular}{p{15cm}}
{\bf 0.71 ``vegetable soup'' HAVE$_1$}\\
{\scriptsize \textbf{MTurk:} \underline{contain(14)}, \underline{be made from(11)}, \underline{have(9)}, \underline{be made of(8)}, \underline{include(5)}, \underline{taste like(4)}, \underline{use(3)}, \underline{consist of(3)}, be composed of(3), come from(3), $\ldots$}\\
{\scriptsize \textbf{Web:} \underline{contain(45)}, \underline{have(33)}, \underline{include(32)}, \underline{be made with(21)}, be(19), \underline{use(15)}, be filled with(10), be loaded with(8), puree(7), \underline{consist of(5)}, $\ldots$}\\
\hline
\end{tabular}\end{center}\end{small}
\begin{small}\begin{center}\begin{tabular}{p{15cm}}
{\bf 0.70 ``food supplies'' NOMINALIZATION:PRODUCT}\\
{\scriptsize \textbf{MTurk:} \underline{include(12)}, \underline{contain(9)}, \underline{consist of(9)}, \underline{be(6)}, \underline{involve(4)}, be made of(4), be made up of(4), \underline{provide(3)}, be comprised of(3), \underline{have(2)}, $\ldots$}\\
{\scriptsize \textbf{Web:} \underline{include(58)}, \underline{be(34)}, \underline{make(17)}, \underline{consist of(11)}, \underline{provide(9)}, be used for(8), \underline{have(6)}, be derived from(6), mean that(5), flow around(5), $\ldots$}\\
\hline
\end{tabular}\end{center}\end{small}
\begin{small}\begin{center}\begin{tabular}{p{15cm}}
{\bf 0.70 ``sex scandal'' ABOUT}\\
{\scriptsize \textbf{MTurk:} \underline{involve(15)}, \underline{be about(9)}, concern(4), \underline{revolve around(3)}, \underline{include(3)}, come from(3), be caused by(3), relate to(2), deal with(2), contain(2), $\ldots$}\\
{\scriptsize \textbf{Web:} \underline{involve(46)}, go beyond(21), \underline{revolve around(3)}, college(3), turn into(3), be(3), \underline{include(2)}, \underline{have(2)}, originate as(2), make(2), $\ldots$}\\
\hline
\end{tabular}\end{center}\end{small}
\begin{small}\begin{center}\begin{tabular}{p{15cm}}
{\bf 0.69 ``drug deaths'' CAUSE$_2$}\\
{\scriptsize \textbf{MTurk:} \underline{be caused by(19)}, \underline{result from(7)}, be due to(6), occur because of(5), \underline{involve(2)}, \underline{come from(2)}, \underline{occur from(2)}, be related to(2), occur due to(2), be induced by(2), $\ldots$}\\
{\scriptsize \textbf{Web:} \underline{be caused by(25)}, be(14), \underline{involve(9)}, \underline{result from(9)}, stand for(8), be fueled by(7), mention(6), occur without(6), \underline{be blamed on(4)}, be associated with(4), $\ldots$}\\
\hline
\end{tabular}\end{center}\end{small}
\begin{small}\begin{center}\begin{tabular}{p{15cm}}
{\bf 0.69 ``college town'' HAVE$_1$}\\
{\scriptsize \textbf{MTurk:} \underline{have(15)}, \underline{contain(15)}, \underline{house(4)}, \underline{surround(4)}, \underline{host(3)}, \underline{support(3)}, \underline{be near(2)}, depend on(2), \underline{center around(1)}, \underline{include(1)}, $\ldots$}\\
{\scriptsize \textbf{Web:} \underline{have(94)}, \underline{be(22)}, \underline{house(10)}, dissect(8), attend(7), \underline{host(6)}, \underline{contain(5)}, \underline{support(5)}, \underline{center around(4)}, accredit(4), $\ldots$}\\
\hline
\end{tabular}\end{center}\end{small}
\begin{small}\begin{center}\begin{tabular}{p{15cm}}
{\bf 0.68 ``grain alcohol'' FROM}\\
{\scriptsize \textbf{MTurk:} \underline{be made from(16)}, come from(10), \underline{be produced from(7)}, be distilled from(5), contain(5), use(3), originate from(2), be made of(2), be fermented from(2), \underline{be derived from(1)}, $\ldots$}\\
{\scriptsize \textbf{Web:} \underline{be made from(13)}, \underline{be produced from(11)}, \underline{be manufactured from(6)}, \underline{be derived from(5)}, \underline{require(4)}, crystallize in(3), raise(3), be called(1), mean(1), produce from(1), $\ldots$}\\
\hline
\end{tabular}\end{center}\end{small}
\begin{small}\begin{center}\begin{tabular}{p{15cm}}
{\bf 0.68 ``flounder fish'' BE}\\
{\scriptsize \textbf{MTurk:} \underline{be(9)}, \underline{look like(6)}, \underline{resemble(3)}, be called(3), be classified as(3), taste like(2), be identified as(2), be named(1), smell like(1), group with(1), $\ldots$}\\
{\scriptsize \textbf{Web:} \underline{look like(9)}, \underline{be(4)}, \underline{resemble(4)}, include(2), be used include(2), substitute for(1), go with(1), be shaped like(1), follow(1), want(1).}\\
\hline
\end{tabular}\end{center}\end{small}
\begin{small}\begin{center}\begin{tabular}{p{15cm}}
{\bf 0.68 ``growth hormone'' CAUSE$_1$}\\
{\scriptsize \textbf{MTurk:} \underline{promote(11)}, \underline{cause(9)}, \underline{stimulate(6)}, \underline{affect(4)}, \underline{encourage(3)}, \underline{help(3)}, \underline{increase(2)}, \underline{regulate(2)}, \underline{enhance(2)}, \underline{produce(2)}, $\ldots$}\\
{\scriptsize \textbf{Web:} \underline{stimulate(403)}, \underline{regulate(222)}, \underline{promote(211)}, \underline{affect(105)}, \underline{control(67)}, \underline{cause(61)}, inhibit(31), \underline{influence(28)}, include(28), \underline{encourage(22)}, $\ldots$}\\
\hline
\end{tabular}\end{center}\end{small}
\begin{small}\begin{center}\begin{tabular}{p{15cm}}
{\bf 0.67 ``government land'' HAVE$_2$}\\
{\scriptsize \textbf{MTurk:} \underline{be owned by(23)}, belong to(15), \underline{be managed by(4)}, \underline{be controlled by(3)}, \underline{be possessed by(3)}, \underline{be administered by(2)}, come from(2), be maintained by(2), \underline{be occupied by(1)}, \underline{be regulated by(1)}, $\ldots$}\\
{\scriptsize \textbf{Web:} \underline{be owned by(205)}, be acquired by(80), have(58), be(41), be taken by(39), be purchased by(34), be granted by(27), be leased from(26), be held by(23), be acquired from(21), $\ldots$}\\
\hline
\end{tabular}\end{center}\end{small}
\begin{small}\begin{center}\begin{tabular}{p{15cm}}
{\bf 0.67 ``picture book'' HAVE$_1$}\\
{\scriptsize \textbf{MTurk:} \underline{contain(19)}, \underline{have(10)}, \underline{include(6)}, \underline{use(4)}, \underline{be composed of(3)}, \underline{be made up of(3)}, \underline{display(3)}, \underline{consist of(3)}, utilize(3), \underline{hold(2)}, $\ldots$}\\
{\scriptsize \textbf{Web:} \underline{have(1146)}, \underline{contain(366)}, \underline{include(345)}, give(290), \underline{show(201)}, \underline{be(157)}, paint(119), \underline{present(68)}, \underline{use(67)}, provide(60), $\ldots$}\\
\hline
\end{tabular}\end{center}\end{small}
\begin{small}\begin{center}\begin{tabular}{p{15cm}}
{\bf 0.66 ``film cutter'' NOMINALIZATION:AGENT}\\
{\scriptsize \textbf{MTurk:} \underline{cut(16)}, edit(11), slice(9), \underline{divide(3)}, trim(3), work with(3), snip(2), splice(2), \underline{chop up(1)}, \underline{separate(1)}, $\ldots$}\\
{\scriptsize \textbf{Web:} \underline{cut(20)}, see(7), \underline{divide(5)}, \underline{chop up(3)}, conform(3), \underline{separate(2)}, \underline{sever(2)}, perforate(2), act in(1), be reciprocated across(1), $\ldots$}\\
\hline
\end{tabular}\end{center}\end{small}
\begin{small}\begin{center}\begin{tabular}{p{15cm}}
{\bf 0.64 ``pet families'' HAVE$_1$}\\
{\scriptsize \textbf{MTurk:} \underline{have(14)}, \underline{own(13)}, \underline{love(8)}, \underline{care for(6)}, \underline{consist of(5)}, \underline{like(4)}, \underline{enjoy(4)}, be composed of(4), contain(4), \underline{include(2)}, $\ldots$}\\
{\scriptsize \textbf{Web:} \underline{have(187)}, \underline{adopt(84)}, lose(77), \underline{love(46)}, want(43), \underline{own(38)}, bring(35), \underline{keep(28)}, \underline{include(24)}, take(24), $\ldots$}\\
\hline
\end{tabular}\end{center}\end{small}
\begin{small}\begin{center}\begin{tabular}{p{15cm}}
{\bf 0.64 ``picture album'' FOR}\\
{\scriptsize \textbf{MTurk:} \underline{contain(22)}, \underline{hold(16)}, \underline{display(11)}, \underline{have(5)}, \underline{protect(4)}, \underline{show(3)}, \underline{store(3)}, organize(3), \underline{be filled with(2)}, show off(2), $\ldots$}\\
{\scriptsize \textbf{Web:} \underline{contain(318)}, \underline{have(310)}, \underline{include(193)}, need(42), be(36), \underline{hold(34)}, feature(31), \underline{show(29)}, \underline{be filled with(28)}, \underline{display(19)}, $\ldots$}\\
\hline
\end{tabular}\end{center}\end{small}
\begin{small}\begin{center}\begin{tabular}{p{15cm}}
{\bf 0.63 ``hand brake'' USE}\\
{\scriptsize \textbf{MTurk:} \underline{be operated by(11)}, \underline{be used by(7)}, \underline{require(6)}, \underline{use(5)}, be activated by(5), \underline{be made for(4)}, \underline{be applied by(3)}, \underline{operate by(3)}, \underline{be controlled by(3)}, need(3), $\ldots$}\\
{\scriptsize \textbf{Web:} \underline{be operated by(112)}, \underline{be applied by(14)}, \underline{be controlled by(6)}, be(6), be set by(5), be released by(5), require that(4), \underline{require(3)}, be of(3), secure(3), $\ldots$}\\
\hline
\end{tabular}\end{center}\end{small}
\begin{small}\begin{center}\begin{tabular}{p{15cm}}
{\bf 0.63 ``cactus plant'' BE}\\
{\scriptsize \textbf{MTurk:} \underline{be(12)}, \underline{look like(9)}, \underline{be called(2)}, \underline{grow like(2)}, be related to(2), contain(2), be classified as(2), \underline{resemble(1)}, originate from(1), be composed of(1), $\ldots$}\\
{\scriptsize \textbf{Web:} \underline{resemble(83)}, \underline{look like(77)}, \underline{be(38)}, include(13), \underline{grow like(7)}, be confused with(5), last(4), be among(4), associate with(3), range from(3), $\ldots$}\\
\hline
\end{tabular}\end{center}\end{small}
\begin{small}\begin{center}\begin{tabular}{p{15cm}}
{\bf 0.62 ``shock treatment'' USE}\\
{\scriptsize \textbf{MTurk:} \underline{use(16)}, utilize(7), \underline{involve(6)}, consist of(6), \underline{cause(3)}, \underline{include(2)}, \underline{induce(2)}, \underline{give(2)}, create(2), be comprised of(2), $\ldots$}\\
{\scriptsize \textbf{Web:} \underline{include(13)}, \underline{use(13)}, \underline{involve(8)}, be(8), \underline{cause(5)}, \underline{produce(4)}, remove(4), be used during(3), \underline{induce(2)}, ensure(2), $\ldots$}\\
\hline
\end{tabular}\end{center}\end{small}
\begin{small}\begin{center}\begin{tabular}{p{15cm}}
{\bf 0.62 ``head noun'' BE}\\
{\scriptsize \textbf{MTurk:} \underline{be(6)}, describe(3), \underline{involve(2)}, be found in(2), refer to(2), be at(2), come at(2), \underline{stand at(1)}, \underline{have(1)}, \underline{form(1)}, $\ldots$}\\
{\scriptsize \textbf{Web:} \underline{be(149)}, \underline{form(24)}, function as(16), occur as(14), appear as(13), modify(10), \underline{act as(7)}, serve as(7), precede(5), be analysed as(4), $\ldots$}\\
\hline
\end{tabular}\end{center}\end{small}
\begin{small}\begin{center}\begin{tabular}{p{15cm}}
{\bf 0.62 ``designer creations'' NOMINALIZATION:PATIENT}\\
{\scriptsize \textbf{MTurk:} \underline{be made by(20)}, \underline{come from(6)}, be designed by(6), \underline{be produced by(5)}, \underline{be created by(4)}, \underline{be presented by(2)}, \underline{be from(2)}, originate from(2), be developed by(2), be thought of by(2), $\ldots$}\\
{\scriptsize \textbf{Web:} \underline{be made by(5)}, \underline{be produced by(4)}, be selected with(4), \underline{be created by(3)}, leave(3), require(3), \underline{come from(2)}, \underline{be from(2)}, land(2), design(2), $\ldots$}\\
\hline
\end{tabular}\end{center}\end{small}
\begin{small}\begin{center}\begin{tabular}{p{15cm}}
{\bf 0.62 ``tear gas'' CAUSE$_1$}\\
{\scriptsize \textbf{MTurk:} \underline{cause(18)}, \underline{create(9)}, \underline{make(7)}, generate(5), \underline{produce(3)}, \underline{induce(3)}, provoke(2), initiate(1), lead to(1), encourage(1), $\ldots$}\\
{\scriptsize \textbf{Web:} \underline{cause(13)}, choke as(8), be like(6), include(4), tamponades(4), be(4), \underline{induce(3)}, be made from(3), trigger(2), bring(2), $\ldots$}\\
\hline
\end{tabular}\end{center}\end{small}
\begin{small}\begin{center}\begin{tabular}{p{15cm}}
{\bf 0.61 ``country butter'' FROM}\\
{\scriptsize \textbf{MTurk:} \underline{be made in(18)}, \underline{come from(14)}, \underline{be produced in(6)}, be associated with(3), be from(3), originate in(2), be churned in(2), taste(1), be styled after(1), stand for(1), $\ldots$}\\
{\scriptsize \textbf{Web:} \underline{be produced in(12)}, \underline{come from(8)}, \underline{be made in(3)}, be preferred in(3), be imported into(3), be picked up around(2), be consumed in(2), be celebrated around(2), enter(1), leave(1), $\ldots$}\\
\hline
\end{tabular}\end{center}\end{small}
\begin{small}\begin{center}\begin{tabular}{p{15cm}}
{\bf 0.60 ``automobile plant'' FOR}\\
{\scriptsize \textbf{MTurk:} \underline{make(17)}, \underline{assemble(15)}, \underline{manufacture(15)}, \underline{build(11)}, \underline{produce(10)}, create(3), \underline{turn out(2)}, put together(2), construct(2), manufatures(1), $\ldots$}\\
{\scriptsize \textbf{Web:} \underline{produce(16)}, be utilized in(14), \underline{build(7)}, \underline{manufacture(7)}, \underline{assemble(5)}, include(5), \underline{make(3)}, \underline{turn out(2)}, be used in(2), process(2), $\ldots$}\\
\hline
\end{tabular}\end{center}\end{small}
\begin{small}\begin{center}\begin{tabular}{p{15cm}}
{\bf 0.59 ``winter season'' BE}\\
{\scriptsize \textbf{MTurk:} \underline{be(7)}, \underline{include(6)}, \underline{occur in(5)}, \underline{be in(5)}, \underline{occur during(4)}, \underline{feel like(2)}, \underline{happen during(2)}, \underline{be called(2)}, \underline{come in(2)}, exist in(2), $\ldots$}\\
{\scriptsize \textbf{Web:} \underline{be(128)}, \underline{occur in(20)}, be during(17), \underline{be in(15)}, follow(15), \underline{coincide with(14)}, \underline{be called(13)}, \underline{include(12)}, join(10), run from(9), $\ldots$}\\
\hline
\end{tabular}\end{center}\end{small}
\begin{small}\begin{center}\begin{tabular}{p{15cm}}
{\bf 0.59 ``rye whiskey'' FROM}\\
{\scriptsize \textbf{MTurk:} \underline{be made from(18)}, come from(7), \underline{contain(5)}, be made of(5), be brewed from(4), \underline{be distilled from(3)}, use(3), be derived from(2), be formulated from(2), be created from(2), $\ldots$}\\
{\scriptsize \textbf{Web:} \underline{be distilled from(4)}, \underline{be made from(3)}, \underline{contain(1)}, call(1), accent(1), have(1).}\\
\hline
\end{tabular}\end{center}\end{small}
\begin{small}\begin{center}\begin{tabular}{p{15cm}}
{\bf 0.58 ``family problems'' IN}\\
{\scriptsize \textbf{MTurk:} \underline{involve(5)}, \underline{affect(5)}, \underline{occur within(3)}, \underline{plague(3)}, \underline{happen in(3)}, \underline{be experienced by(2)}, \underline{occur in(2)}, \underline{be found in(2)}, \underline{be solved by(2)}, \underline{arise in(2)}, $\ldots$}\\
{\scriptsize \textbf{Web:} \underline{affect(318)}, \underline{plague(120)}, \underline{arise in(90)}, bring(90), confront(79), \underline{occur in(51)}, \underline{run in(47)}, \underline{involve(41)}, cause(39), threaten(33), $\ldots$}\\
\hline
\end{tabular}\end{center}\end{small}
\begin{small}\begin{center}\begin{tabular}{p{15cm}}
{\bf 0.56 ``sports activities'' BE}\\
{\scriptsize \textbf{MTurk:} \underline{involve(12)}, \underline{include(7)}, \underline{be(4)}, be related to(4), \underline{use(3)}, \underline{revolve around(2)}, \underline{be comprised of(2)}, contain(2), \underline{play(1)}, \underline{focus on(1)}, $\ldots$}\\
{\scriptsize \textbf{Web:} \underline{include(494)}, \underline{be(118)}, range from(70), \underline{involve(62)}, \underline{use(29)}, be considered(29), bring(16), organize(13), be classified as(13), \underline{be associated with(12)}, $\ldots$}\\
\hline
\end{tabular}\end{center}\end{small}
\begin{small}\begin{center}\begin{tabular}{p{15cm}}
{\bf 0.55 ``worker teams'' MAKE$_2$}\\
{\scriptsize \textbf{MTurk:} be made up of(11), \underline{include(10)}, \underline{be composed of(6)}, be comprised of(6), \underline{have(5)}, \underline{consist of(5)}, \underline{contain(4)}, \underline{comprise(3)}, \underline{employ(3)}, be made from(3), $\ldots$}\\
{\scriptsize \textbf{Web:} \underline{include(407)}, \underline{consist of(94)}, be(31), \underline{have(13)}, \underline{bring(10)}, provide(10), give(8), allow(8), \underline{incorporate(7)}, need(7), $\ldots$}\\
\hline
\end{tabular}\end{center}\end{small}
\begin{small}\begin{center}\begin{tabular}{p{15cm}}
{\bf 0.54 ``mining engineer'' FOR}\\
{\scriptsize \textbf{MTurk:} \underline{work in(9)}, study(9), \underline{specialize in(6)}, know about(6), \underline{do(3)}, \underline{work with(3)}, design(3), \underline{understand(2)}, \underline{know(2)}, investigate(2), $\ldots$}\\
{\scriptsize \textbf{Web:} \underline{specialize in(15)}, \underline{work in(10)}, be(7), stay within(6), work for(6), \underline{do(5)}, \underline{work with(5)}, \underline{build(3)}, select(3), \underline{understand(2)}, $\ldots$}\\
\hline
\end{tabular}\end{center}\end{small}
\begin{small}\begin{center}\begin{tabular}{p{15cm}}
{\bf 0.54 ``bear country'' HAVE$_1$}\\
{\scriptsize \textbf{MTurk:} \underline{have(11)}, contain(9), be inhabited by(6), house(4), harbor(4), support(4), feed(3), \underline{protect(2)}, \underline{supply(2)}, \underline{sustain(2)}, $\ldots$}\\
{\scriptsize \textbf{Web:} \underline{have(33)}, do(15), \underline{protect(10)}, be(6), go(5), \underline{sustain(3)}, be screened in(3), deliver(3), see as(3), allow(3), $\ldots$}\\
\hline
\end{tabular}%
\end{center}\end{small}

\newpage
\section{Comparison by Noun Compound: Using The First Verb Only}
\label{sec:verbs:first}
\vspace{6pt}

\begin{small}\begin{center}%
\end{center}\end{small}

\newpage
\section{Comparison by Class: Using All Human-Proposed Verbs}
\label{sec:verbs:all:class}
\vspace{6pt}

\begin{small}\begin{center}\begin{tabular}{p{15cm}}
{\bf 0.89 CAUSE$_1$}\\
\hline
\end{tabular}%
\end{center}\end{small}

\newpage
\section{Comparison by Class: Using The First Verb Only}
\label{sec:verbs:first:class}
\vspace{6pt}

\begin{small}\begin{center}%
\end{center}\end{small}

\end{document}